\documentclass[11pt, a4paper, oneside]{Thesis} 

\graphicspath{{./Pictures/}} 

\usepackage[utf8]{inputenc}
\usepackage[square, numbers, comma, sort&compress]{natbib} 
\hypersetup{urlcolor=blue, colorlinks=true} 
\title{\ttitle} 

\begin{document}

\frontmatter 

\setstretch{1.3} 

\fancyhead{} 
\rhead{\thepage} 
\lhead{} 

\pagestyle{fancy} 

\newcommand\numberthis{\addtocounter{equation}{1}\tag{\theequation}}
\newcommand{\HRule}{\rule{\linewidth}{0.5mm}} 

\hypersetup{pdftitle={\ttitle}}
\hypersetup{pdfsubject=\subjectname}
\hypersetup{pdfauthor=\authornames}
\hypersetup{pdfkeywords=\keywordnames}


\pagestyle{empty} 


\begin{titlepage}
\begin{center}

\includegraphics[height=0.13\textheight]{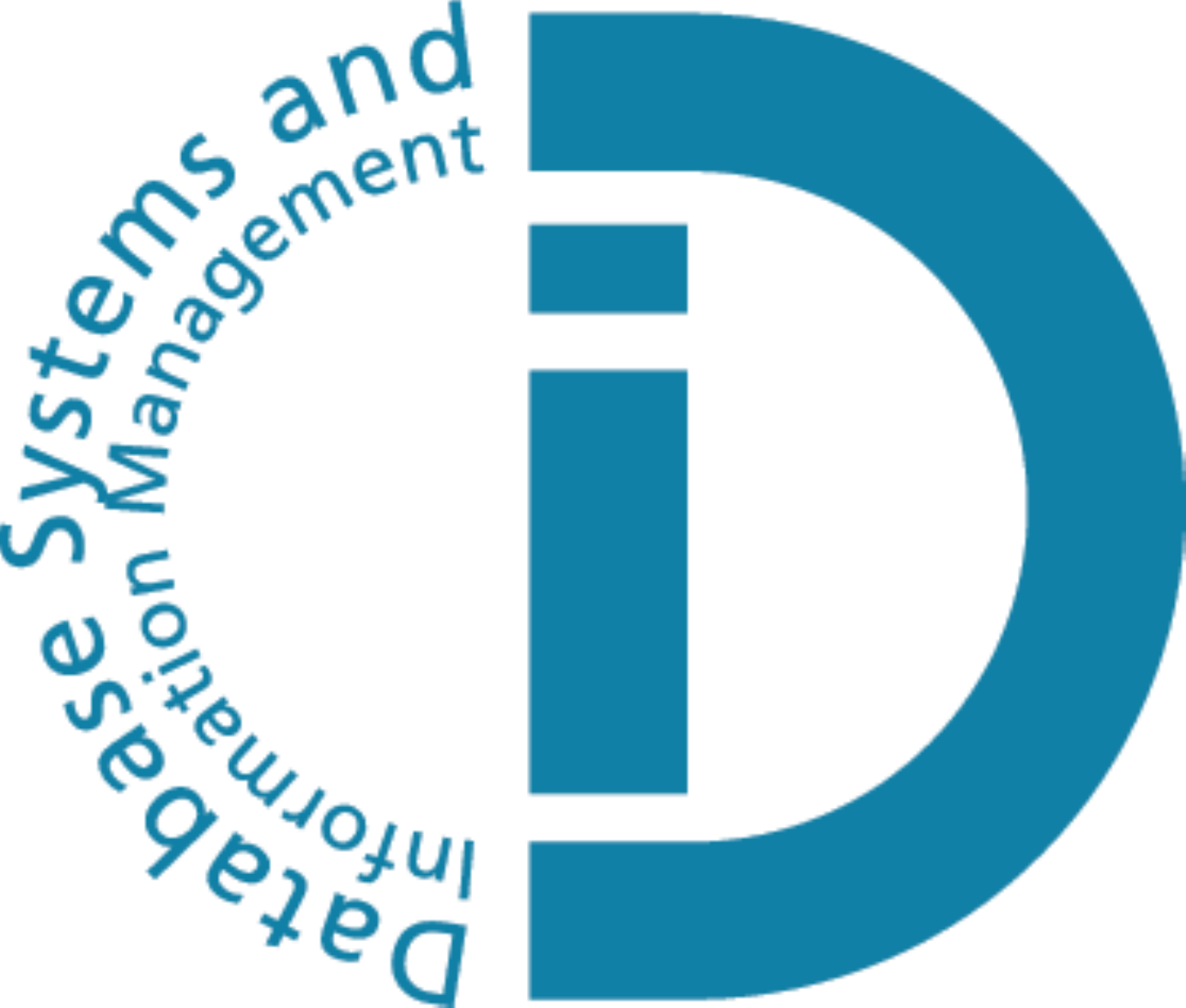}
\hfill
\includegraphics[height=0.15\textheight]{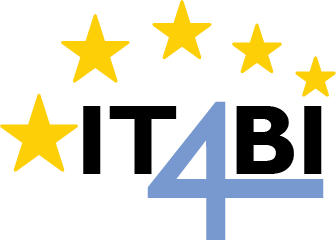}
\hfill
\includegraphics[height=0.13\textheight]{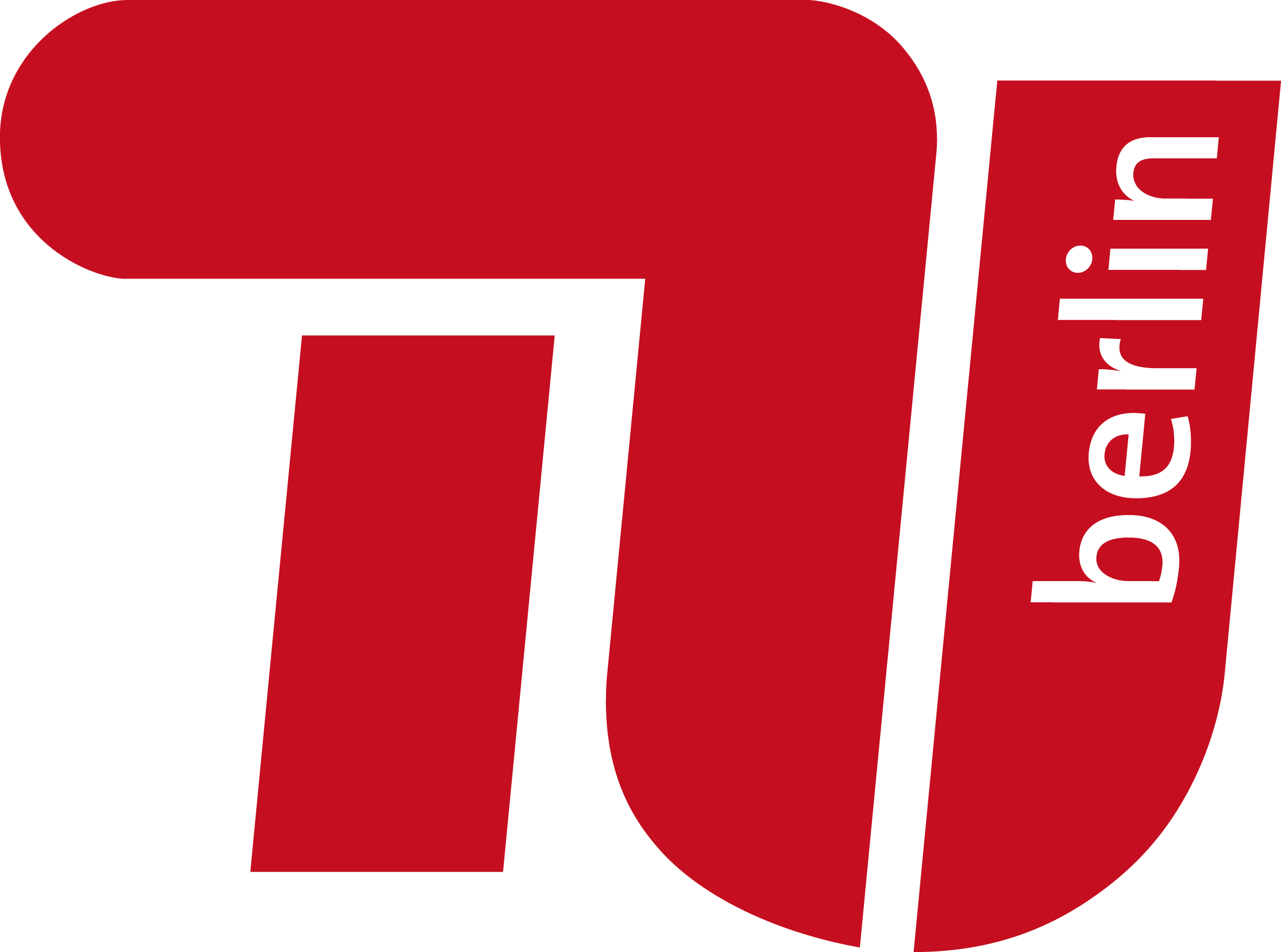}

\vspace*{1.5cm}

\LARGE
\textsc{Online Machine Learning Techniques for Predicting Operator Performance}

\vspace{1cm}

\Large \textsc{Master's Thesis}

\vspace{0.2cm}

by

\vspace{0.4cm}

\textsc{Ahmet Anil Pala}

\vspace{1.0cm}

\vfill

\large 
Submitted to the Faculty IV, Electrical Engineering and Computer Science
Database Systems and Information Management Group
in partial fulfillment of the requirements for the degree of

\textbf{Master of Science in Computer Science}

as part of the ERASMUS MUNDUS programme IT4BI

at the 

\textsc{Technische Universit\"{a}t Berlin} \\

August 15, 2015

\vfill

\begin{flushright} 
\normalsize 
\emph{Thesis Advisor:}\\
\textsc{Max Heimel} \\

\ \\

\emph{Thesis Supervisor:}\\
Prof. Dr. \textsc{Volker Markl}\\
\end{flushright}

\end{center}
\end{titlepage}

\newpage

\ \\ 

\vspace{1.0cm}

\textbf{Eidesstattliche Erkl\"arung}

Ich erkl\"are an Eides statt, dass ich die vorliegende Arbeit selbstst\"andig 
verfasst, andere als die angegebenen Quellen/Hilfsmittel nicht benutzt, 
und die den benutzten Quellen w\"ortlich und inhaltlich entnommenen Stellen 
als solche kenntlich gemacht habe.

\ \\

\textbf{Statutory Declaration}

I declare that I have authored this thesis independently, that I have not used 
other than the declared sources/resources, and that I have explicitly marked 
all material which has been quoted either literally or by content from the used sources.

\ \\ \ \\

\begin{flushright} 

Berlin, August 15, 2015

\ \\

\textsc{Ahmet Anil Pala}

\end{flushright}

\newpage

\pagestyle{empty} 

\null\vfill 



\vfill\vfill\vfill\vfill\vfill\vfill\null 

\clearpage 


\addtotoc{Abstract} 

\abstract{\addtocontents{toc}{\vspace{1em}} 

This thesis explores a number of online machine learning algorithms. From a theoretical perspective, it assesses their employability for a particular function approximation problem where the analytical models fall short. Furthermore, it discusses the application of theoretically suitable learning algorithms to the function approximation problem at hand through an efficient implementation that exploits various computational and mathematical shortcuts. Finally, this thesis work evaluates the implemented learning algorithms according to various evaluation criteria through rigorous testing.
}

\clearpage 

\addtotoc{Abstract} 

\abstract{\addtocontents{toc}{\vspace{1em}} 

Die vorliegende Arbeit untersucht eine Reihe von Online-Algorithmen des Maschinellen Lernens. Aus theoretischer Perspektive wird eine Einschätzung zur Anwendbarkeit dieser Algorithmen für ein bestimmtes Funktionsapproximationsproblem gegeben, bei dem analytische Modelle unzureichend sind. Des Weiteren wird die Anwendung der theoretisch adäquaten Lernalgorithmen auf das gestellte Problem durch eine effiziente Implementierung diskutiert, die diverse mathematische Abkürzungen aufzeigt. 
}

\clearpage 


\setstretch{1.3} 

\acknowledgements{\addtocontents{toc}{\vspace{1em}} 

I would like to express my honest gratitude to my advisor, Max Heimel, who guided me through the thesis work with intellectually stimulating discussions on the topics of the thesis.

I would like to thank Prof. Volker Markl for helping me choose my final year specialization offered in Technische Universit\"at Berlin with his inspiring presentation on the specialization. I would also like to thank all the professors who lectured me during my studies.

Finally, I would like to thank my family for always being supportive of me throughout my entire life.

}
\clearpage 


\pagestyle{fancy} 

\lhead{\emph{Contents}} 
\tableofcontents 

\lhead{\emph{List of Figures}} 
\listoffigures 

\lhead{\emph{List of Tables}} 
\listoftables 

\mainmatter 

\pagestyle{fancy} 



\chapter{Introduction} 

\label{Chapter1} 

\lhead{Chapter 1. \emph{Introduction}} 


\section{Motivation}
\label{motivation}

Never-ending demand for faster data processing has compelled the database community to look for creative ways to accelerate database operations. In a world where the computing hardware is getting more diverse, the idea of involving more diverse hardware to achieve better performance by exploiting parallelism among different processing units gave birth to the use of heterogeneous hardware for data processing. However, there is a price to pay for bringing the hardware heterogeneity into data-processing before enjoying its benefits. Ensuring the program portability of software across different platforms is obviously the first challenge to be dealt with. Frameworks such as OpenCL (Open Computing Language) is developed for this purpose and it is also employed by hardware-oblivious databases \cite{heimel_hardware-oblivious_2013}. However, ensuring the portability is not the only challenge to be tackled. Exploding number of combinations of hardware which could be present together is a game-changer for database-tuning as it makes the hand-tuning of database engines for the all possible hardware combinations more tedious task than ever. This suggests that smarter approaches to database-tuning is necessary. Self-adaptive database engines which can tune themselves to underlying hardware on the fly is the new direction database-tuning is veering off.

Hardware-oblivious self-adaptive databases typically employ machine learning approaches for self-tuning. Tuning procedure involves two decision-making processes. First, enabling database engine to make good decisions on which hardware to use for a given operation. In \cite{heimel_demonstrating_2014}, online learning algorithms are used to discover decision boundaries for choosing the most suitable processing unit at the execution engine’s disposal for offloading operator data on the fly. Second decision process involves choosing the most suitable algorithm for a given task on the hardware that was selected for that task. Fitness criterion used in choosing among different alternatives is typically the running time of the candidate hardware/algorithm given an operator such as scan, join, etc. Therefore, An operator performance estimator is needed which aforementioned decision-making processes can rely on.

From the machine learning perspective, operator performance estimation is a regression task. In a self-adaptive database-optimization, dynamics of the system to be learned continuously changes as a result of complex interplay of the decision-making processes mentioned. This entails dealing with a dynamic environment necessitating the use of online learning approaches rather than offline alternatives. Another reason for online learning stems from a more practical reason. As the performance estimating is rather a tool for better decision-making, it is too secondary to be treated as a standalone learning problem with separate training and testing phases. Therefore, the performance estimator module is expected to estimate the performance and learn on the fly as more queries are passed to the database and processed.

This thesis explores, analyses, evaluates and compares different online machine learning techniques for the online regression problem posed by a hardware-oblivious self-tuning database.

\section{Structure of This Thesis}

This thesis consists of 7 chapters including the \textit{Introduction}. Chapter \ref{Chapter2}  recaps the theoretical foundations which is essential for a good understanding of the thesis work done. This includes topics such as Query Cost Modeling, Statistical Learning Theory and Data Stream Mining. Chapter \ref{Chapter3} presents the problem for which an effective and feasible solution is searched. Chapter \ref{Chapter4} discusses the theoretical details of the suitable approaches to be employed for the problem. A thorough investigation of the existing regression algorithms that can be made to operate in an online fashion is presented in this chapter. Chapter \ref{Chapter5} discusses the implementation details of the online regression algorithms found theoretically suitable for the problem. Chapter \ref{Chapter6} presents evaluation criteria for evaluating the implemented online regression algorithms discussed. It also defines various metrics for testing the employability of the algorithms for the problem at hand according to the evaluation criteria presented. Furthermore it visualizes some of the experiments carried out and draw conclusions about the nature of the online learning algorithms. Chapter \ref{Chapter7} presents a general comparison table for the online learning algorithms that are evaluated in Chapter \ref{Chapter6} in the light of the experiment results presented. Moreover, it discusses the potential improvements and future work on online regression.


\chapter{Foundations} 

\label{Chapter2} 

\lhead{Chapter 2. \emph{Foundations}} 

\section{Query Cost Modeling}

Query cost modeling is the primary task for query performance prediction. The way the queries and the operators which make up them are represented has profound implications on the quality of performance estimations. As the cost depends on the operator algorithms and since the time complexity of the algorithms used for query operators are well-known, intuitively speaking, cost of a particular operator can be easily calculated by evaluating the complexity formula given input sizes in number of tuples and number of blocks. Such analytical cost models predicting the number of I/O operations are commonly employed by query optimizers to select cheapest plan among different alternatives. 

\begin{ex}
Consider the join of two relations namely $R(A,B)$ and $S(A,C)$ on the column $A$. Block sizes of $R$ and $S$ are estimated to be 100 and 150 respectively and number of tuples in the relations are estimated to be 1500 and 2000 each respectively. The analytical cost model for block-based nested loop join is $B(R) + B(S)\times B(R)/(M-1)$ I/Os. As for tuple-based nested loop join, the analytical cost model is $T(R)\times T(S)$ I/Os. With these cost models and the sizes of the relations (in number of blocks and in number of tuples occur in the blocks), one can calculate the cost estimations in number of I/Os as follows. 
\begin{align*}
	\text{cost(block-based join)} & = B(R) + B(S)\times B(R)/(M-1) = 100 + 150\times(100/(M-1)) \\ 
	\text{cost(tuple-based join)} & = T(R)\times T(S) = 1500\times2000
\end{align*} 
\end{ex}
The estimations made by analytical cost models can be good enough to compare algorithms and decide which algorithm runs faster for plan selection. However, for making accurate predictions on the running time of an operator, use of such analytical models often results in inaccurate running time estimations. This is because, the actual cost depends so many environment-dependent parameters that are not possible to be hardcoded into an analytical cost model. Among such environment-dependent parameters are the hit ratio of various caches exist in the computing hardware such as instruction and memory caches of CPUs, cache of non-volatile memory, the number of available buffers at the moment the operator gets to execute, hard drive disk head motion speed characteristics, etc. In heterogeneous environments where the different hardware with different performance characteristics, a reliable analytical cost-model would have to consider even more things such as inter-device transfer speed, I/O costs for the selected device (note that I/O operations can behave very differently depending on the hardware) and so on. As coming up with a reliable analytical cost-model is very difficult and often impossible to achieve, model-learning techniques are employed to learn the underlying environment-dependent cost-models.

When learning approaches are employed, typically two different ways of modeling queries are used namely coarser plan-level models and finer-granular operator-level models. The choice between the two depends on the application for which the cost models will be used. While some applications can utilize both types of query cost models, some others strictly require finer-granular operator-models. For instance, for the applications where the accurate measurement of the total cost of the queries is the only major point of interest such as join ordering or plan-selection, the plan-level cost-models can be used as well as the operator-level models. On the other hand, some other applications might need separate estimations of each different operator which queries are composed of explicitly. For example, an operator offloader module of a database engine deployed on a distributed system needs running times of different operators rather than the total running time estimation of a query to be able to decide which hardware to offload the operators to. \cite{akdere_learning-based_2012} discusses both types of query models and compares them in terms of their capability of making accurate predictions of the total query running time. In addition to these two models that sit at the opposite ends of the spectrum in terms of their modeling granularity, they explore the hybrid models where some portions of the execution tree of a query are considered as a separate operator and modeled as a whole and the operators of the rest of the execution tree are modeled one by one in finer-granular fashion as in the operator-level planning. In this thesis, only the operator-level cost models are the point of interest. {\it{Problem}} section explores the details of the learning problem to be tackled and makes it obvious why strictly the cost models with finer-granulation is needed. 

Regardless of the granularity of the cost model to be learned, from the machine learning perspective, the big picture remains the same. There is an {\it{unknown}} function which is too complex to be modeled analytically and from the past observations of the inputs and the outputs of this function, we attempt to build predictive models to estimate the output of the unknown function given an input.

\section{Statistical Learning}

As Vapnik describes in {\citep[pp. 17-19]{vapnik_nature_2000}, the main objective of a learning process is to minimize the risk functional
\begin{flalign} 
\label{def_2.1}
R_{\alpha}=\int L(y,f(\pmb{x},\alpha)d(p(\pmb{x},y)), \quad f \in \mathbb{F}
\end{flalign} 
on the i.i.d sample of $(\pmb{x},y)$ values where $f$ is the function that maps inputs to target space, $\alpha$ denotes the variable by which the function selection is parameterized in the function space $\mathcal{F}$ that $f$ belongs to, $L$ is the loss function which computes the amount of penalty incurred by comparing the target estimation and the target and finally $p$ is the (unknown) function that returns the joint probability density for a given input point $\bf{x}$ and target $y$.

Risk minimization framework very neatly distinguishes the different components of the learning process. More specifically, the criteria that assigns loss values to the individual predictions is abstracted away from the choice of the function space $\mathcal{F}$ where the optimization over $\alpha$ is done. These two different integral parts of the Risk Minimization framework can be adapted to different classes of learning problems such as \textit{Regression Estimation}, \textit{Density Estimation} and \textit{Pattern Recognition}. Although, these problems are fairly different in their nature, the same risk minimization framework can be used for them thanks to the flexibility of the framework.

\subsection{Choice of Loss Function}

Different kinds of learning problems typically use different loss functions to measure the amount of deviation from what is considered to be \textit{accurate} in the context of a particular learning problem. In the case of classification problems, mostly the loss function used is \textit{0-1 Loss Function}. This function either penalizes an individual prediction by recording the value of 1 as the penalty incurred or not by simply returning 0. For the learning scenarios where the prediction error cannot be easily measured by a quantitative measure or it could be but what matters is whether a prediction is approximately accurate or not rather than how \textit{much} it is off, 0-1 loss function could be a good choice. On the other hand, in some other class of learning problems such as regression estimation, since the amount of error in the prediction is an important consideration and it can be measured quantitatively, the loss functions like \textit{$L_1$ loss} and \textit{$L_2$ loss} are commonly employed.
\begin{align*}
L_1(\pmb{x}, y) = |y-f(\pmb{x}, \alpha)| \qquad L_2(\pmb{x}, y) = (y-f(\pmb{x}, \alpha))^2
\end{align*}
More sophisticated loss functions are also available namely Hinge Loss, $\epsilon$-insensitive loss, Huber's loss function, etc. \citep[pp. 74-75]{gao_probabilistic_2002}, \citep[pp. 6-7]{rosasco_are_2004}

\subsection{Function Space}

Another nice feature of the risk minimization framework is the use of the notion of function space. The function space $\mathcal{F}$ to which $f$ belongs along with the parametrization variable $\alpha$ is defined according to the characteristics of the learning problem to be dealt with. For example, in the case of a classification problem with three possible classes and two boolean features, $\mathcal{F}$ would be the set of all the functions with the range set of possible classes and the domain of all the possible configurations of the boolean features and an $\alpha$ variable can be defined to \textit{scan} the function space. On the other hand, in \textit{Regression Estimation} problems, the range is the set of real numbers and the domain is the set of real-valued n-tuples where $n$ is the number of input dimensions.
\begin{ex}
\begin{align*}
&Range=\{red,green,blue\} & \\
&Domain=\{(false,false),(false,true),(true,false),(true,true)\}&
\end{align*}
For the range and domain sets defined above, the function space $\mathcal{F}$, for the corresponding learning scenario, is the one that exhaustively includes all the possible mappings from the domain set to range set.

$$\mathcal{F}=\{f_{\alpha} : \forall f_{\alpha} : Domain \mapsto Range\} $$

One can parametrize the set $\mathcal{F}$ by a parameter variable $\alpha$. This variable does not have to be single-valued or numerical, it is just a notational device. For the domain and range sets specified above, a simple way to parametrize the function space $\mathcal{F}$, is using n-tuples that simultaneously take on $n$ different values where $n$ denotes the cardinality of the domain set and $m$ denotes that of the range set. 
For instance, $\alpha = (r,r,g,b)$ defines the mapping $f_{(r,r,g,b)}$ that maps the $(false, false)$ to the $red$, the $(false,true)$ to $red$, the $(true, false)$ to $green$ and $(true,true)$ to $blue$.

\end{ex}

The function space $\mathcal{F}$ depends solely on the range and domain sets. In the toy example given above, these sets were countable but in general they don't have to be so. For example in the case both domain and range are real-numbers, one can still imagine a function space which consists of all the real-valued functions. Obviously, then the parameterizing the functions space by indexing as it is done above for the discrete case would not be possible due to the unaccountability of the set that defines the function space. However, this does not mean one cannot find a parametrization variable $\alpha$ that allows to parametrize at least a subset of the function space.

\begin{ex} 
\begin{align*}
&Range=\mathbb{R} \quad Domain=\{\pmb{x}=[x_1,x_2,x_3]^\top :\forall \pmb{x} \in \mathbb{R}^3\}&
\end{align*}
Consider above range and domain sets. Here, the cardinality of the function space $\mathcal{F}$ that contains all the possible mappings from the domain set to the range is uncountable. By contrast to the previous example, there is not a way to easily come up with a parametrization trick that enables us to index all the functions in $\mathcal{F}$.
\end{ex}

Uncountability of the functions in the functions space is very common (e.g all the regression problems) and it poses a challenge when searching for the function that minimizes the risk. However, if the function that generates the data is assumed to belong to a certain family of functions, things are easier. 

\begin{ex} 
\label{ex:regression_formulated}
\begin{align*}
&\pmb{x} = [x_1, x_2]^\top, \quad y = r(\pmb{x}) + \epsilon, \quad y = N(r(\pmb{x}),\sigma_y^2), \quad \epsilon  = N(0,\sigma_y^2) & \\
&r = f(\pmb{x},\alpha_{r})
\end{align*}
Imagine a learning scenario with two real-numbered inputs and one real number target. This learning problem is an example of \textit{multiple regression}. Let $r$ denote the function used for generating the data. Moreover, the data generation process was not noise-free and there is Gaussian distributed noise in the produced data with zero mean and $\sigma_y$ standard deviation. In fact, $r$ is a function that belongs to the function space in consideration under the risk minimization framework (In this case it is the Hilbert space). And the function $r$ can be indexed by the parameter variable $\alpha$. If we assume that $r$ is a linear function, then we can write $y=N(\pmb{w}^{\top} \pmb{x}, \sigma_y)$ where $w = [a,b]^\top$ and since different choices of $a$ and $b$ values allows us to pick any linear function possible, we can use this pair of variables as $\alpha$. Thus, we have $\alpha=\{a,b\}$. However, now the problem is by varying $\alpha$ we do not scan all the functions in the function space defined according to the input and output specified for the multiple regression problem. But do we really need that?
\end{ex} 

When optimizing the risk, the integration of risk functional will yield large numbers for the function $f$ if the domain-range pairs mapped by $f$ have low probability of occurring by the the distribution of the targets given inputs of the data generation process $y = N(r(\pmb{x}),\sigma{_y})$ and the rest of the possible matchings between the domain and range elements that are not mapped by $f$ have a high probability. On the other hand, the functions that generates domain-range mappings that conform to the data distribution will yield less risk. Therefore, considering only the portion of the function space that are \textit{assumed} to entail less risk than any other portion of it will not change the result of the optimization over $\alpha$. In other words, As the risk is a function of $\alpha$ and the optimization task is to determine the $\alpha$ that makes the risk smallest, one can come up with some parameterization of the function space that spans only the functions that produce the mappings which conform to the assumed nature of data generation (which is linear in above example). 

The idea of parameterizing a family of functions and optimizing the risk functional by picking a function from the assumed family is attractive as it makes it possible to parameterize an uncountable domain making the risk minimization possible. This is the motivation behind the \textit{parametric} learning models that are commonly used in machine learning applications. However, there is a catch with assuming a data distribution. What if the assumption was wrong and the data was generated through a function that is not being considered in the risk minimization? For example, if the data was generated through a quadratic function and only the linear functions are considered in the optimization over $\alpha$ then $r = f(\pmb{x},\alpha)$ will not hold for any $\alpha$. Briefly, an erroneous (under)assumption about the data results in missing the risk-minimizing function during the risk optimization. This is called \textit{underfitting} and it is a crucial problem that parametric models suffer from. Some examples of underfitting from the experiments carried out during this thesis work is presented in the Chapter \ref{Chapter6}. 

\subsection{Empirical Risk Minimization (ERM)}
\label{subsection:ERM}

So far, we have assumed that, we can actually evaluate the integral equation in the risk minimization. However, in reality, this is not possible, because we do not actually know the probability density function that appear in the integral.

Vapnik, in \citep[pp. 20-21]{vapnik_nature_2000}, discusses \textit{Empirical Risk Minimization Inductive Principle}. He defines a functional called empirical risk as follows.
\begin{flalign} 
& R_{emp}(\alpha)=\frac{1}{l}\sum_{i=1}^{l}L({\pmb{x},y})
\end{flalign}
By minimizing this functional, on a finite i.i.d $(\pmb{x},y)$ sample over $\alpha$, one can approximate the optimal $\alpha$ value that could be obtained from the risk functional. This is called \textit{consistency} of a learning process \citep[pp. 35-38]{vapnik_nature_2000}. A possible interpretation of consistency is as follows: For a risk minimization method (e.g ERM), if the risk minimized converges to the same \textit{minimal} risk value calculated by the equation \ref{def_2.1} as the size of the i.i.d $(\pmb{x},y)$ sample, $l$, goes to infinity, then the learning minimization method is said to be \textit{consistent}.
\begin{flalign} 
& \underset{l\rightarrow \infty} \lim  R_{emp}(\alpha)=R(\alpha) \label{def_2.3}
\end{flalign}
The details of the proof that consistency of ERM method is available in \citep{vapnik_nature_2000}
%
%

ERM is very commonly used method in machine learning and many classical methods can be derived from it simply by substituting a specific loss function into the ERM risk equation. In the case of $L_1$ loss function, one obtains the standard least-squares formula out of the empirical risk functional.
\begin{flalign} 
& R_{emp}(\alpha)=\frac{1}{l}\sum_{i=1}^{l}(f(\pmb{x}, \alpha) - y)^2
\end{flalign}

ERM principle has a serious problem. When the sample does not reflect the characteristics of the underlying unknown data distribution, the function corresponds to the optimal $\alpha$ value that minimizes the empirical risk is very unlikely to be the one used by the data generation. This is due to the overassumptions of the ERM method. First overassumption is that ERM regards all the data points included in the sample to weigh equally when calculating the risk which might not be the case since the distribution used in the data generating process can have different probability for different points that might appear in the sample. Secondly, and more importantly, ERM relies only on the sampled points meaning that the underlying data distribution of the data is assumed to be the same as that of the sample. This is often not the case due to the sample being not fully representative of the data. As a result, the function learned by the ERM method might perform very badly when tested on another sample from the same data distribution that the learning sample is drawn. 

When the (empirically) minimized risk is close to zero and the function space parametrized by $\alpha$ includes complex functions such as high degree polynomials, the chances are high that, ERM method \textit{tailored} a high-degree function for successfully matching all the data-points occur in the sample. However, taking into the account that the data can be noisy and the sample does not often demonstrate the same distribution as the one used during the data generation, the empirical low risk calculated from a high-degree polynomial is often deceiving and the actual risk with a random sample is much higher. This situation is known as \textit{Overfitting} and it is more likely to happen when the number of possible domain-range pairs for the learning problem at hand is big and the function space under consideration contains complex functions. 

When is the number of possible domain-range pairs big? Consider two regression problems. One problem has  one-dimensional input space and the other one has two-dimensional input space. Assume that the sample we have for each problem \textit{covers}\footnote{By \textit{covering} what is meant is that the sample capturing the representative subset of a contiguous region of the input space and their corresponding response variables} the 90\% of the range of the its input space. Now, the cartesian product of the domain and range sets of the first problem is $\mathbb{R} X \mathbb{R}$ and for the second problem, it is $\mathbb{R}^2 X \mathbb{R}$. One might expect the ERM principle to measure the risk with the same accuracy for both problems. But, this is not true. In the first case, the sample represents 90\% of the data well while in the second case, the sample represents $(90\%)^2 = 81\% $ well. If the problem had 15 input dimensions then the sample could only account for $(90\%)^{15}
 \approx 20\% $ of the data. Although having a good sample for a good range of possible inputs is practically not easy, with many dimensions, it does not guarantee that ERM method will not result in overfitting. This trouble with large number of dimensions is called \textit{Curse of Dimensionality}. 
 
In order to overcome overfitting problem of ERM method, a structural control mechanism through \textit{regularization} is introduced \cite{phillips_technique_1962}, \cite{tikhonov_regularization_1963}, \cite{ivanov_linear_1962}. With this extension over ERM, the risk minimization framework is named as \textit{Structural Risk Minimization} \cite{vapnik_nature_2000}. The structural control refers to limiting the complexity of the learned function. In the risk minimization framework, this idea could be realized by penalizing the candidate functions proportionally to their complexity so that a complex function with small loss entails a comparable risk with a simple function with high loss. This trade-off is known as the \textit{Bias-Variance Tradeoff}. Generally speaking, bias refers to the amount that predictions differ from the targets in the training sample in general and variance is the sensitivity to the small fluctuations of the data in the training set. A high-biased function with low variance is less \textit{funky} and do not account for the noise that causes the data to jump around and making the true data distribution look like more complex than it actually is. That is why, the learned functions with high-bias and low-variance have a better generalization ability. On the other hand, a low-bias high-variance function can successfully fit all the data points that are possibly contaminated by the noise present in the training sample by being a complex function although this could mean the fitted function is just one of the infinite number of functions that crosses the points in the sample dataset and very likely to generalize badly.

The aim of the structural control is not only to favour simpler functions that are known to have a better generalization ability to avoid overfitting but also to penalize extra model complexity fairly so that not the too simple functions with poor data-fit accuracy is chosen by SRM. Simply put, structural control prevents the learning method to favour complex models to just enough extend to avoid overfitting. This extend to which the complexity is penalized is controlled by the regularization constant. Since this constant is rather about the learning framework rather than the learning itself, it is considered as a \textit{hyperparameter} and it should be tuned to find the sweet point between underfitting and overfitting.

\section{Data Stream Learning}
\label{section:data_stream_learning}

Predictive models are algorithmically built upon various assumptions regarding the meta-qualities of the learning environment. Traditionally, the assumptions regarding the learning scenario is restrictive with respect to data availability. More specifically, before building any predictive models, data collection and data preparation should be done. Once the data is ready, one can start training and testing the learning algorithms. This learning scenario is referred to as {\it batch learning} and the assumptions it is based on are listed as follows:

\begin{itemize}
\item Having access to the all training data before the learning process.
\item Finite number of data points in the data set.
\item Data is generated by a static process which results in a fixed conditional distribution of outputs given inputs.
\item Training data sample is i.i.d.
\item Testing and training phases are totally separate.
\item During testing, the actual target values of the test inputs are not available.
\item No strict limits on the time allocated for individual predictions.
\item No strict limits on the space needed for storing the predictive models.
\end{itemize}

The way data is being generated is evolving, so is the way of accessing the data. As explained in \cite[p. 324]{gama_issues_2009}, nowadays, ever-increasing number of different kinds of devices such as sensors, hand-held devices, PCs, workstations, etc continuously generate, send and receive huge amounts of data. Most of the time the data being exchanged is not even persistent. It is consumed as it arrives. This gave rise to the popularity of data streams lately.

With the advent of data streams, the strict assumption about the data availability in traditional learning scenario is relaxed. Furthermore, the continuous data flow demonstrated by the stream data invalidates the other assumptions made in the offline learning setting. This imposes new requirements that the learning algorithm should fulfill in order to be employed in the streaming scenarios. This new learning paradigm is called {\it data stream learning} and it assumes the following.

\begin{itemize}
\item Data arrives one by one through a data stream.
\item Total number of data points is unbounded.
\item Distribution of the data is subject to changes over time.
\item Data does not have to be streamed from i.i.d sample.
\item Testing and Training are allowed to overlap. The learning machine can learn from the previous test points.
\item After a prediction, the target value supposed to be predicted is available (in some online learning scenarios)
\item Data processing rate should be higher than data arrival rate in general so that {\it in-situ analysis} is possible.
\item Space requirements of the learning algorithm used should be bounded by a constant.
\end{itemize}

These essential differences between two learning paradigms are highlighted by a load of previous research in machine learning community \cite{li_towards_2014}, \cite{gama_evaluating_2013}, \cite{domingos_catching_2001}, \cite{nguyen-tuong_incremental_2008}, \cite{vovk_algorithmic_2005} 

While the fundamental conceptual considerations regarding model learning in statistical learning theory such as the tradeoff between bias and variance and the curse of dimensionality are still relevant in data stream learning, some new aspects of the learning needs to be taken into account with the changes in the basic assumptions of the learning scenario. Next, the most important stream learning-relevant consideration, \textit{stationarity}, is discussed.

\subsection{Non-Stationarity of Data Distribution in Streams}
\label{subsection:2.4.1}

As stated in the list of assumptions regarding online learning scenario, the underlying data distribution is subject to changes. This happens when the process that generates the data, for any reason which is not the point of interest for the learning, changes and starts producing data with different characteristics. This phenomenon is called \textit{concept drift}.

\begin{ex}
Imagine a streaming scenario where the data items in the stream consists of two numbers namely the average number of transistor count in the microchips and the year of build of the microchips\begin{align*}
& (x,y)_n = {(\text{avg \# of transistors, year of built})_n}
\end{align*}
This hypothetical data stream started streaming in 1960 and it streams a new data point every year on the first day of January. The underlying distribution of the stream data for the first 30 years did not change (\cite{moore_cramming_1965}, \cite{schaller_moores_1997}). However, for the last two decades, the correlation between year of built and the average number of transistors in microchips seems to have changed significantly (\cite{tuomi_lives_2002}). This \textit{non-stationarity} in the streamed data is a good example of concept drift.
\end{ex}

Conceptual drifts are common in learning scenarios which consumes a live data stream. This is why techniques to deal with them are proposed in the literature mostly under the name of \textit{Online Learning}. However some learning problems that do not feature any concept drifts but have a stream data source are often mistakenly assumed to have concept drifts and the techniques to deal with non-stationarity in data are falsely being applied to these. Therefore, before discussing the ways to handle concept-drift, the distinction between two types of stream learning problems need to be made clear. Next, in order to highlight the difference between the two, learning with a stationary stream scenario will be contrasted with the non-stationary one.

\subsection{Stationary Stream Learning}

This kind of stream learning is structurally same as its non-stationary variant. However, in stationary scenarios, the very crucial difference is that one can assume the underlying data distribution of the stream is fixed. This has deep implications on the way learning should be done. Most obviously, if the data distribution is static, then once the learning algorithm has built a good predictive model, the future data points in the stream is guaranteed to be predicted accurately. This implies, learning does not have to be continuous and one can adapt batch learning algorithms to stream learning scenario. However, the catch is that in some scenarios data distribution can be a function of time that demonstrates repeating patterns. An example of this is weather. Weather data is in fact is stationary (or drifting in a negligible amount \footnote{http://climate.nasa.gov/} due to global warming) although one might think the data distribution changes from one season to another. This is partly true. However, if we look at the big picture, what we see is that as the seasons repeat, so-called changing data distributions also repeat. Therefore, it is more accurate to say data distribution has different local trends and the history contains all the patterns, hence new patterns are not expected to emerge (at least for a long time). Learning problems with this kind of time-dependent local repeating patterns are categorized under the name of \textit{time-series prediction}. The online learning algorithms that incorporates the new data from the stream to capture emerging trends are not well-suited for the time-series prediction problems as when dealing with time-series prediction problems, once the relation between the time and the local data patterns are resolved, application-wise it is no different than batch learning with the exception of predictions still has to be made one the one-by-one basis which is a constraint imposed on by the data stream environment.

\subsection{Online Prediction Protocol}
In order to provide a common way to specify and formulate online learning problems with the emphasis put on the sequentially arriving data and incremental training, \textit{Online Prediction Protocol} is proposed \cite[p. 5]{vovk_-line_2009}. The protocol introduced in the original paper is only for the regression problems and it involves an extra line which is rather about the prediction interval estimation strategy which is irrelevant to our general purpose of defining a protocol for online algorithms. Therefore, a minimally modified version of the online prediction protocol is as follows.

Let $Domain$ be the set of all the possible input values. For the regression with $n$ input variables, $Domain = \mathbb{R}^n$. Let $Range$ be the set of all the possible response (target) values. For the regression problems, $Range = \mathbb{R}$. Data points in the stream are represented as $(\pmb{x}, y)_n=\{\pmb{x}_n,y_n\}$ where $n$ is the position of the tuple in the data stream. Data points with smaller $n$ value arrives earlier. $\hat{y}_n$ and $y$ are respectively the predicted target and target for the $n_{th}$ data point. $UpdateModel$ is a procedure that \textit{incrementally} incorporates new data into the internal learning model which is abstracted away in the online prediction protocol.

As for $err_n$, it is the array of errors computed by the loss function $L$ on given $\hat{y}_n$ and $y$. Capturing the errors this way is not strictly related with the online learning itself. It is included in the loop just to capture the real-time accuracy statistics of the online learning process that is needed for further analysis of the performance of the learner. The loss function $L^*$ should not be confused with the loss function used internally by the learning algorithm to build its internal predictive model to come up with the predictions (e.g least-squares, etc.) as they can be different from each other. In order to avoid confusion, the loss function that is used in the wrapper over the learning algorithm is denoted as $L^*$ while the internal loss function is denoted as $L$. Since for the regression problems most interpretable accuracy metric is the absolute deviation of the prediction from the target value, mostly absolute loss function is used for collecting accuracy statistics. However for a classification problem this is usually 0-1 loss function that returns zero provided that the prediction is correct and returns zero otherwise. This way of error calculation is named as predictive sequential approach and it is discussed thoroughly in Chapter \ref{Chapter6}. 

The pseudocode for online prediction protocol is as follows:

\begin{algorithm}
  \caption{Online Prediction Protocol}\label{alg:opp}
  \begin{algorithmic}[1]
    \Procedure{Predict}{}
    	\While{true}\Comment{Infinite Loop}
    		\State \texttt{Observe the data point $\pmb{x}_n \in Domain$}
        	\State \texttt{Output the prediction $\hat{y_n} \in Range$}
        	\State \texttt{Observe the response $y_n \in Range$}
        	\State \texttt{UpdateModel($\pmb{x}_n$, $\hat{y_n}$, $y_n$)}
        	\State $err_n \gets L^*(y, \hat{y_n})$
    	\EndWhile
		\State \textbf{return}
    \EndProcedure
  \end{algorithmic}
\end{algorithm}

\subsection{Data Horizon and Data Obsolescence}

The online protocol can also be described briefly as \textit{interleaving learning and testing}. The model update call in the loop of online prediction protocol after observing the response value is important when dealing with non-stationary data. New data always need to be utilized. In other words, for online learning algorithms there is no ending to learning, they learn as long as the data stream flows. Often it is pretty challenging task to design learning algorithms that are able update their internal predictive model with the new data without having to build the model from scratch. The term in online learning used to indicate how immediately new data points should be incorporated into the model is \textit{data horizon}. If the predictive model is strictly required to be updated with the observation of the response for the each data point arriving, then the data horizion is very close. On the other hand, if, after predicting a newly arriving data point, there is some time needed to obtain the response or incorporate the new data point with its observed response into the predictive model via $UpdateModel$ call and meanwhile the existing model does not quickly become obsolete, then data horizon is relatively far. 

Another consideration with online learning is \textit{data obsolescence}. When the old stream items which are once used for updating the predictive model is not of any value to the prediction then they should be omitted from the predictive model. The time that takes for a data to become obsolete and its effect on the prediction mechanism should be removed is called data obsolescence time. Removing the effects of the obsolete data points is not easy from an implementation point of view. Especially for the learning algorithms that \textit{absorbs} the data, this become harder as the prediction is not computed by some aggregation of the contributions of separate data points. However, some linear algebraic and computational tricks to this are available and these are discussed in Chapter \ref{Chapter5}. 

\chapter{Problem} 

\label{Chapter3} 

\lhead{Chapter 3. \emph{Problem}} 

\section{Ocelot Overview}

\subsection{Main Design Principles}

Ocelot is a hardware-oblivious parallel database engine. The motivation for opting for \textit{hardware-oblivious} design which is the opposite of the \textit{hardware-aware} design stems from the ongoing shift to heterogeneous architectures that feature various computing devices of possibly different architectures on a single platform. Ocelot fulfills its hardware-oblivious design goal by abstracting away the details of the hardware components at the design time. As discussed in \citep[p. 710]{heimel_hardware-oblivious_2013}, hardware abstraction is needed to deal with following problems of hardware-aware design.

\begin{itemize}
\item{Database vendors focus on only few architectures for hardware specific fine-tuning of the database engine in order to limit the development and maintenance costs.}
\item{It is not possible to use existing operator implementations for a new architecture aimed to be supported. This simply requires reimplementation of the whole set of operators.}
\item{With each additional architecture to be supported, database vendors have to extend their expertise area on a particular hardware beyond their core competences.
}
\end{itemize}

With the hardware details are abstracted away, database-engine developers can use a high-level language to program database operators in a main codebase repository. Then, the compilers that are provided by the hardware vendors compile the operator algorithms available in the main codebase into the binary code. In Ocelot, the high-level programming of the operators is done using the \textit{Kernel Programming Model}. According to \citep[pp. 710-711]{heimel_hardware-oblivious_2013}, in this model, programs consist of a number of kernels each of which define an operation on single element of the input data. These operations then operate on the input in a \textit{lock-free} fashion. Although, by its design, this kernelized programming approach seems to be specifically tailored for GPUs that feature a massive parallelism, they can run on single-core architecture by simply running the kernels sequentially on the input thanks to their highly abstract definition. More examples of the how the kernelized programs are executed in different architectures are presented in \citep[p. 711]{heimel_hardware-oblivious_2013}.

\subsection{Self-Adaptivity}

Kernel Programming model makes it possible to port the programs to different architectures. This is why, it is the main essence of hardware obliviousness. However, the portability is not the only concern. The portable operator programs implemented should also run efficiently on the ported architecture. In order to achieve this efficiency, hardware-dependent optimizations should be done. However, It is not possible to write the operator code tailored for an architecture as it would then break the hardware-oblivious design principle. In order to carry out hardware-dependent optimizations without being hardware-aware, Ocelot features self-adaptivity. Self-adaptivity is achieved in two steps. First, a large number of variants of the operator algorithms either at the design time or dynamically at the run time are prepared. Then, Ocelot attempts to pick the most performant one on the fly. The optimal operator algorithm variant given the features such as input size, etc can be different for different computing devices as the capabilities of different kind of devices are naturally different. Furthermore, the optimal operator variant for a given operator task scenario can also differ within the same kinds of device. Even for the models from the same vendor sharing the same architecture, the optimal variant could be potentially different. Considering the exploding number of combinations of different models of the different hardware can be present together on a platform, picking the most optimal operator variant given the input characteristics (features) for each device at the Ocelot's disposal is a challenging task. Following, the details of this decision-making process is discussed.

Ocelot separates the device selection routine and the algorithm selection routine. Device selection logic considers the performance statistics of the devices for the current operator with given features as well as the other factors such as the current load of the available devices and the transfer times between devices. Once a device for the current operator to be offloaded to is selected, algorithm selection logic decides on the algorithm to be used to obtain the output for the operator. 

The algorithm selection logic is complex. It is based on the runtime predictions for all the choices of algorithms available for the current operator. However, each algorithm is implemented by a big number of kernelized variants. Moreover, for each algorithm, an \textit{active set} of variants that implements the algorithm are maintained. A cost model associated with each variant in the active set is also maintained. When the runtime prediction for an algorithm is requested in order to choose the most performant algorithm for the given input, the lowest of the predictions of these cost models are returned. After an algorithm is chosen, all the variants in the active set is executed and runtime measurements of each of them are recorded. These measurements are used to update the cost models. Initially when the database in its \textit{cold start} phase and not have adapted itself yet to the underlying hardware, the cost models for the active set are empty predictive models hence they do not make good runtime predictions. As more queries are passed to the database engine and more operators are executed, the cost models for the active set variants of the chosen algorithms become mature predictive models and starts delivering good runtime estimations when the algorithm selection logic requests the runtime predictions before choosing an algorithm. This means runtime predictions of a certain algorithm for an operator is reliable only after the algorithm is selected a number of times. This represents a \textit{multi-armed bandit} problem in which there is a trade-off between \textit{exploitation} and \textit{exploration} \citep[p. 618]{heimel_demonstrating_2014}. In Ocelot, decaying $\epsilon$-greedy approach is employed to balance exploitation of the cost models learned for the variants of the algorithm options for an operator to pick the most performant algorithm and the exploration of the performance characteristics of the available algorithms to improve cost models of the underlying active set of variants of them. With $\epsilon$-greedy approach employed, the algorithm selection routine, with probability of $1-\epsilon$, picks the \textit{supposedly} most performant algorithm indicated by the minimum of the estimated runtimes of the active set variants of each algorithm. With probability of $\epsilon$, algorithm selection routine picks a random algorithm other than the supposed most performant one.

The active set of the variants of an operator algorithm is a dynamic set and it is dynamically updated on the fly. When there is a variant in the active set which performs badly in comparison to the other ones, it is replaced by another variant from the library of variants each algorithm has. Alternatively, a variant could be just dropped without an alternative variant taking its place. This dynamically evolving active set is expected to converge to a state with a single-element active set with the only variant that is the optimal one among the variants defined in the variant library. This variant is expected to as close as possible to the \textit{hand-tuned} implementation of the corresponding algorithm of the variant in a hardware-aware database.

How the device selection and algorithm selection work together to realize self-adaptivity goal is sketched in Figure \ref{fig:Ocelot}.

\begin{figure}[htbp]
  \centering
    \includegraphics[width=\linewidth]{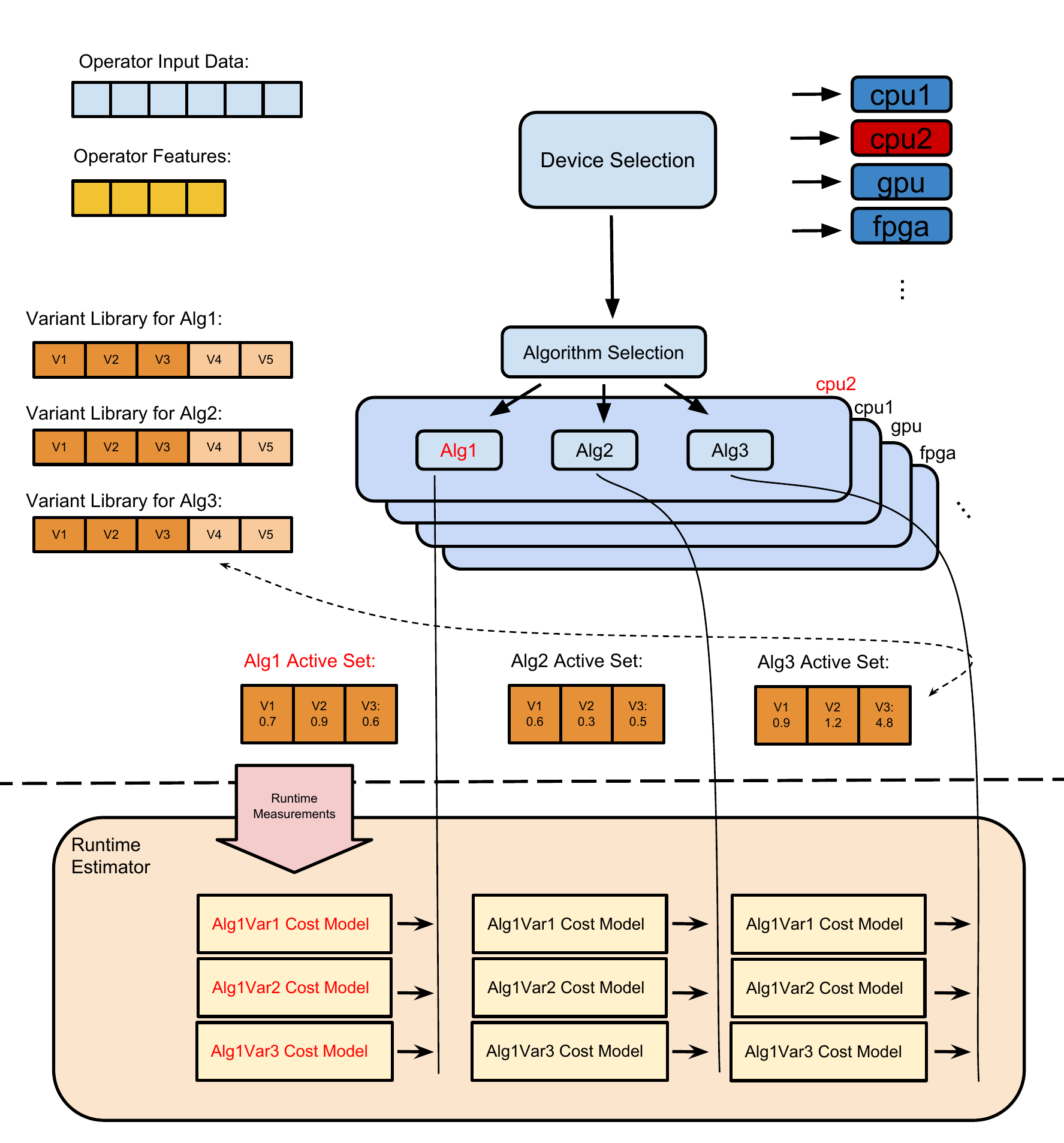}
  \caption{Overview of the device and algorithm selection in Ocelot. Note that the numbers written in the boxes that represent the active algorithm variant set denote the average runtime of the corresponding variant.}
  \label{fig:Ocelot}
\end{figure}

\section{Learning Problem}

\subsection{Correspondence to Data Stream Learning}

In Ocelot, the task of building and updating cost models is abstracted away from the algorithm selection routine. The space above the dotted line in \ref{fig:Ocelot} has an abstract view of the bottom part which depicts the \textit{Runtime Predictor}. Algorithm selection routine is provided with an interface of two functionalities namely update and predict. Whenever an algorithm selection is needed, the algorithm selection routine requests runtime predictions for the algorithms and after the execution of the selected algorithm on the selected device, it updates the runtime estimator with the measured runtime data. 

Runtime estimator module is totally blind to decision-making layers such as algorithm selection and device selection. It can be thought of a utility component with a global state. Within the Runtime Estimator, for each  existing algorithm-variant-hardware combination, there is a corresponding cost-model. This cost model is required to be refined with new measurements and predict the actual cost of its variant accurately as explained in the previous section. Obviously, this is a machine learning problem where an updatable predictive model is needed for fulfilling the requirements of the runtime estimator.

The learning scenario with the runtime estimator module represents the characteristics of stream learning described in \ref{section:data_stream_learning}. Initially empty predictive models (cost models) are first used to make a prediction (runtime estimate) for a feature set of a predefined size (characteristics of the current input) then, they are provided with a real number (the noisy measurement of the runtime) to refine the predictive model. These two operations of predict and update occur repeatedly one after another potentially infinite number of times. \ref{fig:Ocelot2} pictures this particular learning scenario.

\begin{figure}[htbp]
  \centering
    \includegraphics[width=\linewidth]{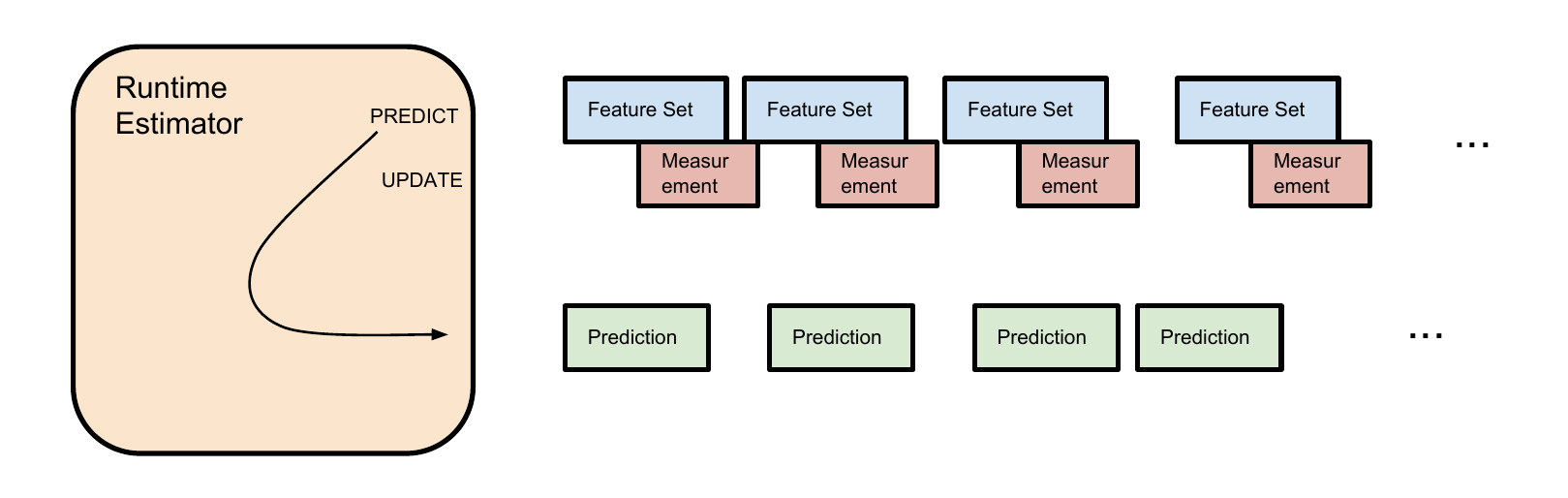}
  \caption{Runtime Predictor is visualized as a stream learner. Note that the irregular spacing between the arriving items as well as between the predictions illustrates that the frequency of runtime predictions are requested by the algorithm selection logic of the Ocelot is variable}
  \label{fig:Ocelot2}
\end{figure}

Data stream in the runtime estimation scenario can be seen as an array of feature sets that describe the characteristics of the inputs to the operators and the measurements. For the sake of adapting the machine learning jargon, the feature sets and the measurements are referred to as \textit{data points} and the \textit{targets} respectively in the rest of the thesis. 

The process that generates the data points and their targets in this scenario is very complex and it depends myriad of things. Among these dependencies of the data generation process, the Ocelot-relevant ones are the performance of the active set variants for an algorithm and the behavior of the algorithm variant replacement logic. Primarily, the former affect the runtimes observed. However, the latter dramatically and abruptly change the measured runtimes. Although, it is stated above that there is a cost model for each possible variant-algorithm-device combination, this is an oversimplification. Since the Runtime Estimator is not aware of the algorithm variant substitutions, it keeps a cost-model for the variant that sits in a certain place in the active set. The implication of this is important in terms of the produced runtime measurement data which is used to update the cost models. When an algorithm variant is replaced, the cost model maintained for the replaced variant \textbf{immediately} becomes obsolete. Thus, through the update calls from the algorithm selection logic, it is expected to become an accurate performance estimator of the new variant substituted in. This corresponds to \textit{learning from non-stationary streams} scenario as discussed in \ref{subsection:2.4.1}. 

Data horizon for this online learning scenario is very close as new measurement after concept drifts are needed to be incorporated into the cost models immediately whereas the data horizon once the cost models become good predictors of the runtimes is infinitely far meaning the new measurements cannot be used to improve the accuracy of the cost models more. Similarly, the data obsolescence time is zero after concept drifts as the cost model that becomes obsolete is built with old data points. However, once the learning algorithm stabilizes and the cost model again become a good predictor of the runtime measurements, the effect of old data points on the predictive models do not need to be removed meaning that data obsolescence time is infinite.

Data generation process also depends on more general things other than the internals of Ocelot. Among these are the potential measurement error, potentially varying performance of the hardware components, number of context switches scheduled by the scheduler of the Operating System running on the platform that Ocelot is deployed. As a result of these, it is not possible to get noise-free runtime measurements for operator algorithm variant runtimes. The noise is assumed to be feature and operator independent (homoscedastic noise). 

\subsection{Required Properties of the Learning Model}

Having described the learning problem posed by the Ocelot's runtime estimation needs in the previous section, in this section, the requirements that the learning algorithm to be employed should meet are discussed.

In the following, these requirements are listed and described.

\begin{itemize}
\label{list:restimator_requirements}

\item \textbf{High accuracy:} The predictive models used as cost models to predict runtimes should make estimations with high accuracy so that the algorithm selection and the device selection routines can make good decisions.
\item \textbf{Robust to Abrupt Concept Drifts:} Whenever the algorithm variant substitution mechanism fires and replaces a variant with a new one from the variant library, the runtime measurements given the features potentially change as explained in the previous chapter. This change can be seen as a concept drift. This is because, the most important factor that affects the distribution of the target values given the features is the performance of the algorithm variant and once the variant itself is different, the underlying data distribution of the features and the targets unavoidably changes. However, when some number of the feature-target pairs are sampled from the old and new underlying distributions, the functions that can fit these samples are expected to have the same growth rate. This is because, although the variants for the same algorithm have different implementations of the same algorithm, since they implement the same logic described in the pseudocode of the algorithm, the performance characteristics of the variants are expected to be similar. 
The learning algorithm used for building predictive models which are used as cost models in the Ocelot context should be robust to abovementioned concept changes. However, this does not impose any kind of restrictions on the update frequency of the predictive model on the contrary to how it is illustrated by the red boxes in \ref{fig:Ocelot2} where the cost model is pictured to be updated after every single runtime estimation made and also how it is implies by the online prediction protocol presented in the pseudocode \ref{alg:opp}. Online learning algorithm has two different mechanism in order to deal with abrupt concept drifts. First, it can either update the predictive model with every single new runtime measurement and never worry about the detection of concept drifts. Second, it can implement a drift detection method and update the model incrementally with the new runtime measurements until an accurate predictive model is obtained only after a concept drift is detected or during the \textit{cold start} phase when the predictive models are empty.
\item \textbf{Robust to Measurement Noise:} As mentioned in the previous section, the measured runtimes that is used to update the predictive models are not noise-free. This requires the learning algorithm to be able to learn from the measurement data contaminated by homoscedastic noise. 
\item \textbf{Provide bounds for estimates:} The sophisticated algorithm selection logic of Ocelot requires more than just a point prediction of the runtime. Under some conditions, which are not covered and considered to be in the scope of this thesis work, it bases its algorithm-selection decision on the potentially minimum or potentially maximum runtime of the algorithm variants estimated for given features. This requirement makes sense when the difficulty of making point predictions with the noisy training data is considered. In short, the learning algorithm should produce estimations given the features that consist of three elements namely lower prediction bound, point prediction and upper prediction bound. The interval marked by the upper and lower prediction bounds should cover the observed runtimes with a high confidence level.
\item \textbf{Efficiency:} Although the amount of time between individual predictions requested by the algorithm selection logic for an operator algorithm is not fixed and it depends on many factors such as the distribution of the different operators that make up the queries that are passed to Ocelot and the device selected by the device selection logic, it is hard to set a minimum required datarate that runtime estimator module can handle. Generally speaking, runtime estimator module should ideally run fast enough so that algorithm selection logic never has to delay its decision. It is hard to foresee the time-efficiency requirements without knowing the maximum call frequency the algorithm selection routine\footnote{During the period this thesis work was being carried out, no such information was available}. Nevertheless, as an initial time-efficiency target to be achieved, the delay that runtime estimator module causes per runtime estimation together with the update operation with the observed runtime should not exceed $1$ ms.

\end{itemize}



\chapter{Approach} 

\label{Chapter4} 

\lhead{Chapter 4. \emph{Approach}} 

The kind of learning problem to be tackled in order to make the desired inferences about operator runtime behavior as discussed in the previous chapter is \textit{regression estimation} among other kinds recognized in Statistical Learning Theory such as density estimation and pattern recognition problems. In this chapter, first, the particular regression estimation problem to be tackled is mathematically modeled. Then, given the formulation of the problem, the regression algorithms with different learning objectives (e.g minimizing the empirical risk, minimizing the expected risk, calculating the most \textit{probable} prediction given the features, etc) or with different techniques to achieve the same objective (e.g finding the optimal parameters that makes the most probable prediction, finding the optimally weighted average of the observed responses that is supposed to be the most probable prediction) are explored. Furthermore, the derivations of the formulas that these algorithms are based on for making predictions and estimating prediction bounds are presented.

\section{Modeling the problem}

In Regression Estimation, the main goal is to approximate the data generation function that is used to generate the data from a sample of data provided. It is also called as \textit{Function Approximation Problem}. This problem is already formulated under the empirical minimization framework in \ref{ex:regression_formulated}. A simpler version of it is given in the following to concentrate solely on the definition of the problem rather than the approach taken to solve it. Later in this chapter, more approach-oriented view of the problem is presented when the algorithms proposed for solving it is discussed.
\begin{flalign}
& \text{Data : } (X,\pmb{y}) = \{(\pmb{x}_i,y_i),\ \forall i < n,\ i \in \mathbb{N}_{>0}, \ \pmb{x}_i \in \mathbb{R}^d,\ y_i \in \mathbb{R}\}, \label{def_4.1} \\
& \quad y_{n} = f(\pmb{x}_n) + \epsilon _n \label{def_4.2} \\
& \text{Find } y_{n+1} | \text{(X,\pmb{y})} \label{def_4.3}
\end{flalign}
Above $n$ denotes the number of training examples (case base) and $d$ denotes the dimensionality of the input space. Moreover, $X$ is a $d$-by-$n$ matrix that is formulated as $X=[\pmb{x}_1,\pmb{x}_2,...\pmb{x}_n]$ and referred to as \textit{design matrix}. \ref{def_4.3} defines the regression estimation problem in a general way and it does not preclude the use of any regression algorithm. It is also worth noting that the learning problem defined does not impose any kind of restrictions on the \textit{logistics} of the learning scenario (e.g batch learning, online learning). In \ref{def_4.1}, $\epsilon _n$ denotes the \textit{additive} noise associated with the $n_{th}$ target observation. In order to properly define the learning problem, this additive noise should be modeled. As mentioned in \ref{list:restimator_requirements}, the measurement noise that corresponds to the additive noise in the above formulation is supposed to be homoscedastic. In other words, the additive noise is not a function of input variable $\pmb{x}$ and it is independently identically distributed. In this case, using the most common noise modeling option, the vector of additive noise quantities, $\pmb{\epsilon}$ is assumed to have a multivariate Gaussian distribution with $0$ mean and the covariance function $\sigma_y^2 I$. 
\begin{flalign}
\epsilon = N(0,\sigma_y^2 I) \label{def_4.4}
\end{flalign}
Having formulated the learning problem, in the following sections, existing methods to solve it are discussed.

\section{Parametric Models}

A common approach to solving regression estimation problem is to employ parametric predictive models. Most commonly, \textit{linear regression} is used as a parametric regression method. Given a loss function $L$, linear regression finds the coefficients, $\pmb{w} \in \mathbb{R}^d$ and assumes the following for the regression function.
\begin{flalign}
& f(\pmb{x})=\pmb{w}^{\top}\pmb{x} \label{def_4.5} \\
& Y_i = N(\pmb{w}^{\top}\pmb{x_i},\sigma_y^2) \ \text{by \ref{def_4.2} and \ref{def_4.4} \label{def_4.6} }
\end{flalign}
Above, $Y_i$'s are independent random variables inheriting the i.i.d noise assumption. The normality assumption on the noise (also called Gaussian distributed noise) allows to model independent target values ($Y_i$) as a Gaussian distribution as well. This is why, linear regression with Gaussian noise assumptions is also called \textit{Gaussian Linear Regression}

\subsubsection{Non-Linear feature space mapping}
\label{subsubsection:nlmap2fs}

Note that, linear regression is linear in terms of $\pmb{w}$. One can still assume a non-linear regression function by mapping the input space to a higher dimensional feature space. In this case, instead of $\pmb{w}^{\top}\pmb{x}$, in \ref{def_4.5} and \ref{def_4.6}, $\pmb{w}^{\top}\phi(\pmb{x})$ where $\pmb{w} \in \mathbb{R}^m$ and $m$ is the number of basis functions defined by $\Phi(\pmb{x})=[\phi_1(\pmb{x}), \phi_2(\pmb{x}), ..., \phi_m(\pmb{x})]^{\top}$ would appear.

How the parametric models finds $\pmb{w}$ is yet to be discussed. As the random variable targets are modeled to be Gaussian distributed, probability density function of normal distribution can be used to find parameters $\pmb{w}$ given the data and the hyperparameters. The hyperparameters are parameter estimation method-dependent. In the following two subsections, the most common two parameter estimation method namely Maximum Likelihood Estimator (\textbf{MLE}) and Maximum A Posteriori Estimator (\textbf{MAP}) are discussed.

\subsection{MLE method}

\subsubsection{MLE-based parameter estimation}

MLE-based parameter estimation is very simple. The only hyperparameter to be set is $\sigma_y$ through which the standard deviation of the target noise around zero is expressed. Thus, we write $\theta = \{\sigma_y\}$. MLE-based parameter estimation is based on finding the parameter, \pmb{w}, configuration that maximizes the \textit{likelihood} of the responses given the design matrix and parameters. In mathematical terms, the expression to be maximized is $p(\pmb{y}|X,\pmb{w})$.

Manipulating the joint probability formula and exploiting pdf of Gaussian distribution, likelihood of the data given parameters is calculated as follows.
\begin{align*}
p(\pmb{y}|X,\pmb{w}) & = p(y_1, y_2, ...y_n|\pmb{x}_1,\pmb{x}_2,...\pmb{x}_n,\pmb{w}) \\ &
=\prod_{i=1}^{n} p(y_i|\pmb{x}_i,\pmb{w}) \\ &
=\prod_{i=1}^{n}\frac{1}{\sqrt{2\pi \sigma_y^2}} exp(-\frac{1}{2 \sigma_y^2} (y_i - \pmb{w}^{\top}x_i)^2) \\ &
=(\frac{1}{\sqrt{2\pi \sigma_y^2}})^n exp(-\frac{1}{2 \sigma_y^2} (\pmb{y} - \pmb{w}^{\top}X)^{\top}(\pmb{y} - \pmb{w}^{\top}X)) \numberthis \label{def_4.7} \\ & \propto -\frac{1}{2 \sigma_y^2} (\pmb{y} - \pmb{w}^{\top}X)^{\top}(\pmb{y} - \pmb{w}^{\top}X)
\end{align*}
Maximizing $(\frac{1}{\sqrt{2\pi \sigma_y^2}})^n exp(-\frac{1}{2 \sigma_y^2} (\pmb{y} - \pmb{w}^{\top}X)^{\top}(\pmb{y} - \pmb{w}^{\top}X))$ can be seen as the equivalent of minimizing\footnote{due to the minus sign in the exponent} $(\pmb{y} - \pmb{w}^{\top}X)^{\top}(\pmb{y} - \pmb{w}^{\top}X)$ as the constants can be ignored in the optimization and. In order to find the $\pmb{w}$ that minimizes the $(\pmb{y} - \pmb{w}^{\top}X)^{\top}(\pmb{y} - \pmb{w}^{\top}X)$ expression, first the critical points of it are found:
\begin{align*}
\mathcal{L} & = (\pmb{y} - \pmb{w}^{\top}X)^{\top}(\pmb{y} - \pmb{w}^{\top}X) \numberthis \label{def_4.8} \\
& = \pmb{y}^{\top}\pmb{y} + 2\pmb{y}^{\top}X^{\top}\pmb{w} + \pmb{w}^{\top}XX^{\top}\pmb{w} \\
\nabla_w \mathcal{L} & = -2X\pmb{y} + 2XX^{\top}\pmb{w}
\end{align*}
Above, setting $\nabla_w \mathcal{L}$ to 0 gives: 
\begin{align*}
-2X\pmb{y} + 2XX^{\top}\pmb{w} & = 0 \\
(XX^{\top})^{-1}X\pmb{y}^{\top} & = \pmb{w} \numberthis \label{def_4.9}
\end{align*}
After finding the critical point, whether it is a maxima or minima is checked:
\begin{align*}
\nabla_w^2 \mathcal{L} & = \nabla_w (\nabla_w \mathcal{L}) \\
& = \nabla_w (-2X\pmb{y} + 2XX^{\top}\pmb{w}) \\
& = 2XX^{\top}
\end{align*}
$2XX^{\top}$, being the Hessian matrix, is a quadratic hence always greater than zero assuming $XX^{\top}$ is invertible. Thus, it is a positive definite matrix making the optimization problem strictly convex. Therefore, one can conclude that the critical point with a strictly positive Hessian is the global minimum of $\mathcal{L}$. This implies $\pmb{w}$ found in \ref{def_4.9} is the MLE estimate of parameters that maximizes the likelihood of the data. 
\begin{flalign}
\pmb{w}_{MLE} = (XX^{\top})^{-1}X\pmb{y} \label{def_4.10}
\end{flalign}
It is worth noting that \ref{def_4.8} can be written as $|\pmb{y} - \pmb{w}^{\top}X|^2$ which is the euclidean distance between the responses vector, $\pmb{y}$, and the predictions vector, $\pmb{w}^{\top}X$. Minimizing this distance corresponds to minimizing the empirical risk with $L_2$ chosen as the loss function as described in \ref{subsection:ERM} which is the idea behind the well-known Ordinary Least Squares (OLS) method.

As mentioned in \ref{subsubsection:nlmap2fs}, it is easy to define a set of basis functions that maps inputs to a higher dimensional space when the unknown regression function $f$ in \ref{def_4.2} is assumed to be non-linear. When this idea is employed for MLE method, the parameter vector, \pmb{w}, that maximizes the likelihood of the data is the following.
\begin{flalign}
\pmb{w}_{MLE} = (\Phi(X)\Phi(X)^{\top})^{-1}\Phi(X)\pmb{y} \label{def_4.11}
\end{flalign}

\subsubsection{Measure of prediction uncertainty in MLE-method}
\label{subsubsection:4.2.1.2}

Finding the parameters, $\pmb{w}$ of the regression function makes it possible to make predictions for a given data point, $\pmb{x}$. Simply calculating $\pmb{w}^{\top}\pmb{x}$ gives the MLE-estimate. However, if an interval with a certain probability containing the unobserved target value is needed instead of a point estimation of the target, finding the regression function parameters alone is not enough. In this case, the asymptotic properties of the OLS method that is essentially the same as MLE method come in handy.

In order to find intervals \textit{probably} covering the target variable for predictions, first the intervals for the parameter values of the regression function should be established. MLE method already computes the parameter estimations but we do not know how close they are to the actual parameters of the data generation function. However, as shown in \ref{def_2.3}, by the \textit{consistency} of the ERM of which OLS is an example, as the sample size goes to infinity, the approximated regression function converges to the regression function implying the parameters estimated for the regression function converges to the true parameters of the regression function. This, along with derivation of the asymptotic normality of the random variable $\pmb{\beta} - \pmb{w}$, is shown below:
\begin{align*}
\text{letting $\pmb{\beta}$ denote the true} & \text{ parameters of the regression function} \\
\text{and } s^{2} \text{ be } = \frac{(\pmb{y} - \pmb{w}^{\top}X)^{\top}(\pmb{y} - \pmb{w}^{\top}X)}{n-d} & \text{ where d is the number of predictors}  \\
\pmb{\beta} - \pmb{w} & = (XX^{\top})^{-1}X(\pmb{y}+\pmb{\epsilon}) - (XX^{\top})^{-1}X\pmb{y} \\
(XX^{\top})^{-1}X\pmb{\epsilon} & = (\frac{1}{n}(XX^{\top})^{-1}(\frac{1}{n}X\pmb{\epsilon}) \\
\text{knowing that;} \ \ \frac{1}{n}(XX^{\top}) = \frac{1}{n}\sum_1^n\pmb{x_i}\pmb{x_i}^{\top} & \rightarrow E[\pmb{x_i}\pmb{x_i}^{\top}] = \frac{XX^{\top}}{n} \numberthis \label{def_4.12} \\
\text{and} \ \ \frac{1}{n}(X\pmb{\epsilon}) = \frac{1}{n}\sum_1^n\pmb{x_i}\pmb{\epsilon_i} & \rightarrow E[\pmb{x_i}\pmb{\epsilon_i}] = 0 \numberthis \label{def_4.13} \\
\text{and} \ \ \frac{1}{\sqrt{n}}\sum_1^n\pmb{x_i}\pmb{\epsilon_i} & \rightarrow N(0,s^{2} \frac{XX^{\top}}{n}) \text{ by C.L.T} \numberthis \label{def_4.14} \\
\text{Therefore; } \\
(\pmb{\beta} - \pmb{w}) = \frac{1}{\sqrt{n}}(\frac{1}{n}(XX^{\top})^{-1}(\frac{1}{\sqrt{n}}X\pmb{\epsilon}) & \rightarrow (XX^{\top})^{-1}\sqrt{n} N(0,s^{2} \frac{XX^{\top}}{n}) \text{ by \ref{def_4.12} and \ref{def_4.14}} \\
& = N(0,s^{2} (XX^{\top})^{-1}) \numberthis \label{def_4.15}
\end{align*}
For a given data point, $\pmb{x}_i^{\top}$, It is easy to manipulate above result to reach the asymptotic normality of the difference of the mean response, $\pmb{x}_i^{\top}\pmb{\beta}$, and the predicted response, $\pmb{x}_i^{\top}\pmb{w}$, as follows:
\begin{align*}
\pmb{x}_i^{\top}\pmb{\beta} - \pmb{x}_i^{\top}\pmb{w} & = \pmb{x}_i^{\top}(\beta-\pmb{w}) \\
\pmb{x}_i^{\top}N(0,s^{2} (XX^{\top})^{-1}) & = N(0,s^{2}\pmb{x}_i^{\top}(XX^{\top})^{-1}\pmb{x}_i) \numberthis \label{def_4.16}
\end{align*}
What \ref{def_4.16} tells is that the variance of the prediction for the data point $\pmb{x}_i$ is $s^{2}\pmb{x}_i^{\top}(XX^{\top})^{-1}\pmb{x}_i$. However, the interval that can be constructed from this will not account for the measurement noise. As soon as the enough number of data points are absorbed into the predictive model that allows parameters to be inferred accurately, the interval will be of zero-length. In order address this problem, the estimation of the variance of the additive noise should be taken in to account. 
\begin{align*}
Var[y_i - \pmb{x}_i^{\top}\pmb{w}] & = Var[\pmb{x}_i^{\top}\pmb{\beta} - \pmb{x}_i^{\top}\pmb{w}] + Var[\epsilon_i] \\
Var[y_i - \pmb{x}_i^{\top}\pmb{w}] & = s^{2}\pmb{x}_i^{\top}(XX^{\top})^{-1}\pmb{x}_i + s^{2} \\
\text{Therefore;} & \\
y_i \in [\pmb{x}_i^{\top}\pmb{w} & \pm z_{1-\frac{\alpha}{2}}\sqrt{s^{2}\pmb{x}_i^{\top}(XX^{\top})^{-1}\pmb{x}_i + s^{2}}] \numberthis \label{def_4.17}
\end{align*}
Above, $z$ is the z-distribution and $\alpha$ is the confidence level desired for the intervals. For instance, if the prediction intervals are supposed to cover the targets with $95\%$ of probability, then $\alpha$ is taken to be $0.05$ making the $z_{1-\frac{\alpha}{2}}$ equal to $1.96$.

\subsection{MAP method}

\subsubsection{MAP-based parameter estimation}
\label{subsubsection:map-based_param_estimation}

As discussed in \ref{subsection:ERM}, Empirical Risk Minimization that MLE-method is based on has a serious overfitting problem. In order to avoid this problem, use of structural risk minimization through regularization is shown in \citep[p. 55]{tikhonov_regularization_1963}, \citep[pp. 137-141]{shalev-shwartz_understanding_2014} \citep[pp. 583-884]{kacprzyk_springer_2015} and \citep[pp. 94-96]{vapnik_nature_2000}. MAP-based parameter estimation features structural control aiming to decrease high-variance that MLE method exhibit in an attempt to avoid overfitting.

MAP-based parameter estimation is pretty similar to MLE method. However, differently from the MLE method, it assumes a parameter distribution in addition to the assumption of the Gaussian distributed noise that is already made in the derivation of MLE-based estimator. Since this assumption reflects the \textit{prior} knowledge about the parameter vector \pmb{w}, the probability density function of the prior distribution assumed is referred to as \textit{prior}. Most commonly, multivariate Gaussian distribution is assumed for the distribution for parameters, $\pmb{w}$. Therefore, the prior is modeled as Gaussian distribution with $0$ mean and $d$-by-$d$ covariance matrix, $\Sigma_w$. Note that, in the case where the independence of input dimensions can be assumed, the covariance matrix for parameters, $\pmb{w}$, is a diagonal matrix. In this case, instead of defining  $\Sigma_w$, a set of variance values for the parameters, $\{\sigma_{w_1}, \sigma_{w_2}, ... \sigma_{w_n}\}$, can be defined. As a result, the hyperparameters set for MAP method can be written as $\theta=\{\sigma_y, \sigma_{w_1}, \sigma_{w_2}, ... \sigma_{w_n}\}$. If there is already some stored data available (e.g training set, case base), the variance-covariance matrix calculated from the data can be set as the covariance matrix of the assumed multivariate Gaussian distribution for the prior. This eliminates the hassle of hyperparameter tuning.

Having already formulated the prior for MLE and having likelihood modeled by multivariate Gaussian distribution, Bayes' Rule defined as follows
\begin{align*}
\texttt{posterior} & = \frac{\texttt{likelihood} \ \times \ \texttt{prior}}{\texttt{marginal likelihood}} = \frac{p(\pmb{y}|X,\pmb{w}) p(\pmb{w})}{p(\pmb{y}|X)} \\
\text{where } p(\pmb{y}|X,\pmb{w}) & = (\frac{1}{\sqrt{2\pi \sigma_y^2}})^n exp(-\frac{1}{2 \sigma_y^2} (\pmb{y} - \pmb{w}^{\top}X)^{\top}(\pmb{y} - \pmb{w}^{\top}X)) \ \text{by \ref{def_4.7}} \\
\text{and } p(\pmb{w}) & = (\frac{1}{\sqrt{2\pi}})^d| \Sigma_w |^{-\frac{1}{2}} exp(-\frac{1}{2} \pmb{w}^{\top}\Sigma_w^{-1}\pmb{w}) \\
\text{and } p(\pmb{y}|X) & = \int p(\pmb{y}|X,\pmb{w} p(\pmb{w}) d\pmb{w}
\end{align*}
is exploited to find the posterior probability that MAP method is based on in the following.
\begin{align*}
& \text{writing only in terms of $\pmb{w}$-dependent terms:} \\
p(\pmb{y}|X,\pmb{w}) & \varpropto -\frac{1}{2 \sigma_y^2} (\pmb{y} - \pmb{w}^{\top}X)^{\top}(\pmb{y} - \pmb{w}^{\top}X) \\
p(\pmb{w})  & \varpropto -\frac{1}{2} \pmb{w}^{\top}\Sigma_w^{-1}\pmb{w} \\
p(\pmb{w}|X,\pmb{y}) & \varpropto -\frac{1}{2 \sigma_y^2} (\pmb{y} - \pmb{w}^{\top}X)^{\top}(\pmb{y} - \pmb{w}^{\top}X)-\frac{1}{2} \pmb{w}^{\top}\Sigma_w^{-1}\pmb{w} \\
& \varpropto -\frac{1}{2}(\frac{1}{\sigma_y^2}(\pmb{y}^{\top}\pmb{y}-2\pmb{w}^{\top}X\pmb{y}+\pmb{w}^{\top}XX^{\top}\pmb{w})+\pmb{w}^{\top}\Sigma^{-1}\pmb{w}) \\
& \varpropto -\frac{1}{2}(\frac{1}{\sigma_y^2}\pmb{y}^{\top}\pmb{y}-2\frac{1}{\sigma_y^2}\pmb{w}^{\top}X\pmb{y}+\pmb{w}^{\top}(\frac{1}{\sigma_y^2}XX^{\top}+\Sigma_w^{-1})\pmb{w}) \numberthis \label{def_4.18} \\
& \text{trying to bring the above expression into the below form:} \\
& -\frac{1}{2}(\pmb{w}-\mu)^{\top}\wedge(\pmb{w}-\mu) \text{ (assuming posterior is Gaussian)} \\
& = \pmb{w}^{\top}\wedge\pmb{w}-2\pmb{w}^{\top}\wedge\mu + \text{constant not depending on $\pmb{w}$} \\
\text{letting } \wedge & = \frac{1}{\sigma_y^2}XX^{\top}+\Sigma_w^{-1} \text{ and } \mu = \frac{1}{\sigma_y^2}\wedge^{-1}X\pmb{y} \text{ gives } \wedge\mu = \frac{1}{\sigma_y^2}X\pmb{y} \\
& \text{substituting $\wedge\mu$ and $\wedge$ into \ref{def_4.18}, we get: } \\
p(\pmb{w}|X,\pmb{y}) & \varpropto -\frac{1}{2}(\pmb{w}-\mu)^{\top}\wedge(\pmb{w}-\mu) \text{ meaning: } \\
p(\pmb{w}|X,\pmb{y}) & \sim N(\mu, \wedge^{-1})
\end{align*}
As the posterior is found to be a Gaussian with mean $\mu$ and covariance matrix $\wedge^{-1}$ (or alternatively precision matrix $\wedge$), it is simple to find the parameters, $\pmb{w}$, that makes the posterior probability $p(\pmb{w}|X,\pmb{y})$ maximum. Since the mean of median of a Gaussian distribution is its maxima, the mean, $\mu$, is simply the MAP estimate.
\begin{align*}
\pmb{w}_{MAP} & = \frac{1}{\sigma_y^2}(\frac{1}{\sigma_y^2}XX^{\top}+\Sigma_w^{-1})^{-1}X\pmb{y} \\
& = (XX^{\top}+\sigma_y^2 \Sigma_w^{-1})^{-1}X\pmb{y} \numberthis \label{def_4.19}
\end{align*}
When \ref{def_4.19} is compared to \ref{def_4.10}, it is observed that, MAP-estimate differs from MLE-estimate only by the additive term $\sigma_y^2 \Sigma_w^{-1}$ in its first factor. This additive term is known as the \textit{regularization} term. As mentioned in the beginning of the discussion of MAP method, regularization helps avoid overfitting by introducing additional bias in the predictions in return of reducing variance of the predictions through structural control discussed in \ref{subsection:ERM}.

Similarly to \ref{def_4.11}, when the non-linear feature-space mapping through a number of basis functions is employed, the MAP-estimate formula changes as follows.
\begin{flalign}
\pmb{w}_{MAP} = (\Phi(X)\Phi(X)^{\top}+\sigma_y^2 \Sigma_w^{-1})^{-1}\Phi(X)\pmb{y}
\end{flalign}

\subsubsection{Measure of prediction uncertainty in MAP method}
\label{subsubsection:map_pred_bounds}

Although MAP method is pretty similar to MLE, due to the modifications on the parameter estimation mechanism (e.g regularization term), nice asymptotic properties of MLE parameter estimation that make it possible to establish prediction intervals around predictions with a specified probability is lost. 

In the literature, bootstrap-based methods for establishing \textit{approximate} asymptotic prediction bounds for regularized least-squares regression variants such as Ridge Regression (MAP method) is proposed (\cite{crivelli_confidence_1995}, \cite{firinguetti_asymptotic_2011}, \cite{jang_two_2006}). However, these approximation methods are based on bootstrapping and require processing the stored data points more than once which makes them unsuitable for the stream learning scenario. In Chapter \ref{Chapter5}, two different prediction interval mechanisms, based on practical ideas lacking theoretical support, for MAP-based learners is presented. Moreover, in \ref{Chapter6} where the experiment results are analyzed, how one of these performed surprisingly well is shown.

\section{Non-Parametric Models}

The aim of the learning with parametric models is to discover the regression function by accurately estimating the parameters whose products with either the input variables or their mapped versions by a set of basis functions are supposed to sum up to target values. Once the parameters are successfully estimated, it is assumed that the function that produces the targets from the data points are known and for a given data point the target it would produce can be predicted by reproducing the data generation scenario. The key point in this approach is assuming the data generation function to belong to some family of functions such as linear, polynomial, etc). In order to accurately model a data generation function as the sum of products of some parameters and the data points (or their mapped versions), the data generation function must be indeed one of the functions that can be expressed by the sum of products of some certain set of parameters (parameters to be estimated) by the inputs. This idea is susceptible to underfitting when the assumption is not true. On the other hand, non-parametric models does not make any assumptions on the data generation function. Instead, these methods, also known as data-driven methods, manipulates the previously observed target values and their corresponding data points to predict the targets for new data points.

\subsection{Gaussian Process Regression}
\label{subsection:gpreg}

For a finite subset of elements of an index-space, $S = \{t_i \ | \ \forall i \in \mathbb{R}\}$ (e.g time), the vector of Gaussian multivariate- distributed random variables $Y_{t_i}$ represent a Gaussian Process. In the regression context in streaming scenario, one can use the index space of the set of natural numbers each of which denotes the order of the data points appear in the stream. In this case, the index space is $S = \{t_1, t_2, ... t_n | \forall i \in \mathbb{N}_{> 0}\, \ \forall i \leq n \}$. As for the vector Gaussian multivariate distributed random variables mentioned, they simply correspond to the vector of targets, , $[y_{t_1}. y_{t_2}, ... y_{t_n}]^{\top}$, in the case of employing Gaussian Process as a regression tool.

The most interesting fact about the Gaussian processes is that, for a given index set and and the target vector each of whose elements are indexed by the elements of the index set, one can define a mean function that maps indices to real numbers and a positive semi-definite covariance function that maps pairs of indices to real numbers, and these two would guarantee the \textit{existence} of the Gaussian Process with the prescribed indices and the targets that represents a multivariate Gaussian distribution.
\begin{align*}
mean : S \rightarrow \mathbb{R} \\
cov : S \times S \rightarrow \mathbb{R}
\end{align*}
In the regression scenerio, instead of using index values or the cartesian product of index values as the domain sets of these mean and covariance functions respectively, the set of possible input vectors that can be indexed by the index set can be used. This way, the mean and covariance functions can be written as follows.
\begin{align*}
mean(t_i) & = \mu(\pmb{x}_i) \\
cov(t_i, t_{i+k}) & = k(\pmb{x}_i, \pmb{x}_{i+k}) \numberthis \label{def_4.21}
\end{align*}
Above both $k$ and $cov$ are positive semi-definite functions. It is worth noting that the covariance of a pair of targets are computed by a function of an input pair. This establishes the way of relating inputs to the targets which is essential for making predictions for the data points whose targets are not (yet) known.

As previously said the target vector is (multivariate) Gaussian distributed and it is described by a mean and a covariance function.
\begin{flalign}
\begin{bmatrix}y_{t_1}\\y_{t_2}\\...\\y_{t_n}\end{bmatrix} \sim N(
\begin{bmatrix}\mu(\pmb{x}_1) \\ \mu(\pmb{x}_2) \\ ... \\ \mu(\pmb{x}_n)\end{bmatrix}
\begin{bmatrix}k(\pmb{x}_1,\pmb{x}_1)&k(\pmb{x}_1,\pmb{x}_2)&...&k(\pmb{x}_1,\pmb{x}_n)\\k(\pmb{x}_2,\pmb{x}_1)&k(\pmb{x}_2,\pmb{x}_2)&...&k(\pmb{x}_2,\pmb{x}_n)\\...&...&...&...\\ k(\pmb{x}_n,\pmb{x}_1)&k(\pmb{x}_2,\pmb{x}_n)&...&k(\pmb{x}_n,\pmb{x}_n) \end{bmatrix}) \label{def_4.22}
\end{flalign}
The mean vector and covariance matrix above is computed by the mean and covariance functions respectively. The individual elements of the targets vector above is (single-variate) Gaussian distributed. Thus, the following can be written: 
\begin{flalign}
y_i \sim N(\mu(\pmb{x}_i), k(\pmb{x}_i, \pmb{x}_{i})) \label{def_4.23}
\end{flalign}
What \ref{def_4.22} tells is that the vector of targets (observations) is simply a \textit{single} n-dimensional point that is sampled from some multi-variate Gaussian distribution specified by a mean vector and a covariance matrix. How the individual elements of targets are related to the input data points through mean and covariance functions is also clearly seen in \ref{def_4.22}. This is conceptually  one of the key points in making predictions with Gaussian processes. The nature of this relation is \textit{shaped} by the mean and covariance functions. Thus, it can be argued that a Gaussian Process (specified by a certain mean and covariance) picks the functions that can \textit{simulate} the abovementioned sampling of single n-dimensional point from some multi-variate Gaussian distribution given inputs (data points). Here, $n$ denotes the number of data points and their corresponding targets (observations). There are infinite number of functions that can realize the mapping between some given datapoints and given targets and also infinite others that cannot. By \textit{implicitly} specifying these functions that can fit the (training) data, the unobserved target for a given data point can be predicted as the aggregation of the return values of the functions that are in consideration. By doing exactly this, Gaussian Processes provides a very powerful tool to predict the targets for the future data points. Next, the mechanics of this prediction mechanism is discussed.

For a datapoint, $\pmb{x}_i$, \ref{def_4.23} seems to provide a way to predict the target. The predicted target could trivially be what mean function for $\pmb{x_i}$ returns with the upper and lower prediction bounds that could simply be the square root of the value that covariance function returns added to and subtracted from the point prediction respectively. However, this way the existing observed targets and the datapoints are not used for the prediction. The MAP method analog of this would be drawing a random set of parameters from the prior assumed for the parameters without fitting the existing training data. However, as discussed in the previous paragraph, only the functions that can map the existing data points to the observed targets should be considered. In order to do that, the conditioning property of multivariate Gaussian of joint random variables are exploited.
\begin{align*}
K(X,X) & = \begin{bmatrix}k(\pmb{x}_1,\pmb{x}_1)&k(\pmb{x}_1,\pmb{x}_2)&...&k(\pmb{x}_1,\pmb{x}_n)\\k(\pmb{x}_2,\pmb{x}_1)&k(\pmb{x}_2,\pmb{x}_2)&...&k(\pmb{x}_2,\pmb{x}_n)\\...&...&...&...\\ k(\pmb{x}_n,\pmb{x}_1)&k(\pmb{x}_2,\pmb{x}_n)&...&k(\pmb{x}_n,\pmb{x}_n) \end{bmatrix} \\ \numberthis \label{def_4.24}
K(X,\pmb{x}_{n+1}) & = \begin{bmatrix}k(\pmb{x}_1,\pmb{x}_{n+1})& k(\pmb{x}_2,\pmb{x}_{n+1})&...&k(\pmb{x}_n,\pmb{x}_{n+1})\end{bmatrix}^{\top} \\
K(\pmb{x}_{n+1},X) & = \begin{bmatrix}k(\pmb{x}_{n+1},\pmb{x}_1)& k(\pmb{x}_{n+1},\pmb{x}_2)&...&k(\pmb{x}_{n+1},\pmb{x}_n)\end{bmatrix} \\
K(\pmb{x}_{n+1},\pmb{x}_{n+1}) & = k(\pmb{x}_{n+1},\pmb{x}_{n+1}) \\
\end{align*}
By using the above definitions, \ref{def_4.22} can be written in a partitioned way with an extra data point along with its unobserved target to be predicted as follows.
\begin{align*}
\begin{bmatrix}\pmb{y}\\y_{n+1}\end{bmatrix} \sim \ & N(\begin{bmatrix}\pmb{M} \\  \mu_{n+1}\end{bmatrix}, \begin{bmatrix} K(X,X) & K(X,\pmb{x}_{n+1}) \\ K(\pmb{x}_{n+1},X) & K(\pmb{x}_{n+1},\pmb{x}_{n+1}) \end{bmatrix}) \\
\text{where } \pmb{y} & = \begin{bmatrix}y_1&y_2&...&y_n\end{bmatrix}^{\top} \\
\text {and } \pmb{M} & = \begin{bmatrix}\mu_1&\mu_2&...&\mu_n\end{bmatrix}^{\top}
\end{align*}
Applying the conditioning property of Gaussian distributions gives the conditional predictive distribution that is a Gaussian distribution.
\begin{align*}
y_{n+1}|\pmb{y} \sim N & (\mu(\pmb{x}_{n+1}) + K(\pmb{x}_{n+1},X)K(X,X)^{-1}(\pmb{y}-\pmb{M}), \\ & K(\pmb{x}_{n+1},X)K(X,X)^{-1}K(X,\pmb{x}_{n+1}))) \numberthis \label{def_4.25}
\end{align*}

\ref{def_4.25} provides a way to make predictions for a given data point while using the past observed targets and their corresponding data points. It is worth noting that, the variance for the prediction is already provided. Therefore, calculating the prediction bounds are easy. Multiplying the square root of the prediction variance with the z-value that corresponds to the confidence level desired gives the symmetric double-sided pointwise prediction interval whose upper and lower limit can be used as prediction bounds. However, in order to use this nice normal distributed prediction formula, a mean function and, more crucially, a covariance function should be defined. Next, these two are discussed.

Mean function can simply be chosen as the constant function that always returns $0$. What this actually means is that modeling the observed data purely by a Gaussian Process. Using a zero mean function is valid and a common choice. However, sometimes it is reasonable to try to fit the data with a parametric model and model the residuals with a Gaussian Process. This idea is advocated in \cite{blight_bayesian_1975}. With zero means or not, the prediction mechanism of Gaussian Processes remains the same. With a slight modification to \ref{def_4.25}, the conditional predictive distribution in zero-mean case can be obtained.
\begin{align*}
y_{n+1}|\pmb{y} \sim N & (K(\pmb{x}_{n+1},X)K(X,X)^{-1}\pmb{y}), \\ & K(\pmb{x}_{n+1},X)K(X,X)^{-1}K(X,\pmb{x}_{n+1}))) \numberthis \label{def_4.26}
\end{align*}
Regarding the predictions based on Gaussian processes, the choice for the covariance function is more crucial than choosing a mean function. What a covariance function essentially specifies is the \textit{smoothness} of the functions that can fit the data. If, in the data set, neighboring data points are associated with nearby target values, the functions that can fit these data are expected to be smooth ones. However, if the covariance function returns high values for close data points, the functions that are considered by the Gaussian Process to fit the data and make a prediction for new data points will be rather \textit{funky} ones. Thus, the predictions are very likely to be much higher or lower than the target that is not yet observed. On the other hand, if the covariance function returns too low values for data points that are moderately far from each other, this time the function considered by the Gaussian Process will be too smooth to capture local fluctuations (if there is any) that the data generation function features. Therefore, the parameters of the covariance function that specifies the smoothness should be chosen carefully\footnote{This is done through hyperparameter tuning. \label{hyperparam_tuning}}.

As shown in \ref{def_4.21} and explained in the proceeding paragraph, a covariance function in the index space can be defined as the kernel function in the input space. Most widely used kernel function, Squared Exponential, is given below.
\begin{align*}
k(\pmb{x}_p,\pmb{x}_q) = \sigma_w^2exp(-\frac{1}{2}(\pmb{x}_p-\pmb{x}_p)^{\top}D^{-2}(\pmb{x}_p-\pmb{x}_p) + \sigma_y^2\delta_{pq} \numberthis \label{def_4.27}
\end{align*}
Above, $D$ is a diagonal matrix with diagonal entries $\pmb{l}=\{l_1,l_2,..l_d\}$ where $d$ denotes the number of dimensions in the input space. In Squared Exponential kernel, there are $2+d$ parameters namely $\sigma_w$, $\sigma_y$, and elements of $\pmb{l}$. The parameter $\sigma_w$ is called \textit{signal variance} and it can be considered to be a multiplying factor for what is computed by the term with the exponent. $\sigma_y$ is called \textit{noise variance}. It specifies the amount of additive variance in the case the kernel measurement between two identical data points is calculated. The reason why this additive noise is only added to the kernel measure of the identical points have to do with the assumed Gaussian noise in \ref{def_4.4}. Since the noise is assumed to be independently distributed, the additive noise in the targets that correspond to different data points should be \textit{uncorrelated}. Therefore, the contribution of the noise covariance to the non-diagonal entries of the covariance matrix should be zero. Hence, the covariance function which is used to compute the covariance matrix includes the dirac delta function which returns always 0 unless the data points whose covariance to be calculated is identical otherwise 1 as the coefficient of the term noise variance. As for, $\pmb{l}$, it defines \textit{lengthscales} for input data points. Lengthscales are very important because they define how \textit{close} is close enough to make the exponent with a negative sign in \ref{def_4.26} to attain a relatively high value that can make the whole expression evaluate to a high covariance value. A bad choice for this parameter results in the wrong smoothness level causing bad predictions as discussed previously. Another interesting feature about the lengthscale matrix is that, if the lengthscales are correctly chosen, no matter how far the data points are from each other in an irrelevant dimension, the kernel measurement computed for them remains unaffected by the irrelevant dimension as the high-valued lengthscale that correspond to that dimension would diminish its effect from the kernel computation. This is known as ARD (Automatic Relevance Determination) \cite{neal_bayesian_2012}. In addition to detecting the irrelevant dimensions, setting the right lengthscales also helps deal with normalizing dimensions with wide-spread or too close values.

The parameters for the kernel function is actually secondary to the learning process. Moreover, they are supposed to be set prior to learning. Hence, it is reasonable to argue that they are different than the parameters which the parametric models based on. To differentiate them from the parameters used in the sense of coefficients used in parametric-learning, the kernel function parameters are called hyperparameters. The set of hyperparameters for Gaussian Process regression with Squared Exponential kernel function is $\theta=\{\sigma_w, \sigma_y, l_1, l_2, ... l_d\}$

In the following section, how the hyperparameters for Gaussian Process can be tuned for regression is discussed.

\subsubsection{Optimizing Hyperparameters}
\label{subsubsection:gpreg_opt_hyperparams}

As discussed in the previous section, when the hyperparameters of the kernel function are not set correctly, the predictions made with the Gaussian Process with the said kernel function might not be accurate. However, first what is meant by setting the hyperparameters \textit{correctly} should be described in mathematical terms so that the task of choosing the right hyperparameters can be considered to be an optimization problem. Fortunately, the marginal likelihood within the Bayesian inference framework provides a good measure of the \textit{desirability} of the choice of hyperparameters for given inputs and  targets.

In the case of function-space view in which the Gaussian Processes are discussed in the previous section, the marginal likelihood term in level-1 Bayesian inference is defined as follows:
\begin{align*}
p(\pmb{y}|X) = \int p(\pmb{y}|f,X)p(f|X)df \numberthis \label{def_4.28}
\end{align*}

Varying the hyperparameters of the kernel function changes the marginal likelihood term that can be interpreted as the likelihood of the data (targets given the design matrix). For a vector of observed targets and also the stored data points, it is evident that the marginal likelihood probability should be close to unity. The hyperparameter setting that makes it evaluate to a value which is closest to 1 is the optimal. 

Evaluating the integral on the right hand side of \ref{def_4.28} can be tricky. However, luckily we know that given X, $\pmb{y}$ is (multivariate) normal distributed by \ref{def_4.22}. Using the probability density function definition for the multivariate Gaussian Distribution and then taking the logarithm of it, the following is obtained.
\begin{align*}
p(\pmb{y}|X,\theta) & = \frac{1}{(2\pi)^{\frac{n}{2}}|K(X,X)|} exp(-\frac{1}{2}(\pmb{y}-\pmb{M})^{\top}K(X,X)(\pmb{y}-\pmb{M})) \\
log(p(\pmb{y}|X,\theta)) & = -\frac{n}{2}log(2\pi) - \frac{1}{2}log(|K(X,X)|) - \frac{1}{2}(\pmb{y}-\pmb{M})^{\top}K(X,X)(\pmb{y}-\pmb{M}) \numberthis \label{def_4.29}
\end{align*}
The equation \ref{def_4.29}, called \textit{log likelihood} formula, provides an expression to be maximized over different choices of hyperparameter configurations. Moreover, each term in this expression has interpretable contributions to the overall sum. The first negative term is a constant for a fixed-size training set (or a case base). The second negative term specifies the \textit{complexity penalty}. Complex models have higher variance hence the entries of their covariance matrices are bigger resulting in a higher complexity penalty. The third negative term is for the data-fit. Better the algorithm can fit the data, closer this term gets to zero. Note that the trade-off between the complex models that can fit the data usually better and the simpler models that have less risk of overfitting is reflected in the log likelihood computation. 

Fortunately, \ref{def_4.29} expression is differentiable with respect to all hyperparameters. Thus, the optimization algorithms that require the derivative of the target function can be employed. A good choice for the optimization scenario described is using the gradient-based unconstrained optimization methods such as gradient-descend (gradient-ascend in the case target function is a loss function), conjugate gradients method, RPRop (\cite{blum_optimization_2013}, \cite{rasmussen_gaussian_2004}).

All the listed optimization methods are based on the derivative of the log likelihood with respect to hyperparameters. Hence, in order to employ any of these methods, the mentioned derivatives should be derived.
\begin{align*}
\frac{\log(\mathbf{y}|X,\mathbf{\theta})}{d\theta}=\frac{1}{2} \ \mathrm{trace}(K^{-1}\frac{dK}{d\theta})+\frac{1}{2}((\pmb{y}-\pmb{M})\frac{dK}{d\theta}K^{-1}\frac{dK}{d\theta}(\pmb{y}-\pmb{M})) \numberthis \label{def_4.30}
\end{align*}
Above, $K$ is a shorthand for $K(X,X)$ defined in \ref{def_4.22}. It simply denotes the covariance matrix computed by the kernel function of every pair of data points stored in the case base. In \ref{def_4.30}, it appears that the derivative of the covariance matrix with respect to the hyperparameters should also be computed to be able to evaluate the derivative of the log likelihood. Assuming $M$ is diagonal, there are $2+d$ number of hyperparameters namely $\sigma_w$, $\sigma_y$ and lengthscales for each dimension in the input space. This is why, $2+d$ different gradient equations should be derived so that the gradient-based optimizer can optimize different hyperparameters simultaneously. These different gradient equations of different hyperparameters differ only in their $\theta$-dependent terms. Only such term (occurring three times) in equation \ref{def_4.30} is $\frac{dK}{d\theta}$. Therefore, it suffices to find $\frac{dK}{\sigma_w}$, $\frac{dK}{\sigma_y}$ and gradients of $K$ with respect to each lengthscale, $\frac{dK}{\sigma_{l_i}}$, where $i$ denotes the order of the corresponding dimension for which the lengthscale to be tuned. 

Although the hyperparameters are in fact $\sigma_w$, $\sigma_y$ and the lengthscales, taking derivatives with respect to them might result in finding hyperparameter configurations possibly involving negative valued terms when an unconstrained optimization method is employed like gradient-descent. A negative value does not make sense as a signal variance, a noise variance or a lengthscale value. In order to avoid negative hyperparameters, a proxy variable for each of them is introduced in such a way that $e$ to the power of the proxy variable would equate to its corresponding actual hyperparameter. The idea is formulated as follows.
\begin{align*}
\sigma_w = e^{px_w} ,\ \sigma_y = e^{px_y} ,\ \sigma_{l_i} = e^{px_{l_i}}
\end{align*}
Taking the derivative of matrix $K$ means taking the derivatives of the individual elements of the matrix $K$. Since the matrix elements are computed by the squared exponential kernel function and this kernel function has a term that is non-zero only when its inputs are identical, this term is effective only for the diagonal elements, First the the derivatives of the kernel function with respect to different hyperparameters as partial functions is written.
\begin{align*}
& \text{Writing the kernel function in terms of the proxy variables as a piecewise function; } \\
& \qquad \qquad k(\pmb{x_p}, \pmb{x_q}) = \begin{cases} 
      e^{2px_w} + e^{2px_y} & p = q \\
      e^{2px_w} exp(-\frac{1}{2}(\pmb{x}_p-\pmb{x}_q)^{\top})D^{-2}(\pmb{x}_p-\pmb{x}_q) & p \neq q
   \end{cases} \\
   & \text{Note that the diagonal entries of the diagonal-matrix $D$ is} \{e^{px_{l_1}}, e^{px_{l_2}}, ...e^{px_{l_d}}\} \\
   & \qquad \qquad \frac{dk(\pmb{x}_p, \pmb{x}_q)}{d(px_w)} = \begin{cases} 2e^{2px_w} & p = q \\ 
   										2k(\pmb{x}_p, \pmb{x}_q) & p \neq q \end{cases} \numberthis \label{def_4.31} \\
   & \qquad \qquad \frac{dk(\pmb{x}_p, \pmb{x}_q)}{d(px_y)} = \begin{cases} 2e^{2px_y} & p = q \\ 
   										0 & p \neq q \end{cases} \numberthis \label{def_4.32} \\
   & \qquad \qquad \frac{dk(\pmb{x}_p, \pmb{x}_q)}{d(px_{l_i})} = \begin{cases} 0 & p = q \\ 
   										-2e^{-2px_{l_i}}(\pmb{x}_{p_i} - \pmb{x}_{q_i})^2k(\pmb{x}_p, \pmb{x}_q) & p \neq q \end{cases} \numberthis \label{def_4.33}
\end{align*}
Utilizing \ref{def_4.31}, \ref{def_4.32} and \ref{def_4.33}, the derivative of matrix K with respect to one of the hyperparameters can be easily calculated element-wise as the derivative of matrix K with respect to hyperparameter $\theta$ can be written as follows.
\begin{flalign}
\frac{dK}{d\theta} & = \begin{bmatrix}\frac{dk(\pmb{x}_1,\pmb{x}_1)}{d\theta}&\frac{dk(\pmb{x}_1,\pmb{x}_2)}{d\theta}&...&\frac{dk(\pmb{x}_1,\pmb{x}_n)}{d\theta}\\ \frac{dk(\pmb{x}_2,\pmb{x}_1)}{d\theta}&\frac{dk(\pmb{x}_2,\pmb{x}_2)}{d\theta}&...&\frac{dk(\pmb{x}_2,\pmb{x}_n)}{d\theta}\\...&...&...&...\\ \frac{dk(\pmb{x}_n,\pmb{x}_1)}{d\theta}&\frac{dk(\pmb{x}_2,\pmb{x}_n)}{d\theta}&...&\frac{dk(\pmb{x}_n,\pmb{x}_n)}{d\theta} \label{def_4.34} \end{bmatrix}
\end{flalign}

A very important consideration about the optimization task is its non-convexity. As pointed out in \citep[115]{rasmussen_gaussian_2005}, the optimization log likelihood over hyperparameter configurations poses a non-convex optimization problem. In order to deal with non-convexity, use of optimization schemes based on random-search and grid-search are advocated in\cite{bergstra_random_2012}.

The choice for the hyperparameter optimization method heavily depends on the time-efficiency requirements of the online learning application. An exhaustive grid search method might find a better hyperparameter configuration than a gradient-based optimizer which gets stuck at the local optima. However, the grid search takes much more time as it has to scan a big portion\footnote{Big portion refers to a subset of the search-space where the viable parameters are assumed to lie within} of a potentially infinitely big search-space. This problem is aggravated with the growing number of input dimensions as it would require more lengthscale parameters to be tuned. In the next chapter, the choice of hyperparameter optimization method used for Gaussian Process regression is discussed from a more practical point of view.

\subsection{Kernel Regression}

Kernel Regression is a data-driven non-parametric regression method that is based on Kernel Density Estimation. In this section, before delving into the derivation of Kernel Regression prediction formula and its details, Kernel Density Estimation is briefly summarized.

\subsubsection{Kernel Density Estimation}

Suppose that we have a finite sample from an unknown distribution such as $\{\pmb{x}_1, \pmb{x}_2, ... \pmb{x}_n\}$. The aim of Kernel Density Estimation is to recover the probability density function of the unknown distribution by using the sample available. The most basic non-parametric way of estimating the probability density is building histograms. The histogram building is a very intuitive and simple technique based on defining bins that are uniformly spaced in the each direction of the space where the data points are defined and counting the number of samples that fall into the bins. The distribution of the number of data samples in the bins represent the frequency distribution of the sampled points. This is an approximation for the probability density of the distribution where the samples is drawn from. It is worth noting that the estimated probability density estimate can be seen as the collection of partial constant functions that returns the same value for all the points that fall into the same bin. The approximated density function can be written as follows.
\begin{align*}
m & = n\prod_{i=1}^d l_i \numberthis \label{def_4.35} \\
\hat{f}(\pmb{x}) & = \frac{1}{nm} \sum_{i=1}^{n} \sum_{j=1}^m I(\pmb{x} \in b_j) I(\pmb{x_i} \in b_j) \numberthis \label{def_4.36}
\end{align*}
In \ref{def_4.35}, $l_i$ specifies the spacing in the $i_{th}$ dimension of the space where the data points are defined in and $m$ is the number of bins defined in this space. In \ref{def_4.36}, $\pmb{x}$ is the data point for which the probability density is to be estimated. Moreover, $\pmb{x}_i$ is the $i_{th}$ sample in the sample set. Finally, $b_j$ denotes the $j_{th}$ bin and the function $I$ returns $1$ if the condition specified within its parenthesis space holds and $0$ otherwise. The right hand side of the equation \ref{def_4.36} simply counts the number of the data points in the sample that fall in the same bin as the data point for which the density is estimated and divides this number by the number of bins multiplied by the number of items.

The problem with estimating the probability density of a given data point by \ref{def_4.36} is that, as mentioned earlier, it returns the same density estimation for the neighboring data points. In a naive attempt to solve this problem, one might try shrinking the bin size by decreasing the spacing parameters $i$'s. However, as the bins get smaller, the chances for the data point for which a kernel density estimation to be made to be in the same bin as the ones in the sampled set decreases. This means no matter how a data point is close to the sampled data points (if not identical), the density for it will be predicted as zero implying the contribution to the estimated density of both far and close data points both same and zero. Therefore, it can be argued that the density estimation with smaller bins do not produce desirable results.

A good solution to the problem with density estimation by histogram is using \textit{overlapping} bins. Differently from the bins used for histogram building, this time every data point in the sample has its own bin. Moreover, an interesting feature of these bins is that their \textit{containment effect} fades when moved away from their corresponding data point and it peaks at the data points for which the bins are defined. Containment can be easily understood if one imagines the $I$ functions appear in \ref{def_4.35} returning containment of either $0$ or $1$ for a pair of data points. Differently from the density estimation by histograms, in the kernel-based density estimation, whenever a density estimate should be computed for a data point, containment of it by different bins of the data points in the sample which can be floating values between $0$ and $1$ are added up. The containment by the different data points in the sample can be seen as the contribution of them to the density estimate for the data point that the distribution density to be estimated. This way, for the points with no matching identical data-point in the sample, still a density estimate other than $0$ could be calculated. The mentioned \textit{fading} effect is modeled by what is called the \textit{kernel}. The kernel is a function of $(\pmb{x}_i-\pmb{x})H^{-1}$ where $\pmb{x}_i$ is the sampled data point, \pmb{x} is the point for which the density is computed and $H$ is the lengthscale matrix which is explained later. The contribution to the density estimate by a single bin to the all possible different points should sum up to unity. In other words, the kernel function should satisfy the following constraint.
\begin{align*}
\int_{-\infty}^{\infty} k((\pmb{x}_i-\pmb{x})H^{-1})d\pmb{x} = 1
\end{align*}
Since, there are $n$ points in the training set (case base), calculating the kernel density as the summation of the contributions by the \textit{bins} of $n$ data points potentially result in a bigger density estimate than what would be calculated by the unknown probability density function. This becomes clear, if we write the following by manipulating the above constraint.
\begin{align*}
\int_{-\infty}^{\infty} \sum_{i=1}^{n} k((\pmb{x}_i-\pmb{x})H^{-1})d\pmb{x} = n|H|
\end{align*}
In order to have a \textit{sane} approximate probability function whose integral from negative infinity to positive infinity, a normalizing constant which is exactly $\frac{1}{n|H|}$ is put in front of the kernel density estimation formula which is shown below.
\begin{flalign}
\hat{f}(\pmb{x}) = \frac{1}{n|H|} \sum_{i=1}^n k((\pmb{x}_i-\pmb{x})H^{-1}) \label{def_4.37}
\end{flalign}
Above, $H$ is a $d$-by-$d$ nonsingular bandwidth matrix and its diagonal entries correspond to lengthscales of each dimension in the $d$-dimensional space where the data points are defined. Very similarly to the lengthscales used in the Squared Exponential function used in Gaussian Process Regression, lengthscales in the Kernel Regression context also are used to normalize the values of the data points in different dimensions with different variance. The reason why the lengthscales are needed and what happens when they are not used or not optimized is discussed in the next subsection.

A widely used kernel function is Gaussian kernel and it is based on the probability density function of Gaussian distribution.
\begin{flalign}
k_H(\pmb{x}_i-\pmb{x}) = \frac{1}{(2\pi)^{\frac{d}{2}}} exp((\pmb{x}_i-\pmb{x})^{\top}(\pmb{x}_i-\pmb{x})) \label{def_4.38}
\end{flalign}
Plugging \ref{def_4.38} into \ref{def_4.37}, the formula for kernel density estimation using Gaussian kernel is obtained.
\begin{flalign}
\hat{f}(\pmb{x}) = \sum_{i=1}^n \frac{1}{n|H|(2\pi)^{\frac{d}{2}}} exp(((\pmb{x}_i-\pmb{x})H^{-1})^{\top}((\pmb{x}_i-\pmb{x})H^{-1}))
\end{flalign}

Next, how the kernel density estimate for regression can be used is discussed.

\subsubsection{Nadaraya-Watson Estimator}

Most common way of making predictions given datapoints by using Kernel Density Estimation is employing Nadaraya-Watson Estimator. The prediction formula of Nadaraya-Watson is derived from the definition of conditional expectation as presented in \citep[p. 89]{hardle_nonparametric_2012}:
\begin{flalign}
E(y| X = \pmb{x}) & = \int_{-\infty}^{\infty} p(y|\pmb{x})ydy = \frac{\int_{-\infty}^{\infty} p(y,\pmb{x}) y dy}{p(\pmb{x})} \label{def_4.40}
\end{flalign}
Above, $p(y,\pmb{x})$ and $p(\pmb{x})$ are unknown. Instead, the kernel density estimate functions can be used. Writing the kernel density estimate for $\pmb{x}$ is easy. That of $(y,\pmb{x})$ is a bit tricky. \citep[p. 89]{hardle_nonparametric_2012} suggests using the what is known as \textit{product kernel} which is defined for two different random variables that have the same number of samples in the sample set. The product kernel for $(y,\pmb{x})$ is written as follows.
\begin{flalign}
\hat{f}(\pmb{x},y) = \frac{1}{n|H_1|H_2} \sum_{i=1}^n k((\pmb{x}_i-\pmb{x})H_1^{-1}) k(\frac{y_i-y}{H_2}) \label{def_4.41}
\end{flalign}
Plugging the density estimates for the unknown probability functions in \ref{def_4.40}, we have the following.
\begin{align*}
\frac{\int_{-\infty}^{\infty} p(y,\pmb{x}) y dy}{p(\pmb{x})} & = \frac{\frac{1}{n} \sum_{i=1}^n \int_{-\infty}^{\infty} \frac{1}{|H_1|} k((\pmb{x}_i-\pmb{x})H_1^{-1}) \frac{1}{H_2} k(\frac{y_i-y}{H_2}) y dy}{\frac{1}{n|H1|} \sum_{i=1}^n k((\pmb{x}_i-\pmb{x})H_1^{-1})} \\
& = \frac{\sum_{i=1}^n k((\pmb{x}_i-\pmb{x})H_1^{-1}) \int_{-\infty}^{\infty} \frac{1}{H_2} k(\frac{y_i-y}{H_2}) y dy}{\sum_{i=1}^n k((\pmb{x}_i-\pmb{x})H_1^{-1})} \\
& = \frac{ \sum_{i=1}^n k((\pmb{x}_i-\pmb{x})H_1^{-1}) \int_{-\infty}^{\infty} (y_i-H_2s) k(-s) ds}{ \sum_{i=1}^n k((\pmb{x}_i-\pmb{x})H_1^{-1})} \ (s = \frac{y-y_i}{H_2},\ ds = \frac{dy}{H_2}) \\
& \text{ (knowing that } \int_{-\infty}^{\infty} sk(-s) ds = 0 \text{ and } \int_{-\infty}^{\infty} k_{H_2}(s) ds = 1 ) \\
& = \frac{ \sum_{i=1}^n k((\pmb{x}_i-\pmb{x})H_1^{-1}) y_i}{ \sum_{i=1}^n k((\pmb{x}_i-\pmb{x})H_1^{-1})} \numberthis \label{def_4.42}
\end{align*}
Finally, the prediction formula used by the Nadarya-Watson estimator is derived in \ref{def_4.42}

\subsubsection{Prediction Bounds Estimation}

Both \citep[pp 35-36]{yatchew_semiparametric_2003} and \citep[p. 119]{hardle_nonparametric_2012} provide a way to calculate the confidence intervals for the predictions made by the Nadarya-Watson estimator. The variance is calculated as follows.
\begin{flalign}
Var[\hat{y}|\pmb{x}] = \frac{||K||_2^2 ASE}{\hat{f}(\pmb{x})} \label{def_4.43}
\end{flalign}
In above formula, ASE denotes the average squared error and $\hat{f}(\pmb{x})$ is the kernel density estimate for the data point, \pmb{x}, which the prediction is made for. As for $||K||_2^2$, it depends on the kernel function choice and it is calculated by $\int_{-\infty}^{\infty} k(\pmb{x})^2d\pmb{x}$. For the Gaussian Kernel, this calculation yields $(4\pi)^{\frac{-d}{2}}$.

Once the prediction variance is known, the prediction bounds could be found straightforwardly for a desired confidence level as discussed in the section where Gaussian Process Regression is covered.

\subsubsection{Optimizing Hyperparameters}
\label{subsubsection:kreg_tuning_approach}

Only hyperparameter to be tuned is the lengthscale matrix, $H$. Without using lengthscales explicitly (equivalent of using an identity matrix for $H$) or bad choices for lengthscales, the kernel values computed by the kernel function can be always too high or too low no matter for what pair of data points the kernel is being calculated rendering the whole kernel density estimation idea useless. This results in inaccurate predictions for targets given the datapoints by kernel regression. Therefore, lengthscales are crucial to adapt the kernel function to the distribution of the values in the input dimensions. 

Unlike the Gaussian Process Regression, there is no derivable loss function for Kernel Regression. Therefore, instead of turning the hyperparameter tuning problem into an optimization problem, validation techniques are typically used to find near-optimal values for the entries of $H$. It is assumed that an optimal $H$ matrix would be the one that minimizes the empirical error. In order to calculate the empirical error, an error measure should be chosen among various options such as MSE, ISE, MISE, ASE and MASE. In \cite[p. 110]{hardle_nonparametric_2012}, ASE is chosen for the same purpose and in \cite{hardle_optimal_1985} it is stated that ASE, ISE and MISE asymptotically lead to the same choices of length-scales\footnote{The authors adopted the term smoothing as different choices of lengthscales result in different levels of smoothness of the line that connects the predictions of the all possible input points} choice when employed in a validation scheme to tune the lengthscales for Nadarya-Watson Estimator. Therefore, as the error measure, ASE is preferred.

Having chosen the error measure, The aim of the hyperparameter tuning can be now explicitly stated as finding the matrix $H$ that minimizes the ASE error defined as follows
\begin{align*}
ASE(H)=\frac{1}{n}\sum_{i=1}^{n}(\hat{y}_i - y_i)^2
\end{align*}
The problem with trying to minimize the above equation is that $y_i$ itself is used for computing $\hat{y}_i$. Therefore, making the lengthscales infinitely small would make ASE equal to $0$ as the only point used in calculating the prediction $\hat{y_i}$ would be $y_i$. This, of course, is not a good validation strategy. In order to solve this problem, the use of hold-out-one estimator is proposed in \citep[pp. 112-113]{hardle_nonparametric_2012}. The idea is, when calculating the error, the predictions for the target is made without using the target point that is in the training set (case base). Hold-out-one Nadarya-Watson estimator is formulated as follows.
\begin{align*}
\hat{f}(\pmb{x}_i)_{-i} = \frac{\sum_{i\neq j} k((\pmb{x_i}-\pmb{x_j})H^{-1})y_j}{\sum_{i\neq j} k((\pmb{x_i}-\pmb{x_j})H^{-1})}
\end{align*}
Using this estimator cross-validation error can be estimated using the following formula:
\begin{align*}
CV(H) = \frac{1}{n}\sum_{i=1}^{n}(\hat{f}(\pmb{x}_i)_{-i} - y_i)^2
\end{align*}
The $H$ that minimizes $CV(H)$ is the optimum choice for the lengthscale matrix. In the absence of the derivative of the $CV(H)$ with respect to $H$, a better idea than a random search for the optimal $H$ is suggested in \cite[p. 175]{silverman_density_1986}. The author argues that the optimal values of the lengthscales which are essentially used to normalize the values of the inputs in different dimensions of the input space could be inferred from the data itself and suggests that the lengthscales matrix should be in the same \textit{shape} as the data. Therefore, it is estimated that the optimal lengthscale matrix $H$ should be a multiple of the variance-covariance matrix of the data that could be calculated as follows.
\begin{align*}
COV = \frac{(X^{\top}-\frac{11^{\top}X^{\top}}{n})^{\top}(X^{\top}-\frac{11^{\top}X^{\top}}{n})}{n}
\end{align*}
Above, $1$ is a column vector with $n$ $1$s and $X$ is the $n$-by-$d$ design matrix.

After finding the variance-covariance matrix of the data points available, the hyperparameter tuning problem for Kernel Regression can be formulated as follows:
\begin{align*}
& \text{for a predefined $\alpha_{min}$, $\alpha_{max}$ and $\alpha_{step}$} \\
& \text{find: } argmin_{\alpha} CV(\alpha COV)
\end{align*}
Above defined problem can be trivially solved by scanning different values of $\alpha$ by starting from $\alpha_{min}$ and incrementing it by $\alpha_{step}$ at each step until $\alpha_{max}$ is reached and picking the $\alpha$ that makes $CV(\alpha COV)$ the smallest $CV$ calculated. This way, the nearly-optimal choice for $H$ is found to be $\alpha COV$. 

\chapter{Implementation} 
\label{Chapter5} 

\lhead{Chapter 5. \emph{Implementation}} 

This chapter explores the implementation aspects of the regression algorithms discussed in Chapter \ref{Chapter4}. The implementation goal is to have \textit{learners} (running learning algorithm instances), based on the previously covered regression algorithms, that fulfill the requirements listed in \ref{list:restimator_requirements}. Most of the effort put in the implementation of the learners is to meet the second requirement which is \textit{robustness to abrupt concept drifts}. As discussed in \ref{subsection:2.4.1}, learners with the ability to adapt to non-stationary streaming data are called online learners. From the implementation point of view, the most essential feature of an online learner is the \textit{incremental update mechanism} that affords the required adaptivity to the non-stationarity. Therefore, the main focus of this Chapter is on the implementation of the incremental update mechanism for the the regression algorithms discussed previously. 

Since none of the regression algorithms covered in the previous chapter is originally proposed with an in-built incremental update mechanism, their prediction mechanism should be \textit{reimagined} so that they can incrementally update their internal predictive models with the new data points arriving from the data stream. Undoubtedly, the incremental update mechanism for different learning algorithm should be implemented using different techniques due to the different prediction mechanisms they feature. Hence, before discussing the implementation of update mechanism employed in the implemented online learners, a categorization of the regression algorithms based on the way they build their predictive model and come up with prediction bounds is presented. 

\section{Categorization of Regression Algorithms}

In how regression algorithms treat the input to build a predictive model, they are divided into two classes namely \textit{absorbing methods} and \textit{accumulative methods}. The parametric algorithms discussed in Chapter \ref{Chapter4}, MAP-method and MLE-method, are absorbing methods. They infer the parameters of the regression function from the observed data and make predictions with the parameters inferred. More precisely, the observed data play no direct role in making predictions. On the other hand. non-parametric algorithms covered, Gaussian Process Regression and Kernel Regression, are data-driven methods. They accumulate the data to deliver predictions. More precisely, In non-parametric regression algorithms, the predicted target for a new point can be formulated as the weighted average of the observed targets for the accumulated data points that.

For the main regression algorithms considered in this thesis, the absorbing methods and the accumulative methods nicely correspond to parametric learning models and non-parametric ones respectively. However, this holds only when the algorithms are used to make point predictions. If prediction bounds along with the point predictions are needed, the prediction bounds estimation mechanism should also be considered before deciding an algorithm needs to store the historic data or not. 

From the non-parametric learning models, Gaussian Process Regression finds the predictive variance by using only the design matrix which is composed of only the inputs of the past data (\ref{def_4.25}). On the other hand, Kernel Regression needs to calculate the \textit{instantaneous} squared residual error\footnote{Instantaneous squared residual error is the measure of amount of error that the updated version of the online learner \textit{would} make for the past data points.} for the historic data points prior to the estimation for new data point (\ref{def_4.42}) necessitating the bookkeeping of the past observed targets. However, for making the point predictions, since both of these algorithms already need to store both the inputs and the observed targets of the historic data (\ref{def_4.25}, \ref{def_4.41}), their prediction bound estimation mechanism do not impose any additional data storage requirements. 

For the parametric models, as their prediction mechanism itself does not require storing any of the past data, the choice for the prediction bound estimation mechanism determines the data storage requirements. Finding the asymptotic prediction intervals for MLE-method discussed in \ref{subsubsection:4.2.1.2} requires the computation of the instantaneous squared residual error just like the prediction bounds estimation method of Kernel Regression. Since computing the instantaneous squared residual error computation requires the observed targets of the past data points, the absorbing methods which use this prediction bound estimation method have to store them. As an alternative to the asymptotic prediction bounds, an ad-hoc prediction bounds estimation method\footnote{This is a fully practical method without any theoretical support and it is referred to as ad-hoc prediction bounds estimation method. Its details is discussed in \ref{subsubsection:impl_predict_mleforgetting}} that is not based on the instantaneous squared residual error is implemented to be employed for parametric algorithms. Due to these two options for the prediction bound estimation method, two different sets of online learners based on parametric models are implemented. One set implements the ad-hoc prediction bounds method that do not require storing any of the historical data and the other set implements the asymptotic prediction bounds estimation method.

\section{Categorization of Online Learners}

Having discussed the dependencies of the online learning algorithms on the data when making point predictions and estimating prediction bounds, it is possible to propose two different types of design for the online learners to be implemented to handle stream data with potential non-stationaries. These are the forgetting-based design and sliding-windowed design.

\subsection{Forgetting-based Design}

Forgetting-based design is for the learning algorithms that do not need to store the past data for their point prediction and prediction bounds estimation mechanism. This kind of design is applicable to online learners based on parametric regression algorithms. The main idea is to discount the effect of the less recent data points on the calculation of the parameters by means of a forgetting-factor parameter, denoted as $\alpha$. In order to make this possible a recursive parameter computation method parameterized with $\alpha$ is needed. Borrowing some of the ideas presented in \citep[pp. 161-164]{bontempi_statistical_2011}, the needed formula is derived as follows.
\begin{align*}
\text{let } \pmb{w}_{old} & = (XX^{\top})^{-1}X\pmb{y} = M1_{old}M2_{old} \numberthis \label{def_5.1} \\
\pmb{w}_{new} & = ([X \ \pmb{x}_{new}][X \ \pmb{x}_{new}]^{\top})^{-1}[X \ \pmb{x}_{new}] [\pmb{y} \ y_{new}]^{\top} \\
M1_{new}^{-1} & = [X \ \pmb{x}_{new}][X \ \pmb{x}_{new}] = XX^{\top} + \pmb{x}_{new}\pmb{x}_{new}^{\top} \\ 
& = M1_{old}^{-1} + \pmb{x}_{new}\pmb{x}_{new}^{\top} \text{ (rank-1 update) } \numberthis \label{def_5.2} \\
M2_{new} & = [X \ \pmb{x}_{new}][\pmb{y} \ y_{new}]^{\top} = M2_{old} + \pmb{x}_{new}y_{new} \\
M1_{old}^{-1}\pmb{w}_{old} & = (XX^{\top})^{-1}(XX^{\top})^{-1}X\pmb{y} = X\pmb{y} \\
M1_{new}^{-1}\pmb{w}_{new} & =  [X \ \pmb{x}_{new}][\pmb{y} \ y_{new}] = M1_{old}^{-1}\pmb{w}_{old} + \pmb{x}_{new}y_{new} \\
& = (M1_{new}^{-1} - \pmb{x}_{new}\pmb{x}_{new}^{\top})\pmb{w}_{old} + \pmb{x}_{new}y_{new} \text{ (using } \ref{def_5.2}) \\
& = (M1_{new}^{-1}\pmb{w}_{old} - \pmb{x}_{new}\pmb{x}_{new}^{\top}\pmb{w}_{old} + \pmb{x}_{new}y_{new} \\
& = (M1_{new}^{-1}\pmb{w}_{old} - \pmb{x}_{new}(y_{new} - \pmb{x}_{new}^{\top}\pmb{w}_{old}) \\
\pmb{w}_{new} & = \pmb{w}_{old} - M1_{new}\pmb{x}_{new}(y_{new} - \pmb{x}_{new}^{\top}\pmb{w}_{old}) \text{ where; }   \numberthis \label{def_5.3} \\
M1_{new} & = (M1_{old}^{-1} + \pmb{x}_{new}\pmb{x}_{new}^{\top})^{-1} \text{ (using } \ref{def_5.2}) \\
M1_{new} & = M1_{old} - \frac{M1_{old}\pmb{x}_{new}\pmb{x}_{new}^{\top}M1_{old}}{1+\pmb{x}_{new}^{\top}M1_{old}\pmb{x}_{new}} \text{ (using matrix inversion lemma \cite{bartlett_inverse_1951})} \numberthis \label{def_5.4} \\
& \text{Plugging \ref{def_5.4} into \ref{def_5.3}; } \\
\pmb{w}_{new} & = \pmb{w}_{old} - (XX^{\top})^{-1} - \frac{(XX^{\top})^{-1}\pmb{x}_{new}\pmb{x}_{new}^{\top}(XX^{\top})^{-1}}{1+\pmb{x}_{new}^{\top}(XX^{\top})^{-1}\pmb{x}_{new}})\pmb{x}_{new}(y_{new} - \pmb{x}_{new}^{\top}\pmb{w}_{old}) \numberthis \label{def_5.5}
\end{align*}
The equation \ref{def_5.5} is recursive, therefore a new data point (denoted as $\pmb{x}_{new}$ can be easily used to update the parameters from $\pmb{w}_{old}$ to $\pmb{w}_{new}$. However, the forgetting factor, $\alpha$, is not used in \ref{def_5.5} to discount the effect of less recent data points on the parameters. The following modification of the formula that calculates the first multiplicand, $M1$, of the parameter vector to be estimated is done to introduce the forgetting factor into the calculation.
\begin{align*}
M1_{new} & = \frac{1}{1-\alpha}(M1_{old} - \frac{M1_{old}\pmb{x}_{new}\pmb{x}_{new}^{\top}M1_{old}}{1+\pmb{x}_{new}^{\top}M1_{old}\pmb{x}_{new}}) \numberthis \label{def_5.6}
\end{align*}
Plugging \ref{def_5.6} into \ref{def_5.3} and writing only in terms of data points and their corresponding observed targets, the following parameter estimation formula parametrized with a forgetting factor $\alpha$ is obtained.
\begin{flalign}
\pmb{w}_{new} & = \pmb{w}_{old} - \frac{1}{1-\alpha}((XX^{\top})^{-1} - \frac{(XX^{\top})^{-1}\pmb{x}_{new}\pmb{x}_{new}^{\top}(XX^{\top})^{-1}}{1+\pmb{x}_{new}^{\top}(XX^{\top})^{-1}\pmb{x}_{new}})\pmb{x}_{new}(y_{new} - \pmb{x}_{new}^{\top}\pmb{w}_{old})) \label{def_5.7}
\end{flalign}

Forgetting-based design ensures the adaptivity to the emerging concepts by discounting the effect of the $m_{th}$ most recent data point by $\alpha^{m-1}$ favoring the newer data points over the older ones.

Online learners using forgetting-based design cannot tune themselves as all the tuning schemes require access to data.

\subsection{Sliding-Window Design}

Sliding-window design can be used for learning algorithms that depend on the past data for either making point predictions or estimating the prediction bounds. The main idea is to store most recent $k$ data points in most-recent to least-recent order in a structure called sliding window. Moreover, with every new data point arrive from the stream, the least recent data point from the tail of the window is dropped and the window items are slid by one position to the tail making space for the most recent data point in the head position of the stream. The data available to the learning algorithm for delivering predictions with prediction bounds is the data stored in the sliding-window. As sliding window stores the most recent data points, the adaptation to emerging concepts is automatic. Furthermore, as the sliding-window size is fixed throughout the lifetime of the online learner, the space it occupies can always be bounded by a constant which is a desirable property in online learning scenario.

The operation of the sliding-window on a data stream is illustrated in \ref{fig:streaming}

\begin{figure}[htbp]
  \centering
    \includegraphics[width=\linewidth]{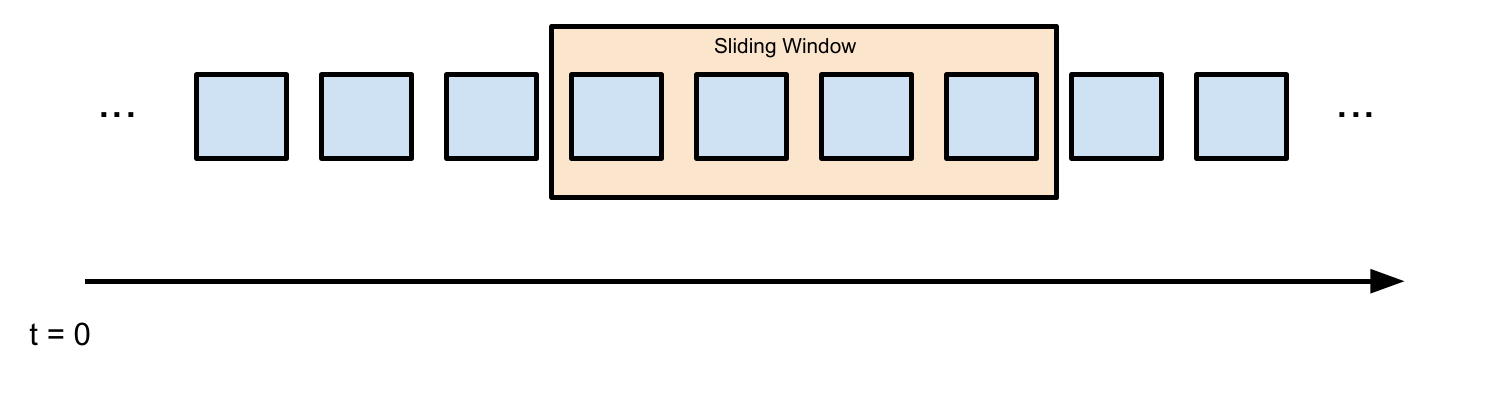}
  \caption{Sliding window on a data stream is visualized}
  \label{fig:streaming}
\end{figure}

In order to realize the mentioned \textit{sliding} operation (costly if implemented in a naive way) with the new data points arriving, a ring buffer is employed making it possible to \textit{slide} the window by only modifying two pointers that point to beginning end the end of the buffer.

\section{Online Learner Semantics}
\label{section:online_learner_semantics}

There are two basic functionality that all the online learners should implement. These are \texttt{predict}, and \texttt{update}. Moreover, there is one more optional operation that is implemented only by the sliding-windowed online learners. This operations is called \texttt{tune} and it \textit{calibrates} the online learner by using the data stored in the sliding window. The behavior of the online learners are determined by their current state. The frequency they carry out these two mandatory and one optional operations change from one state to another. In order to capture these different behaviors, state transitions for forgetting-based learners and the sliding-windowed learners are presented.

\begin{figure}[htbp]
  \centering
    \includegraphics[width=\linewidth]{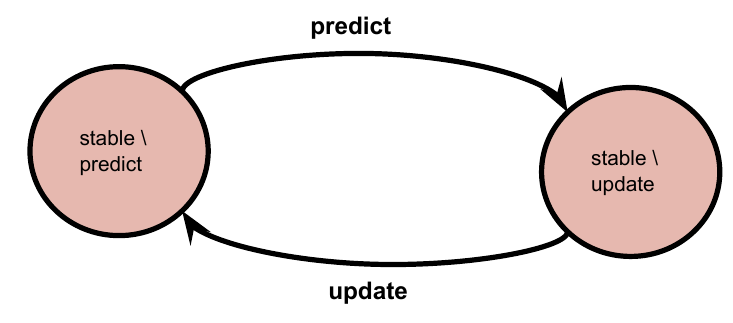}
  \caption{State-transition diagram of the forgetting-based learners}
  \label{fig:forgetting-based_states}
\end{figure}

Figure \ref{fig:forgetting-based_states} describes the states, state transition invoking operations and the order by which different kind of operations follow each other for the forgetting-based learner implementations. According to the diagram, there are simply two states where the learner is allowed to carry out one of the \texttt{update} or \texttt{predict} operations. These two operations always follow each other. It is easy to \textit{play} the online prediction protocol whose pseudocode is presented in pseudocode \ref{alg:opp} on this state diagram.

\begin{figure}[htbp]
  \centering
    \includegraphics[width=\linewidth]{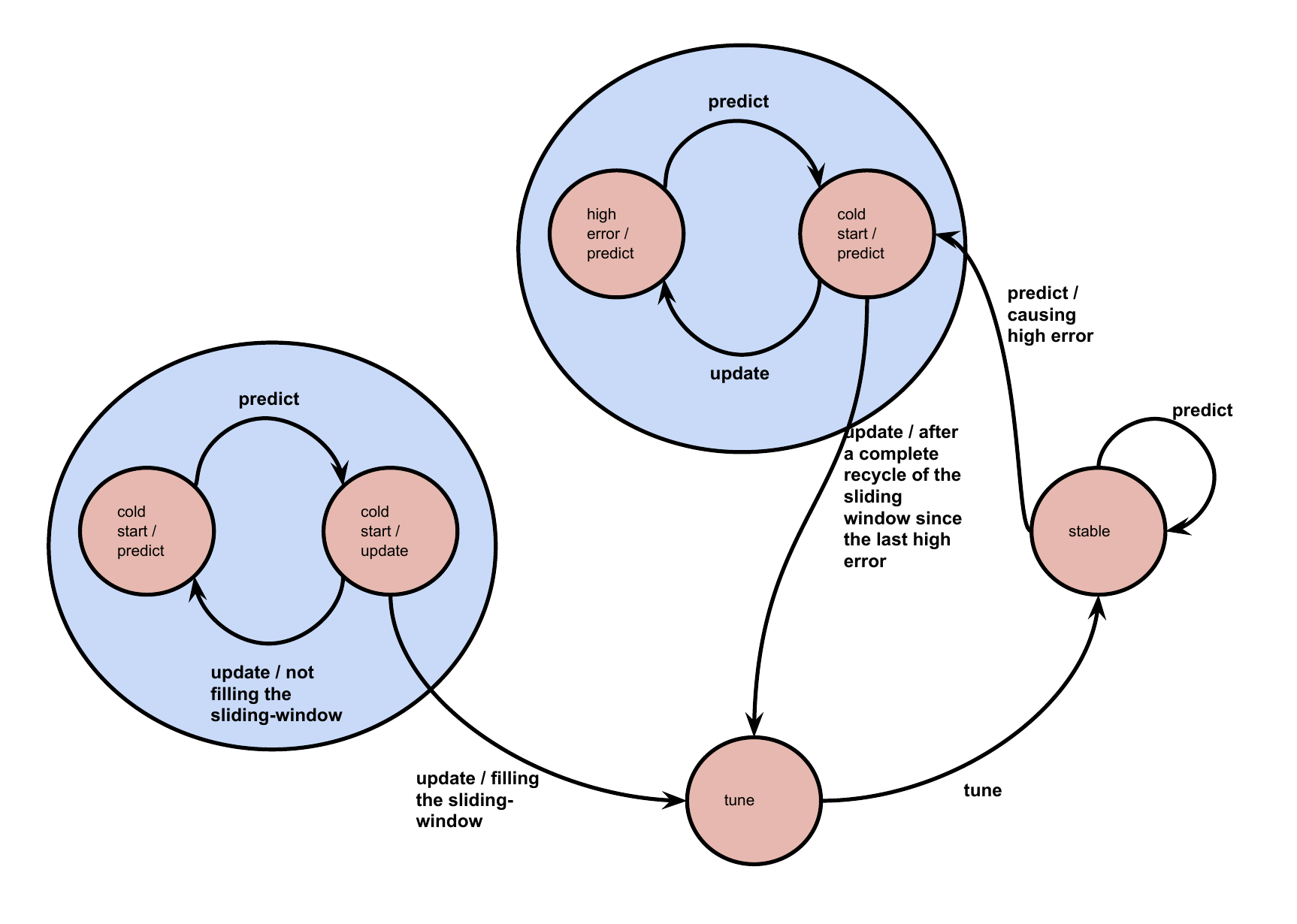}
  \caption{State-transition diagram of the sliding-windowed learners}
  \label{fig:sliding-windowed_states}
\end{figure}

Figure \ref{fig:forgetting-based_states} describes the states, state transition invoking operations and the order by which different kind of operations follow each other for the sliding-windowed learner implementations. Their state-transition diagram is more complex than that of forgetting-based learners. According to the diagram, a sliding-windowed learner is in one of the four main state namely cold start, stable, high-error and tune. There are two mini-states within the cold start and high-error states where the predictions and updates successively follow each other conforming to online prediction protocol (pseudocode \ref{alg:opp}). However, in the state stable after transiting from the one-operation state tune, the learner is not updated with the new data as long as its predictions are accurate. The reason for such a design choice is that for the sliding-windowed learners the \texttt{update} is a costly operation involving matrix manipulations. Therefore, whenever it is not necessary, it can be skipped to speed up the online learner. When the prediction error for predicted data points is low, this means that the learner has adapted to the current concept and its current model will perform well until a new concept is encountered in the data stream. When a concept drift occurs, this is detected by the increasing prediction error and the learner transits to high-error state where each \texttt{predict} operation is followed by an \texttt{update} operation. And this high-error state lasts until all the data points left from the pre-high-error detection is replaced by the newer ones. Then, tuning takes place where the hyperparameters of the learning algorithms are fine-tuned. The cold-start state is very similar to high-error state however, differently, in the cold-start state, the sliding window implemented by a ring buffer only incorporates the data points arriving from the stream and do not drop any until the sliding window gets full. Then the learner tunes itself and winds up in the stable state.

\section{Implementation of Online Learner Operations}

In this section, the implementation of the \texttt{predict}, \texttt{update} and \texttt{tune} operations for each online regression algorithm discussed in Chapter \ref{Chapter3} are presented. Moreover, the computational shortcuts proposed in the literature for recursive fast-update routines are discussed for the applicable regression algorithms. Furthermore, the prediction bounds estimation methods that are not covered in Chapter \ref{Chapter4} due to the their lack of theoretical support is discussed here.

\subsection{MLE method operations}

MLE method, being a parametric algorithm, could be implemented using either forgetting-factor design or sliding-windowed design. Obviously, the implementation of the \textit{predict}, \textit{update} and \textit{tune} operations depend on this design choice. Next, the implementation of each operator for both of the design choices is discussed.

\subsubsection{Implementation of \texttt{predict} for \texttt{BayesianMLEForgetting}}
\label{subsubsection:impl_predict_mleforgetting}

Point predictions mechanism of \texttt{BayesianMLEForgetting} is simple once the regression parameters vector $\pmb{w}$ is known. The point prediction is simply the product of parameters, $pmb{w}$ and the data point, $\pmb{x}$, for which the target is predicted.

In the case feature-space mapping is opted in, the input is mapped to the feature space by the set of base functions defined. The set of base functions implemented is formulated as follows.
\begin{align*}
 \Phi(\pmb{x}) = \{\phi: y = \phi(\pmb{x}) = \pmb{x}_i^a\pmb{x}_j^b \ | \ \forall y \in \mathcal{R} \wedge \forall \pmb{x} \in \mathcal{R}^d \wedge \forall i,j \in \mathcal{R}_{> 0} \\ \wedge \ \exists a,b \in \mathcal{R}_{\geq 0} \wedge i \leq d \wedge j \leq d \wedge 1 \leq a+b \leq 2\}
 \end{align*}
Above, i and j are the indices of the values that sit at respectively the $i_{th}$ and $j_{th}$ dimension of the input vector \pmb{x}. In addition to the base functions expressed above, logarithm and square root of each dimension of the input vector is used as additional base functions. Feature-space mapping is oblivious to learning algorithm used. It is used by the half of the parametric learners. When it is used the number of parameters increase from $d$ to the cardinality of above described set plus $2d$ (because of the addition of the logarithm and square root base functions). In the rest of this chapter, the feature-space mapping implementation is not separately discussed for different learners other than \texttt{BayesianMLEForgetting} as they all share the same implementation.

As for estimating the prediction bounds for the data point which a point prediction is made for, the asymptotic prediction intervals by the MLE-method is not applicable as that method requires access to the past observed targets and forgetting-based method do not store any historic data. This is why, the previously mentioned ad-hoc prediction bounds idea is proposed to make prediction bound estimations without needing the historic data. The idea is maintaining three versions of the online learner one of which is called \textit{base} learner and the other two are upper bound learner and lower bound learner. The learner system consists of these three learners is called \textit{3-learner ensemble}. The point predictions made by the base learner is returned as the point prediction for a data point for which the target prediction is requested. As for bound learners, the point prediction of the upper bound learner constitutes the upper bound of the prediction. Likewise, the point prediction by the lower bound learner constitutes the lower bound of the prediction. In order to make this prediction scenario possible, these three learners should be updated with different set of data points after a predefined \textit{burn-in} time. After that, the upper bound learner is updated with the data points with observed targets that are higher than the the lower bound prediction made made by the 3-learner ensemble. Similarly, the lower bound learner is updated with the data points with observed targets that are lower than the the upper bound prediction made by the 3-learner ensemble. This idea is sketched in \ref{fig:ad-hoc_prediction_bounds}.

\begin{figure}[htbp]
  \centering
    \includegraphics[width=\linewidth]{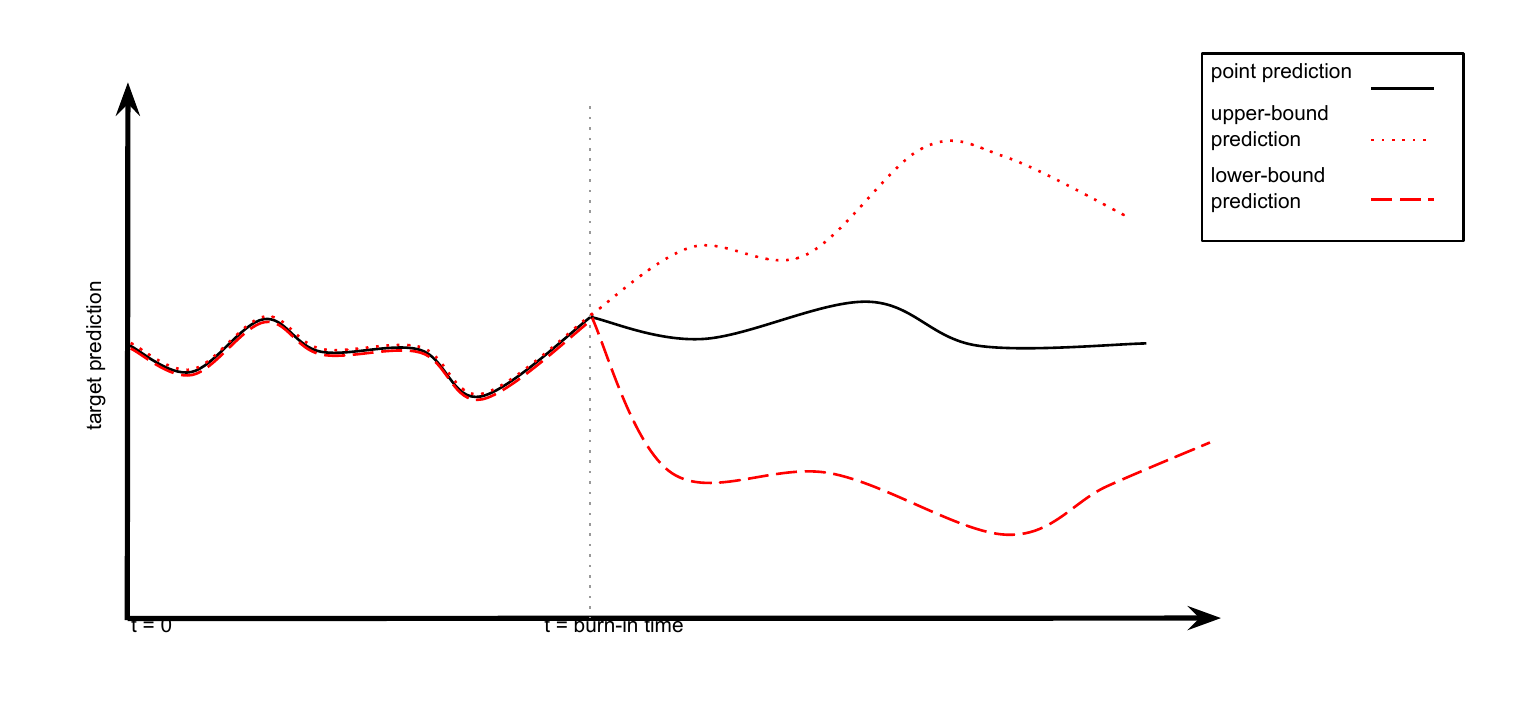}
  \caption{Ad-Hoc prediction bounds method is illustrated}
  \label{fig:ad-hoc_prediction_bounds}
\end{figure}

\subsubsection{Implementation of \texttt{update} for \texttt{BayesianMLEForgetting}}
\label{subsubsection_impl_update_bmleforgetting}

Implementation of \texttt{update} operation depends on the previously derived recursive version of parameter calculation formula, \ref{def_5.7}. However, the concrete implementation should define an initial matrix for $X_0X_0^{\top}$. \citep[pp. 162-163]{bontempi_statistical_2011} suggests that the initial $X_0X_0^{\top}$ matrix should be in the form of $kI$ where $k$ is some constant. It is also suggested that $k$ should be chosen to be high if the parameters estimated using the recursive formulation is preferred to diverge from the initial conditions quickly. Therefore, the constant $k$ is chosen to be $10000$. It is also worth noting that starting with $kI$ as the initial matrix guarantees that $XX^{\top}$ will be invertible in the following recursive steps.

\subsubsection{Implementation of \texttt{predict} for \texttt{BayesianMLEWindowed}}

Once the parameters $\pmb{w}$ is ready, making a point prediction takes simply multiplying the parameters and the data point whose unobserved target needs to be estimated.

Estimating the prediction bounds using the asymptotic prediction bound estimation method is formulated in \ref{def_4.17}. As previously discussed, \ref{def_4.17} includes the term $s^2$ which is the instantaneous squared residual error. Assuming the instantaneous squared residual error is computed in the previous \textit{update} operation, finding the prediction bounds by evaluating \ref{def_4.17} is straightforward.

\subsubsection{Implementation of \texttt{update} for \texttt{BayesianMLEWindowed}}
\label{subsubsection:update_bmlewindowed}

The operation \texttt{update} for \texttt{BayesianMLEWindowed} involves two different sub-operations. These are updating the parameters and recomputing the instantaneous squared residual error. The first operation is twofold. First, the effect of the oldest data point in the (already full) sliding window should be removed from the parameters. Then, the new data point should be \textit{absorbed} into the parameters. These two sub-operations can be done using rank-1 downdate and rank-1 update of $M_1^{-1}$ matrix respectively. Additionally $M2$ should be downdated and updated accordingly. The derived parameter update formula in \ref{def_5.5} only addresses the \textit{update} part of the operation and it lacks the downdate part. Therefore, the following parameter downdate equation is derived.
\begin{align*}
\pmb{w}_{old} & = (XX^{\top})^{-1}X\pmb{y} = M1_{old}M2_{old} \\
M1_{old}^{-1} & = [\pmb{x}_k \ \pmb{x}_{k+1} \ ... \ \pmb{x}_{k+w-1}][\pmb{x}_k \ \pmb{x}_{k+1} \ ... \ \pmb{x}_{k+w-1}]^{\top} \\
M1_{new}^{-1} & = [\pmb{x}_{k+1} \ ... \ \pmb{x}_{k+w-1}][\pmb{x}_{k+1} \ ... \ \pmb{x}_{k+w-1}]^{\top} \\
& = M1_{old}^{-1} - \pmb{x}_{k}\pmb{x}_{k}^{\top} \text{ rank-1 downdate, analogous to \ref{def_5.2} } \numberthis \label{def_5.8} \\
M1_{new} &= M1_{old} + \frac{M1_{old} \pmb{x}_{k}\pmb{x}_{k}^{\top} M1_{old}}{1 - \pmb{x}_{k}^{\top} M1_{old} \pmb{x}_{k}} \text{ (using matrix inversion lemma \cite{bartlett_inverse_1951} on \ref{def_5.8})} \numberthis \\
M2_{old} & = [\pmb{x}_k \ \pmb{x}_{k+1} \ ... \ \pmb{x}_{k+w-1}][y_k \ y_{k+1} \ ... \ y_{k+w-1}]^{\top} \\
M2_{new} & = [\pmb{x}_{k+1} \ ... \ \pmb{x}_{k+w-1}][y_{k+1} \ ... \ y_{k+w-1}]^{\top} \\
& = M2_{old} - \pmb{x}_k y_k \numberthis \label{def_5.10} \\
\pmb{w}_{new} & = M1_{new}M2_{new} \numberthis \label{def_5.11}
\end{align*}
Above, w denotes the size of the sliding window. For downdating \ref{def_5.11} can be used. Thus, the downdating requires to recursively update two stored matrices namely $M1$ and $M2$. The initialization of $M1$ is done as explained in \ref{subsubsection_impl_update_bmleforgetting}. On the other hand, $M2$ can be initialized as a zero column matrix. It is worth noting that the derived downdate formula can be adapted to rank-1 update scenario for updating as an alternative to the \ref{def_5.5} formula for updating. Two formulas are essentially the same.

Only variable the time spent by update and downdate sub-operations for the matrix operations involved in \ref{def_5.10}} and \ref{def_5.2} depends on is $n$ which is the input space dimensionality. Since the for a given learning problem, $n$ is fixed, the time spent by these operations can be bounded by a constant.

As for the second operation mentioned, recalculation of the instantaneous squared residual error is straightforward. After obtaining the new parameters following the downdate and update mentioned, for each data point stored, a new point prediction is made and the squared difference of each of the predictions from the corresponding observed stored response is summed up. This requires one pass over the stored data points and their corresponding responses. The length of this pass is simply the size of the sliding window which is $w$. Since the sliding window-size is a constant, the time required for the described one-pass routine can be bounded by a constant.

\subsubsection{Implementation of \texttt{tune} for \texttt{BayesianMLEWindowed}}

\texttt{BayesianMLEWindowed} does not depend on a hyperparameter to be tuned. Although the learner inherits all the features of sliding-windowed learners and also follows the state transition diagram of them, its tuning method is an empty method.

\subsection{MAP method operations}

MAP-method implementation is very similar to that of MLE-method. Hence, this section frequently refers to the previous section.

\subsubsection{Implementation of \texttt{predict} for \texttt{BayesianMAPForgetting}}

The \texttt{predict} operation does the same as its \texttt{BayesianMLEForgetting} counterpart for making point predictions. It simply returns the multiplication of the parameter vector that represents the previously calculated parameters and transpose of the matrix that represents the datapoint. Furthermore, for estimating the prediction bounds,  \texttt{BayesianMAPForgetting} uses the ad-hoc prediction bounds estimation method like \texttt{BayesianMLEForgetting} does.

\subsubsection{Implementation of \texttt{update} for \texttt{BayesianMAPForgetting}}

The implementation of the \textit{update} operator of \texttt{BayesianMAPForgetting} is exactly same as that of \texttt{BayesianMLEForgetting}. However, how the matrix $X_0X_0^{\top}$ is initialized is different. Instead of using $kI$ as the initial $X_0X_0^{\top}$ matrix, \texttt{BayesianMAPForgetting} uses $\sigma_y^2 \Sigma_w^{-1}$ as implied by the equation \ref{def_4.19}. This difference, initializing the $XX^{\top}$ with the regularization term does not have any procedural implications. In other words, \texttt{update} operation runs exactly in the same way as it does in \texttt{BayesianMLEForgetting}.

\subsubsection{Implementation of \texttt{predict} for \texttt{BayesianMAPWindowed}}

\texttt{BayesianMAPWindowed} learners make point predictions by calculating the product of parameters and the data point just as any other parametric learner does. As for estimating prediction bounds, the available theoretically sound methods for MAP-method are not practically suitable for the online learning scenario due to their high complexity as discussed in \ref{subsubsection:map_pred_bounds}. As an alternative, the asymptotic prediction bound estimation method for MLE-method is used.

\subsubsection{Implementation of \texttt{update} for \texttt{BayesianMAPWindowed}}

The implementation of \texttt{update} operation for \texttt{BayesianMAPWindowed} is exactly same as the implementation of \texttt{update} operation for \texttt{BayesianMLEWindowed}. Once the initialization of $XX^{\top}$ is done with the regularization term (different than the initialization for \texttt{BayesianMLEWindowed}) as discussed in \ref{subsubsection:update_bmlewindowed}, the entire updating process is the same.

\subsubsection{Implementation of \texttt{tune} for \texttt{BayesianMAPWindowed}}

For \texttt{BayesianMAPWindowed}, there are two hyperparameters to tune. These are $\sigma_y$ and $\Sigma_w$. Thus, tuning routine consists of two steps. First, $\Sigma_w$ is tuned. As discussed in \ref{subsubsection:map-based_param_estimation}, covariance-variance matrix that can be obtained from the stored data points is in fact what $\Sigma_w$ should be. This is because, $\Sigma_w$ is the prior for the parameters and it reflects the belief that how much different parameters change together (covariance) and what their variance is. Thus, setting variance-covariance matrix as the parameters prior, $\Sigma$, is the best tuning option.

As for tuning noise variance hyperparameter, $\sigma_y$, an exhaustive search method is employed. For defined minimum, maximum values and step size for $\sigma_y$, Starting with the minimum as the experimental noise standard deviation value and incrementing it by the step size in every iteration, the residual error the learner \textit{would} made for the data points stored in the sliding window is computed. The experimental noise standard deviation that produced lowest residual error is set as the the noise standard deviation to be used for the \texttt{BayesianMAPWinodwed} learner. 

After finding the optimized parameters, the parameters for the MAP-method should be recomputed from scratch. For this, the right hand side of the equation \ref{def_4.19} is evaluated. In order to speed up the matrix inversions involved in the referred equation, a fast inversion method based on the Cholesky decomposition is used. This method computes the Cholesky decomposition of the matrix to be inverted and produces a lower triangular matrix. It is possible to invert a triangular matrix by using back-substitution method which is much faster than using a standard matrix inversion algorithm. Multiplication of the transpose of the inverted lower triangular matrix and inverted lower triangular matrix gives the inverted matrix needed. One precondition for this method to work is that the matrix to be inverted in the beginning must be positive semi-definite. The variance-covariance matrix and its scalar multiplication with the tuned noise standard deviation parameter are supposed to be positive semi-definite as the parametric models with sliding window never allows a repeated data point in their sliding window by effectively detecting the duplicate data points in the beginning of the update routine and update only the stored target associated with them.

\subsection{Gaussian Process Regression operations}

The following operations discussed depend on the inverse of the kernel matrix for the data points stored in the sliding window. This is a $w$-by-$w$ where $w$ denotes the size of the sliding window. Moreover, for the reasons which becomes clear with the discussion of the \texttt{update} operation below, the kernel matrix itself, another $w$-by-$w$ matrix, should be stored. Furthermore, an array of mean values for the data points in the sliding window for the \texttt{GPRegression} variants that does not use a non-zero mean function is needed to be stored.

\subsubsection{Implementation of \texttt{predict} for \texttt{GPRegression}}

Gaussian Process Regression models the target to be predicted as (single-variate) Gaussian distributed random variable with the mean and variance specified in \ref{def_4.25}. Once the required terms namely the new data point, $\pmb{x}_{n+1}$; design matrix, X; observed targets, $\pmb{y}$; means calculated for past data points, $\pmb{M}$ are plugged into \ref{def_4.25}, the point prediction and prediction variance that could be used for finding prediction bounds by simply multiplying the square root of it with the z-value that corresponds to the confidence level desired. However, in order to have a concrete implementation of this logic, mean and covariance functions have to be defined. 

The covariance function choice is already discussed in \ref{subsection:gpreg}. As for the mean function, it is rather secondary to the nature of making predictions by Gaussian Processes. Mean function is only used to fit the data roughly with a relatively simply predictive model and model the residuals with Gaussian Process. Popular choices for mean function are zero mean, average mean and ordinary-least-squares mean. Using a zero mean function obviously means not employing a mean function and for the predictions in this particular case \ref{def_4.26} can be used where neither the mean function nor the past mean values vector appear. Average mean function is also a simple choice like zero mean function. Average mean function, as its name suggests, returns the average of the observed targets until the point where a new prediction is requested making its return value independent of its input just like zero mean function. A more complex option for calculating means for data points is fitting a linear model for the data using MLE-method described. This surely adds to the complexity of the \texttt{predict} and \texttt{update} routines. Nevertheless, as discussed in MLE-method implementation, its prediction cost is constant. Evaluating \ref{def_4.25} also requires constant time assuming that the inverse of the kernel matrix, $K(X,X)^{-1}$, is precomputed. Next section which discuses the implementation of \texttt{update} operator describes how the inverse of the kernel matrix is maintained.

\subsubsection{Implementation of \texttt{update} for \texttt{GPRegression}}

The \texttt{update} operation for \texttt{GPRegression} involves two sub-operations. First sub-operation is to update the inverse of the kernel matrix. This sub-operation is twofold. First, the effect of the dropped data point from the sliding-window should be removed from the inverse of the kernel matrix, then the effect of the the new data point should be added.

Removing the oldest data point from the kernel matrix can straightforwardly done by removing the first row and first column. However, how this affects the inverse of the kernel matrix is not as trivial to imagine. \citep[p. 792]{van_vaerenbergh_sliding-window_2006} presents the formulations by which the inverse of the kernel matrix with removed first column and first row.
\begin{align*}
& \text{let matrix $K$ partitioned as } \begin{bmatrix} a & \pmb{b}^{\top} \\ \pmb{b} & K_{new}  \end{bmatrix} \\
& \text{and matrix $K^{-1}$ partitioned as } \begin{bmatrix} e & \pmb{f}^{\top} \\ \pmb{f} & G \end{bmatrix} \\
& \qquad \qquad  K_{new}^{-1} =  G - \frac{\pmb{f}\pmb{f}^{\top}}{e} \numberthis \label{def_5.12} \\
& \ \text{ satisfying; } \\ 
& \pmb{b}e + K_{new}\pmb{f} = 0 \qquad \pmb{b}\pmb{f}^{\top} + K_{new}G = \pmb{I}
\end{align*}
Similarly, adding the new data point to the kernel matrix can be easily done by inserting a last column and a last row. How this changes the inverse of the kernel matrix is formulated in the following.
\begin{align*}
& \text{let matrix $K_{new}$ partitioned as } \begin{bmatrix} K & \pmb{b} \\ \pmb{b}^{\top} & k(\pmb{x}_{new},\pmb{x}_{new}) \end{bmatrix} \\
& \text{and matrix $K_{new}^{-1}$ partitioned as } \begin{bmatrix} E & \pmb{f} \\ \pmb{f}^{\top} & g \end{bmatrix} \\
& \qquad \qquad  E = K^{-1}(\pmb{I}+\pmb{b}\pmb{b}^{\top}K^{-1H}g) \label{def_5.13} \numberthis \\
& \qquad \qquad  \pmb{f} =  -K^{-1}\pmb{b}g \label{def_5.14} \numberthis \\
& \qquad \qquad  g = (k(\pmb{x}_{new},\pmb{x}_{new}) - \pmb{b}^{\top}A^{-1}\pmb{b})^{-1} \label{def_5.15} \numberthis \\
& \ \text{ satisfying; } \\ 
& KE + \pmb{f}^{\top} = \pmb{I} \qquad K\pmb{f} + \pmb{b}g  = \pmb{0} \qquad \pmb{b}^{\top}\pmb{f} + k(\pmb{x}_{new},\pmb{x}_{new})g = 1
\end{align*}
Applying \ref{def_5.13}, \ref{def_5.14} and \ref{def_5.15}, the needed inverse of the kernel matrix is obtained. What is common in equation \ref{def_5.12} that calculates the inverted kernel matrix when the oldest data point is dropped from the sliding window and \ref{def_5.13}, \ref{def_5.14} and \ref{def_5.15} equations together that calculate the inverted kernel matrix when a new data point is added to the sliding window is that both set of formulations are recursive. In other words, they use the the \textit{old} kernel matrix and its inverse to find the updated inverse of it. Thanks to this recursive nature of the update equations, inverse of the matrix do not have to be recomputed meaning the update mechanism devised is truly \textit{incremental}.

Two parts of the first sub-operation requires several matrix multiplications. Time complexity of the implemented naive matrix multiplication algorithm is $\mathcal{O}(n^3)$. Since the matrix sizes are determined by the sliding-window size, w, which is a constant. Therefore, it can be argued that the time first update sub-operation spends can always bounded by a constant.

The second sub-operation of \text{update} is computing the mean value for the new data point and adding it to the tail of the mean values vector while dropping the mean value that is the mean of the oldest data point which is dropped from the sliding window. For the mean values vector, similarly to the sliding window that accommodates the data points and their corresponding targets, a ring buffer is used. Computing the mean value and adding it to the tail of the ring buffer are both constant-time spending operations. 

\subsubsection{Implementation of \texttt{tune} for \texttt{GPRegression}}
\label{subsubsection:impl_tune_gpreg}

For Gaussian Process Regression, there are $k+2$ hyperparameters to tune ($k$ is the number of input dimensions). For a detailed discussion on the hyperparameters, their derivatives, the \textit{proxy} trick to keep them positive in a gradient-based optimization process, see the subsection \ref{subsubsection:gpreg_opt_hyperparams} in previous chapter.

In this chapter, the optimization method employed to tune $k+2$ is presented. In order to deal with the non-convexity of the optimization problem, a hybrid optimization routine that includes both random search ideas and gradient-based log likelihood optimization method is implemented. The implementation is presented in the pseudocode given in \ref{alg:gpreg_hyperparam_tuning}. Briefly, the tuning algorithm randomly picks the hyperparameters. This corresponds to a random point in the hyperparameter search space. For this point, it computes the gradient vector for hyperparameters and moves in the opposite direction of it (search line). It adjusts the step size to take take a step that improves the log likelihood score along the search line. Once, by moving along the search line finding a point with a higher log likelihood score, it computes the new gradient vector and repeats the previous steps. If no good step size is found that takes the random point to a point producing higher log likelihood, hyperparameters are randomized again. This continues until a point is found with a very low (approximately 0) gradient values are found or maximum number of iterations set is reached. In the latter case, algorithm terminates and in the case last visited point has a lower log likelihood than the initial point, the initial hyperparameter settings are recovered.

Tuning algorithm described is implemented by a fairly heavy routine. It involves many costly matrix operations such as matrix multiplications, matrix inversions that involve back-substitutions. It also traverses the arrays which matrices are stored many times for finding derivate kernel matrices and for assignments to the temporary test matrices. However, most costly operation among these is matrix multiplication\footnote{Note that the fast-matrix inversion based on Cholesky decomposition also involves a matrix multiplication in addition to back substitution}. Hence, for a descent estimation of the runtime, only the number of matrix multiplications can be counted. Line number 2 is always executed once and involves 2 matrix multiplications, line number 17 which involves 1 matrix multiplication and line number 26 which involves 15 matrix multiplications are executed as many as the \texttt{maximum\_iteration\_count} set at the worst case. Line number 32 which involves 1 matrix multiplication is as executed as many as the \texttt{maximum\_iteration\_count} times \texttt{maximum\_decay\_count}. Thus, at the worst-case scenario, $2+16\times\texttt{maximum\_iteration\_count} + \texttt{maximum\_iteration\_count}\times\texttt{maximum\_decay\_count}$ matrix multiplications take place. For the implementation, 50 for \texttt{maximum\_iteration\_count} and 10 for \texttt{maximum\_decay\_count} is chosen. Thus, the asymptotic time complexity of the tuning algorithm is $\mathcal{O}((2+ 50\times16 + 50\times10)\times n^3) = \mathcal{O}(n^3)$ where n denotes the dimensions of the square matrices which is the size of sliding-window $w$. Since this number is a constant, for a given sliding-window, the time tuning routine spends can be bounded by a constant. However, this does not mean that it is a fast operation. It is only safe to say the tuning cost will not increase as more data points are streamed from the data stream.
\begin{algorithm}
  \caption{GPRegression Hyperparameter Tuning}\label{alg:gpreg_hyperparam_tuning}
  \begin{algorithmic}[1]
    \Procedure{Predict}{\texttt{proxy\_hyperparams}, $K$, $K^{-1}$}
    	\State $\texttt{init\_proxy\_hyperparams} \gets \texttt{proxy\_hyperparams}$
    	\State $\texttt{init\_loglhood} \gets \textsc{calc\_loglhood}(K,K^{-1})$
    	\State $\texttt{target\_loglhood} \gets -\infty$
    	\State $\texttt{iteration\_count} \gets 0$
    	\While {\texttt{true}}
    		\If {$\texttt{target\_loglhood} < \texttt{cur\_loglhood}$}
    			\If {$\texttt{iteration\_count > max\_iteration\_count}$}
    				\If {$\texttt{cur\_loglhood} < \texttt{init\_loglhood}$}
    					\State $\texttt{proxy\_hyperparams} \gets \texttt{init\_proxy\_hyperparams}$
    					\State $K \gets \textsc{compute\_kernel\_matrix}(\texttt{proxy\_hyperparams})$
 	 		  			\State $K^{-1} \gets \textsc{fast\_invert\_psd\_matrix}(K)$
 	 		  		\EndIf
    				\State $\textbf{return}$
    			\EndIf
  				\State $\texttt{proxy\_hyperparams} \gets rand()$
  				\State $K \gets \textsc{compute\_kernel\_matrix}(\texttt{proxy\_hyperparams})$
    			\State $K^{-1} \gets \textsc{fast\_invert\_psd\_matrix}(K)$
    			\State $\texttt{cur\_loglhood} \gets \textsc{calc\_loglhood}(K,K^{-1})$
    		\EndIf
    		\If {$\texttt{target\_loglhood} \geq \texttt{cur\_loglhood}$}
    			\State $\texttt{init\_loglhood} \gets \texttt{target\_loglhood}$
    			\If {$\texttt{gradients} < \texttt{approximate\_local\_optima\_gradients}$}
    				\State \textbf{return}		
    			\EndIf
    		\EndIf
    		\State $\texttt{gradients} \gets \textsc{compute\_gradients}(\texttt{proxy\_hyperparams},K,K^{-1})$
    		\State $\texttt{step\_size} \gets \texttt{max\_step\_size}$
    		\State $\texttt{decay\_count} \gets 0$
    		\While {\texttt{true}}
    			\State $\texttt{ test\_proxy\_hyperparams} \gets \texttt{proxy\_hyperparams} - \texttt{step\_size*gradients}$
    			\State $K_{test} \gets \textsc{compute\_kernel\_matrix}(\texttt{test\_proxy\_hyperparams})$
    			\State $K_{test}^{-1} \gets \textsc{fast\_invert\_psd\_matrix}(K_{test})$
    			\State $\texttt{target\_loglhood} \gets \textsc{calc\_loglhood}(K_{test},K^{-1}_{test})$
    			\If {$\texttt{target\_loglhood} < \texttt{cur\_loglhood}$}
    				\If {$\texttt{decay\_count > max\_decay\_count}$}
    					\State $\textbf{return}$
    				\EndIf
    				\State $\texttt{step\_size} \gets \texttt{step\_size*decayer}$
    				\State $\texttt{decay\_count} \gets \texttt{decay\_count}+1$
    			\EndIf
    			\If {$\texttt{target\_loglhood} \geq \texttt{cur\_loglhood}$}
    				\State $K \gets K_{test}$
    				\State $K^{-1} \gets K_{test}^{-1}$
    				\State \textbf{break}
    			\EndIf
    		\EndWhile	
    		\State $\texttt{iteration\_count} \gets \texttt{iteration\_count}+1$
    	\EndWhile
    	\State \textbf{return}
    \EndProcedure
  \end{algorithmic}
\end{algorithm}

\subsection{Kernel Regression operations}

For its operations, Kernel Regression algorithm stores two sliding-windows in addition to the inherited sliding window that stores the recent data points and their corresponding responses. First one, called \texttt{density\_estimates}, keeps the estimated density in the input space for the data points in the sliding window. Second one, called \texttt{contributions}, maintains, for each data point in the sliding window, the contributed total weighted observed responses from all the data points in the sliding window. Note that, for a data point, the values kept in these two collections correspond to the nominator and denominator of the prediction formula given in \ref{def_4.42}. Moreover, \texttt{KernelRegression} learners maintain ASE (Average Squared Error) score of their predictions and the inverse of the lengthscales matrix, $H^{-1}$.

\subsubsection{Implementation of \texttt{predict} for \texttt{KernelRegression}}

In order to make a prediction for a new data point, the density estimate for it and the sum of the weighted contributions of the data points in the sliding window should be computed. These can be computed using the prediction formula derived in \ref{def_4.42}. This computation requires one pass over the sliding window elements. Thus the time the \texttt{update} operator consumes is proportional to the size of the sliding-window.

As for the prediction bounds estimations, the formula \ref{def_4.43} can be evaluated directly as all the terms  \ref{def_4.43} is known prior to estimation of the prediction bounds.

\subsubsection{Implementation of \texttt{update} for \texttt{KernelRegression}}

Implementation of the \texttt{update} operation involves 2 loops over the elements of the \texttt{density\_estimates} and \texttt{contributions} collections each. The loops over the \texttt{density\_estimates} is done in order to discount the contribution to the density estimates of the data points in the sliding window by the dropped data point and add the contribution to the density estimates by the new data point. The loops over the \texttt{contributions} is to remove the weighted contribution of the dropped data point from the sum of weighted target contributions of the data points in the sliding window and to add the weighted contribution of the response of the new data point to the data points in the sliding window. Apart from the mentioned loops, the update routine also needs to update ASE with the new data point and its prediction. This is easy as the prediction previously made by the learner is provided back to it with the invocation of the \texttt{update} operation following the \texttt{predict} along with the observed response. The pseudocode for \texttt{update} operation is presented in \ref{alg:kreg_update_routine}.

\begin{algorithm}
  \caption{\texttt{KernelRegression} Update}\label{alg:kreg_update_routine}
  \begin{algorithmic}[1]
    \Procedure{Update}{\texttt{new\_dp}, \texttt{observed\_target}, \texttt{predicted\_target}}
    	\State $\texttt{dropped} \gets \textsc{poll\_tail}(sliding\_window)$
    	\State $\texttt{dropped\_dp} \gets \textsc{get\_dp}(\texttt{dropped})$
    	\State $\texttt{dropped\_target} \gets \textsc{get\_target}(\texttt{dropped})$
    	\While {for (\texttt{each, each\_density, each\_contribution}) in (\texttt{sliding\_window},\texttt{density\_estimates}, \texttt{contributions})}
    		\State $\texttt{dp} \gets \textsc{get\_dp}(\texttt{each})$
    		\State $\texttt{kernel\_measurement} \gets \textsc{kernel\_func}((dropped\_dp-dp)H^{-1})$
    		\State $\texttt{each\_density} \gets \texttt{each\_density} - \texttt{kernel\_measurement}$
    		\State $\texttt{each\_contribution} \gets \texttt{each\_contribution} - \texttt{kernel\_measurement}\times\texttt{dropped\_target}$
    	\EndWhile
  		\State $\texttt{new\_dp\_density} \gets 0$
  		\State $\texttt{new\_dp\_contribution} \gets 0$
    	\While {for (\texttt{each, each\_density, each\_contribution}) in (\texttt{sliding\_window},\texttt{density\_estimates}, \texttt{contributions})}
    		\State $\texttt{target} \gets \textsc{get\_target}(\texttt{each})$
    		\State $\texttt{dp} \gets \textsc{get\_dp}(\texttt{each})$
    		\State $\texttt{kernel\_measurement} \gets \textsc{kernel\_func}((new\_dp-dp)H^{-1})$
    		\State $\texttt{each\_density} \gets \texttt{each\_density} + \texttt{kernel\_measurement}$
    		\State $\texttt{each\_contribution} \gets \texttt{each\_contribution} + \texttt{kernel\_measurement}\times\texttt{observed\_target}$
    		\State $\texttt{new\_dp\_density} \gets \texttt{new\_dp\_density} + \texttt{kernel\_measurement}$
    		\State $\texttt{new\_dp\_contribution} \gets \texttt{new\_dp\_contribution} + \texttt{kernel\_measurement}\times\texttt{target}$
    	\EndWhile
    	
    	\State $\texttt{kernel\_measurement} \gets \textsc{kernel\_func}((new\_dp-new\_dp)=\pmb{0})$
    	\State $\texttt{new\_dp\_density} \gets \texttt{new\_dp\_density} + \texttt{kernel\_measurement}$
    	\State $\texttt{new\_dp\_contribution} \gets \texttt{new\_dp\_contribution} + \texttt{kernel\_measurement}\times\texttt{observed\_target}$
    	\State $\textsc{add\_to\_head}(\texttt{sliding\_window}, (\texttt{new\_dp}, \texttt{observed\_y}))$
		\State $\textsc{add\_to\_head}(\texttt{density\_estimates}, \texttt{new\_dp\_density})$
		\State $\textsc{add\_to\_head}(\texttt{contributions}, \texttt{new\_dp\_contibution})$
		\State $\textsc{update\_ase}(\texttt{observed\_target}, \texttt{predicted\_target})$
   \EndProcedure
  \end{algorithmic}
\end{algorithm}

The heaviest operation that occur within the loops are the computation of the kernel measurement between data points. The kernel measurement for a pair of points involves multiplication of the difference of two input points (as vectors) with the inverse of the lengthscales matrix. Inverse of the lengthscales matrix is stored. Thus, there is no need to take the inverse of it. However, the mentioned multiplication is necessary. It is repeated for every data point in the sliding window twice (once for computing the kernel measure between the dropped data point and sliding-window data points and once for computing the kernel measure between the new data point and sliding-window data points). Since the sliding-window is a fixed-size collection, time required for the update operation can be bounded by a constant.

\subsubsection{Implementation of \texttt{tune} for \texttt{KernelRegression}}
\label{subsubsection:impl_tune_kreg}

\ref{subsubsection:kreg_tuning_approach} already covers the details of the method employed for finding a nearly optimal lengthscales matrix, \texttt{H}. In this section only the pseudocode that provides a high-level overview of the implementation is presented.

As shown in \ref{alg:kreg_tune_routine}, within the tuning routine, slave functions such as \textsc{get\_hold\_out\_one\_ase} and \textsc{get\_hold\_out\_one\_estimate} are implemented. In this hierarchical function calling design, the main routine, \textsc{tune}, calls \textsc{get\_hold\_out\_one\_ase} as many times as there are steps defined between $\alpha_{min}$ and $\alpha_{max}$. In the implementation, $\alpha_{min}$, $\alpha_{max}$ and $\alpha_{step}$ is defined as $0.05$, $2.00$ and $0.01$ respectively. Thus, for the tuning operation, \textsc{get\_hold\_out\_one\_ase} routine gets called $195$ times. Furthermore, \textsc{get\_hold\_out\_one\_ase} calls \textsc{get\_hold\_out\_one\_estimate} $w$ times. Each call to \textsc{get\_hold\_out\_one\_estimate} take same time as a single \texttt{tune} operation does. It calls the routine that calculates the kernel measurement for a pair of input vectors, $w-1$ times. Kernel measurement itself and the multiplication of the subtraction of two input vector by the inverse of the lengthscales matrix takes a constant amount of time considering that input width (dimensionality of input space is constant during the runtime of the learner). This constant-time consuming operation is called $195\times w \times (w-1)$ times. Since sliding-window has a fixed-length, the time \texttt{tune} operation consumes is fixed and does not grow as more data points are streamed from the data stream. However, obviously, it is much more costly operation than the \texttt{predict} operation.

\begin{algorithm}
  \caption{\texttt{KernelRegression} Hyperparameter Tuning}\label{alg:kreg_tune_routine}
  \begin{algorithmic}[1]
	\Procedure{Tune}{$X$, $H^{-1}$}
		\State $\texttt{COV} \gets \textsc{get\_var\_cov\_matrix(X)}$
		\State $\texttt{target\_ase} \gets \textsc{get\_hold\_out\_one\_ase}(H^{-1})$
		\State $\alpha_{current} \gets \alpha_{min}$
		\While {$\alpha_{current} < \alpha_{max}$}
			\State $H_{experimental}^{-1} \gets \textsc{fast\_invert\_psd\_matrix}(\alpha_{current}\texttt{COV})$
			\State $\texttt{current\_ase} \gets \textsc{get\_hold\_out\_one\_ase}(H_{experimental}^{-1})$
			\If {${\texttt{current\_ase} < \texttt{target\_ase}}$}
				\State $H^{-1} \gets H_{experimental}^{-1}$
				\State $\texttt{target\_ase} \gets \texttt{current\_ase}$
			\EndIf
			\State $\alpha_{current} \gets \alpha_{current} + \alpha_{step}$
		\EndWhile
	\EndProcedure	
  \end{algorithmic}
  \begin{algorithmic}[1]
	\Procedure{get\_hold\_out\_one\_ase}{$H^{-1}$}
		\State $\texttt{squared\_error} \gets 0$
		\While {for \texttt{each} in \texttt{sliding\_window}}
			\State $\texttt{target} \gets \textsc{get\_target}(\texttt{each})$
  			\State $\texttt{dp} \gets \textsc{get\_dp}(\texttt{each})$
    		\State $\texttt{target\_estimate} \gets \textsc{hold\_out\_one\_estimate}(\texttt{dp}, H^{-1})$
    		\State $\texttt{squared\_error} \gets \texttt{squared\_error} + (\texttt{target\_estimate}-\texttt{target})^2$
    	\EndWhile
    \State \textbf{return} \texttt{squared\_error/w}
	\EndProcedure	
   \end{algorithmic}
   \begin{algorithmic}[1]
	\Procedure{get\_hold\_out\_one\_estimate}{$\texttt{dp\_to\_be\_held\_out}, H^{-1}$}
	\State $\texttt{density\_estimate} \gets 0$
	\State $\texttt{weighted\_target\_contribution} \gets 0$
		\While {for \texttt{each} in \texttt{sliding\_window}}
			\State $\texttt{dp} \gets \textsc{get\_dp}(\texttt{each})$
			\If {$\texttt{dp}\neq \texttt{dp\_to\_be\_held\_out}$}
				\State $\texttt{target} \gets \textsc{get\_target}(\texttt{each})$
    			\State $\texttt{kernel\_measurement} \gets \textsc{kernel\_func}((dp\_to\_be\_held\_out-dp)H^{-1})$
    			\State $\texttt{density\_estimate} \gets \texttt{density\_estimate} + \texttt{kernel\_measurement}$
    			\State $\texttt{weighted\_target\_contribution} \gets \texttt{weighted\_target\_contribution} + \texttt{kernel\_measurement}\times\texttt{target}$
    		\EndIf
		\EndWhile
	\State \textbf{return} \texttt{weighted\_target\_contribution/density\_estimate}
	\EndProcedure
	\end{algorithmic}
\end{algorithm}


\chapter{Evaluation} 

\label{Chapter6} 

\lhead{Chapter 6. \emph{Evaluation}} 

Model Evaluation in Machine Learning is used in two different context. First, \textit{within} a learning algorithm while searching for a good hypothesis (e.g risk minimization\footnote{optimization over $\alpha$ discussed in Chapter \ref{Chapter2}.}, hypothesis assessment), an evaluation mechanism is needed. Second, when assessing the applicability of a particular learning algorithm to a particular learning problem, again an evaluation mechanism is needed \cite{gama_evaluating_2013}. Evaluation in it first meaning is already discussed in \ref{Chapter2}.

This chapter covers the evaluation of the online learning methods that were found to be theoretically suitable for the operator runtime estimation problem through the experiments conducted on the synthetic data sets and real runtime data sets obtained from Ocelot instances installed on various platforms with different hardware. 

\section{Evaluation Methodology}

The main concern of model evaluation when testing the applicability of a learner to a problem is measuring the \textit{generalization power} of the learner. \citep[p. 320]{gama_evaluating_2013} describes the generalization power as the effectiveness of a predictive model to capture the true underlying concept. As argued in \cite{gama_evaluating_2013} and \cite{gama_issues_2009}, traditional techniques used for measuring the generalization power of batch learners are not always suitable for the stream learning algorithms. For instance, cross-validation variants such as hold-out validation and bootstrapping require several passes over the data. Therefore, they are not feasible when the dataset is potentially infinitely big which is the fundamental assumption in stream learning scenarios. Another problem with the cross-validation techniques when employed in a stream learning setting is that they are oblivious to the ordering of the data points \cite[p. 128]{ikonomovska_regression_2009}. The order of the data items does not matter in the case of non-drifting streams since every data point in the stream would have been generated through the same process with fixed characteristics. However, it is not allowed to make non-drifting concept assumptions in stream learning. The trouble when an evaluation strategy which is insensitive to input ordering is that it does not consider different concepts by potentially shuffling the data of different concepts into the same holdout test sample. Furthermore, it disregards the evolution of the predictive model as different concepts occur in an order and instead evaluates the final\footnote{The adjective final is used in the sense that the current model at the time the model evaluation is requested while the stream keeps running.} model that would have been constructed with the all past data points except for the holdout sample.

Proposed in \cite{dawid_present_1984}, \textit{predictive sequential} or \textit{prequential} approach provides input-ordering sensitive evaluation method which simply computes the cumulative sum of the loss function and divides it by the number of stream items predicted until the point the model evaluation is requested. 
\begin{flalign} 
& S = \sum_{i=1}^{n}L^*(y_i,\hat{y}_i) \label{def_6.1}
\end{flalign}
In above formula, $n$ is the number of data points appeared in the data stream until the evaluation, $\hat{y}_i$ and $y_i$ are the observed target value and the predicted target respectively for the $i_{th}$ data point in the stream. $S$ denotes the total accumulated loss  \footnote{Loss here refers to the error computed by the loss function $L^*$.} the cumulative mean loss is calculated as $M=\frac{S}{n}$

Prequential approach is well-suited for stream learning and the use of it for stream learning algorithms are encouraged by previous research (\cite{ikonomovska_regression_2009}, \cite{vovk_-line_2009}, \cite{gama_issues_2009}, \cite{gama_evaluating_2013}). Its feasibility stems from how it computes the loss term. It computes the loss \textit{on the fly} using all the data points from the start. Moreover, unlike the validation techniques, instead of separate procedures to come up with validation and training sets and retrained versions of the online learner being evaluated, it only uses the prediction of the target value and the target value itself to accumulate loss terms. That is why, computation of the generalization error this way is as easy as adding a line of code in the prediction loop of the stream learning algorithms (including online learning) as shown in the pseudocode \ref{alg:opp}. 
%
%

One might possibly question the validity of the error estimation by the prequential method as it does not explicitly employ a mechanism to avoid the optimistic \textit{bias} in the estimation of the error on contrary to unbiased cross-validation error estimate which uses a validation set. However, when the online prediction protocol discussed in \ref{Chapter2} is carefully examined, it is easy to see that testing is always done on \textit{unseen} data points hence there is no need for a validation set \citep[p. 111]{lemaire_survey_2015}.

Prequential approach also has desirable asymptotic properties. Authors of \citep[p. 331]{gama_issues_2009} argue that for any loss function $L^*$ (used in 6.1), the calculated cumulative mean loss $M=\frac{S}{n}$ can be bounded by an application of Chernoff bound. Letting $M\pm \epsilon$ be the probability of error, we have the following: 
\begin{align*} 
& \epsilon_c = \sqrt{\frac{3\times{ln(2/\delta)}\hat{\mu}}{n}} \\
& \underset{n\rightarrow \infty} \lim {\epsilon_c} = 0
\end{align*} 
Above  $\delta$ denotes the user-defined confidence level. Highlighted in the formula, what is computed by the prequential method converges to the possibility of predictive error while the error term in the estimation of predictive error converges to $0$. 

Another asymptotic feature of the prequential approach (that also holds for validation techniques) is its convergence to Bayes error\footnote{Bayes error is the actual risk in Risk Minimization framework which is discussed in Chapter \ref{Chapter2}.}. In \citep[pp. 323-324]{gama_evaluating_2013}, it is proved that the error in the measured error of a consistent learning algorithm by both holdout validation error and prequential error asymptotically converges to Bayes error.

An obvious problem with prequential approach is that it is a pessimistic error estimator. \cite{ikonomovska_regression_2009}. Due to the accumulation of loss terms from the very beginning of the stream, it considers the entire history of the predictions made by the learner under the test instead of assessing the learner's current error potential. This means prequential approach disregards the evolution of a learner which can be serious problem in the case of drifting streams where the evolution of the predictive models are necessary and expected. In order to overcome this problem with the prequential approach, slightly modified versions of it are proposed in \cite{gama_issues_2009}, \cite{gama_evaluating_2013}.

One of the extensions made to the prequential method is using an error discount factor to discount the past loss terms accumulated in the sum $S$ in the equation \ref{def_6.1}. This way, for the calculation of total loss, the loss in the recent predictions weighs more than the loss in older ones. This idea can be formulated by the following recurrence relations.
\begin{flalign} 
& S_{\delta}(i) = L^*(y_i,\hat{y_i}) + \delta S_{\delta}(i-1) \\
& N_{\delta}(i) = 1 + \delta N_{\delta}(i-1)
\end{flalign}
Above $\delta$ is the error discount factor which is a constant smaller than $1$. When $\delta$ is equal to 1, the fading vanishes and the evaluation method reverts to the base prequential approach. Obviously, with a discount factor close to $1$, error discount factor extension of the prequential method forgets the past error terms less and with a low discount factor, it forgets them very quickly. In the latter case, the most recent data is taken more into the consideration when estimating the predictive error. An advantage of using error discount factor in the prequential evaluation is that it is the \textit{minimal} extension over the original method and it is a memory-less approach without requiring any extra memory for its calculations. As it is discussed later in this section memory consumption is an important consideration in stream learning.

Another proposed extension to the prequential method for evaluating learning algorithms that learn from non-stationary data is keeping a sliding error window where a number of most recent losses are stored and aggregating the values in the window to calculate the mean loss. Similarly to the fading factor extension, this extension has also a parameter to be set manually: window size. The effect of the window size on the estimated predictive error is intuitive. Larger the window size is the more of the recent loss terms are considered. However, this has a deeper implication in the case of evaluation of learners because of the concept drifts. With smaller sliding window, sliding error window extension of the prequential method responds faster to the predictive model evaluation when a concept drift is encountered. This is because, with a smaller sliding-window, it quicker forgets the high loss terms occur while the predictive model is adapting itself to the new concept. On the other hand, when it is equipped with a big window size, it has the advantage of having low-variance in its predictive error estimations. Therefore, it is more stable when evaluating the learning algorithm while it is consuming stationary data from a stream.

Sliding error window extension of the prequential evaluation contrasts the error discount factor extension of it in the way it forgets the past loss terms. With sliding error window as many most recent loss terms as the size of the window is taken into account equally in the calculation of the predictive error and the other older loss terms are ignored altogether whereas in the fading factor method, the forgetting is gradual. All the error terms are always considered in the calculation of predictive error but the contribution of each loss term are different. Older a loss term is, less it contributes to the error prediction sum. The discount of the most recent $i_{th}$ term is $\delta^(i-1)$. As the discount term is computed via an exponential function, the loss terms older than some point in history makes a negligible effect in the total sum.

The extensions over the prequential models are nice ways to solve the problem of evaluation with a non-stationary data. Furthermore, asymptotically they are shown to have the same desirable properties as the prequential approach \cite{gama_evaluating_2013}. However, they bring about the problem of choosing parameters such as the error discount factor or window size. In the same paper in which these extensions are discussed thoroughly, some (somewhat less formal) ideas to tune the parameters of them are presented. In the case of error discount factor extension, it is argued that having known the maximum error in the predictive error measurement and the number of most recent loss terms to be considered, one can find the optimal error discount factor by using the equality $\delta = e^{\frac{ln(\epsilon)}{i}}$ where $i$ is the number of recent loss terms that affect the predictive error estimation and $\epsilon$ is the maximum limit by which the error in the error prediction is bounded. It is worth noting that while it is relatively easy to set a certain maximum error bound, it is hard to come up with the number of most recent loss terms to be taken into account for the error calculation. As for the sliding error window extension, the parameter tuning is even more tricky because it requires to know the variance of the error in the calculation of predictive error before starting the evaluation. In \cite{gama_evaluating_2013}, the authors assumed a binomial distribution for the error and since they were evaluating a binary classifier with $0-1$ loss function, they were able to assume a reasonably good variance value for the error. However, since when evaluating learning algorithms, the variance of error in the predictive error is supposed to be proportional to the target variance which is subject to changes as the concepts drift, tuning the parameters of the evaluation method prior to the learning is not feasible.

As discussed above, extensions of prequential evaluation method needs tuning and although there are sound methods to do that, they are not feasible when there is too little prior information about the data in the stream to learn from. This is exactly the case with the problem dealt with in the scope of this thesis. Any presumptions on the stream data except for its input dimensions should not be made. This makes the task of tuning of the evaluation method parameters harder. However, since these parameters need to be tuned only for the sake of evaluation (they do not have anything to do with training or updating the learning algorithms), coming up with the most optimal parameters is not critically important making it a secondary issue. Therefore, the parameters window-size for the sliding error window extension of prequential method and the error discount factor for the error discount factor extension are regarded as \textit{resolution} parameters for plotting accuracy comparison graphs and usually chosen to be the values that produces the most interpretable plots helping understand the behavior of the learners in the case of the stationary or non-stationary stream data.

Having compared and contrasted different kinds of methods for evaluating the error of the online learning algorithms that are used for the function estimation problem considered in the scope of this thesis, the base prequential method for averaging the error estimation of a learner over a large number of tests is employed. Furthermore, for side-by-side comparison of two learners both prequential evaluation extensions were available as feasible options. As these evaluation methods are learner-type-oblivious, the choice for their resolution parameters (either the window size or the error discount factor) do not necessarily have to inherit the the similarly named parameters of the concept-drift adaptation methods implemented within the corresponding learner. As a result there was a complete freedom to opt in one extension over another as they do not have clear advantages over each other. For the tests, sliding error window is employed for side-by-side accuracy comparisons.

An evaluation subtask which is particularly important for evaluating online learning algorithms used for runtime estimation is the quality of prediction bounds. As discussed in \ref{Chapter3}, not only producing a good point prediction of the runtime but also finding finding good upper and lower bounds for the target variable (which is the runtime measurement) is important for the decision-making layer of the Ocelot to discover good operator algorithm variants for the hardware on which it is running. To this end, evaluation metrics for the prediction bounds that learning algorithm produces per data point is needed. These are discussed in the next subsection. Prequential approach is used for producing the summary statistics regarding prediction bounds quality from the recorded per-data point values using the prediction bound metrics.

Apart from estimating the error and the quality of prediction bounds, there are also other dimensions of evaluation that need to be considered in stream learning even though they are not relevant in batch learning settings. These are the \textit{space} and \textit{learning time} as pointed out in \citep[p. 320]{gama_evaluating_2013}. The reason that makes them relevant to stream learning is the \textit{streaming} nature of stream learning setting. Since an online learning algorithm is a continuously running computer program that maintains an internal state as long as stream items keep arriving, the space its state occupies in the main memory becomes relevant. Even the algorithms with the humblest space complexity can cause troubles when working with streams if the space they occupy is a function of the input size. This is the key difference of stream learning algorithms from their batch counterparts which are static algorithms hence their memory consumption can always be bounded by a constant as the input size is known prior to learning process. A similar observation regarding the relevance of learning time of batch learners and stream learners can be made. The time needed for a static algorithm again can be bounded by a constant as the input size is fixed and known. On the other hand, in the streaming scenario where the learner has to keep up with the rate of data items arriving, how fast it can learn from the current data point before the next data point arrives from the stream is important.

Before discussing different evaluation metrics used to assess the feasibility of the online regression algorithms, it is worth explaining how the chosen evaluation methodology, the base prequential approach and its sliding error window extension, can be used to compute the statistics needed for the assessment.

\subsection{Prequential Approach in Action}
As mentioned in the discussion of the prequential approach and also as seen in the pseudocode \ref{alg:opp} for the Online Prediction Protocol, the statistics being computed is always assumed to be the prediction error-related ones. However, this does not have to be the case. Replacing the line that accumulates the error in the Online Prediction Protocol with piece of code that records other values regarding the current prediction such as prediction time, update time and the change in occupied memory suffices to obtain different statistics than error-related statistic. This method is used for obtaining the statistics for most of the metrics discussed in the following section. However, while for some of them, core prequential approach is used, for the rest the sliding error window extension of it is used. The decision which is a better choice for a certain metric is made by considering the following criteria listed below.

\begin{itemize}
\item Preferred format of the statistic: Whether it is preferred to be a single summary value over the whole history or an array of values with different values accounting for different points in the history of the stream.
\item Sensitivity of the statistic to the conceptual drifts.
\item Necessity of the aggregation of the statistic from different tests.
\end{itemize}

It is worth mentioning the difference between the plain prequential method and its extensions with respect to above criteria. The plain prequential method can be used to obtain evaluation statistics that belong to different points in time in the history of the stream or optionally one single value that summarizes the error potential of the evaluated learning algorithm taking the whole history of the prediction over the stream into account. On the other hand, its sliding error window extension produces array of numbers each of which accounts for a \textit{local} evaluation of the learner under the test at a different point in the history of the stream. This is an important difference as one can, without any trouble, aggregate a single valued summary statistics over different tests on which a particular learner was put while aggregating an array carrying time-sensitive data does not make sense.

In the following subsection, metrics for the previously four dimensions of the model evaluation that are feasible for calculating by using prequential model or its sliding error window extension are listed and described.

\subsection{Error Metrics}

\texttt{RMSE:} Root Mean Squared Error is a standard statistical metric used for measuring the error of predictive models. Despite its sensitivity to outliers in data, it is a preferable choice for the runtime estimation problem because the penalty it imposes on  the predictions demonstrates a quadratic growth as opposed to its alternative Mean Absolute Error (MAE) in which the bad predictions does not get enough penalty in comparison to slightly off predictions. \texttt{RMSE} is formulated below.
$$\texttt{RMSE}=\sqrt{\frac{1}{n}\sum_{i=1}^{n}(\hat{y}\_i - y_i)^2}$$
Above n is the number of target prediction and target pairs. This metric can be used for side-by-side comparisons to compare the prediction behavior of the learning algorithms over the course of the stream. Hence, it is calculated via the sliding window extension of prequential method resulting in an array of numbers allowing to plot a graph for comparison.

\texttt{RMSE\_ST:} When evaluating the online learners featuring sliding windows, the learning algorithm gradually stabilizes as the sliding window gets full. Until the windows is full, bad predictions are expected due to the low number of incremental training the algorithm received. As explained in detail in \ref{Chapter5}, this situation is even worse if the algorithm needs to be tuned which is the case after the first time the sliding windows is full and also the contents of the sliding window is completely recycled after a concept drift. The error is usually high in both of these periods. In smaller windowed algorithms, since the stabilization is faster due to the low size of the sliding window, the effect of this on the total \texttt{RMSE} result is expected to be low while high window-sized algorithms are thought to suffer a bigger drop in their \texttt{RMSE} score due to late stabilization. In order to capture this expected phenomenon in the experiments, \texttt{RMSE\_ST} metric that accounts for only the prediction errors in the \textit{stable} periods of the stream is used along with \texttt{RMSE}. Similar to \texttt{RMSE}, the sliding error window extension of the prequential method is used to plot changing values of this metric of learners over streams.

\texttt{SMSE:} The problem with \texttt{RMSE} is that the final \texttt{RMSE} value calculated is sensitive to the target variance of the responses observed in the data stream. If the targets have high variance, then \texttt{RMSE} tends to be bigger. Provided that the different tests prepared all have the same variance for the target, averaging the \texttt{RMSE} results and compare the averaged \texttt{RMSE}s of different learning algorithms would work fine. However, by simply dividing the \texttt{MSE} result (square of \texttt{RMSE}) by the target variance, abovementioned fixed target variance restriction on the different tests can be lifted. To this end, Standardized Mean Squared Error is used. Averaging the \texttt{SMSE} values from different tests with different target variances is possible. It is worth noting that, a trivial example of regression algorithm that always predicts the target value to be the mean of the all the targets appear in the training set, \texttt{SMSE} value of it would be approximately 1. Moreover, accurate predictors of target value will result in \texttt{SMSE} values closer to 0 while for bad predictors, there is no upper bound. The use of \texttt{SMSE} is advocated in \cite[p. 23]{rasmussen_gaussian_2005}.
$$\texttt{SMSE} = \frac{\texttt{MSE}}{\text{Var}[y]}$$
As this metric is suitable for aggregating over different tests for a general comparison, it is calculated by the base prequential method to produce one single value.

\texttt{SMSE\_ST:} Target-variance insensitive version of \texttt{RMSE\_ST}. Similarly to \texttt{SMSE}, the base prequential method is used to calculate this metric allowing aggregations across different tests of a stream learner.

\subsection{Prediction Bound Metrics}

\texttt{ICR:} \texttt{ICR} stands for Interval Containment Ratio. It is simply the ratio of the number of times provided prediction bounds by the learning algorithm contains the observed target to the number of total predictions.
$$\ \texttt{ICR}=\frac{\text{\# of predictions satisfying } y_i \in [\hat{y_i}^{l},\hat{y_i}^{u}]}{\text{\# of total predictions}} $$
Above $\hat{y_i}^{l}$ and $\hat{y_i}^{u}$ denote the lower and upper bound provided by the learning algorithm for the $i_{th}$ prediction.

Since the \texttt{ICR} metric is mostly preferred to be single valued and not considered to be concept drift sensitive, and finally because it is a metric that need to be averaged over different tests, the core prequential method is used for calculating it.

\texttt{AIW:} \texttt{AIW} stands for Average Interval Width. It has a self-descriptive name. Since \texttt{ICR} metric can be deceiving in case the interval width is too large or too narrow (any learning algorithm producing infinite intervals such as [-inf, +inf] will have good \texttt{ICR} scores on the other hand algorithms returning narrow bounds can get poor \texttt{ICR} scores although they might be more preferable to their alternatives with better \texttt{ICR}s), use of \texttt{AIW} in conjunction with \texttt{ICR} as prediction bound metrics is a more reasonable choice than using \texttt{ICR} alone.
$$\texttt{AIW}=\frac{1}{n}\sum_{i=1}^{n}(\hat{y_i}^{u} - \hat{y_i}^{l})$$
\texttt{AIW} value is considered to be a single-valued statistic. Hence, it is evaluated via the core prequential model.

\texttt{SAIW:} Similar to \texttt{RMSE}, \texttt{AIW} is also sensitive to target characteristics. More concretely, when the mean of the target variable is high, the predictive intervals tend to be high. That is why, it does not make sense to aggregate the \texttt{SAIW} values obtained through different tests with different target variances without normalizing them. Henceforth, a new metric which is a simple extension of \texttt{AIW} is used. It is trivially computed as follows:
$$\texttt{SAIW} = \frac{\texttt{AIW}}{\mu _y}$$
\texttt{SAIW} is an aggregatable metric and it is not interesting to track its behavior at the times of conceptual drifts. Thus, the core prequential method is used for calculating it.   

\subsection{Space Efficiency Metrics}

Space efficiency is a very serious issue for stream learning algorithms as discussed in the previous section. In Chapter \ref{Chapter5} where the implementation details of the learning algorithms used discussed, it is mentioned that the memory complexity of the online learners investigated in the scope of this thesis is all constant. In other words, the space they occupy in the main memory of the platform where they are running on does not increase with the number of training points (which is the number of data points in the stream meaning asymptotically infinite). Having described the implementation details of the implemented learners and made clear that they store fixed amount of data for their operations and knowing the practical difficulties involved in measuring the memory consumption of algorithms implemented in a language featuring a lazy garbage collection\footnote{Note that the learning algorithms are implemented in a testbench project for benchmarking and they are not deployed to Ocelot which is rather an engineering task and require reimplementation of the online learners in a low level programming language such as C.}, no space efficiency measurements are done. 

\subsection{Time Efficiency Metrics}

As discussed previously, for a stream learner the rate of processing which consists of predicting, observing the true target and updating the internal predictive model has to be  at least as high as the rate of the stream so that it is possible to employ the stream learner in real-time scenarios that require in-situ analysis. Another reason for this high processing rate is to be able to facilitate all the data coming from the stream and avoid lagging behind it since if the processing rate is lower than the stream rate than the unprocessed but already arrived data points will grow larger and larger resulting in underutilization of data and failing to meet real-time processing requirements. Real-time processing requirements are also present in the Ocelot decision-making layer as discussed in Chapter \ref{Chapter3}. Therefore, measuring the time it takes for an online regression algorithm to be evaluated to predict, update its internal predictive model and if applicable to tune its hyperparameters is important. To this end, various time efficiency metrics are used. These have self-descriptive names and they are listed below.

\begin{itemize}
\item \texttt{APT}: Average Prediction Time in $ms$.
\item \texttt{HPT}: Highest Prediction Time in $ms$.
\item \texttt{TPT}: Total Prediction Time in $ms$.
\item \texttt{AUT}: Average Update Time in $ms$.
\item \texttt{HUT}: Highest Update Time in $ms$.
\item \texttt{TUT}: Total Update Time in $ms$.
\item \texttt{ATT}: Average Tuning Time in $ms$.
\item \texttt{HTT}: Highest Tuning Time in $ms$.
\item \texttt{TTT}: Total Tuning Time in $ms$.
\item \texttt{TT}: Total Time.
\item \texttt{ATPI}: Average Time Per Item in $ms$.
\end{itemize}

Majority of the time efficiency metrics are not expected to fluctuate during concept drifts as the implementations of all the operations but the \texttt{tune} operations are drift-aware. Moreover, they all are aggregatable as they are not sensitive to target variable characteristics. This is why, the core prequential method is used for evaluating them.

\section{Evaluation Setting}

Employing the evaluation methodology discussed in the previous section, a battery of tests that simulate a stream with pregenerated synthetic data and expect predictions from the online learners in succession just like in a real-world stream learning scenario are prepared. The implementation details of the testing and stream simulation platform that is a generic implementation\footnote{Source is available at https://github.com/aanilpala/online-regression} of the Online Prediction Protocol discussed in Chapter \ref{Chapter2} is not the main focus of this section. The simulation logic is implemented following the following ideas:
Data sets which accommodates the synthetic data has a built-in order which might be relevant in the evaluation scenarios regarding the concept drifts. Without changing this order, data points are presented one by one to the learner and the response for the predictions are made available only after the learner makes its prediction for the most recent data point it is provided with. 
%
%

The evaluation involves the comparison of all the algorithms that discussed in Chapter \ref{Chapter4} and theoretically found to be suitable for the requirements outlined in \ref{list:restimator_requirements} for the learning problem. The tests used for the evaluation are stream simulations that are generated using the synthetic data with a vast selection of different characteristics that are explained below as well as the actual operator running time data from running Ocelot installations on the platforms geared up with different hardware. Next subsections outline the actual implementations of online learning algorithms to be tested and the characteristics of datasets used in streams.

\subsection{Evaluated Online Learners}

The codenames of the main regression algorithms whose variants (excluding the set of batch variants which is the control group) tested are listed below with the descriptions explaining the variants of each.

\begin{itemize}
\item \texttt{BayesianMLEForgetting}: Bayesian Maximum Likelihood Estimation Algorithm discussed in Chapter \ref{Chapter4} with a forgetting mechanism employed for concept drift adaptation. Two categories of variants are implemented. One category features a mapping from the input space to higher dimensional feature space and the other category does not. All the implemented variants have one of the following forgetting factors: 0.0 (no forgetting), 0.05 and 0.1. Parametrized with forgetting factor variable, each category has 3 variants summing up to total of 6 variants.
\item \texttt{BayesianMLEWindowed}: Bayesian Maximum Likelihood Estimation Algorithm discussed in Chapter \ref{Chapter4} featuring a sliding window for concept drift adaptation. Two categories of variants are implemented. One category features a mapping from the input space to higher dimensional feature space and the other category does not. All the implemented variants have one of the following sliding window sizes: 32, 48, 64, 96 and 128. Parametrized with the sliding window size variable, each category has 3 variants summing up to total of 10 variants.
\item \texttt{BayesianMAPForgetting}: Bayesian Maximum A Posteriori Estimation Algorithm discussed in Chapter \ref{Chapter4} with a forgetting mechanism employed for concept drift adaptation. Two categories of variants are implemented. One category features a mapping from the input space to higher dimensional feature space and the other category does not. All the implemented variants have one of the following forgetting factors: 0.0 (no forgetting), 0.05 and 0.1. Parametrized with forgetting factor variable, each category has 3 variants summing up to total of 6 variants.
\item \texttt{BayesianMAPWindowed}: Bayesian Maximum Likelihood Estimation Algorithm discussed in Chapter\ref{Chapter4} featuring a sliding window for concept drift adaptation. Two categories of variants are implemented. One category features a mapping from the input space to higher dimensional feature space and the other category does not. All the implemented variants have one of the following sliding window sizes: 32, 48, 64, 96 and 128. Parametrized with the sliding window size variable, each category has 5 variants summing up to total of 10 variants.
\item \texttt{GPRegression}: Gaussian Process Regression Algorithm using Gaussian Kernel discussed in \ref{Chapter4}. Three different variants are based on Gaussian Processes with different mean functions. The mean functions used are the zero mean function that always returns 0, average mean function that returns the average of the targets observed until the point of prediction and OLS mean function that uses the Ordinary Least Squares (MLE-method) prediction for a given data point as the mean and let the Gaussian Process model the residuals. With three different mean functions used and 5 different choices of the window sizes of 32, 48, 64, 96 and 128, there are 15 different variants of Gaussian Process Regression.
\item \texttt{KernelRegression}: Nadarya-Watson Estimator. 5 different choices for the window sizes namely 32, 48, 64, 96 and 128 have resulted in 5 different variants of Nadarya-Watson Estimator.
\end{itemize}

The number of different incremental regression algorithm variants that were put under test during the evaluations is $6\times 2 + 10\times 2 + 15 + 5 = 52$. Moreover, as the control group in the experiments, batch versions of them are also tested. Batch algorithms used are listed below.

\begin{itemize}
\item \texttt{BayesianMLEBatch}: Ordinary Least Squares (MLE-method). Two variants one of which featuring a high-dimensional input mapping mechanism similar to the one employed in Online \texttt{BayesianMLE} and \texttt{BayesianMAP} families are implemented. During the testing, three different runs with different sizes of training sets are used. The training set sizes used are 32, 64 and 128. In total, 6 different instantiations of the algorithm are prepared for testing.
\item \texttt{GPRegressionBatch}: Batch version of Gaussian Process Regression based on Gaussian kernel. For constructing the covariance matrix which is vital for Gaussian Process as discussed in \ref{Chapter3}, training sets of 32, 64 and 128 items are used.
\item \texttt{KernelRegressionBatch}: Kernel Regression algorithm based on Gaussian kernel function and zero mean function is implemented with a static case base which is constructed from the training sets of 32, 64 and 128.
\end{itemize}

In total 12 different instantiations of these 3 batch algorithms are tested along with the 52 incremental variants bringing the number of learners to be evaluated up to $64$.

\subsection{Synthetic Data Characteristics}

The synthetic data sets generated for stream simulation have different varying properties from one set to another. These are the number of input dimensions, scale of randomly generated real-valued inputs, variance of homoscedastic Gaussian distributed noise, properties of the function used for generation of target variables from inputs and conceptual drift characteristics if the stream that will be simulated from the data set is supposed to demonstrate concept drift. Among the properties of the function used for target value generation are whether the function has a discontinuity or not, if it has a discontinuity the growth rate (linear, logarithmic, quadratic etc.) target variable in the different continuous regions of the discontinuous functions. Following list summarizes the choices of the mentioned properties made when generating the synthetic data.

\begin{itemize}
\item \textbf{Input Dimensionality}: Synthetic Data sets generated have either 1, 2 or 4 real-valued inputs.
\item \textbf{Input Scale}: Inputs are drawn from standard uniform distribution via pseudorandom number generator of the programming language used for implementing the testbench mentioned previously. By default, pseudorandom generator produces a real number representation between 0 and 1. Scaling the numbers drawn from this interval by a constant gives the desired uniformly distributed input values between 0 and the scaling factor used. For the data sets, one of the following scaling factor is used: 10, 50, 100.
\item \textbf{Noise Variance}: As explained in Chapter \ref{Chapter4}, homoscedastic normal distributed noise is assumed in the design of learning algorithms. Similarly, the noise is generated in the same way (to fulfil this assumption) and added to the synthetic data. The noise variances used for additive noise generation are 0.0 (no noise), 1.0, 3.0 and 5.0.
\item \textbf{Continuity of Underlying Function}: For the half of the data sets prepared for the stream simulation, the function used for data generation was continuous everywhere in the input space while for the other half, the functions with discontinuities are used. Discontinuity is important especially for mocking the behaviour of many database operators that demonstrate different cost characteristics when the inputs cannot be accommodated in the main memory resulting in swapping in and out of the input data that is responsible for high runtimes. Examples of this behavior are observed in the experiments presented in \cite[p. 717]{heimel_hardware-oblivious_2013}. To model this discontinuity behaviour of the runtime in the synthetic data,  two generating functions used to model different cost characteristics of database operators. The function with the lower growth rate is used to calculate the noise-free targets (additive noise to be added later) for the data points with the sum of the inputs that is less than half of the sum of the scaling factors (the maximum value an input variable can take). As for the noise-free targets for the datapoints that fall into the remaining part of the input space, the function with the bigger growth rate is used. It is worth noting that the data distribution on the region with the discontinuity which is a hyperplane do not necessarily have to be smooth (e.g noise-free targets might jump to high values abruptly as the inputs grow).
\item \textbf{Growth Rate of Underlying functions}: Functions with three different growth rates are employed for data generation. These are listed and formulated below where $\bf{x}$ is vectorized input, $d$ denotes the dimensionality of the input space and $\pmb{b}$ denotes the vector of coefficients (real numbers between 0 and 10 drawn randomly) of size $d$.
\begin{itemize}
\item Linear Growth: $f(\pmb{x}) = \pmb{x}^{\top}\pmb{b}$
\item Log-Linear Growth: $f(\pmb{x}) = sum\times log(sum) \text{ where } sum = \pmb{x}^{\top}\pmb{b}$
\item Quadratic Growth Variant 1: $f(\pmb{x}) = (\pmb{x}^2)^{\top}\pmb{b}$
\item Quadratic Growth Variant 2: $f(\pmb{x}) = sum^2 \text{ where } sum = \pmb{x}^{\top}\pmb{b}$
\end{itemize}

For the data sets for which a continuous function is supposed to be used for the data generation, only one of this 4 different growth functions is chosen. If the data generation function is supposed to feature a discontinuity, then the partial function to model the smaller sized half of the inputs is chosen to be different than the partial function that is to model the higher-sized input space. The choice of these two growth functions is not random in order to allow higher-sized inputs to have higher targets. Allowed combinations in the said order are listed in the following.
\begin{itemize}
\item Linear Growth / Log-Linear Growth
\item Linear Growth / Quadratic Growth Variant 1
\item Linear Growth / Quadratic Growth Variant 2
\item Log-Linear Growth / Quadratic Growth Variant 1
\item Log-Linear Growth / Quadratic Growth Variant 2
\end{itemize}

When generating the synthetic data without discontinuity, 4 different growth rates are used. In the continuous case, for the one-dimensional input space, only 3 of the 5 growth rate combinations listed above are used as the quadratic growth variants imply the same growth-rate when inputs consist of one variable ($\pmb{x} = [x]$) and for the input spaces used in datasets having 2 or 4 dimensions spaces, all 5 possible growth rate combinations are used.

\item \textbf{Stationarity}: Half of the stream simulations to be used in the evaluation is supposed to feature a drifting data distribution after a certain point in the stream has been passed. In order to simulate this phenomenon called concept drift, simply the first half of the data points in the data set that is used for stream simulation is generated with a different function than the function used to generate the second half. Underlying function properties discussed above kept unchanged but the coefficients are changed (by randomly picking numbers between 0 and 10). Although, keeping the growth rate same when changing from the original concept to the new one and only varying the coefficients might sound restrictive, it is indeed  how the underlying function is expected to change in the case of conceptual drift that runtime data producing mechanism which is primarily the choice of the concrete algorithm variant choice done by the Ocelot as discussed in \ref{Chapter3}. On the other hand, for the other half of the stream simulations, the function (either continuous or not) used to generate the data is kept the same within the process of generating the data sets.
\end{itemize}

Following the ideas presented above for synthetic data generation, the numbers of different data sets with varying noise levels and input scales grouped by their input dimensions and the growth rates of the underlying functions are listed below along with the codenames of the datasets. 

The naming pattern used for dataset naming is:

\texttt{SYNTH\_(D|ND)\_(CD|NCD)\_SIZE\_INPDIM\_INPSCALE\_NOISEVAR\_FGROWTH1\_FGROWTH2}

where \texttt{(D|ND)} denotes either the underlying function is discontinuous or continuous, \texttt{(CD|NCD)} tells if the simulation to be generated using the formulated data set is supposed to demonstrate an abrupt conceptual drift or not. \texttt{SIZE, INPDIM, INPSCALE} and \texttt{NOISEVAR} simply refer to the size, number of input dimensions, input scale and the noise variance respectively. \texttt{FGROWTH1} and \texttt{FGROWTH2} denote the growth rate of the partial underlying functions if the underlying function is non-continuous and if not, both are the same and denote the growth rate of the underlying function.

The groups of data sets by their input dimensions and the growth rates of the underlying functions:

\begin{itemize}
\item 36 \texttt{SYNTH\_ND\_NCD\_2000\_1\_INPSCALE\_NOISEVAR\_FGROWTH1\_FGROWTH2}
\item 36 \texttt{SYNTH\_D\_NCD\_2000\_1\_INPSCALE\_NOISEVAR\_FGROWTH1\_FGROWTH2}
\item 36 \texttt{SYNTH\_ND\_CD\_2000\_1\_INPSCALE\_NOISEVAR\_FGROWTH1\_FGROWTH2}
\item 36 \texttt{SYNTH\_D\_CD\_2000\_1\_INPSCALE\_NOISEVAR\_FGROWTH1\_FGROWTH2}

\item 48 \texttt{SYNTH\_ND\_NCD\_2000\_2\_INPSCALE\_NOISEVAR\_FGROWTH1\_FGROWTH2}
\item 60 \texttt{SYNTH\_D\_NCD\_2000\_2\_INPSCALE\_NOISEVAR\_FGROWTH1\_FGROWTH2}
\item 48 \texttt{SYNTH\_ND\_CD\_2000\_2\_INPSCALE\_NOISEVAR\_FGROWTH1\_FGROWTH2}
\item 60 \texttt{SYNTH\_D\_CD\_2000\_2\_INPSCALE\_NOISEVAR\_FGROWTH1\_FGROWTH2}

\item 48 \texttt{SYNTH\_ND\_NCD\_2000\_4\_INPSCALE\_NOISEVAR\_FGROWTH1\_FGROWTH2}
\item 60 \texttt{SYNTH\_D\_NCD\_2000\_4\_INPSCALE\_NOISEVAR\_FGROWTH1\_FGROWTH2}
\item 48 \texttt{SYNTH\_ND\_CD\_2000\_4\_INPSCALE\_NOISEVAR\_FGROWTH1\_FGROWTH2}
\item 60 \texttt{SYNTH\_D\_CD\_2000\_4\_INPSCALE\_NOISEVAR\_FGROWTH1\_FGROWTH2}
\end{itemize}

The calculation of above numbers are simple. The number of different \texttt{INPSCALE, NOISEVAR} combinations are $3\times 4=12$. The number of different combinations of growth rates used in the functions depends on \texttt{INPDIM} and \texttt{D|ND}. For the single dimensional-input, this number is 3 as implied where the growth rates are discussed above making the number of such data sets for each group with different \texttt{(D|ND\_CD|NCD)}. $12\times 3=36$ gives the number 36. As for the groups with 2 and 4 dimensional input spaces, the number of different possible combinations of growth rates is 5 for discontinuous cases and 4 for non-discontinuous cases as explicitly mentioned in the discussion about the growth rates above. Thus, the number of datasets for the groups with discontinuous and continuous functions are $12\times 4=48$ and $12\times 5=60$ respectively.

Note that, since the codenames of the datasets uniquely identify the stream simulated from their corresponding dataset, sometimes the simulated streams are also addressed by the codename of their dataset in the rest of this section.

\subsection{Ocelot Runtime Measurement Data Characteristics}

Unlike the synthetic data sets, there is little known about the characteristics of the measurement data logged from running Ocelot instances. Among the unknown characteristics of the data are noise variance, continuity of underlying function, growth rate of underlying partial functions, stationarity of the data. On the other hand, the input dimensionality is known. The runtime measurement data includes data points with either 1 or 2 inputs. Moreover, once the measurement data is obtained, before running the stream simulations, the inputs can be examined to see their distributions. When discussing the characteristics of the synthetic data, only the input scale is mentioned regarding the input characteristics. This is because, the synthetic data sets are sampled from a uniform random distribution and scaled by a scaling factor. However, since the inputs occur in the measurement data are not drawn from a uniform random distribution, the coverage of the input space by the inputs should be considered when discussing the runtime data characteristics. In the measurement data, the number of distinct input points occur in the input space is very low. This is not surprising because the runtime measurement data is obtained by a number of fixed size TPC-H benchmarks each of which queries a database instance with fixed size tables resulting in operator calls with fixed size inputs.

\section{Analysis of Test Results}

In this section, the results of tests only on the streams that are simulated from the synthetically generated data are analyzed. The tests with the actual measurement data is saved for the \textit{verification} of the the effective and performant learners that are picked from a relatively large set of all the implemented learners. The reason for using the synthetic data and actual measurement data for different purposes is that the extra information available about the the properties of synthetic data which is lacking in the case of the real measurement data such as stationary, underlying function, noise levels, etc. This allows to identify the weakness and strengths of the online learning algorithms better before deciding to employ which one to deploy on a hardware-oblivious database. Moreover, since the actual data sets available by the time of the preparation of this thesis by no means represent the diversity of the stream data that online learners for runtime prediction problem can be confronted in a running environment. On the other hand, the synthetic tests are carefully generated in an attempt to provide a systematic, rigorous and comprehensive testing strategy.

Total number of data sets generated is $(36+48+60)\times 4 = 576$. The $52$ different online learners are tested on total of 576 streams that are simulated from 576 synthetic data sets. Moreover $12$ batch algorithms are tested on the same 576 data sets. Hence, in the synthetic data experiments, $576\times (52+12)=36864$ \textit{session} results which contains the evaluation statistics that are discussed previously. 

As it is not an efficient and effective way to interpret each of these $36864$ results, first, general aggregated results of online learners which are the variants of different family of learning algorithms are presented. Whenever a general picture does not allow to make good observations, more detailed analysis through either filtering out some set of test cases or algorithms or drilling down to details (e.g grouping the results by algorithm family instead of grouping them into only two categories namely online and batch). The presentation of the test results are structured into subsections where different conclusions regarding the nature of online learning algorithms in stream learning environments are drawn.

\subsection{Online Learners vs Batch Learners}

\begin{figure}[htbp]
  \centering
    \includegraphics[width=\linewidth]{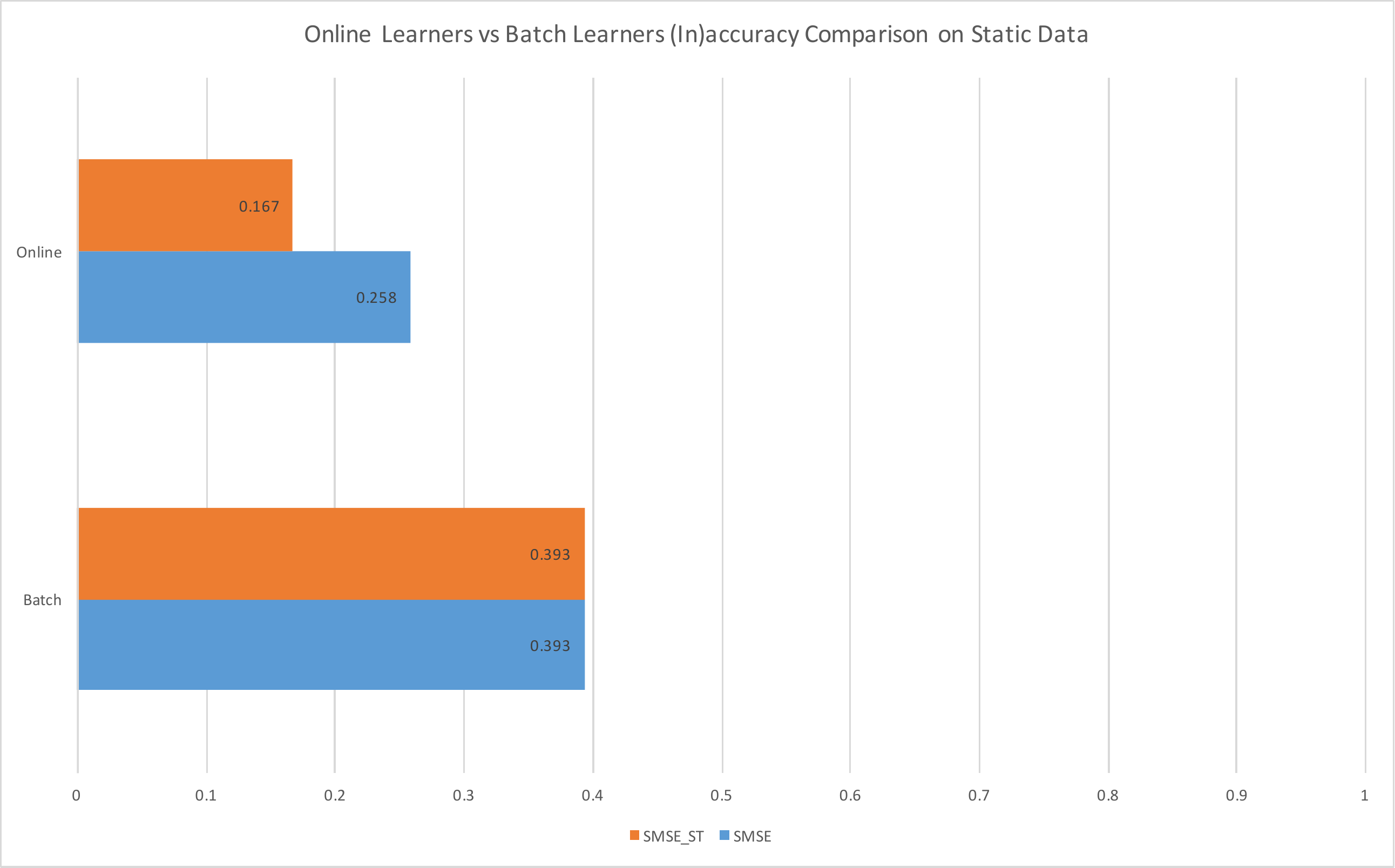}
  \caption{Inaccuracy Comparison of Batch and Online Learners. The bar values are obtained through aggregating the corresponding statistics over 576 stream simulations on 36 online learners and 576 tests on 12 batch learners respectively.}
  \label{fig:batch_vs_online_inaccuracy_comp1}
\end{figure}

In \ref{fig:batch_vs_online_inaccuracy_comp1}, what is shown is a \textit{rough} comparison of the accuracy of online learners and batch learners. It is a rough comparison because the first of two \texttt{SMSE} values shown are actually the average of already aggregated (over 576 streams) \texttt{SMSE} values of 52 online learners. As for the second value which is for the offline learners, it is similarly the average of already aggregated (over 576 synthetic data sets) \texttt{SMSE} values of 12 batch learners. According to this highly general results, the average accuracy difference of batch algorithms and online algorithms seems not to be very large in both \texttt{SMSE} and \texttt{SMSE\_ST} terms. Naturally, the average \texttt{SMSE} and \texttt{SMSE\_ST} values for Batch Learners are the same as they do not take any time to stabilize during the testing unlike their online counterparts. By looking at this results only, one might question the effort needed to design and implement online versions of batch learning algorithms since although the improvement is accuracy is still substantial with the online learners, batch ones did not have a worse \texttt{SMSE} than 0.4 which is acceptable. However, as said in the beginning, this rough results can be deceiving.

\begin{figure}[htbp]
  \centering
    \includegraphics[width=\linewidth]{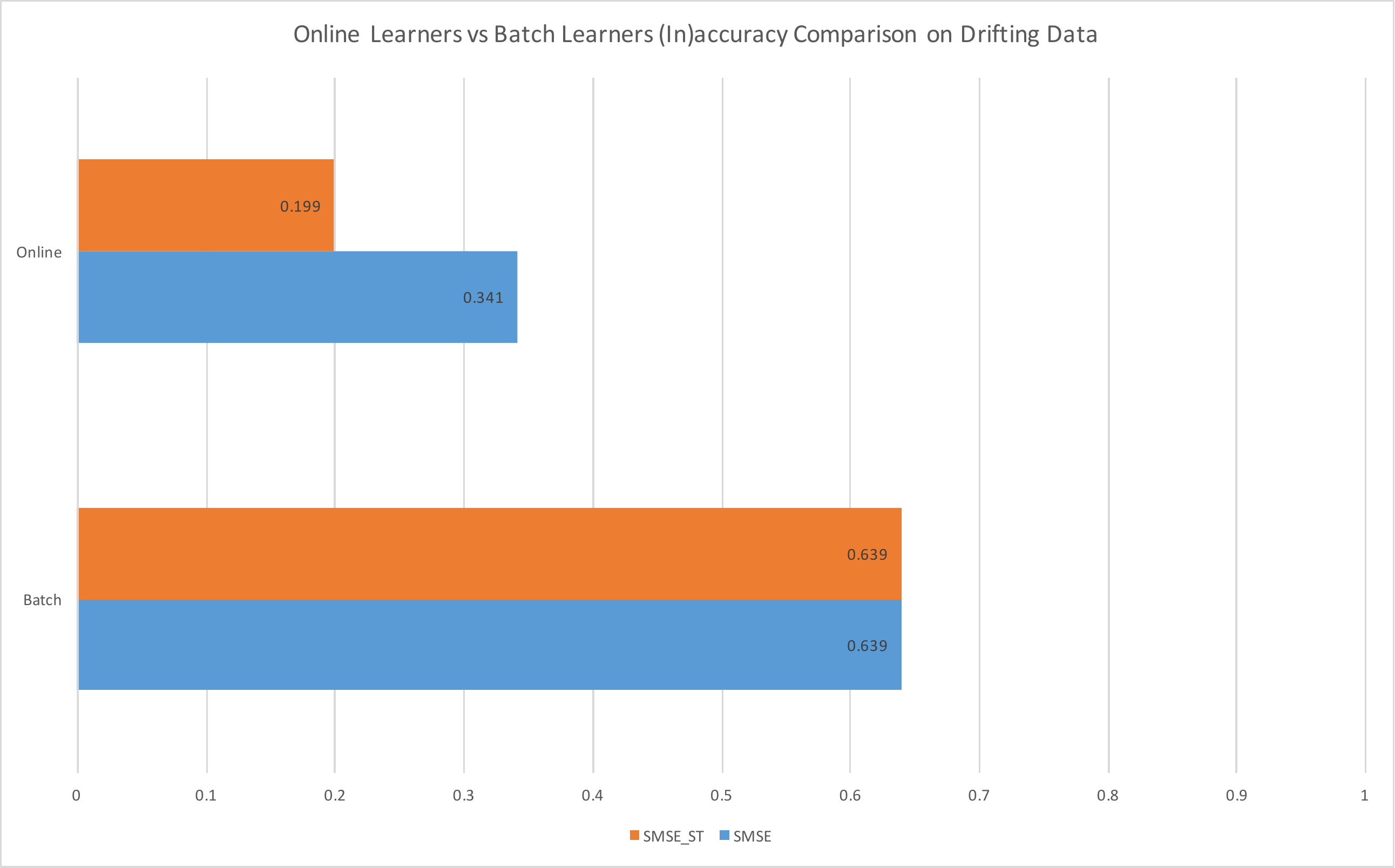}
  \caption{Inaccuracy Comparison of Batch and Online Learners on Non-Static Data. The bar values are obtained through aggregating the corresponding statistics over 276 stream simulations on 52 online learners and 276 tests on 12 batch learners respectively.}
  \label{fig:batch_vs_online_inaccuracy_comp2}
\end{figure}

\begin{figure}[htbp]
  \centering
    \includegraphics[width=\linewidth]{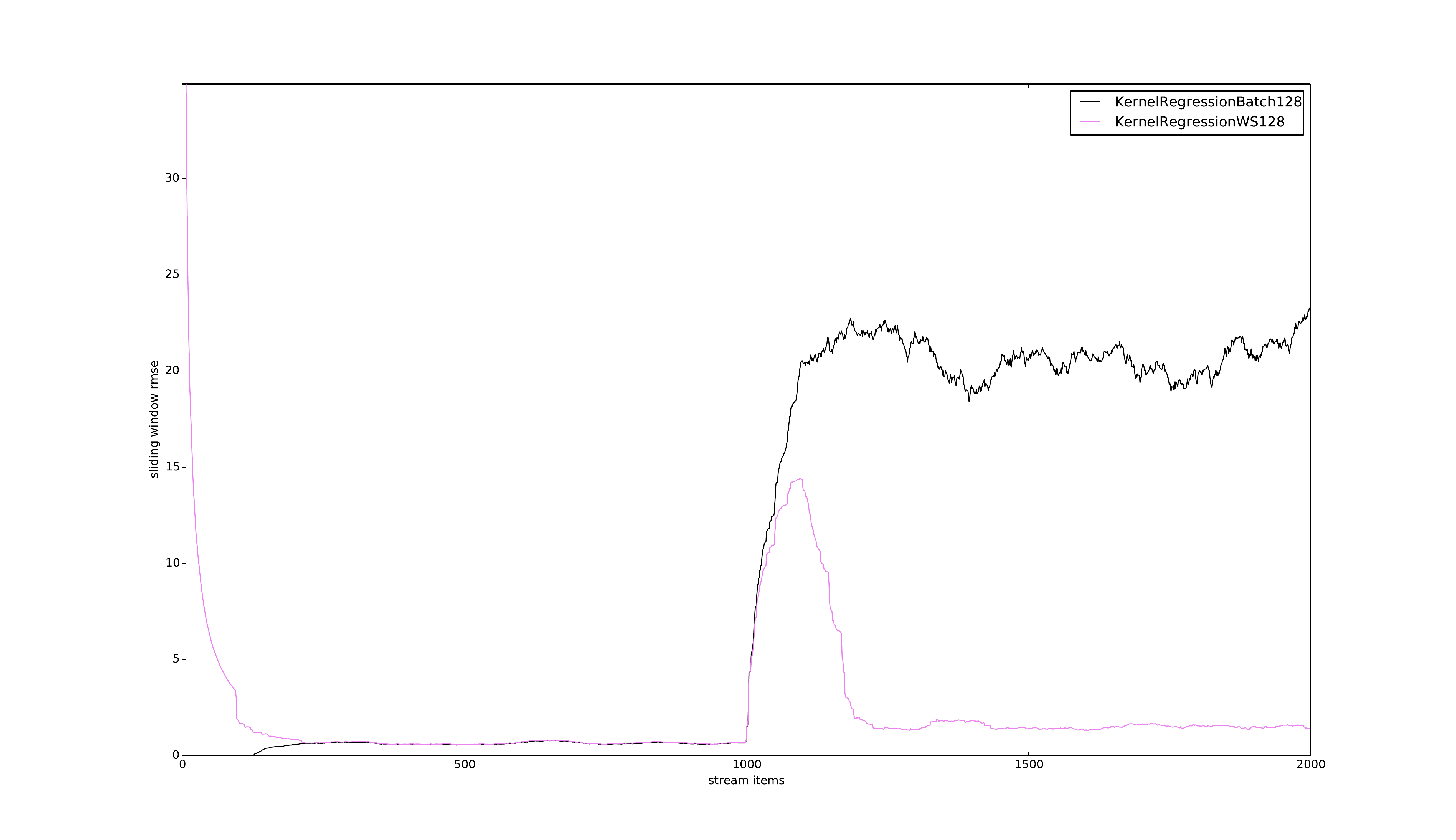}
  \caption{Side-by-side Accuracy Comparison of Batch and Online versions of Kernel Regression with training set of 96 and sliding window size of 96 respectively. The test data used is SYNTH\_ND\_CD\_2000\_2\_10\_1\_22 and the resolution of the sliding error window accuracy evaluation method is set to 96}
  \label{fig:batch_vs_online_sidebyside_comp_res96}
\end{figure}

If the accuracy results aggregated over tests with non-stationary data is analyzed, we see a totally different picture than \ref{fig:batch_vs_online_inaccuracy_comp1}. In \ref{fig:batch_vs_online_inaccuracy_comp2}, the accuracy gap between batch and online learners for non-static data tests is huge. What this means is if the changes in data distributions are expected during the online learning process, using a batch learner which is trained with test data beforehand is a very bad idea as the concept changes render the batch learners useless. One can reach this conclusion by looking at a side-by-side \texttt{RMSE} comparison of an individual batch learner and individual online learner on a single data stream \ref{fig:batch_vs_online_sidebyside_comp_res96}. In the simulated stream a concept change has affected the data points after the $1000_{th}$ stream item. We see that both algorithms perform equally well up until the concept drift. After the the drift occurs, the online algorithm was able to adapt itself to the change and recover its \textit{sliding window} error measured by the prequential method extension with sliding window of 128 while batch algorithm becomes \textit{useless}.

This first and the most fundamental observation validates the necessity of the use of online learning algorithms in streams. Rest of the analysis will rather concentrate on the comparison within the online learners.

\subsection{A general picture}

\begin{figure}[htbp]
  \centering
    \includegraphics[width=\linewidth]{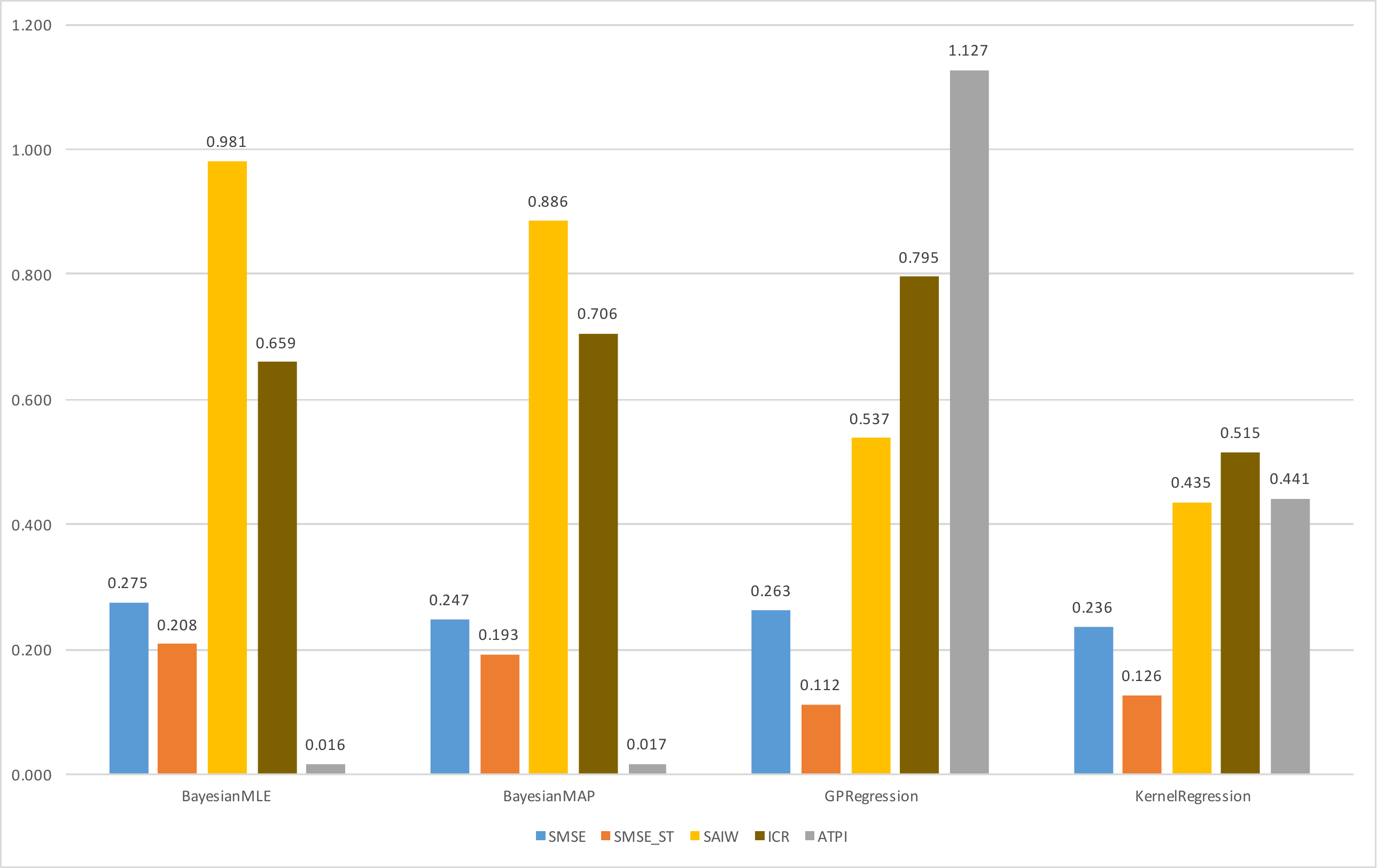}
  \caption{A General Comparison of Online Algorithms. The values on columns are aggregated over 576 stream simulations on 52 online learners and grouped by the algorithm family.}
  \label{fig:gen_alg_comparison}
\end{figure}

In \ref{fig:gen_alg_comparison}, we see the aggregated results of different family of online learning algorithms with 4 different metrics. The metrics used are the most critical ones in deciding which algorithm to deploy. Among the metrics used are \texttt{SMSE} that summarizes the accuracy, \texttt{SAIW} and \texttt{ICR} which together gives an overall opinion about the feasibility of the prediction bounds and lastly \texttt{ATPI} that allows to decide which algorithm category is quicker than the other in predicting the new data point. 

At first glance, it looks like the general accuracy scores (\texttt{SMSE}) of the algorithms are similar whereas in terms of static accuracy (\texttt{SMSE\_ST}), learners with parametric algorithms, \texttt{BayesianMLE} and \texttt{BayesianMAP}, are trailing behind of the non-parametric learners. All the families having the aggregate \texttt{SMSE} value lower than $0.28$ means that online learners were effective learning from streams in general so that all of them performed way better than the hypothetical trivial regression algorithm with \texttt{SMSE} of $1.0$ which returns the average of the targets observed until the point of prediction.

As for a general comparison of prediction bounds quality, \texttt{KernelRegression} family provided the tightest prediction intervals for the predictions with the average \texttt{SAIW} score of $0.435$. However, in terms of the target coverage of its prediction bounds, it falls behind that of the algorithms of other families. This is not surprising as intuitively wider the prediction bounds are, higher the interval containment ratio is. This is also expected theoretically as discussed in Chapter \ref{Chapter3}, KernelRegression provides confidence intervals that are used as the prediction bounds rather than the prediction intervals which are known to be wider. This graph alone does not say anything conclusive on the quality of prediction bounds that \texttt{KernelRegression} Family provides. Both \texttt{BayesianMLE} and \texttt{BayesianMAP} families have very high \texttt{SAIW} scores (between $0.85$ and $1.0$) meaning that their prediction bounds are almost as big as the overall scale of the target variable. For a single algorithm variant, this alone would be enough to conclude that it is unsuitable for a an application where the online learner to be deployed to does not have a tolerance for wide prediction bounds. However, in the graph, results are aggregated over different variants within the individual algorithm families presented. That is why, it is hard to conclude anything from it. Moreover, \texttt{GPRegression} family has the best \texttt{ICR} score and counterintuitively its \texttt{SAIW} score, $0.535$, is not very high. This indicates that, generally speaking, in terms of prediction bounds, \texttt{GPRegression} family is able to produce good prediction intervals.

In terms of average prediction time which is a very crucial consideration in stream learning scenarios, parametric families, \texttt{BayesianMLE} and \texttt{BayesianMAP}, proved to be very quick with the average ATPI score on the ballpark of 15 $\mu s$. Non-parametric models are relatively slower. \texttt{KernelRegression} family could process stream items with 441 $\mu s$ latency on average, Not so surprisingly due to its sophisticated prediction and update mechanism involving heavy matrix operations together with costly tuning routine that involves a non-convex optimization task, \texttt{GPRegression} family is the slowest with the average item processing latency of 1127 $\mu s$.

As pointed out, it is not possible to draw any significant and concrete conclusions from this general analysis. This is why, it is necessary to further examine the test data by drilling down to more details. To this end, the following sections include a more detailed analysis on the behavior of learners with changing qualities of either themselves (e.g window size, forgetting factor) or of the learning environment (e.g input dimensionality, noise variance, target scale).

\subsection{Sliding window size}

\begin{figure}[htbp]
  \centering
    \includegraphics[width=\linewidth]{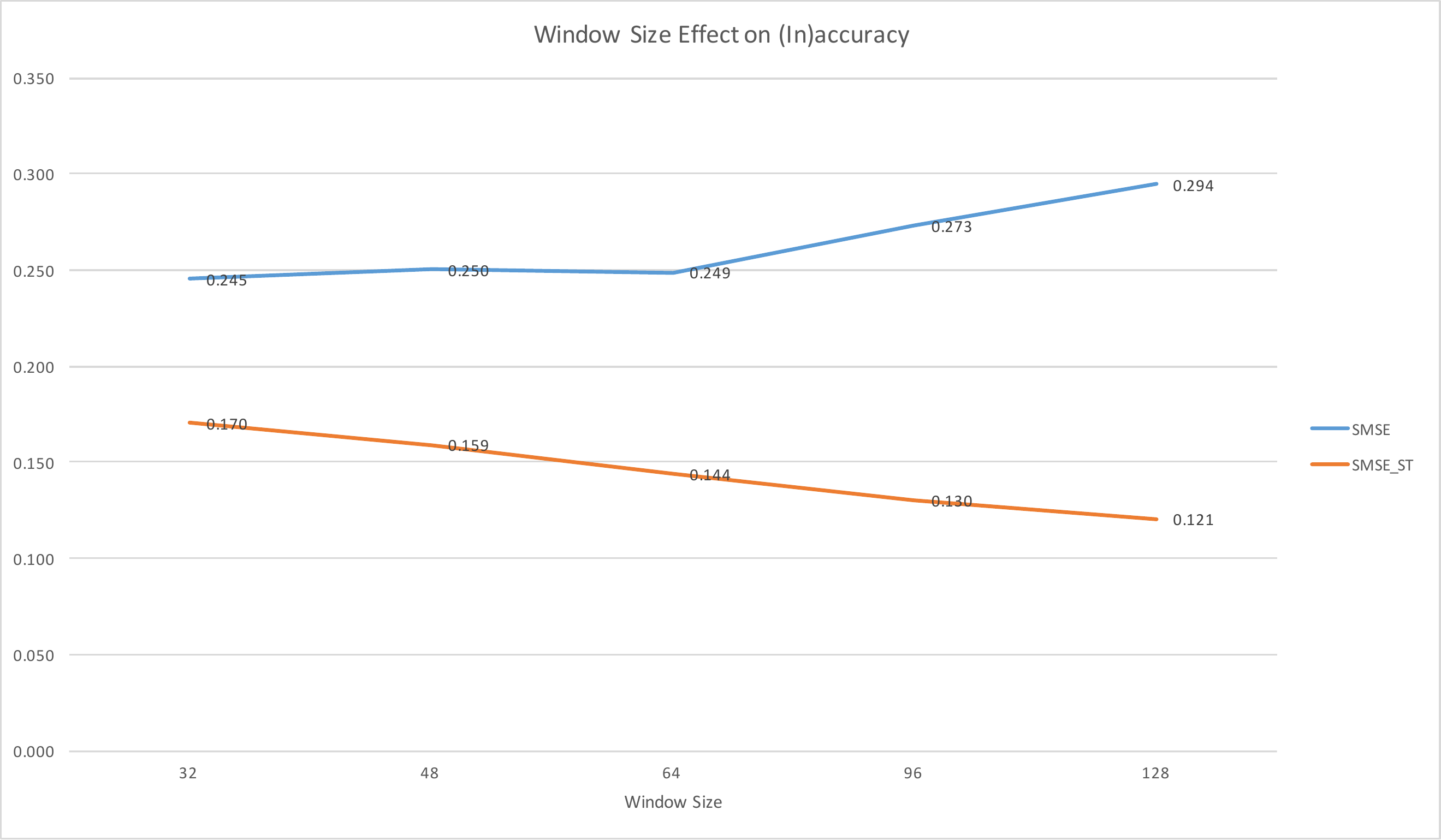}
  \caption{Accuracy comparison of sliding windowed learners with varying window sizes. \texttt{SMSE} and \texttt{SMSE\_ST} values are aggregated over 576 stream simulations on 8 sliding windowed online learners (2 BayesianMAP, 2 BayesianMLE, 3 GPRegression and 1 KernelRegression variants) for each window size}
  \label{fig:ws_on_accuracy1}
\end{figure}

\begin{figure}[htbp]
  \centering
    \includegraphics[width=\linewidth]{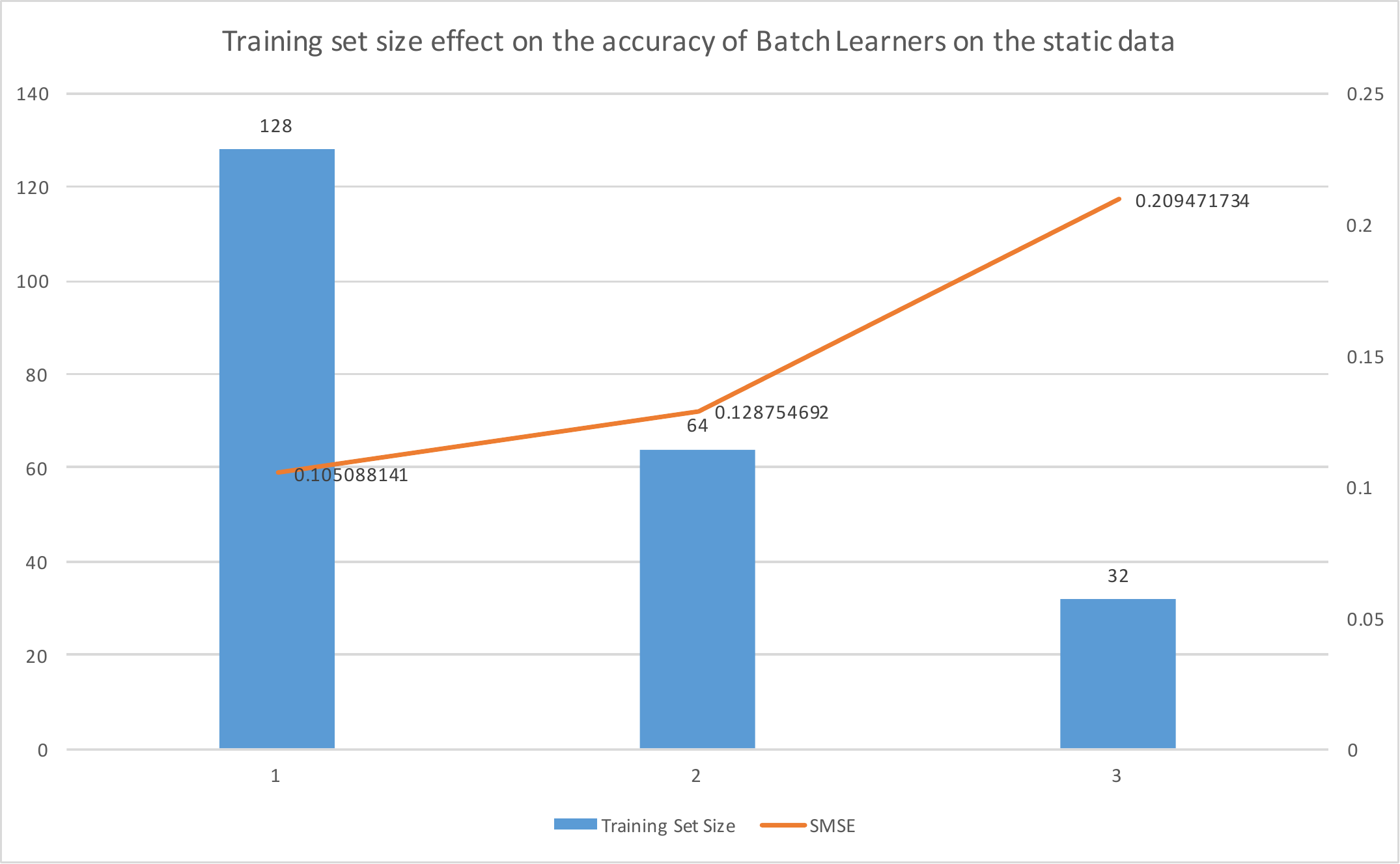}
  \caption{Accuracy comparison of batch learners with varying training set sizes. \texttt{SMSE} values are aggregated over 288 tests (with static data) on 4 different batch algorithm for each training set size}
  \label{fig:batch_trainingsetsize_on_accuracy}
\end{figure}

In \ref{fig:ws_on_accuracy1}, on the x- axis we see five different window sizes used in online learners featuring sliding window mechanism. \texttt{SMSE\_ST} values improve (decreases) as the window size gets bigger expectedly. This must be due to the bigger \textit{case base} upon which the predictions of the online algorithm are based. A similar situation exists in batch algorithms with the changing training set size. As shown in \ref{fig:batch_trainingsetsize_on_accuracy}. This is a simple implication of sliding window semantically being a \textit{dynamically} evolving training set. However, this relation between the sliding window size and the accuracy seem to hold only in the case of \texttt{SMSE\_ST} statistic which excludes the prediction errors that online learners make during the \textit{adaptation} periods either before its sliding window gets full or after an abrupt concept drift. 

If we examine the change in \texttt{SMSE} values with changing window size, something which cannot be explained trivially is observed: The highest choice of the sliding window size, being $128$, resulted in the poorest \texttt{SMSE} result in the averaged test results. Knowing the difference between the computed statistics \texttt{SMSE} and \texttt{SMSE\_ST}, one could link this phenomenon of accuracy drop as the window size increases to the relatively higher stabilization time learners with bigger sliding windows need after concept drifts. In order to validate this claim, one needs to analyze the evolution of the accuracy as a function of time. As discussed previously, the proposed extensions to the prequential evaluation method extensions provide such time-dependent analysis. However, this time the analysis will be confined to an individual test case (single stream simulation) as these extensions forbid the aggregations over different tests. 

\begin{figure}[htbp]
  \centering
    \includegraphics[width=\linewidth]{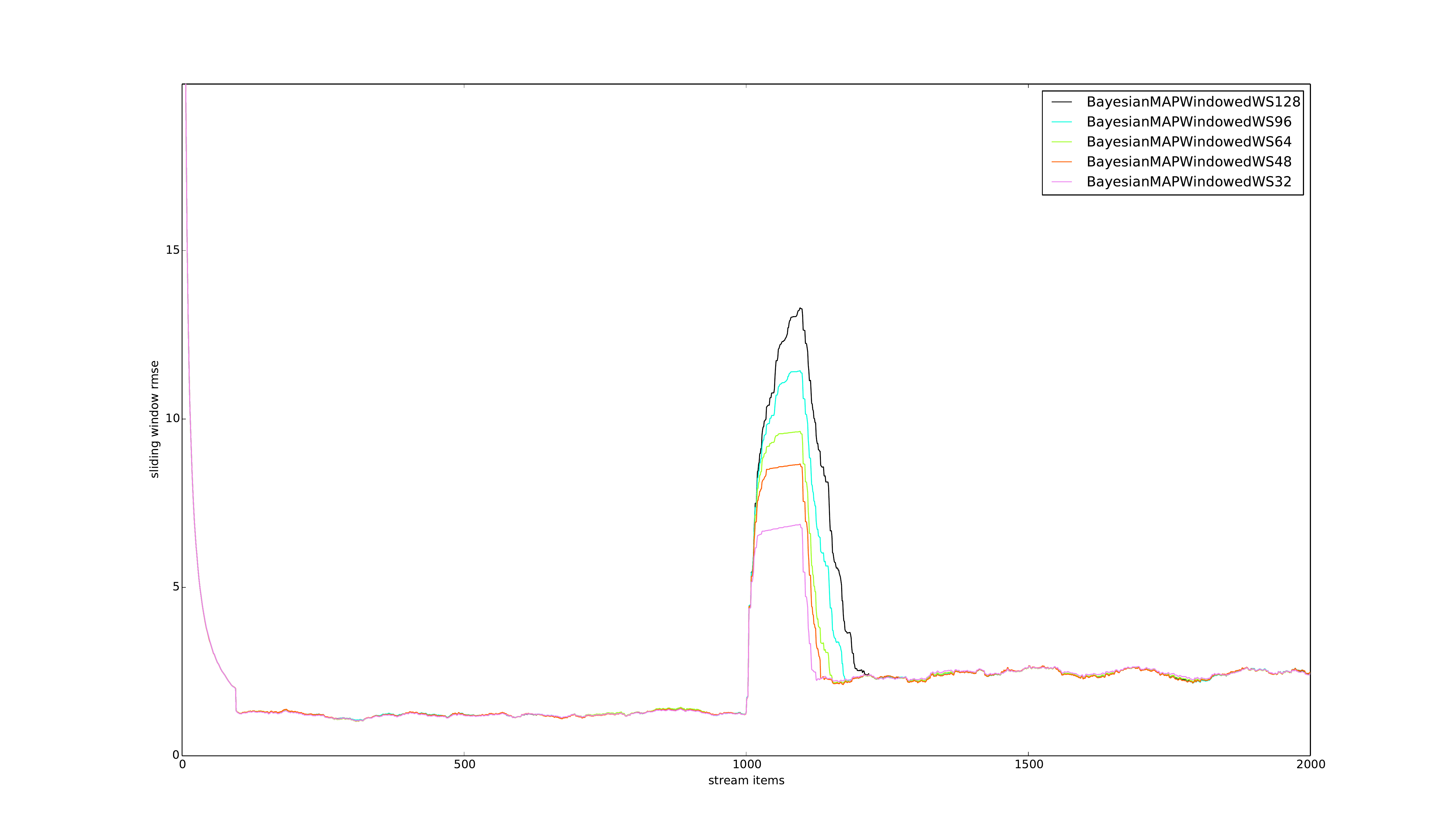}
  \caption{Side-by-side Accuracy Comparison of \texttt{BayesianMAPWindowed} with different window sizes. The test data used is \texttt{SYNTH\_ND\_CD\_2000\_2\_10\_1\_22} and the resolution of the sliding error window accuracy evaluation method is set to 96}
  \label{fig:wsize_on_stabilization_sidebyside_comp_res96}
\end{figure}

\begin{figure}[htbp]
  \centering
    \includegraphics[width=\linewidth]{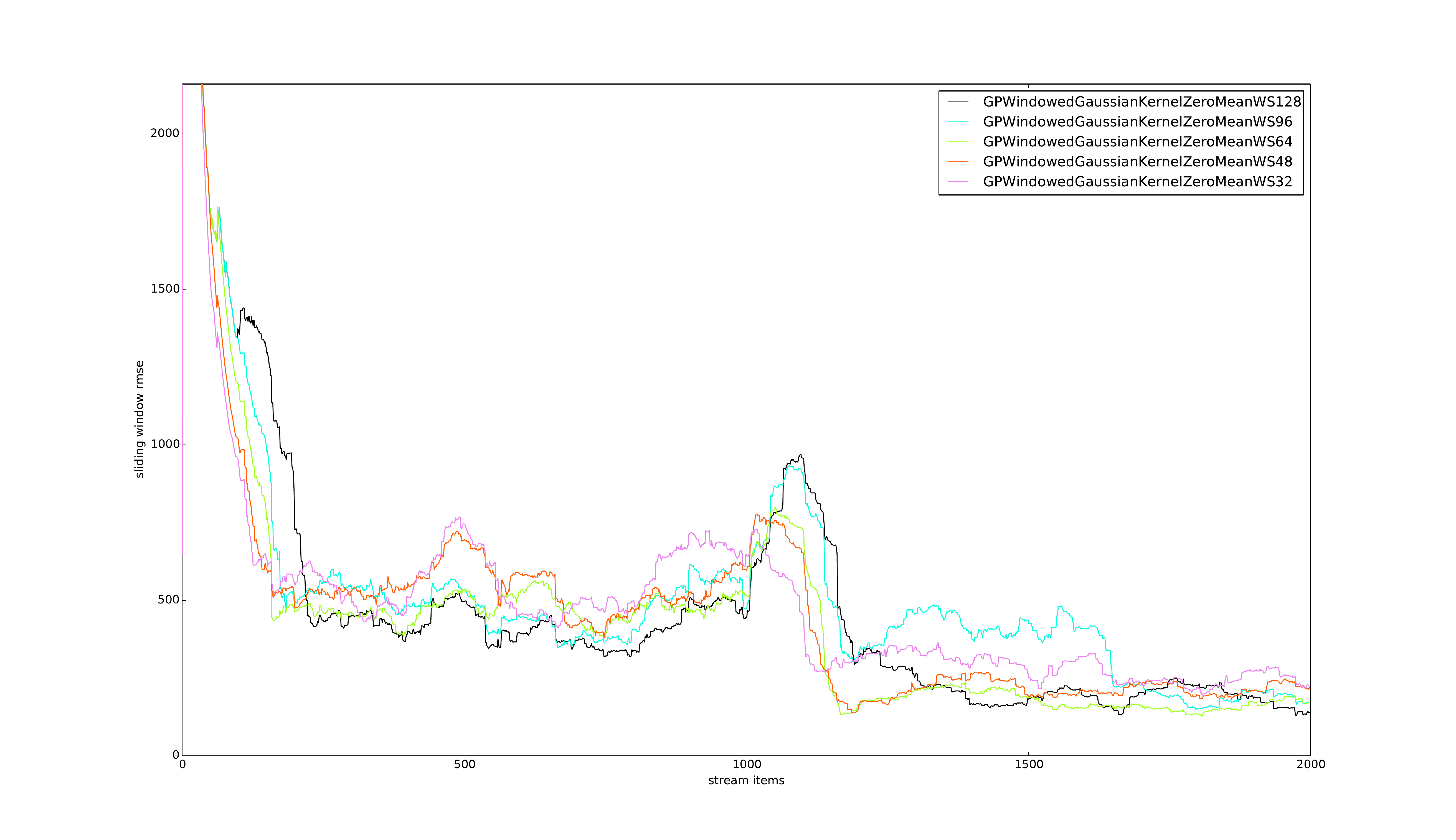}
  \caption{Side-by-side Accuracy Comparison of Online \texttt{GPWindowedGaussianKernelZeroMean} variants with different window sizes. The test data used is \texttt{SYNTH\_D\_CD\_2000\_4\_10\_1\_24} and the resolution of the sliding error window accuracy evaluation method is set to 96}
 \label{fig:wsize_on_stabilization2_sidebyside_comp_res96}
\end{figure}

In \ref{fig:wsize_on_stabilization_sidebyside_comp_res96} and \ref{fig:wsize_on_stabilization2_sidebyside_comp_res96}, it is clearly seen that after the concept drift the learner with the biggest window size took the longest time to adapt to the changing data distribution. This is not surprising because the sliding window algorithms implemented incorporate the new items one by one after their update mechanism is triggered by high error and until the \textit{deceiving} the data points left in the window from the previous concept are replaced with the new ones, the predictions will suffer from big errors as what is \textit{assumed} to be the reference data points (case base) in building the predictive model internally by the learner are far off in the mentioned unstable phase. This may or may not result in poor general accuracy scores (\texttt{SMSE}) depending on the how big the increased accumulated error term is during the unstable learning periods and the overall accuracy improvement in the stable learning periods with the higher window size. In the experiments carried out, it is observed that the average \texttt{SMSE} result remained the same when the window size is increased from $48$ to $64$ possibly because of the increased stable accuracy compensating the longer unstable period producing larger number of erroneous predictions resulting in unchanged general accuracy. The explained relations between the stable and total accuracy with the window size are inherently present in the case of any sliding-window algorithm regardless of the algorithm used to build a predictive model from the data points stored in the sliding window. 

\begin{figure}[htbp]
  \centering
    \includegraphics[width=\linewidth]{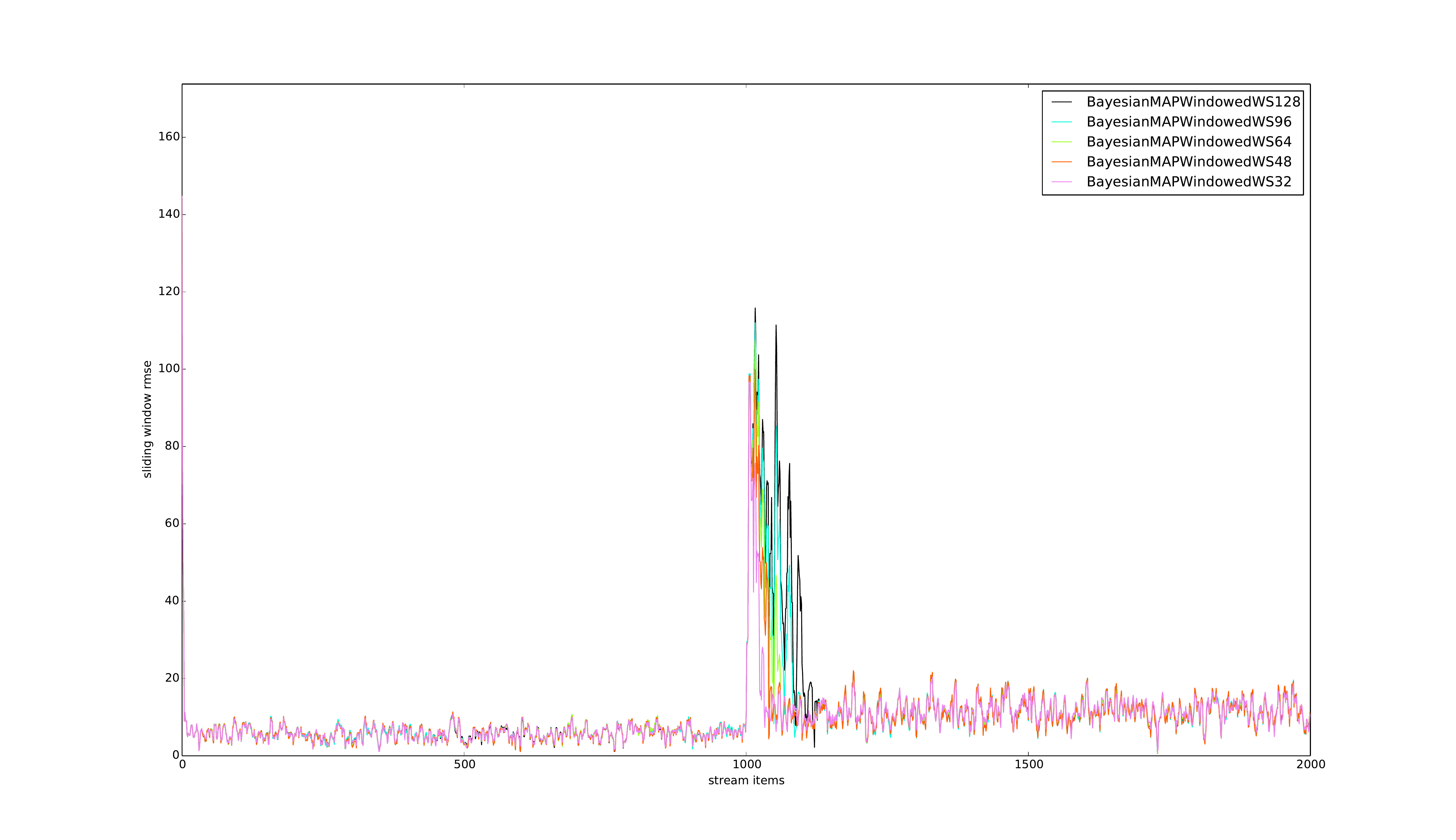}
  \caption{Side-by-side Accuracy Comparison of \texttt{BayesianMAPWindowed} variants with different window sizes. The test data used is \texttt{SYNTH\_ND\_CD\_2000\_2\_10\_1\_22} and the resolution of the sliding error window accuracy evaluation method is set to 4}
  \label{fig:wsize_on_stabilization_sidebyside_comp_res4}
\end{figure}

\begin{figure}[htbp]
  \centering
    \includegraphics[width=\linewidth]{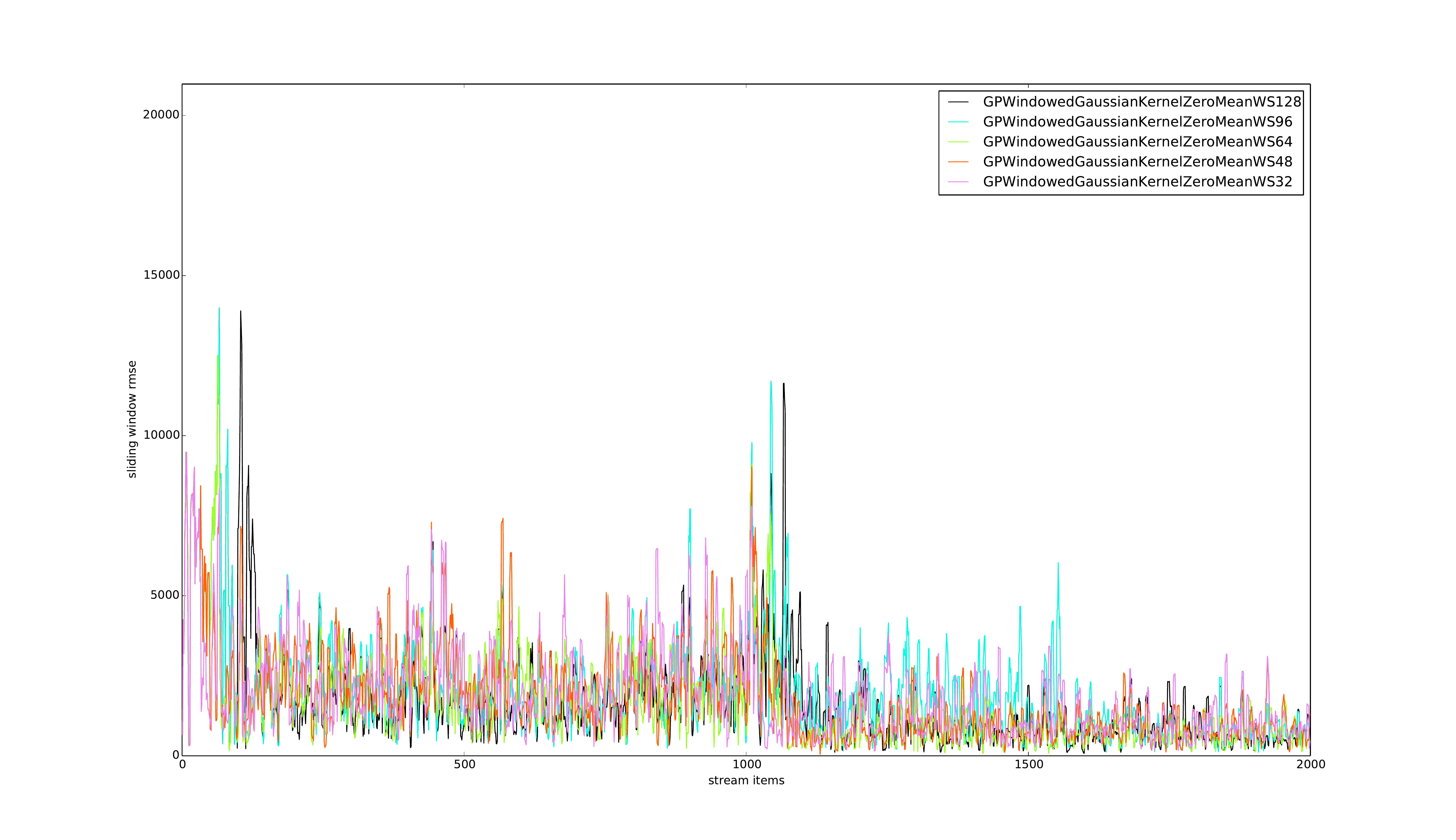}
  \caption{Side-by-side Accuracy Comparison of \texttt{GPWindowedGaussianKernelZeroMean} variants with different window sizes. The test data used is \texttt{SYNTH\_D\_CD\_2000\_4\_10\_1\_24} and the resolution of the sliding error window accuracy evaluation method is set to 4}
  \label{fig:wsize_on_stabilization2_sidebyside_comp_res4}
\end{figure}

Another observation could be made from \ref{fig:wsize_on_stabilization_sidebyside_comp_res96} and \ref{fig:wsize_on_stabilization_sidebyside_comp_res4} is that the errors that the predictions of the online learner with the biggest window size contained during its unstable phase seems to be larger than that of the learners with smaller window sizes. However, this is an illusion of the sliding error window accuracy evaluation method employed for plotting the accuracy as a function of time. If we set the resolution parameter to a relatively low value, we do not get any clear picture of which learner have suffered from greater errors during their unstable phases as seen in \ref{fig:wsize_on_stabilization_sidebyside_comp_res4} and \ref{fig:wsize_on_stabilization2_sidebyside_comp_res4}.

\begin{figure}[htbp]
  \centering
    \includegraphics[width=\linewidth]{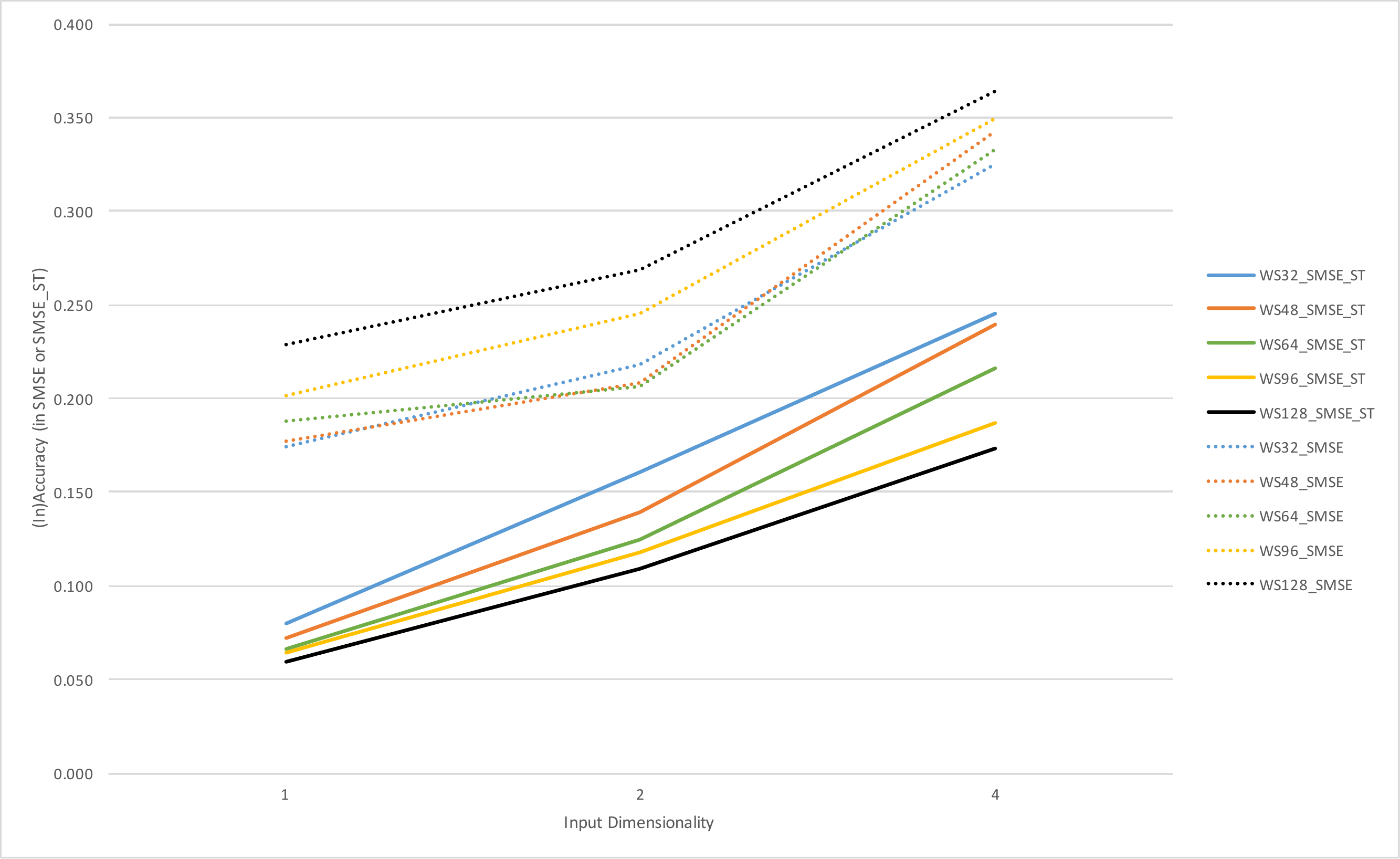}
  \caption{Comparison of the increase in accuracy for different window sizes. \texttt{SMSE\_ST} values are aggregated over 576 stream simulations on 8 online learners (2 BayesianMAP, 2 BayesianMLE, 3 GPRegression and 1 KernelRegression variants) for each window size}
  \label{fig:curse_of_dim}
\end{figure}

Examining how the aggregate \texttt{SMSE\_ST} statistics change as the number of predictor variables increases. Figure \ref{fig:curse_of_dim} clearly shows that both statistics increase (meaning stable and total accuracy decreases) when there are more number of predictors. In the same figure, we also see that the learners with the smallest sliding window achieved a better stable accuracy on single dimensional data (one predictor) than the learners with the biggest sliding window on both two and four dimensional data. In other words, even though with the window size is increased from $36$ to $128$, when the number of dimensions are doubled, the accuracy drops. This suggests that increasing dimensions in the input space are not compensated by increasing the \textit{dynamic} training set as the same rate. This observation is a manifestation of \textit{Curse of Dimensionality}.

Similarly to \texttt{SMSE\_ST}, the change in \texttt{SMSE} results follow the same increasing pattern. However, as \texttt{SMSE} results are contaminated by the high errors during the unstable phases of online learners, drawing conclusions from them is less preferable to using \textit{SMSE\_ST} results if it is a relation between an environmental variable (e.g input dimensionality, input noise etc) and the accuracy in general that is being investigated. Nevertheless, \texttt{SMSE} results are also shown in \ref{fig:curse_of_dim} with by the dotted lines for the records.

The sliding window size affects the stream processing rate of online learners which are important considerations when learning from streams due to input-size dependent time complexity of learning algorithms. As described in \ref{Chapter5}, the lifecycle of an online learner featuring a sliding window is complex and depending on its state, it can take different amount of time to process a single stream item. This makes it harder to evaluate time-efficiency of the online learners. Thus, first a good time-efficiency evaluation strategy is discussed next, considering the variable state of the learners, then this strategy is applied to experiment data to make observations about the sliding window effects on the time-efficiency.

Whenever a significant error increase in the predictions is detected by its update mechanism, its update costs dominate its prediction cost (until it recycles one full window of data points and stabilizes) resulting in a significant amount of time spent on incorporating new points besides making predictions for new data points. This causes the average time to process data points to be higher than the average prediction time in the online algorithms in general. Thus, it can be argued that the pace at which an online learner with error-triggered update mechanism can learn is subject to changes. Therefore, when calculating the maximum data rate that it can learn from, its slowest processing speed should be taken into account. However, this would be a suitable time-efficiency evaluation strategy if the most naive approach dealing with streams is employed. A smarter one would be calculating the maximum stream rate according to the weighted average of the prediction, update and tuning times (weights being the number of times each operation took place until the point an evaluation is requested) and for the periods when the online learner is doing multiple operations on the incoming data stream items making use of a \textit{buffer} to be freed again once the online learner stabilizes on a concept and stops updating and tuning. Another sophisticated one would be using multi-threading to carry out update and tuning operations while new data points are arriving and being predicted. However, in this thesis, by assuming the availability of efficient stream buffering and multi-threading techniques to handle streams, for evaluating the consumption of time resources by the online learners, the weighted average of the prediction, update and tuning average is used as the criterion. In the testing setup employed, an easy way to compute the mentioned weighted average is simply to divide the total time spent on the stream by the learner by the number of data points\footnote{\label{not_violate_assumption}Note that this does not violate the \textit{possibly infinitely long} assumption of the data streams. only the number of stream items until an evaluation on the learner's performance is requested should be known to compute the weighted average.}. Thus, the time-efficiency metric \texttt{ATPI} is the major point of interest in the following time-efficiency analysis of the experiment data although in order to point out remarkable differences in prediction, update and tuning times separately between different learners with different algorithms and/or different window sizes \texttt{APT}, \texttt{TPT}, \texttt{AUT}, \texttt{TUT}, \texttt{ATT}, and \texttt{TTT} metrics are also used.

\begin{figure}[htbp]
  \centering
    \includegraphics[width=\linewidth]{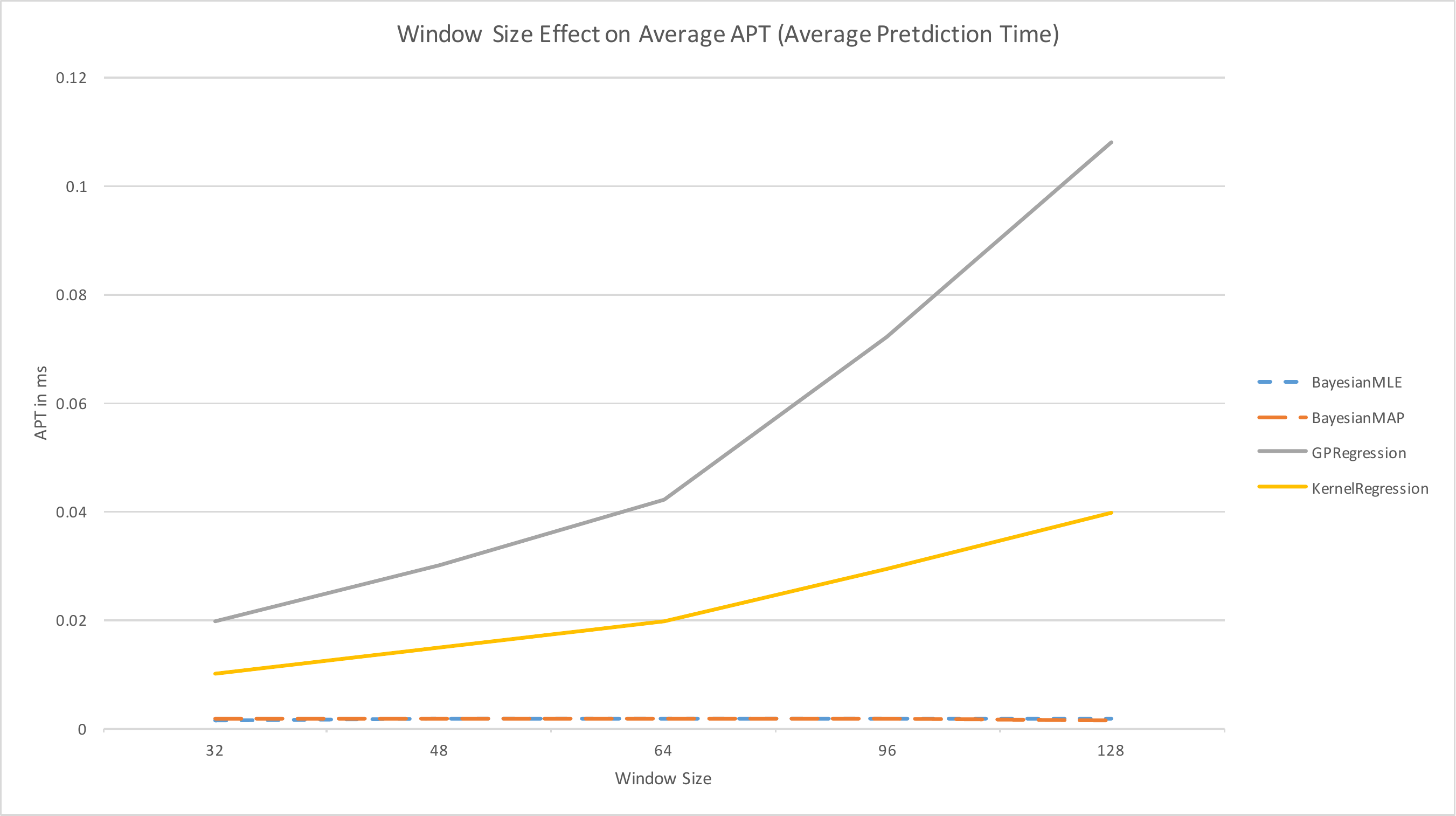}
  \caption{Comparison of the increase in average \texttt{APT} for different window sizes and algorithms. APT values are aggregated over 144, 216 and 216 streams which are simulated from data sets with 1,2 and 4 input dimensions respectively on 8 sliding windowed online learning algorithms for each window size}
  \label{fig:ws_on_apt_per_alg}
\end{figure}

\begin{figure}[htbp]
  \centering
    \includegraphics[width=\linewidth]{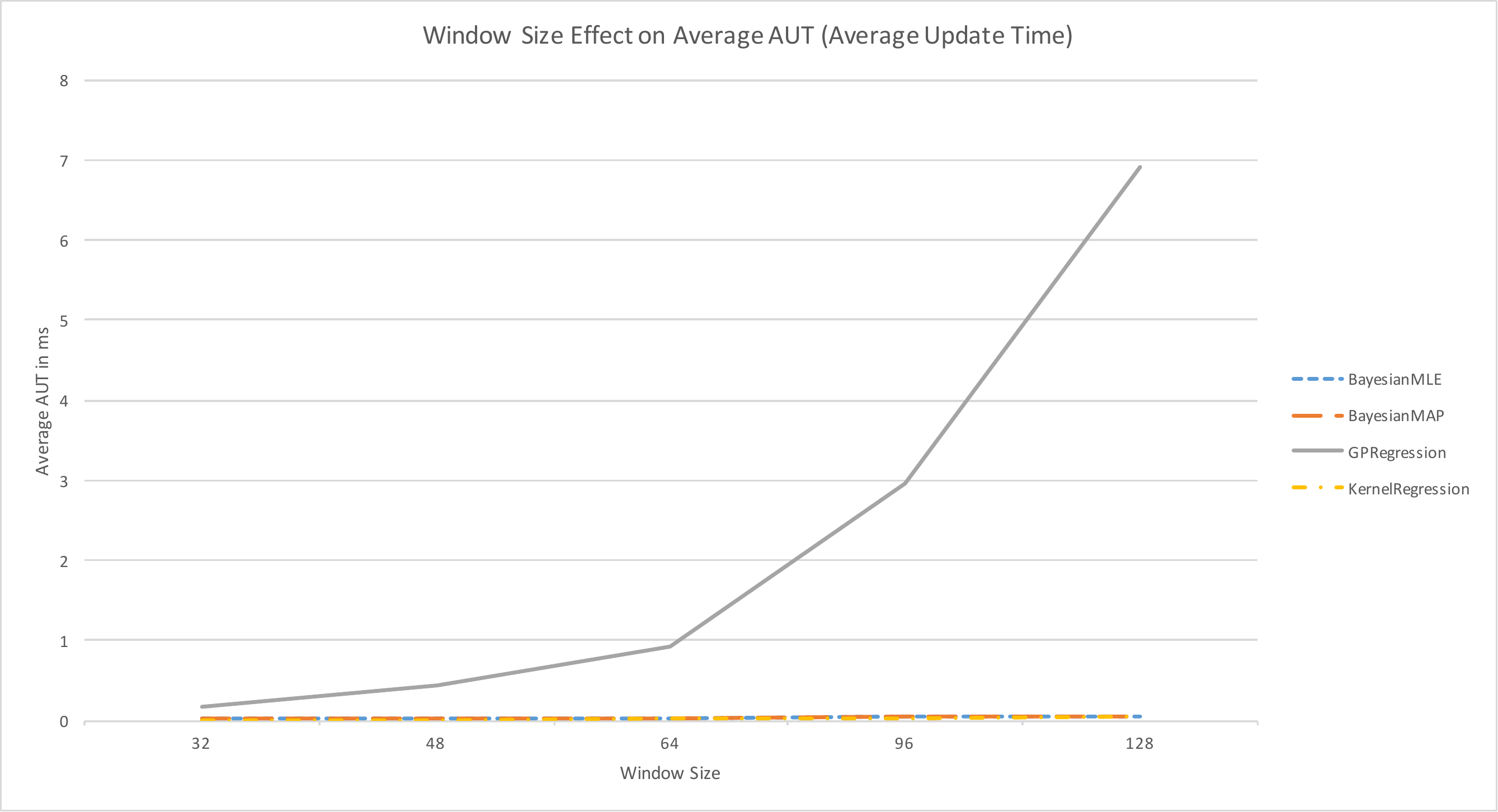}
  \caption{Comparison of the increase in average \texttt{AUT} for different window sizes and algorithms. APT values are aggregated over 576 stream simulations on 8 online learners (2 BayesianMAP, 2 BayesianMLE, 3 GPRegression and 1 KernelRegression variants) for each window size}
  \label{fig:ws_on_aut_per_alg}
\end{figure}

In \ref{fig:ws_on_apt_per_alg} and \ref{fig:ws_on_aut_per_alg}, we see how the average prediction time and the average update time respectively changes as the window size changes. Generally speaking, windows size comes at the cost of a slow-down in both prediction and update times however the amount of slow-down depends heavily on the algorithm choice. For example, \texttt{BayesianMLE} and \texttt{BayesianMAP} families are almost window-size insensitive in terms of prediction speed while \texttt{KernelRegression} and \texttt{GPRegression} learners exhibit a significant slow-down with larger window-sizes. As for the update speed, the amount of slow-down with all the algorithms but \texttt{GPRegression} are on the scale of $\mu$ seconds. On the other hand, \texttt{GPRegression} slow-down is measured in $m$ seconds making the points representing the average update time for all the algorithms except for \texttt{GPRegression} connect with an almost horizontal line. 

\begin{figure}[htbp]
  \centering
    \includegraphics[width=\linewidth]{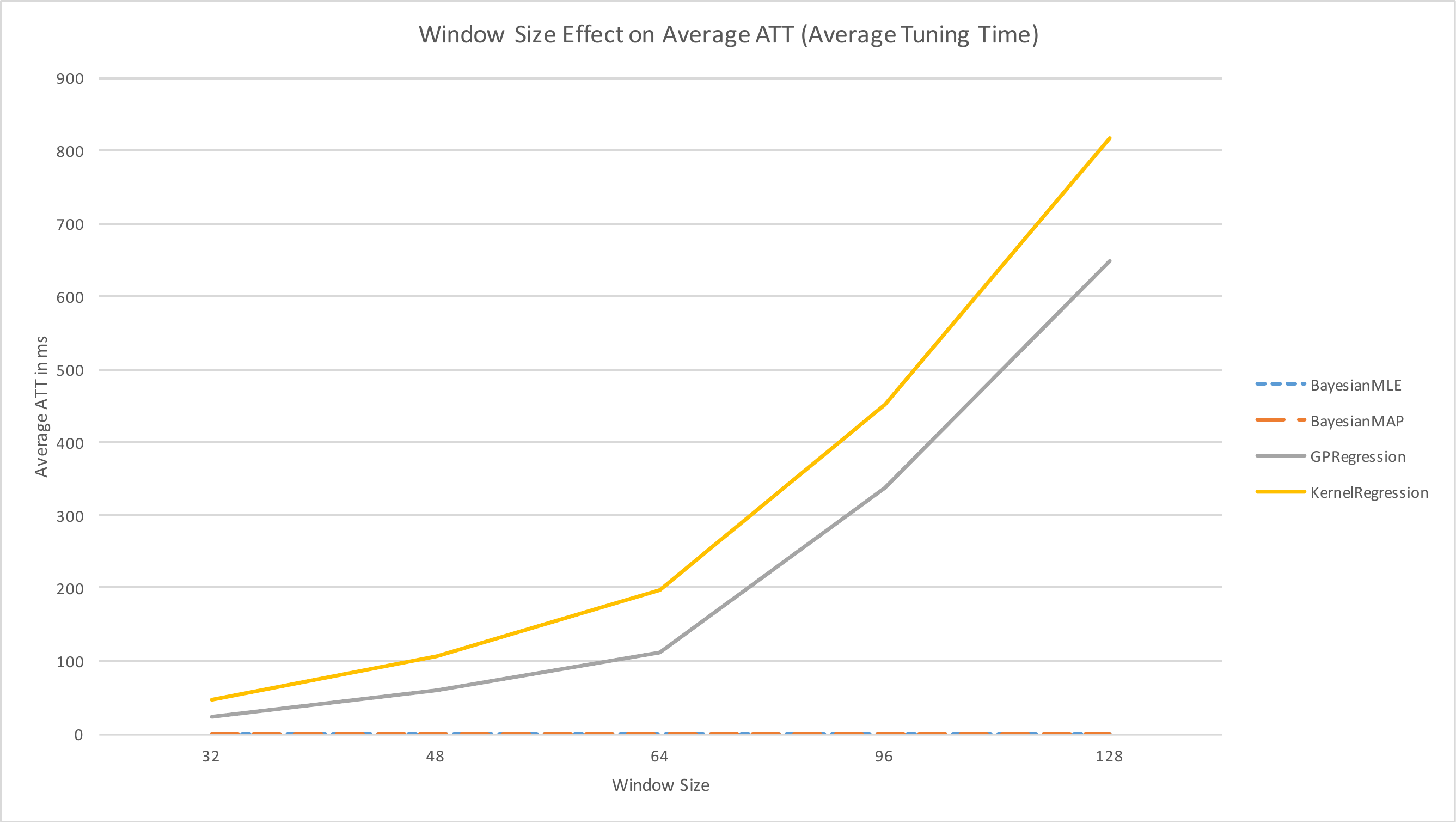}
  \caption{Comparison of the increase in average \texttt{ATT} for different window sizes and algorithms. APT values are aggregated over 576 stream simulations on 8 online learners (2 BayesianMAP, 2 BayesianMLE, 3 GPRegression and 1 KernelRegression variants) for each window size}
  \label{fig:ws_on_att_per_alg}
\end{figure}

As the online learners with sliding window mechanism are tuned at least once after their window gets full and later conditional on the increasing error, tuning time is an important consideration. Figure \ref{fig:ws_on_att_per_alg} shows how the average \texttt{ATT} changes as the window size increases. \texttt{BayesianMLE} and \texttt{BayesianMAP} family learners appear to be insensitive to window size on the $ms$ scale. As for the learners using non-parametric algorithms exhibits a dramatic increase in their average tuning times. This is not surprising as explained in \ref{subsubsection:impl_tune_gpreg}, the tuning algorithm employed in \texttt{GPRegression} learners attempts to find the optimal hyperparameter configurations by manipulating the the gram-matrices that store as many items as the square of the number of items in the sliding window and as explained in \ref{subsubsection:impl_tune_kreg}, the tuning algorithm used by \texttt{KernelRegression} does a number of passes on the sliding-window which is proportional to the square of the sliding window size. On an important note, \texttt{KernelRegression} tuning routine takes more time than that of \texttt{GPRegression} according to averaged test results.

\begin{figure}[htbp]
  \centering
    \includegraphics[width=\linewidth]{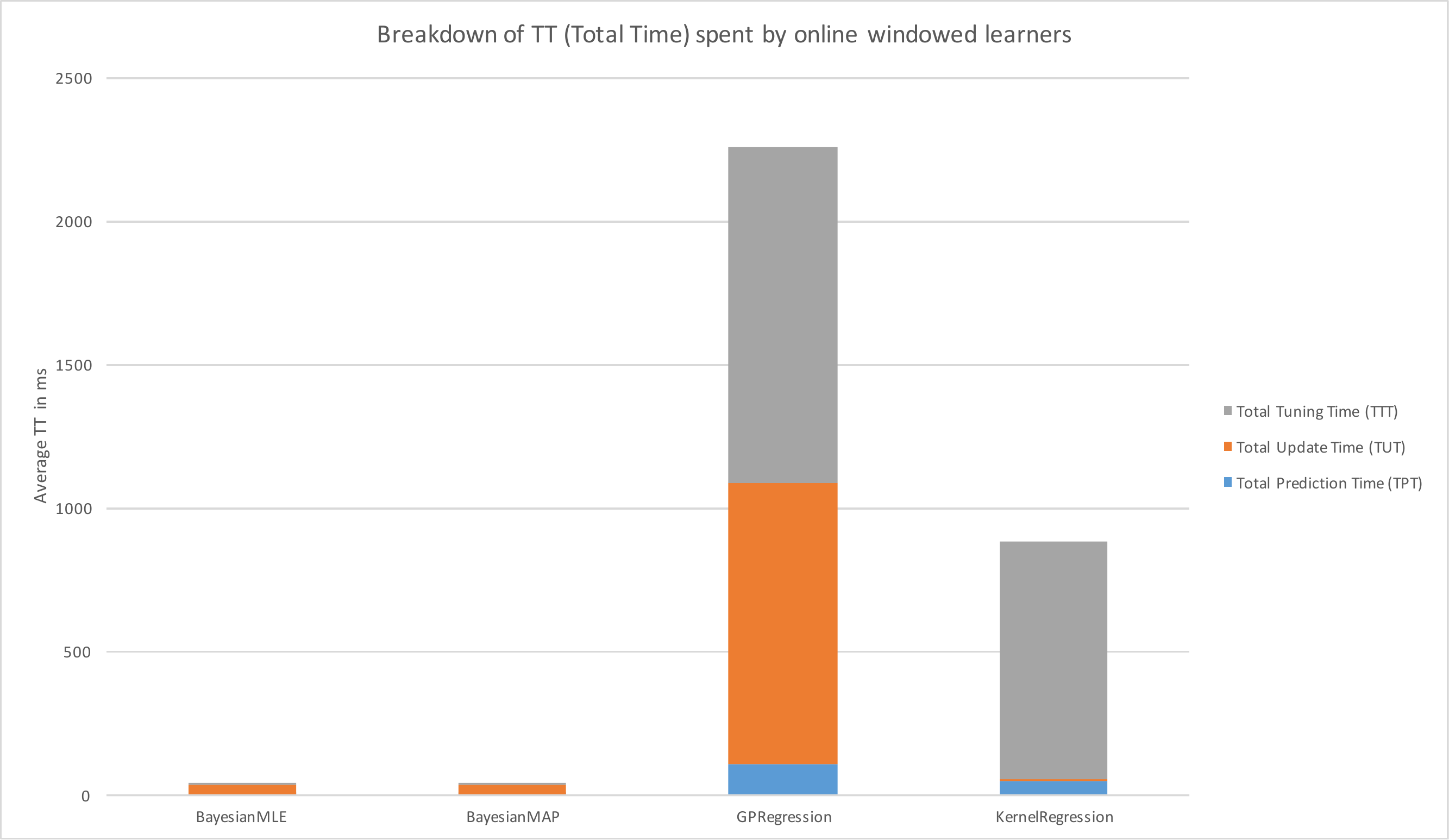}
  \caption{Average Total time spent per stream simulation shown as the summation of average prediction, update and tuning times. Sum of TPT, TUT and TTT are aggregated over 576 stream simulations on 6 BayesianMLE learners (2 variant parametrized with 3 window sizes), 6 BayesianMAP learners (2 variant parametrized with 3 window sizes), 9 GPRegression learners (3 variant parametrized with 3 window sizes) and 3 KernelRegression learners (Single variant parametrized with 3 window sizes)}
  \label{fig:total_time_breakdown}
\end{figure}

In order to see how each of the different operations, prediction, update and tuning, contributes to the total time spend by an online learner, a stacked column graph showing the breakdown of the total time spent is presented in \ref{fig:total_time_breakdown}. \texttt{GPRegression} and \texttt{KernelRegression} families spent substantial amount of time for tuning on average. As for \texttt{BayesianMLE} and \texttt{BayesianMAP} families, total time spent for tuning was negligible according to aggregated test results.

Having investigated how different window sizes affect different dimensions of the evaluation such as accuracy and the time efficiency, the conclusion about the window size to be made is that it has to be chosen very carefully. While with small window sizes, the high speed is guaranteed, big window size does not guarantee higher general accuracy always. Nevertheless, there is a sweet point between high window sizes and low window sizes that need to be picked in order to achieve a good accuracy without suffering from substantial slow-down that prevents the algorithm to be employed from the stream learning scenarios. However, where this sweet point lies depends on the algorithm choice. 

\subsubsection{\texttt{GPRegression} Window Size}
\label{section:gp_reg_win_size}

\begin{figure}[htbp]
  \centering
    \includegraphics[width=\linewidth]{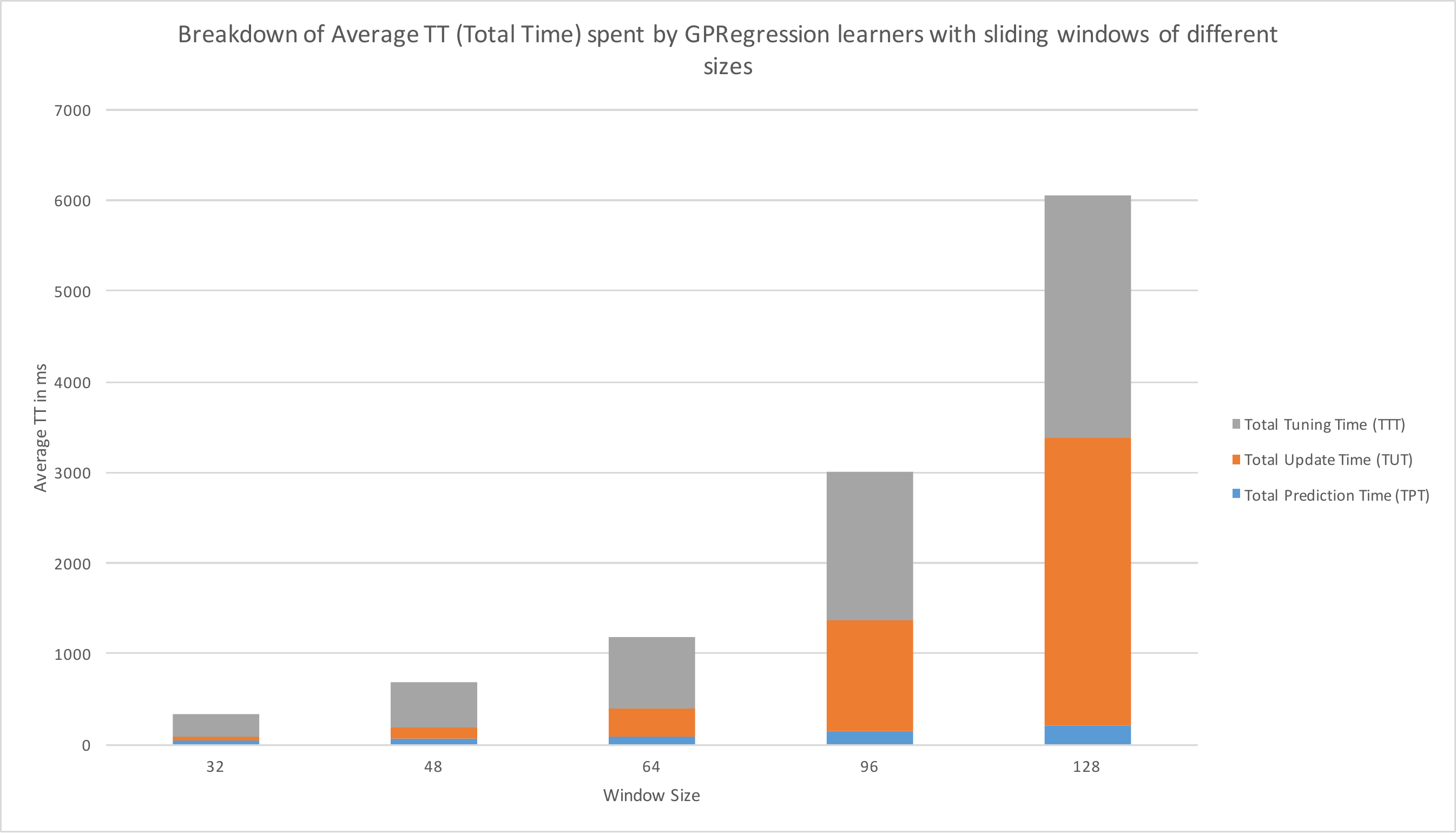}
  \caption{Average Total time spent per stream simulation shown as the summation of average prediction, update and tuning times. Sum of TPT, TUT and TTT are aggregated over 576 stream simulations on 3 KernelRegression variants for each different window sizes)}
  \label{fig:ws_on_gpreg_tt}
\end{figure}

As shown in \ref{fig:ws_on_gpreg_tt}, total time spent by the \texttt{GPRegression} learners grows with the increasing window size. \texttt{ATPI} statistic which can be used to calculate the maximum data rate an online learner can handle as explained previously is obtained by simply dividing the \texttt{TT} by the number of data points appeared in the stream until the performance evaluation of the online learner is requested. In the test setup, the streams are simulated from $2000$ data points, in this case, if \texttt{TT} is known, \texttt{ATPI},  maximum data rate, denoted as \texttt{DRmax} and measured in items per $ms$ can be handled by an online learner, can be obtained using the following formula.
\begin{flalign} 
& \texttt{ATPI} = \frac{\texttt{TT}}{2000} \\
& \texttt{DRmax} = \texttt{ATPI}^{-1}
\end{flalign}

\begin{figure}[htbp]
  \centering
    \includegraphics[width=\linewidth]{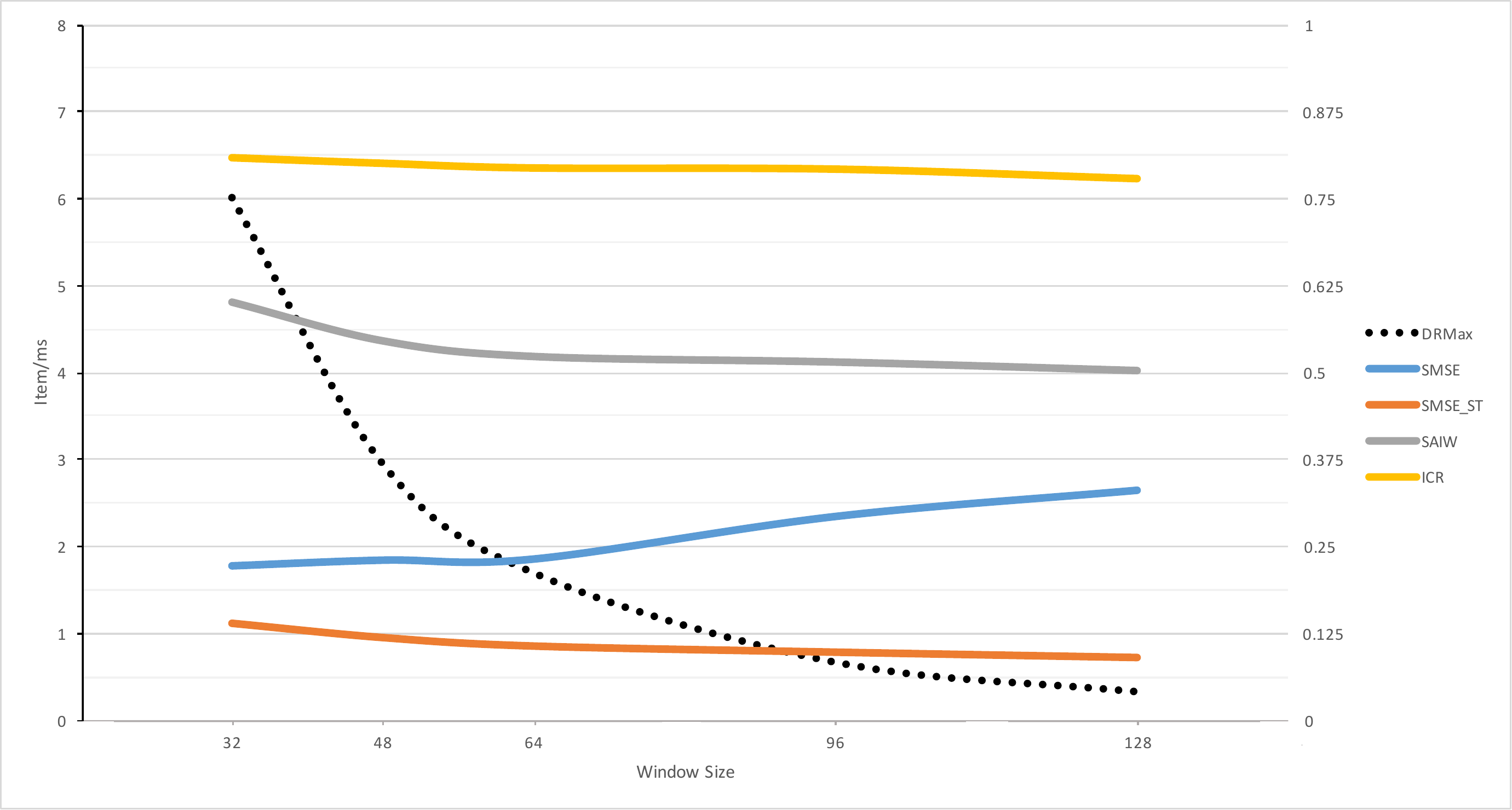}
  \caption{Average \texttt{DRmax}, \texttt{SMSE}, \texttt{SMSE\_ST}, \texttt{SAIW} and \texttt{ICR} scores for \texttt{GPRegression} learners are graphed. First statistic \texttt{DRmax} uses the primary axis with the interval $[0,8]$ while the rest use the secondary y- axis with the $[0,1]$ interval. The statistics are aggregated from 576 stream simulations on 3 KernelRegression variants for each different window sizes)}
  \label{fig:gpreg_wsize_sweet_pt}
\end{figure}

Applying above formula to the \texttt{GPRegression} learners to obtain the average \texttt{DRMax} scores of them over different tests and three different variants of \texttt{GPRegression} and using the average \texttt{SMSE} presented previously, we get Figure \ref{fig:gpreg_wsize_sweet_pt} which helps spot the sweet point for the size of the sliding window that \texttt{GPRegression} learners have. The figure indicates that the prediction bound coverage are nearly the same for all the window sizes. In terms of the average width of prediction bounds, all the window sizes produced reasonably sized bounds although the prediction bounds of the learners with the smallest choice of window size was a bit larger than the rest. Having verified that prediction bounds and their coverage are good for all window sizes, the trade-off between accuracy and the maximum data rate can be handled is the major consideration to pick a good window size. 

In the same graph, we see that there is actually no trade-off between the general accuracy \texttt{SMSE} and \texttt{DRmax}. The smallest window size appears to dominate the other window sizes by providing the lowest \texttt{SMSE} and highest \texttt{DRmax} values. However, when evaluating the \texttt{SMSE} scores one should keep in mind that in addition to the window size, the general accuracy scores depend on the number of concept drifts and the number of stream items arrived until the moment online learner performance statistics are computed in the lifetime of the stream. In the stream simulations, these parameters are all fixed. As explained previously in this chapter, half of the stream simulations feature an abrupt concept drift at the $1000_{th}$ data point. In a real-life stream learning scenario, it is not possible to know how many concept changes how frequently will occur. Therefore, instead of \texttt{SMSE}, looking at the \texttt{SMSE\_ST} scores which is a measure of the stable (in)accuracy while still keeping in mind that high window sizes have longer adaptation times is better when finding the sweet point for the window sizes. 

When the \texttt{SMSE\_ST} scores are examined in \ref{fig:gpreg_wsize_sweet_pt}, there indeed seems to be a true trade-off between the stable accuracy and maximum data rate. Higher the stable accuracy is lower the maximum data rate is. The highest window sizes namely $96$ and $128$ having the superior stable accuracy to the other window sizes can handle streams up to approximately $0.33$ and $0.67$ items per millisecond. Since, the Ocelot runtime predictor is expected to get performance estimation requests once in a millisecond on average, window size of $96$ and $128$ are not good choices. After $96$ and $128$, the window size choice of $64$ has the best stable accuracy and it allows to process stream items approximately at $1.69$ items per millisecond meeting the data rate requirement. Therefore, the sweet point for the window size choice in \texttt{GPRegression} learners is determined to be $64$.

\subsubsection{\texttt{KernelRegression} Window Size}

\begin{figure}[htbp]
  \centering
    \includegraphics[width=\linewidth]{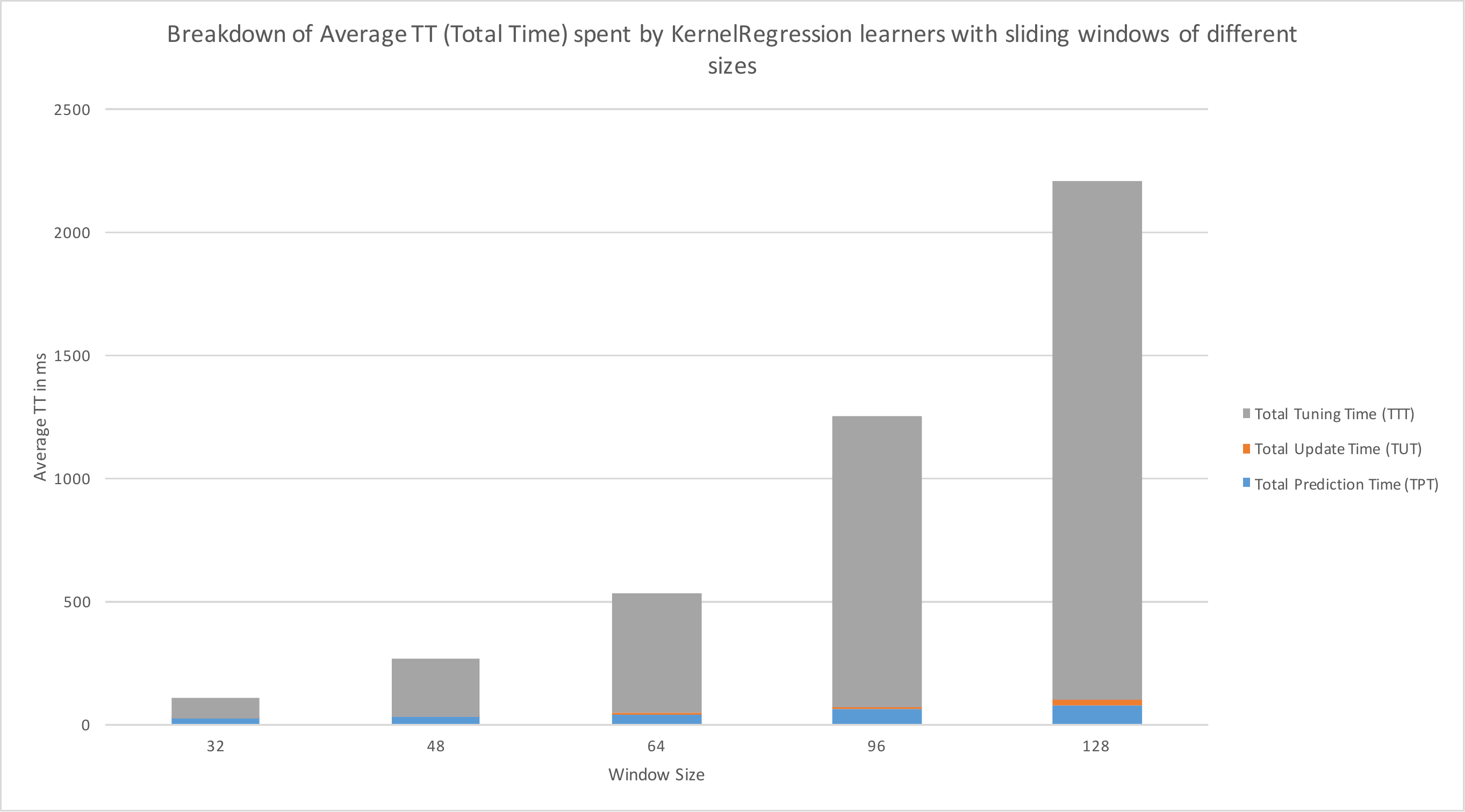}
  \caption{Average Total time spent per stream simulation shown as the summation of average prediction, update and tuning times. Sum of TPT, TUT and TTT are aggregated over 576 stream simulations on the only existing KernelRegression variant for each different window sizes)}
  \label{fig:ws_on_kreg_tt}
\end{figure}

For \texttt{KernelRegresison} learners, Figure \ref{fig:ws_on_kreg_tt} allows for a detailed analysis on their time-efficiency. Similarly to \texttt{GPRegression} learners, the average total time spent grows with the increasing window size although differently from \texttt{GPRegression} the biggest contribution to the total time spent is from the tuning costs. Also similarly to \texttt{GPRegression}, the average \texttt{ICR} scores appear to be insensitive to the window size choice. However, they show that the interval coverage rate of \texttt{KernelRegression} is not sufficient. Nevertheless, when deciding for a good window size, this deficiency is not considered and in the next section a good workaround for it is proposed. As for the average \texttt{SAIW} scores, except for the lowest choice of window size $32$, they seem to be pretty low meaning that prediction bounds were tight which is a desirable result in general. Moreover, with the highest window size option $128$, \texttt{SAIW} shows prediction bounds were even tighter.

The discussion in \ref{section:gp_reg_win_size} regarding the choice between \texttt{SMSE} and \texttt{SMSE\_ST} as the (in)accuracy criteria also holds here. When the average \texttt{SMSE\_ST} and \texttt{DRmax} scores are compared between different window sizes, performance-accuracy trade-off catches attention. Simply, the higher stable accuracy, lower the maximum data rate can be handled. \texttt{GPRegression} learners with the largest window size choice $128$ can process stream items approximately at $0.95$ items per millisecond. This is just below the expected data rate in Ocelot operator runtime performance behavior. Also considering the evident late-stabilization times that large-windowed learners have, the largest window size among the other options is not the best choice. The second biggest window-size among the available options $96$ can handle stream items arriving $1.69$ items per millisecond making it employable for Ocelot operator runtime predictor.
\begin{figure}[htbp]
  \centering
    \includegraphics[width=\linewidth]{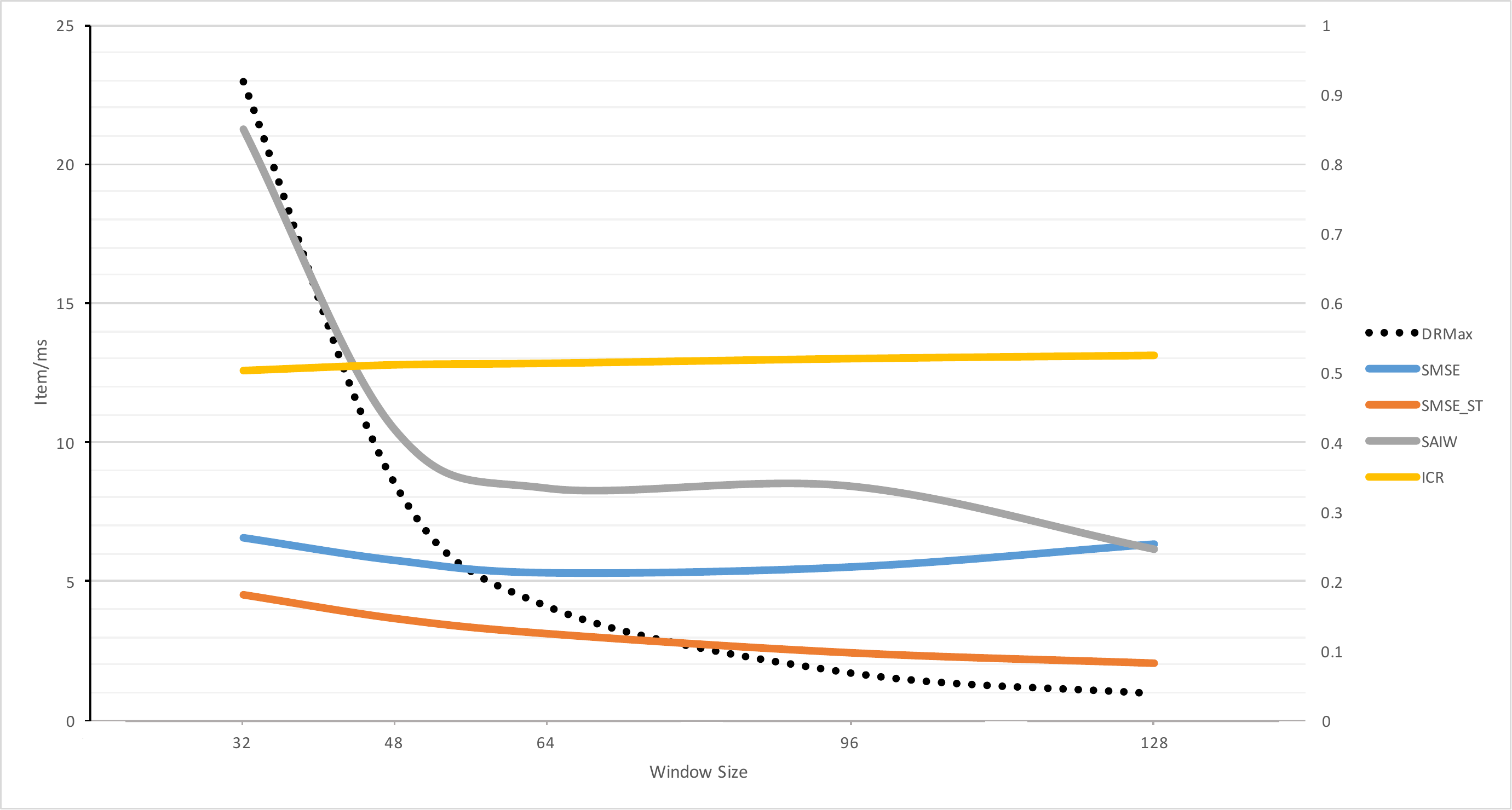}
  \caption{Average \texttt{DRmax}, \texttt{SMSE}, \texttt{SMSE\_ST}, \texttt{SAIW} and \texttt{ICR} scores for \texttt{GPRegression} learners are graphed. First statistic \texttt{DRmax} uses the primary axis with the interval $[0,25]$ while the rest use the secondary y- axis with the $[0,1]$ interval. The statistics are aggregated from 576 stream simulations on 3 KernelRegression variants for each different window sizes)}
  \label{fig:kreg_wsize_sweet_pt}
\end{figure}

When the biggest window size among the available options are employed, KernelRegression is still quicker than the GPRegression with the window size of $64$. Despite this, it may not be the best option to stick with the largest choices for the window size due to the accuracy problems that sliding windowed algorithms with big windows have more during their are unstable periods than their smaller windowed versions. Hence, the preferred window sizes for the \texttt{KernelRegression} is $64$ and $96$.

\subsubsection{\texttt{BayesianMLE} and \texttt{BayesianMAP} Window Size}

\begin{figure}[htbp]
  \centering
    \includegraphics[width=\linewidth]{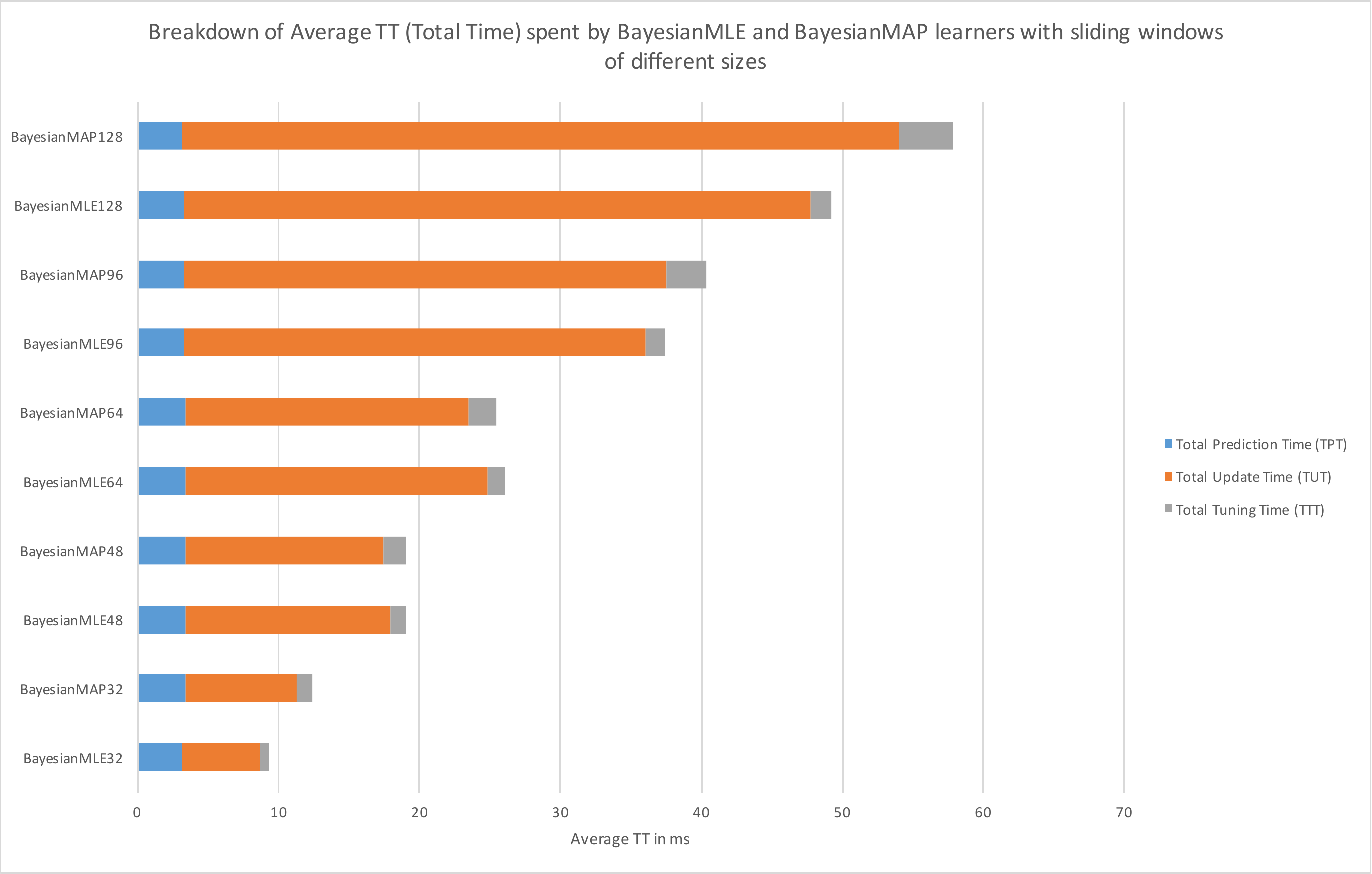}
  \caption{Average Total time spent per stream simulation shown as the summation of average prediction, update and tuning times. Sum of TPT, TUT and TTT are aggregated over 576 stream simulations on 2 BayesianMLE and 2 BayesianMAP variants with three different window sizes each)}
  \label{fig:ws_on_bmle_bmap_tt}
\end{figure}

Although the same relation between the window size and the total time spent observed in parametric learners is also present in the case of non-parametric learners, in \ref{fig:ws_on_bmle_bmap_tt}, we see that even the learners with the window size of $128$ is way quicker than their counterparts from other algorithms with the window size of $32$. As a result, the streams that can be handled with parametric learners can be way faster. For example, \texttt{BayesianMAP} learner with the sliding window of $128$ data points can handle streams up to $(\frac{57.87}{2000})^{-1}\approx 34.56$ items per millisecond. Therefore, performance-wise, any window-size from 32 to 128 is fine for \texttt{BayesianMLE} and \texttt{BayesianMAP} to be employed as the runtime predictor module in Ocelot.

\begin{figure}[htbp]
  \centering
    \includegraphics[width=\linewidth]{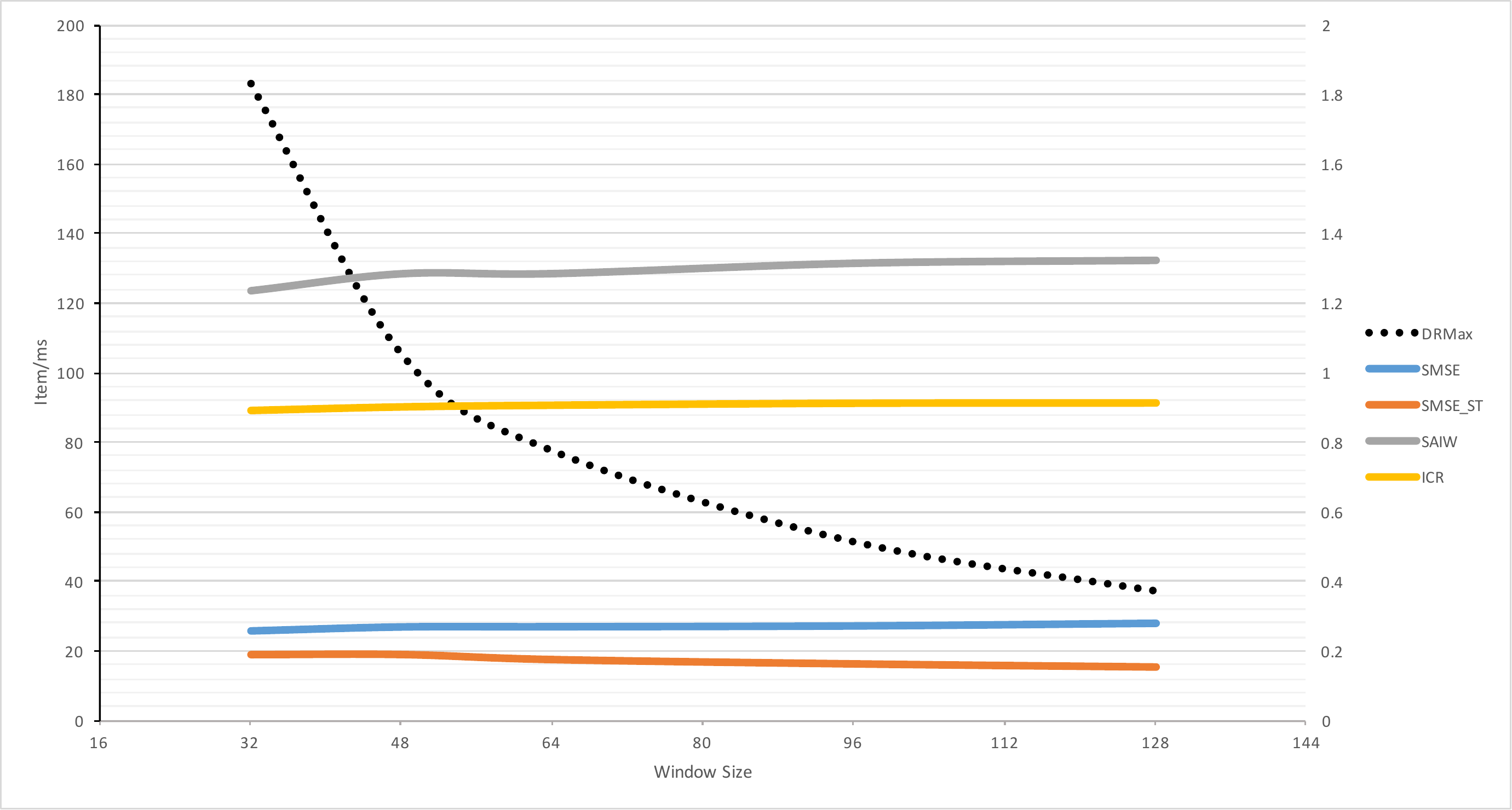}
  \caption{Average \texttt{DRmax}, \texttt{SMSE}, \texttt{SMSE\_ST}, \texttt{SAIW} and \texttt{ICR} scores for \texttt{GPRegression} learners are graphed. First statistic \texttt{DRmax} uses the primary axis with the interval $[0,200]$ while the rest use the secondary y- axis with the $[0,2]$ interval. The statistics are aggregated from 576 stream simulations on 2 \texttt{BayesianMLE} and 2 \texttt{BayesianMAP} variants for each different window sizes)}
  \label{fig:gpreg_wsize_sweet_pt}
\end{figure}

In terms of prediction bound related statistics namely \texttt{ICR} and \texttt{SAIW}, window size appeared to have no effect. Therefore, these metrics are not considered when choosing the window-size.

Since all the window sizes offer satisfactory processing speeds, there is no trade-off between the accuracy and the performance. One can simply opt for the the window size that affords the best performance. However, the inherited unstable phase accuracy drop problem is more troublesome with the larger window sizes similar to any other sliding-windowed learner. Therefore, it is not wisest choice to opt for the highest window size possible. Thus the moderate sizes such as $64$ and $96$ are considered for the further evaluation.

\subsection{Forgetting-factor}

\begin{figure}[htbp]
  \centering
    \includegraphics[width=\linewidth]{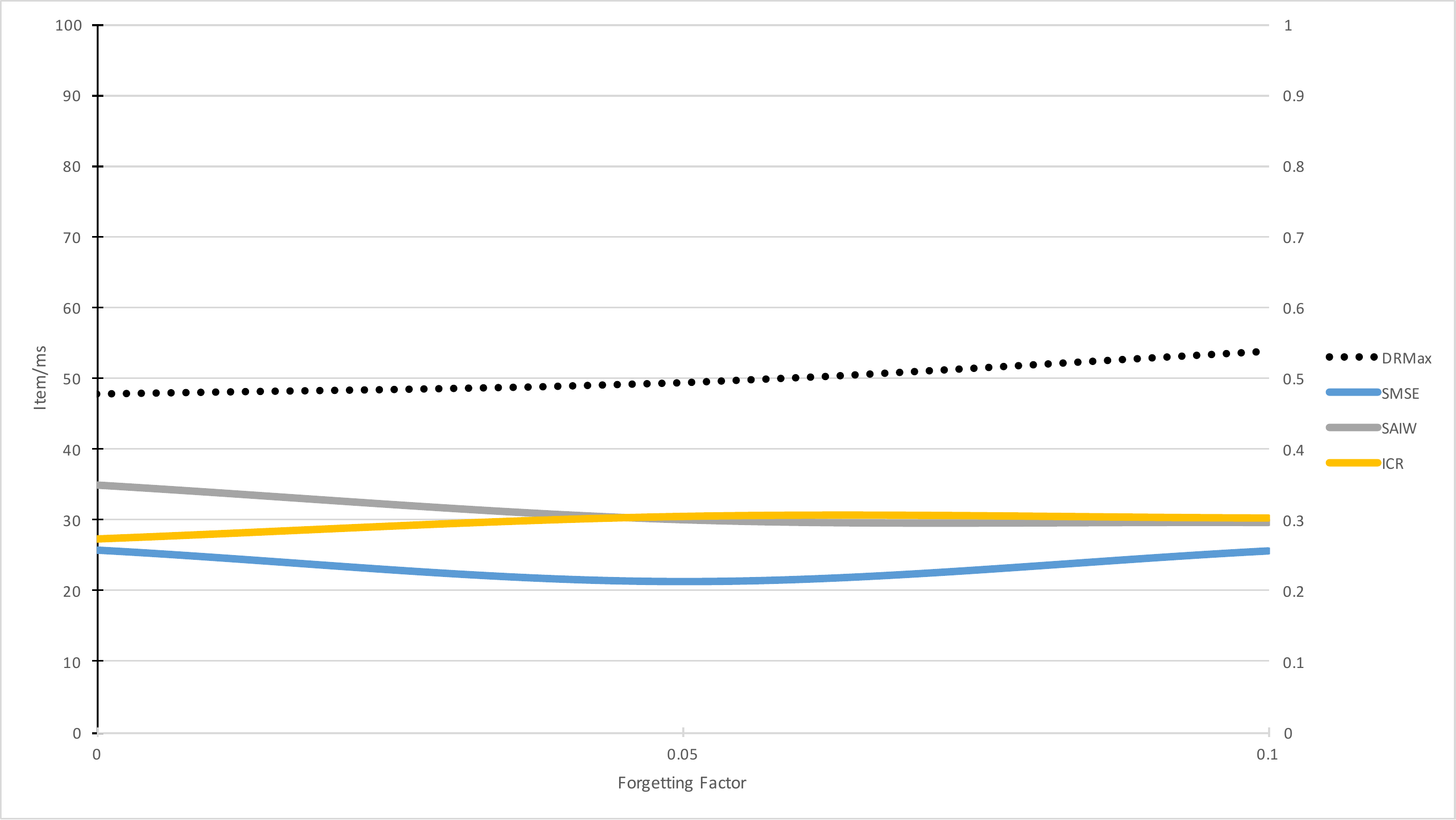}
  \caption{Average \texttt{DRmax}, \texttt{SMSE}, \texttt{SAIW} and \texttt{ICR} scores for \texttt{BayesianMAP} and \texttt{BayesianMLE} learners are graphed. First statistic \texttt{DRmax} uses the primary axis with the interval $[0,100]$ while the rest use the secondary y- axis with the $[0,1]$ interval. The statistics are aggregated from 576 stream simulations on 2 \texttt{BayesianMLE} and 2 \texttt{BayesianMAP} variants for each forgetting factors)}
  \label{fig:bmle_bmap_ff_sweet_pt}
\end{figure}

A similar parameter to the sliding window size for the online learners that do not feature a sliding window is the forgetting factor. Out of 52 online learner, 12 use a forgetting mechanism to adapt to potential concept drifts. Similar to sliding-windowed approaches, the forgetting factor, being the essential parameter in forgetting-factor based learners is expected to affect the evaluation metrics. In order to discover how it affects the learning performance (accuracy, time costs, prediction bounds quality) of the forgetting-windowed learners, the figure \ref{fig:bmle_bmap_ff_sweet_pt} is prepared.

According to the test results presented in the graph, the average prediction bounds scores namely \texttt{SAIW} and \texttt{ICR} were almost the same for the forgetting factors $0.05$ and $0.1$. For the $0$, the average \texttt{SAIW} was higher and the average \texttt{ICR} was lower than those of the learners with other forgetting factors although the differences were not significantly high. As for the performance implications, the change in forgetting factor did not cause a remarkable change in the maximum data rate can be handled.

The evaluation dimensions such as time-efficiency and the prediction bounds did not give a clue which forgetting factor is a better choice. In terms of the accuracy, similarly to the window-size parameter, there seems to be a sweet point for the forgetting factor to be chosen. Among the three forgetting factor choices used in the experiments namely $0$, $0.05$ and $0.1$, the forgetting factor of $0.05$ proved to be the best choice in terms of the accuracy\footnote{Note that there is no distinction of stable and general accuracy in the case of online learners not featuring a sliding window} of the predictions that the learners produced. This is not surprising since at the lower end using $0$ as the forgetting factor means not allowing algorithm to forget anything which obviously makes the adaptation of algorithm to the new concepts last forever. Setting forgetting-factor too high would result in underfacilitating the training examples that the online learner has observed making the algorithm always undertrained. 

As a result, when the experiment data is analyzed, it is observed that the sweet point for the forgetting factor lies somewhere in between $0.0$ and $0.1$. Thus, the forgetting factor $0.05$ works best among the available forgetting factor options.

\subsection{Feature Space Mapping}

As explained in Chapter \ref{Chapter4}, learners using a parametric learning algorithm have the option to map the input data point to a feature space which has a higher dimensionality than the input space. In the subsection, the implications of the feature space mapping on the evaluation criteria namely accuracy, prediction bounds and time-efficiency are discussed.

\begin{figure}[htbp]
  \centering
    \includegraphics[width=\linewidth]{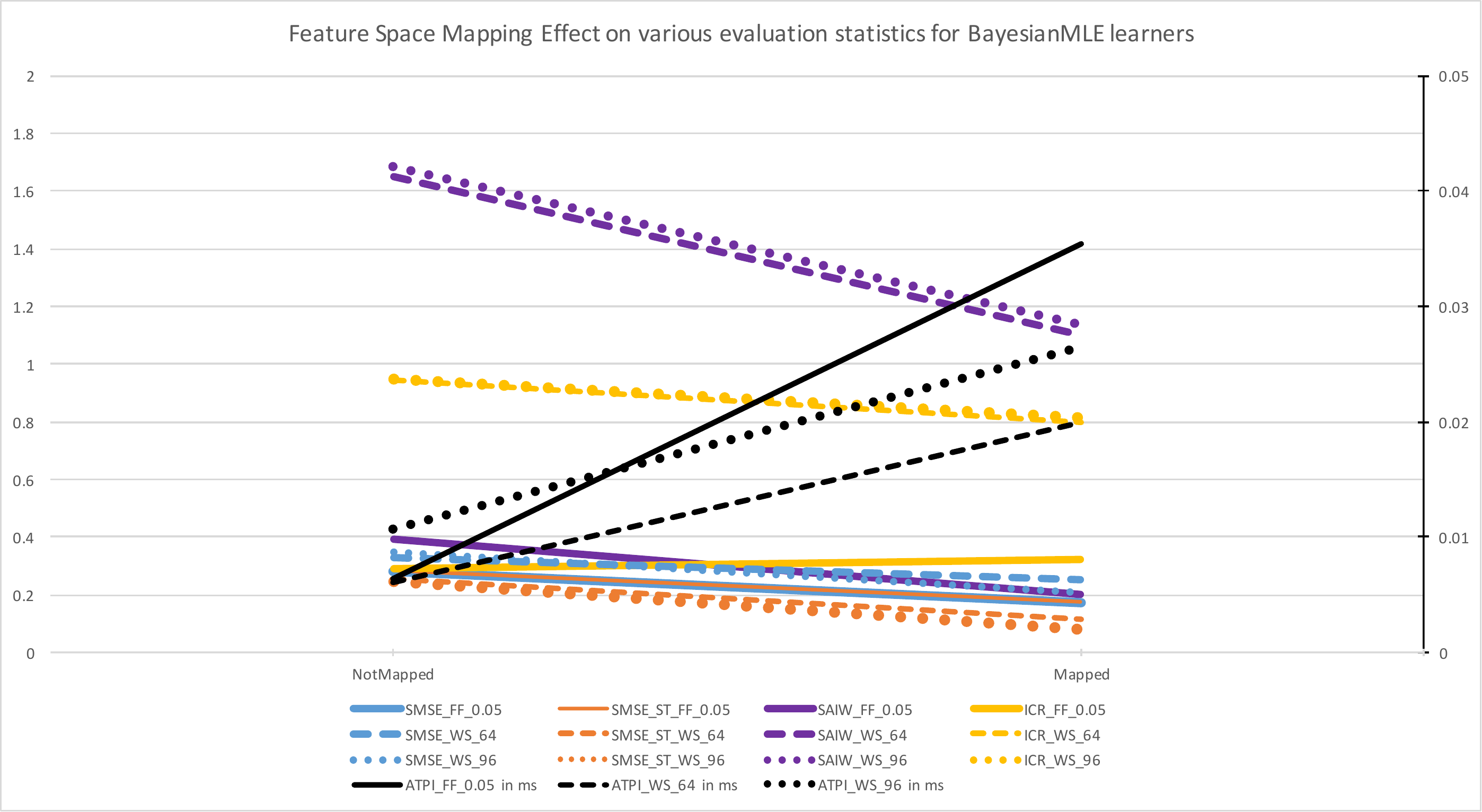}
  \caption{\texttt{SMSE}, \texttt{SAIW},\texttt{ICR} and \texttt{ATPI} scores for 6 different \texttt{BayesianMLE} learners are graphed. \texttt{ATPI} scores measured with respect to the secondary y-axis with the interval $[0,0.005]$ while the rest use the primary y-axis with the $[0,2]$ interval. The statistics are aggregated from 576 stream simulations on \texttt{BayesianMLEForgetting\_FF0.05}, \texttt{BayesianMLEWindowed\_WS64} and \texttt{BayesianMLEWindowed\_WS96} learners (each having one mapped and one non-mapped version)}
  \label{fig:fsm_effect_bmle}
\end{figure}

In Figure \ref{fig:fsm_effect_bmle}, how feature space mapping affects the accuracy, prediction bounds quality and time-efficiency of \texttt{BayesianMLE} learners with the window sizes and forgetting factors chosen in the previous subsections is shown. Except for \texttt{ICR}, feature space mapping has affected the other statistics in the same way for all the \texttt{BayesianMLE} learners analyzed to discover feature space mapping implications. \texttt{SMSE}, \texttt{SMSE\_ST} and \texttt{SAIW} appear to be lower when the feature space mapping is employed. This means, feature-space mapping boosts accuracy and tightens the prediction bounds. In terms of the time-efficiency, feature-space mapping has increased the \texttt{ATPI} meaning that more processing time per stream item is needed with the feature-space mapping. However, the average increased processing times for all the learners considered in the figure are all still below $0.04$ millisecond which is still way lower than that of than non-parametric models.  As for \texttt{ICR}, feature-space mapping caused a slight drop in the case of windowed learners making the coverage of the prediction bounds lower prediction bound coverage. However, the \texttt{ICR} values when the feature space is enabled is above $0.8$ for the windowed learners considered in the graph meaning that with feature space mapping, the the prediction bounds coverage of the unobserved targets are still satisfactory. As for the only forgetting factor variant considered, feature-space mapping has improved its interval coverage although the difference is negligible.

Feature-space mapping is observed to boost accuracy and tighten the prediction bounds while keeping the prediction bound coverage at the acceptable levels. This comes at the cost of slight increase in the average processing time. Having also observed that the mapped \texttt{BayesianMLE} learners are still way faster than what is required (all the mapped versions considered can process data streams faster than $25$ item per ms while the requirement is set to be around $1$ item per ms), \texttt{BayesianMLE} learners with feature-space mapping are preferred over their counterparts not featuring feature-space mapping.

\begin{figure}[htbp]
  \centering
    \includegraphics[width=\linewidth]{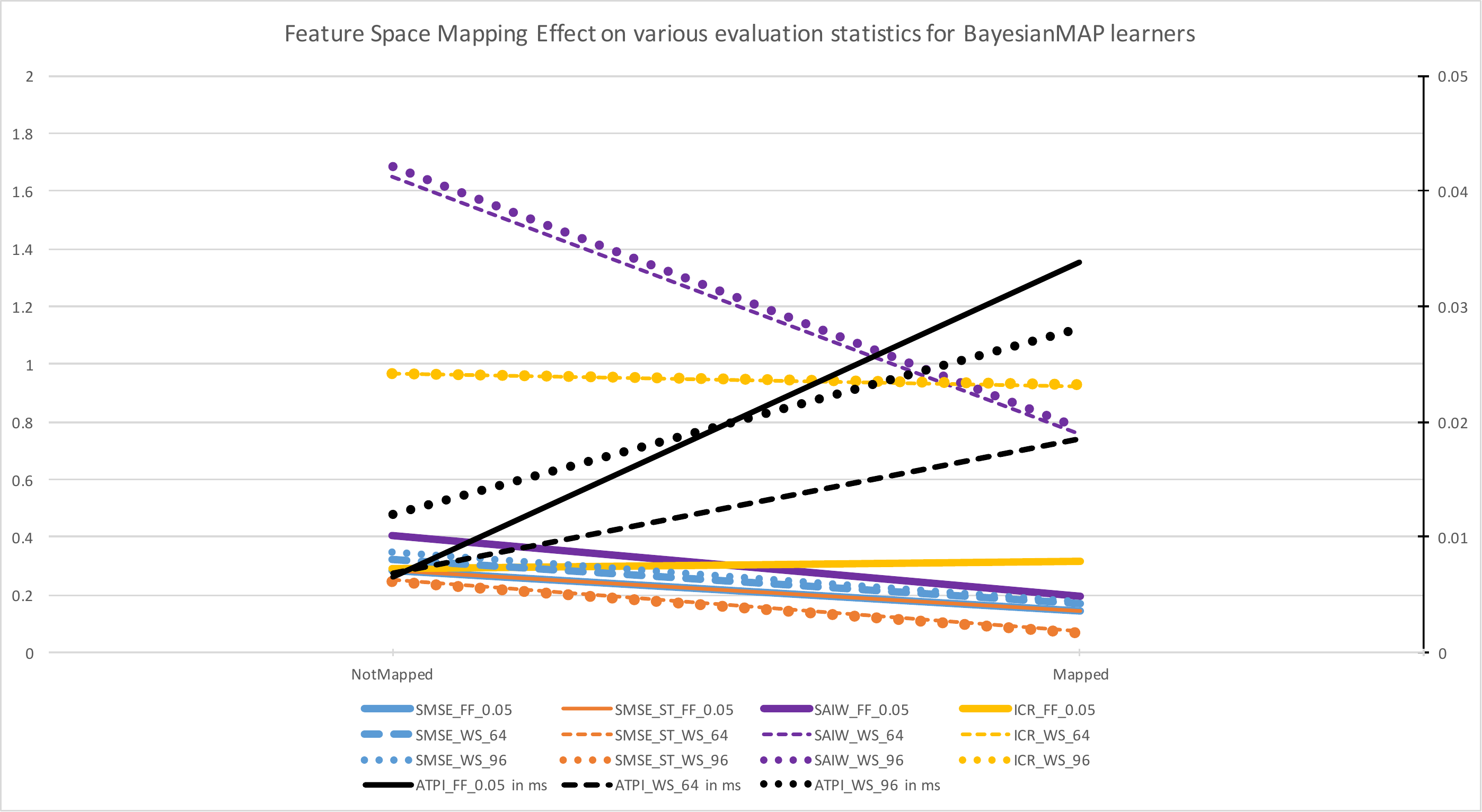}
  \caption{\texttt{SMSE}, \texttt{SAIW},\texttt{ICR} and \texttt{ATPI} scores for 6 different \texttt{BayesianMLE} learners are graphed. \texttt{ATPI} scores measured with respect to the secondary y-axis with the interval $[0,0.005]$ while the rest use the primary y-axis with the $[0,2]$ interval. The statistics are aggregated from 576 stream simulations on \texttt{BayesianMLEForgetting\_FF0.05}, \texttt{BayesianMLEWindowed\_WS64} and \texttt{BayesianMLEWindowed\_WS96} learners (each having one mapped and one non-mapped version)}
  \label{fig:fsm_effect_bmap}
\end{figure}

In Figure \ref{fig:fsm_effect_bmap}, how feature space mapping affects the accuracy, prediction bounds quality and time-efficiency of \texttt{BayesianMAP} learners with the window sizes and forgetting factors chosen in the previous subsections is shown. The same patterns observed as in the relation between the feature-space mapping and the accuracy, prediction bounds and time-efficiency scores appeared for the \texttt{BayesianMLE} learners. Briefly, feature-space mapping has improved the accuracy and without sacrificing a good coverage of the prediction bounds, it tightens them. In return, the average time needed for processing a single data stream item increased although this does not change the employability of the considered \texttt{BayesianMAP} learners with the feature-space mapping. Thus, the option of feature-space mapping is opted in for the considered \texttt{BayesianMAP} namely \texttt{BayesianMAPForgetting\_FF0.05}, \texttt{BayesianMAPWindowed\_WS64} and \texttt{BayesianMAPWindowed\_WS96}.

\subsection{Algorithm comparison}

Having uncovered the effect of sliding window size and forgetting-factor parameters on the qualities of the predictions such as accuracy, prediction bounds quality and time-efficiency and picking more preferable choices for these parameters, now a more detailed comparison of the algorithms than the previous section can be made as a significant number of learners are filtered out. This allows to make more confident conclusions on the applicability of the learners to the runtime prediction problem.

The learners compared are listed as follows:

\begin{itemize}
\label{final_11}
\item \texttt{BayesianMLEForgettingMapped\_FF0.05}
\item \texttt{BayesianMLEWindowedMapped\_WS64} 
\item \texttt{BayesianMLEWindowedMapped\_WS96}
\item \texttt{BayesianMAPForgettingMapped\_FF0.05}
\item \texttt{BayesianMAPWindowedMapped\_WS64} 
\item \texttt{BayesianMAPWindowedMapped\_WS96}
\item \texttt{GPRegressionGaussianKernelZeroMean\_WS64}
\item \texttt{GPRegressionGaussianKernelAvgMean\_WS64}
\item \texttt{GPRegressionGaussianKernelOLSMean\_WS64}
\item \texttt{KernelRegression\_WS64}
\item \texttt{KernelRegression\_WS96}
\end{itemize}

\begin{figure}[htbp]
  \centering
    \includegraphics[width=\linewidth]{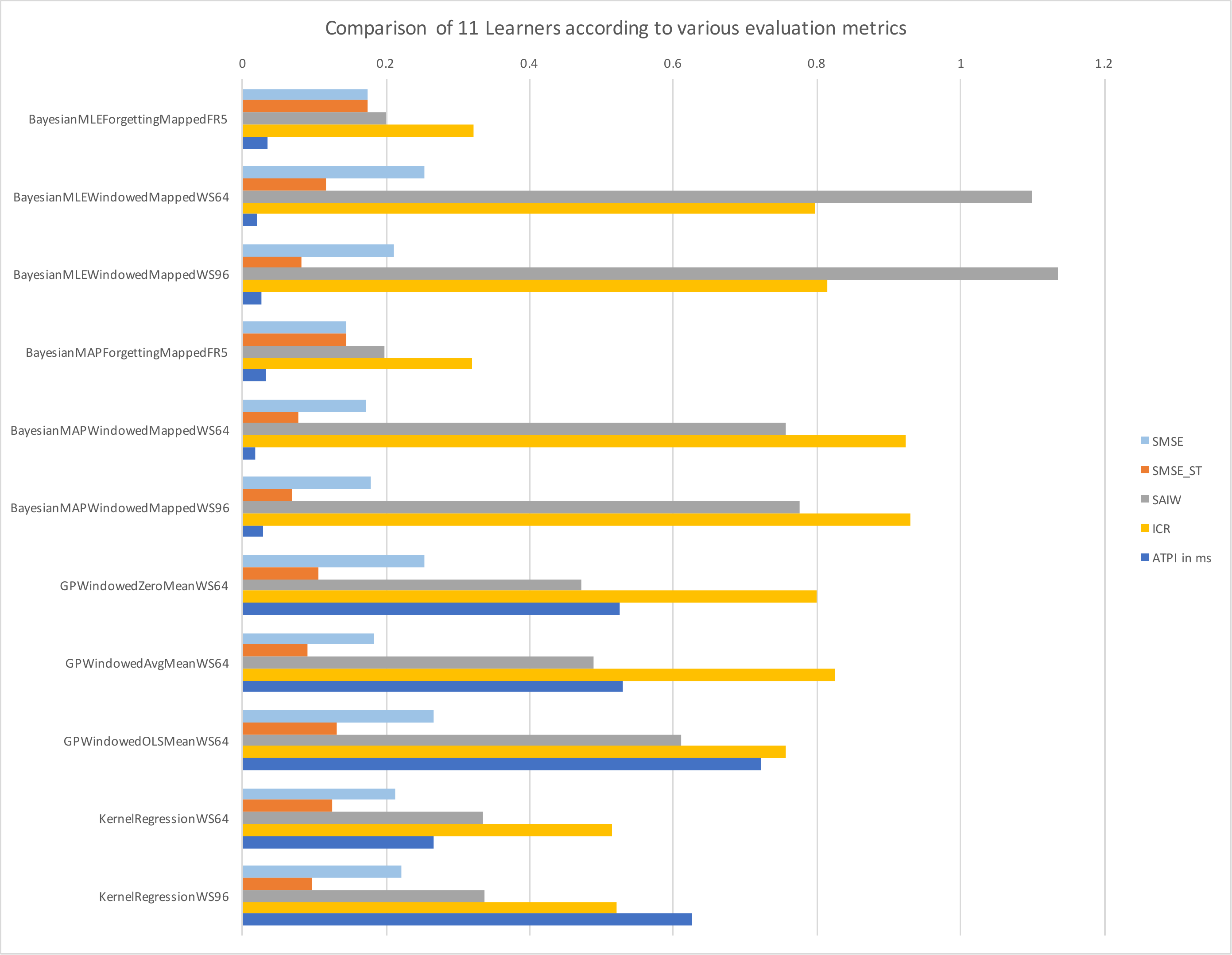}
  \caption{Comparison of \texttt{SMSE}, \texttt{SAIW}, \texttt{ICR} and PIT statistics for 11 different learners whose full codenames are shown on the x- axis. The results are aggregated over the 576 session results.}
  \label{fig:final_11_comparison}
\end{figure}

\begin{figure}[htbp]
  \centering
    \includegraphics[width=\linewidth]{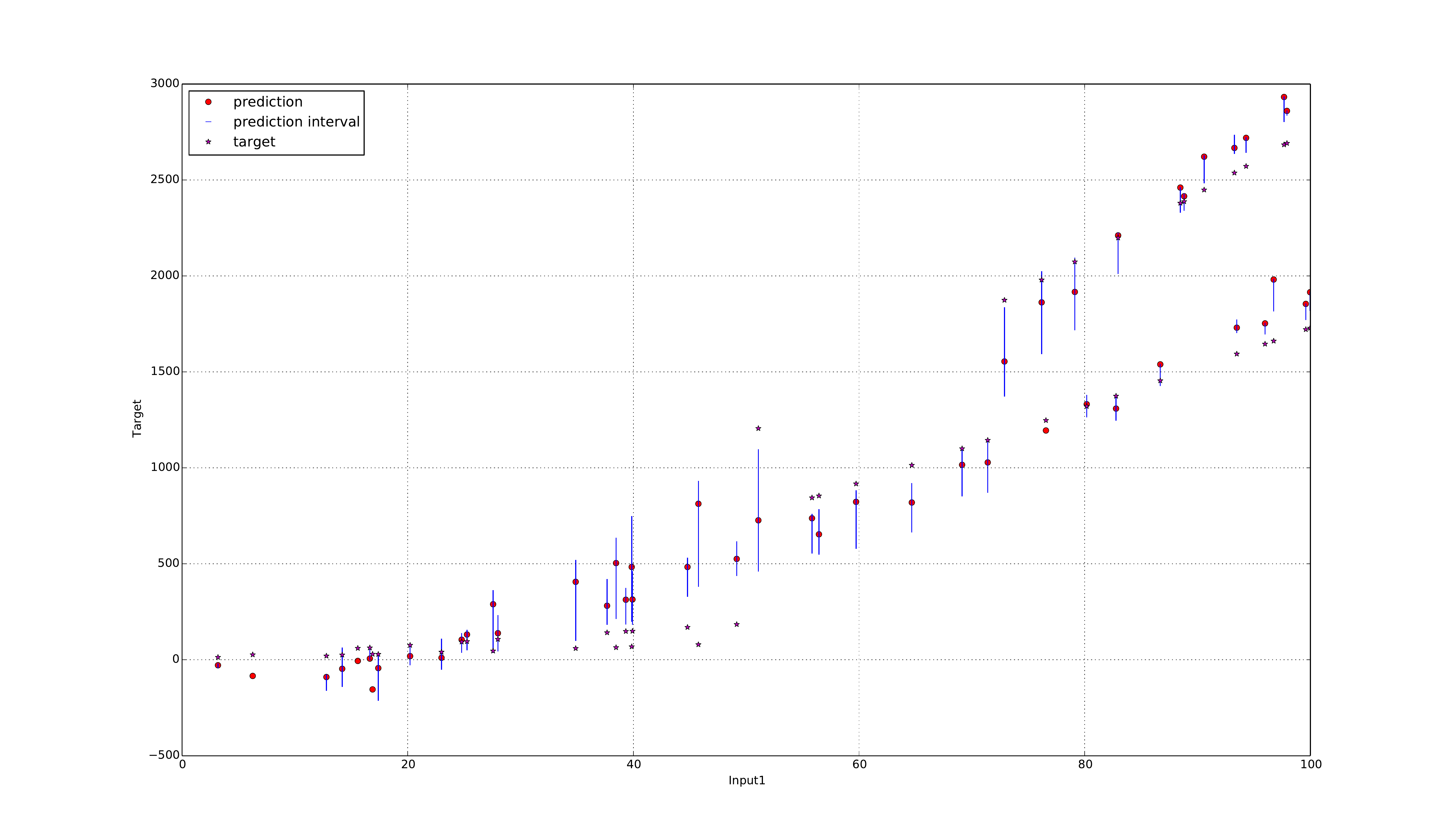}
  \caption{\texttt{BayesianMLEForgettingMapped\_FF0.05} tested on SYNTH\_D\_CD\_2000\_1\_100\_1\_12. 50 sample predictions along with their corresponding targets and prediction intervals are shown}
 \label{fig:sampled_50_predictions_bmle_mapped_fr5_SYNTH_D_CD_2000_1_100_1_12}
\end{figure}

\begin{figure}[htbp]
  \centering
    \includegraphics[width=\linewidth]{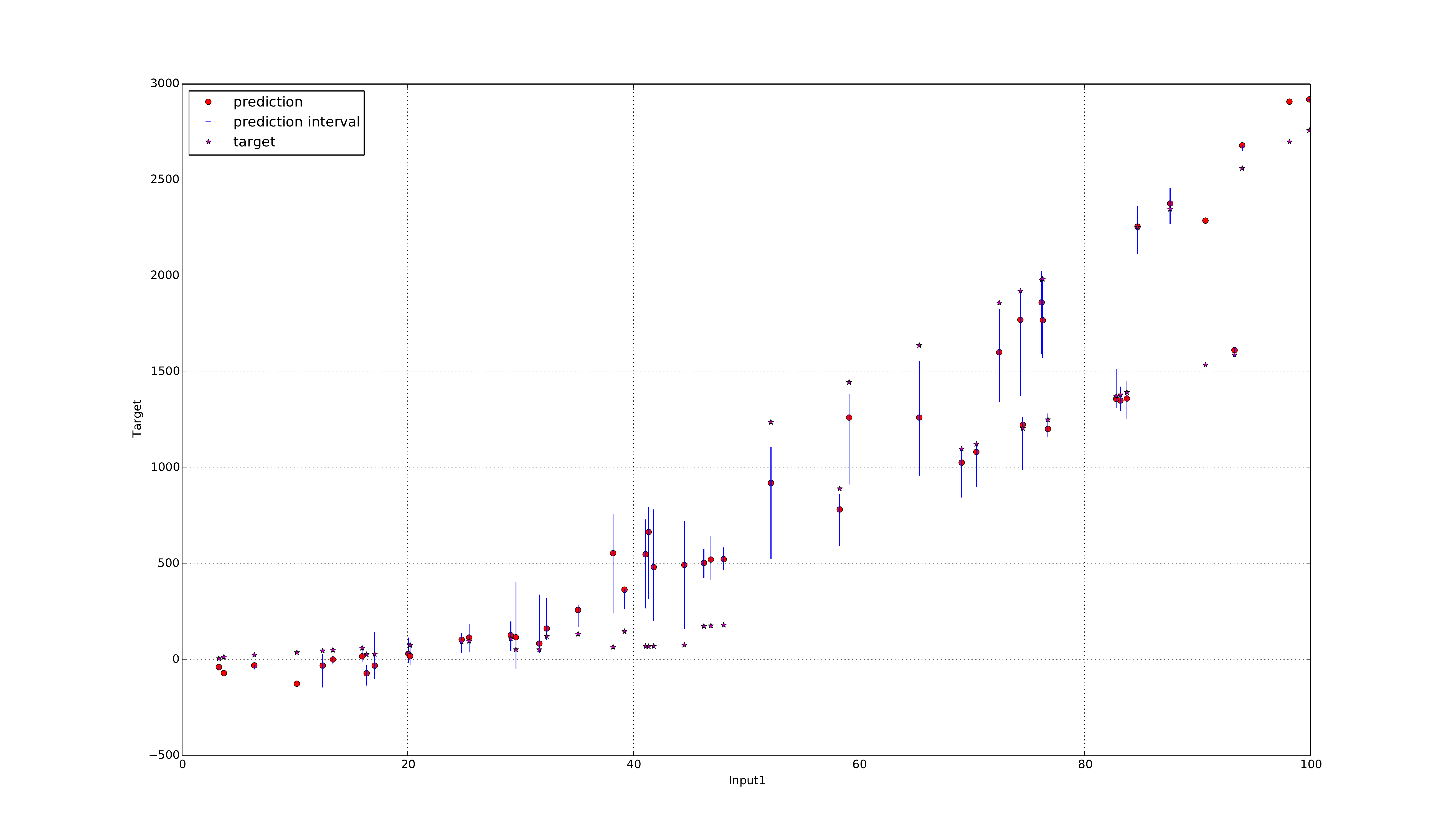}
  \caption{\texttt{BayesianMAPForgettingMapped\_FF0.05} tested on SYNTH\_D\_CD\_2000\_1\_100\_1\_12. 50 sample predictions along with their corresponding targets and prediction intervals are shown}
 \label{fig:sampled_50_predictions_bmap_mapped_fr5_SYNTH_D_CD_2000_1_100_1_12}
\end{figure}

When the \texttt{ICR} and \texttt{SAIW} results examined in \ref{fig:final_11_comparison}, something quite wrong with the the forgetting-factor based learners from the above list is observed. Although their average \texttt{SAIW} results seem to be acceptable, their \texttt{ICR} results are very poor. This means, the prediction bounds the chosen forgetting-based algorithms produce along with the point predictions do not cover the observed targets well. In order to visualize the problem, randomly selected $50$ data points from a simulated stream together with their corresponding target, prediction and prediction bounds estimated by the \texttt{BayesianMLEForgettingMapped\_FF0.05} and \texttt{BayesianMAPForgettingMapped\_FF0.05} are shown in \ref{fig:sampled_50_predictions_bmle_mapped_fr5_SYNTH_D_CD_2000_1_100_1_12} and \ref{fig:sampled_50_predictions_bmap_mapped_fr5_SYNTH_D_CD_2000_1_100_1_12}.

Evidently, forgetting-factor based learners does not provide prediction bounds having good coverage of the target values. Thus, using forgetting-factor based learners for the runtime estimation is not a good idea despite their desirable properties also shared by sliding-windowed learners such as high prediction and update speeds and acceptable accuracy scores.

\begin{figure}[htbp]
  \centering
    \includegraphics[width=\linewidth]{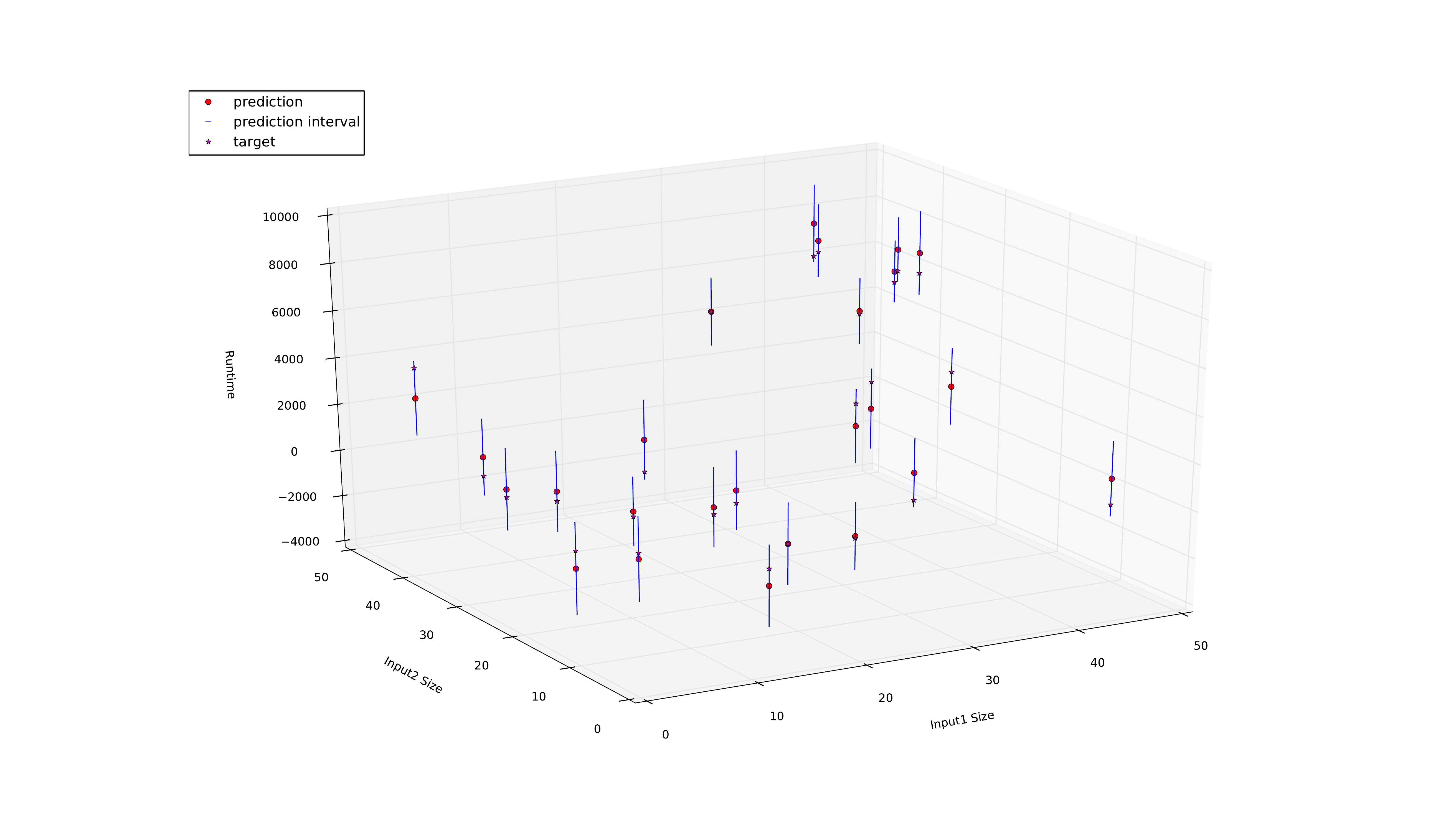}
  \caption{\texttt{BayesianMLEWindowedMapped\_WS64} tested on \texttt{SYNTH\_N\_NCD\_2000\_2\_50\_3\_12}. 25 sample predictions along with their corresponding targets and prediction intervals are shown}
 \label{fig:sampled_25_predictions_bmle_mapped_ws64_SYNTH_D_NCD_2000_2_50_3_12}
\end{figure}

\begin{figure}[htbp]
  \centering
    \includegraphics[width=\linewidth]{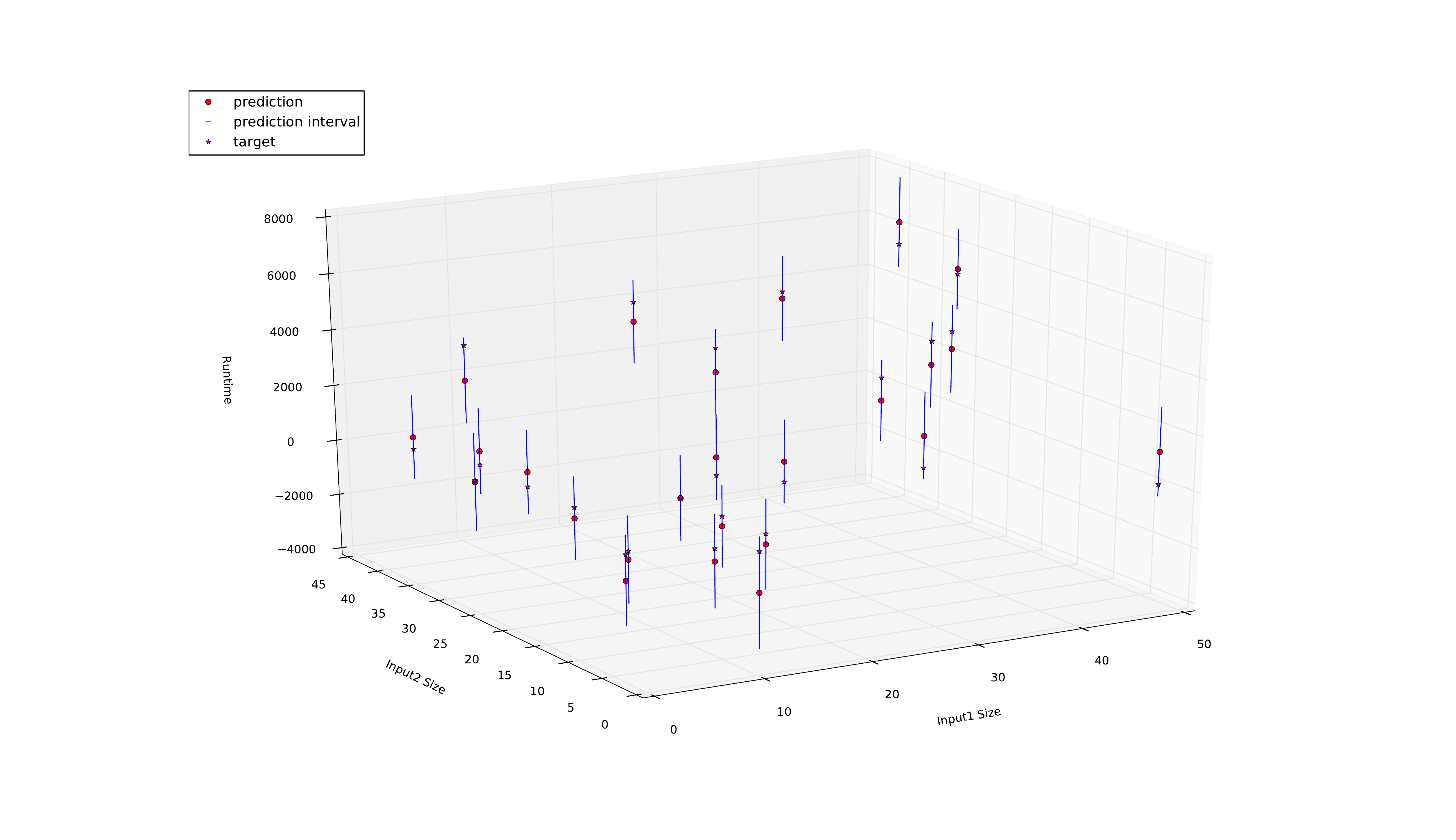}
  \caption{\texttt{BayesianMLEWindowedMapped\_WS96} tested on \texttt{SYNTH\_N\_NCD\_2000\_2\_50\_3\_12}. 25 sample predictions along with their corresponding targets and prediction intervals are shown}
 \label{fig:sampled_25_predictions_bmle_mapped_ws96_SYNTH_D_NCD_2000_2_50_3_12}
\end{figure}

\begin{figure}[htbp]
  \centering
    \includegraphics[width=\linewidth]{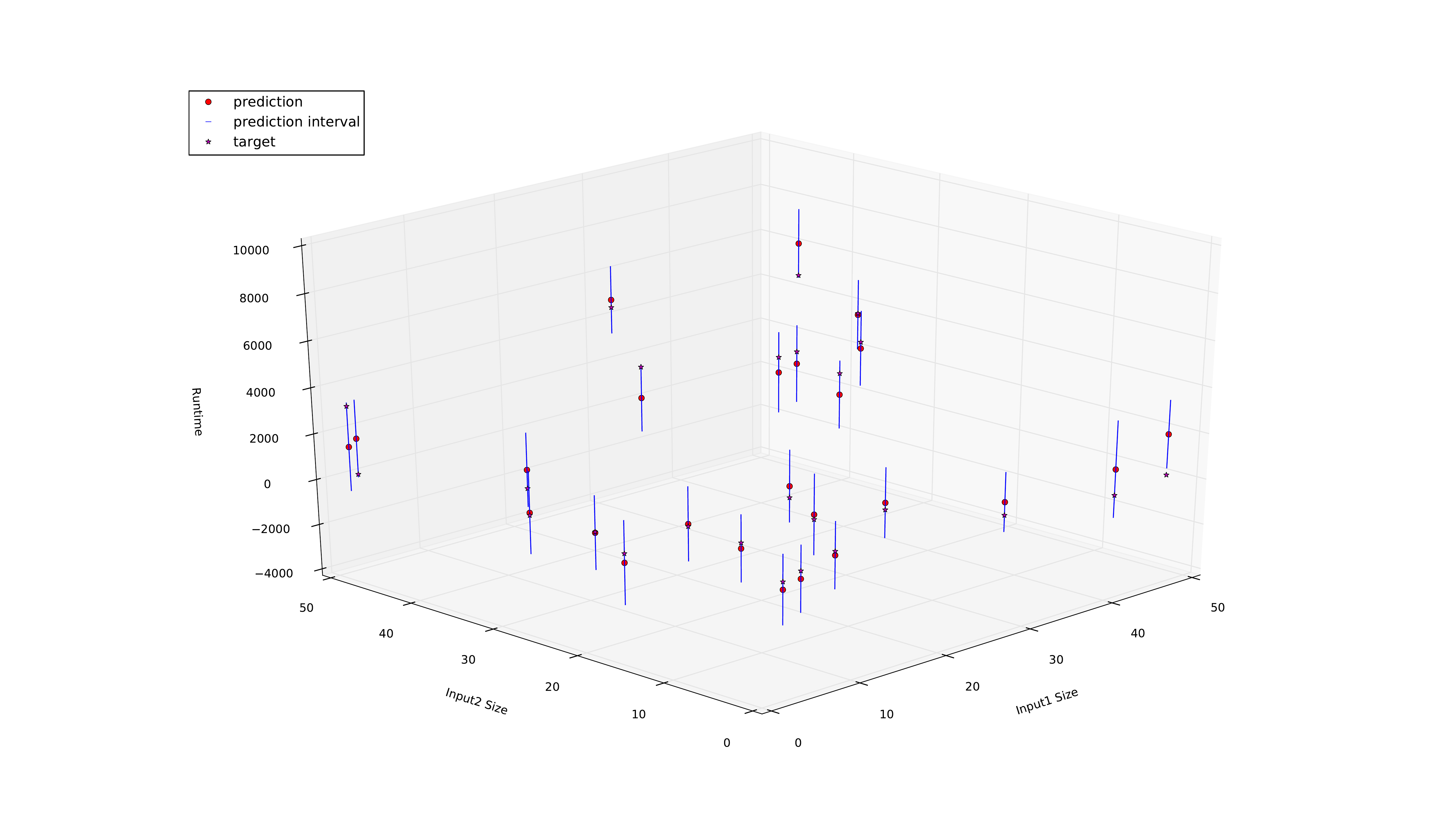}
  \caption{\texttt{BayesianMAPWindowedMapped\_WS64} tested on \texttt{SYNTH\_N\_NCD\_2000\_2\_50\_3\_12}. 25 sample predictions along with their corresponding targets and prediction intervals are shown}
 \label{fig:sampled_25_predictions_bmap_mapped_ws64_SYNTH_D_NCD_2000_2_50_3_12}
\end{figure}

\begin{figure}[htbp]
  \centering
    \includegraphics[width=\linewidth]{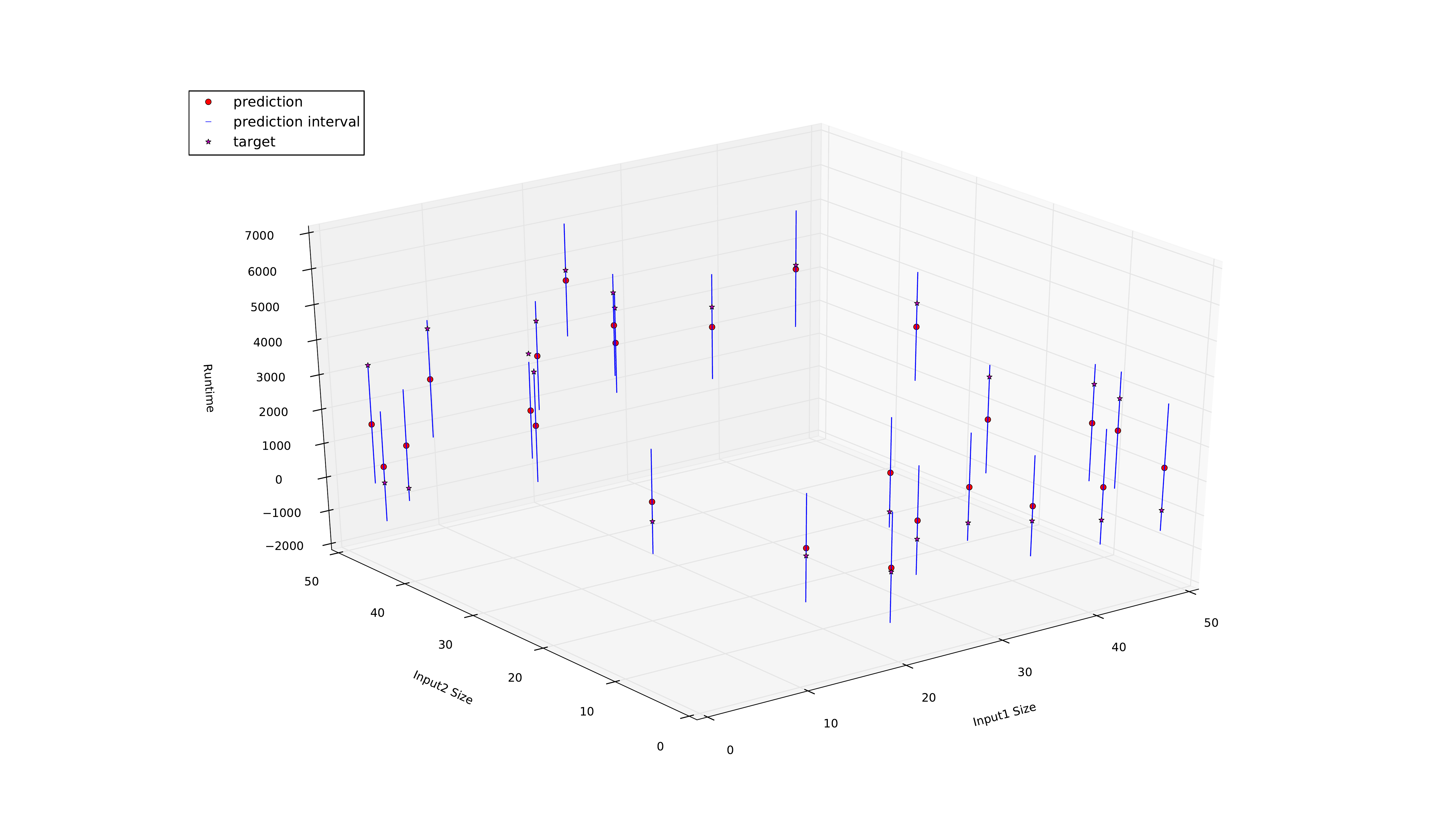}
  \caption{\texttt{BayesianMAPWindowedMapped\_WS96} tested on \texttt{SYNTH\_N\_NCD\_2000\_2\_50\_3\_12}. 25 sample predictions along with their corresponding targets and prediction intervals are shown}
 \label{fig:sampled_25_predictions_bmap_mapped_ws96_SYNTH_D_NCD_2000_2_50_3_12}
\end{figure}

Other parametric learners apart from the forgetting-based ones also exhibit problems with the prediction bounds they produce. In Figure \ref{fig:final_11_comparison}, it appears that \texttt{SAIW} scores of \texttt{BayesianMLE} algorithms are above 1.0 and the \texttt{SAIW} scores of \texttt{BayesianMAP} are close to $0.8$. These \texttt{SAIW} scores indicate that the average gap between the upper and the lower prediction bounds is very high. For reference, one can imagine having the \texttt{SAIW} score if $1.0$ being equivalent of producing lower and upper prediction bounds that are as far as the magnitude of the target mean apart from each other. In order to visualize how high that is, randomly chosen $25$ data points occurred on a simulated stream which each of the sliding-windowed parametric algorithms from the list \ref{final_11} are tested along with the target, prediction and prediction bounds of them are visualized in \ref{fig:sampled_25_predictions_bmle_mapped_ws64_SYNTH_D_NCD_2000_2_50_3_12},
\ref{fig:sampled_25_predictions_bmle_mapped_ws96_SYNTH_D_NCD_2000_2_50_3_12},
\ref{fig:sampled_25_predictions_bmap_mapped_ws64_SYNTH_D_NCD_2000_2_50_3_12},
and \ref{fig:sampled_25_predictions_bmap_mapped_ws96_SYNTH_D_NCD_2000_2_50_3_12}.
Some of the prediction intervals shown with the blue lines in the graphs are too big that lower bound of some of them are even negative which do not make sense for runtime predictions.

For the sake of explaining why the theoretically sound prediction interval estimation method (unlike the one used by the forgetting-factor based learners) employed by the sliding-windowed \texttt{BayesianMAP} and \texttt{BayesianMLE} variants did not work as expected, the experiment data is analyzed from the streams simulated by the data which is generated by a function with discontinuity and without discontinuity separately. This is because, the only data quality that may have caused a parametric regression algorithm fail to make accurate predictions with tight prediction bounds is discontinuity of the function that is used to generate the data. To this end, a graph with showing only \texttt{ICR} and \texttt{SAIW} scores of the sliding window algorithms considered in \ref{final_11} on the streams of different natures in terms of the continuity of the data generation function used for the generation of the data set which they are simulated from is prepared.

\begin{figure}[htbp]
  \centering
    \includegraphics[width=\linewidth]{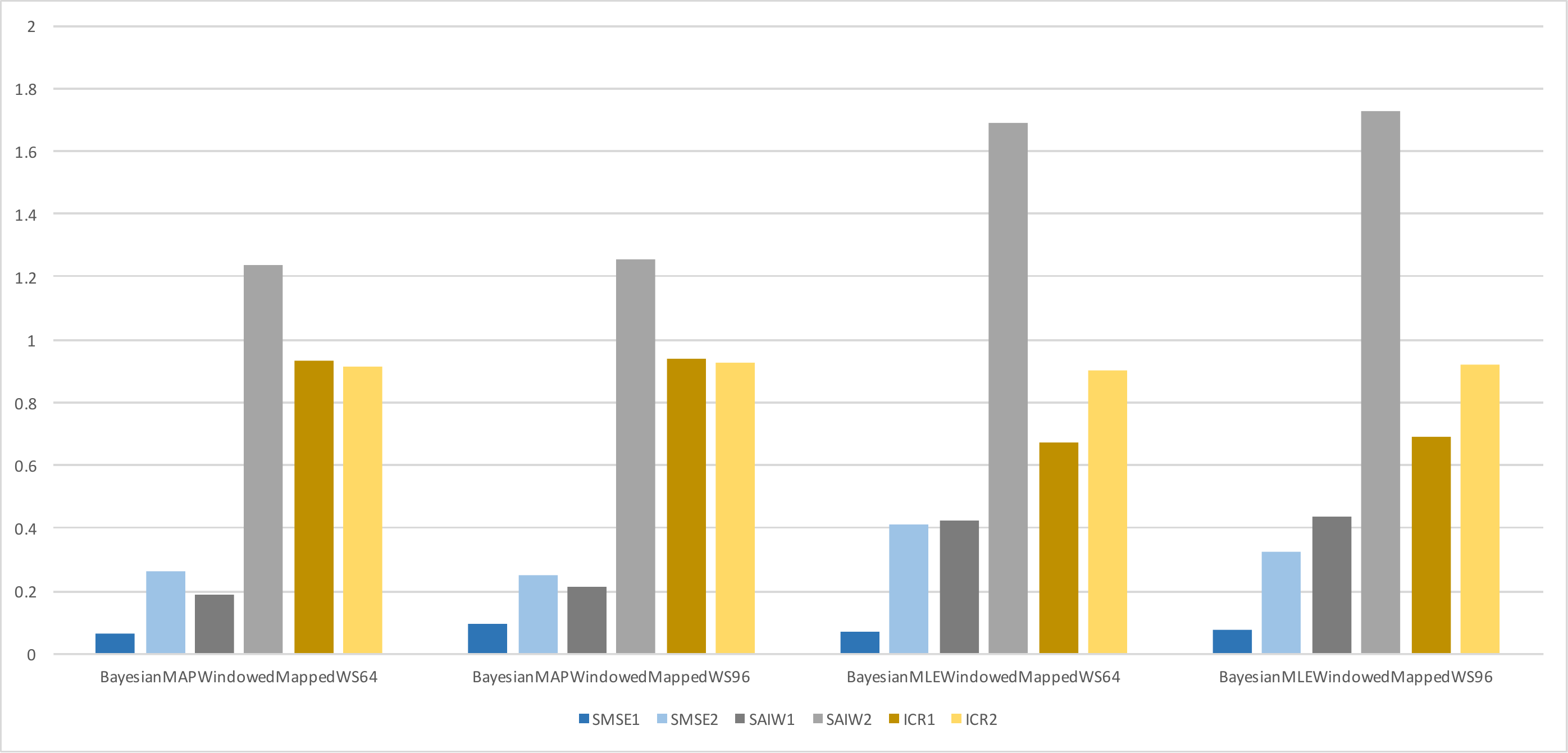}
  \caption{A comparison of \texttt{SMSE}, \texttt{SAIW} and \texttt{ICR} metrics for 4 different learners whose full codenames are shown on the x- axis on the streams simulated of different nature in terms of the continuity of the function that generated the data set which they are simulated from. The values are obtained through aggregation of session results of 276 stream simulations for each class}
  \label{fig:disc_nondisc_bmap_bmap_mapped_ws64_ws96}
\end{figure}

In \ref{fig:disc_nondisc_bmap_bmap_mapped_ws64_ws96}, the statistics represented by darker colors (and also denoted by a name ending with 1 in the legend) are obtained from the tests containing data from continuous functions whereas the light ones (in the legend, having names ending with 2) are obtained from the test data generated by discontinuous functions. Looking only at the difference between \texttt{SMSE}1 and \texttt{SMSE}2 values makes is clear the data from the discontinuous functions was the culprit. This is a clear manifestation of \textit{underfitting} explained in Chapter \ref{Chapter2}. Parametric models fail when the data they learn from generated from a function that belongs to a different class than the one assumed by the parametric model. And due to this failure causing large errors for data points, prediction interval estimation method based on the errors accumulated produces very large prediction intervals making \texttt{SAIW} to jump to very high values from dark blue columns to light blue ones in the graph. 

Since the growth of the runtime of the database operators usually change characteristics as the inputs growth as pointed out in Chapter \ref{Chapter3}, synthetic data generated to mock this situation is important. This is why, parametric models that fails to provide prediction bounds of good quality are found unsuitable for runtime estimation problem. 

\begin{figure}[htbp]
  \centering
    \includegraphics[width=\linewidth]{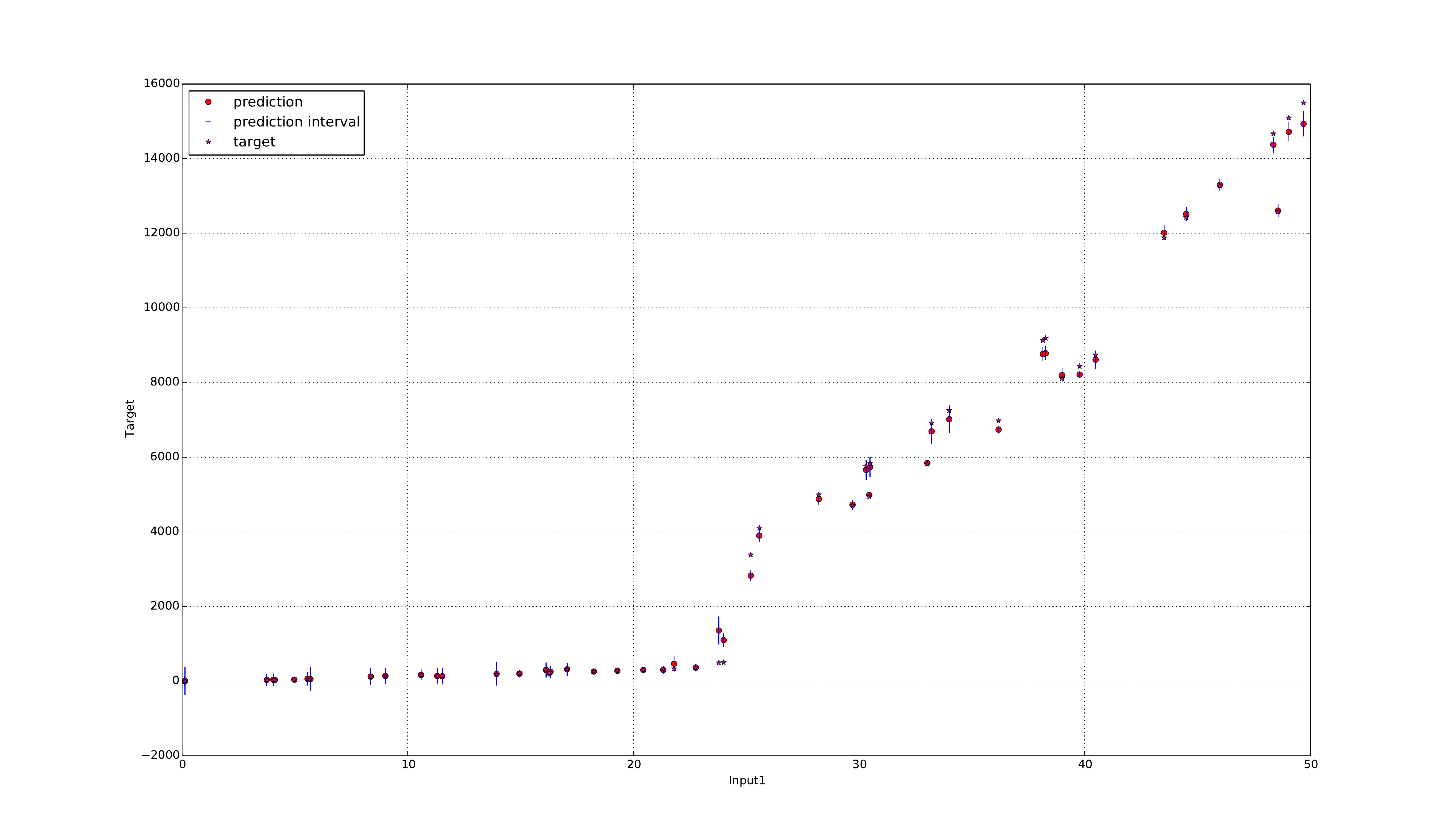}
  \caption{\texttt{KernelRegression\_WS64} tested on \texttt{SYNTH\_D\_CD\_2000\_1\_50\_3\_23}. 50 sample predictions along with their corresponding targets and prediction intervals are shown}
 \label{fig:sampled_50_predictions_kreg_ws64_SYNTH_D_CD_2000_1_50_3_23}
\end{figure}

\begin{figure}[htbp]
  \centering
    \includegraphics[width=\linewidth]{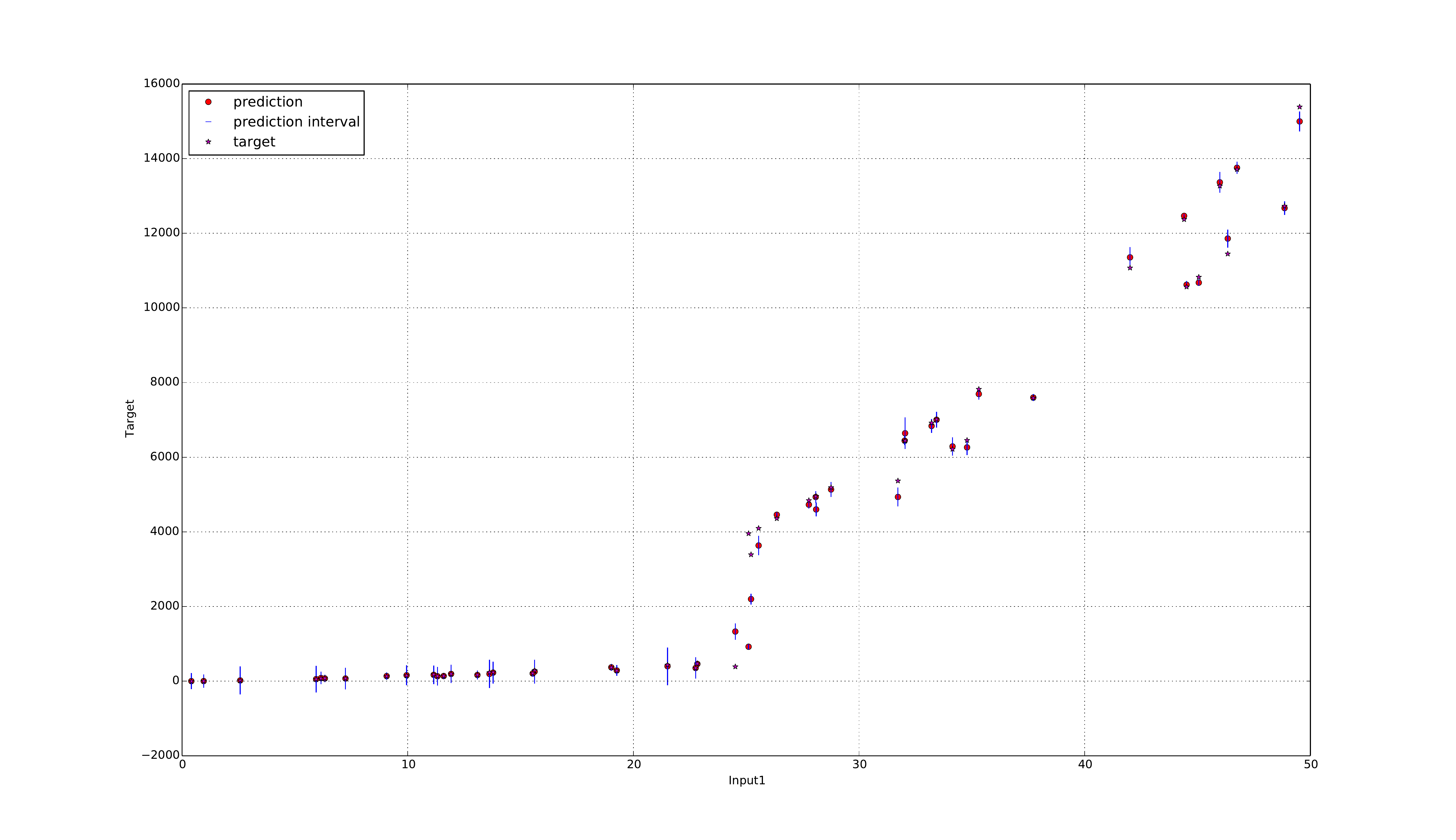}
  \caption{\texttt{KernelRegression\_WS96} tested on \texttt{SYNTH\_D\_CD\_2000\_1\_50\_3\_23}. 50 sample predictions along with their corresponding targets and prediction intervals are shown}
 \label{fig:sampled_50_predictions_kreg_ws96_SYNTH_D_CD_2000_1_50_3_23}
\end{figure}

In Figure \ref{fig:final_11_comparison}, the bottom $2$ bars indicate that \texttt{KernelRegression} learners do not produce prediction bounds that provide high target variable coverage just like forgetting-based parametric learners. In order to visualize the problem with the prediction bounds that \texttt{KernelRegression} learners have, prediction bounds they provided along with the predictions and targets for randomly chosen $50$ data points from a test stream is displayed in \ref{fig:sampled_50_predictions_kreg_ws64_SYNTH_D_CD_2000_1_50_3_23} and \ref{fig:sampled_50_predictions_kreg_ws96_SYNTH_D_CD_2000_1_50_3_23}. 

When \ref{fig:sampled_50_predictions_kreg_ws64_SYNTH_D_CD_2000_1_50_3_23} and \ref{fig:sampled_50_predictions_kreg_ws96_SYNTH_D_CD_2000_1_50_3_23} are examined carefully, what is observed from the figures is that although the point predictions are pretty close to targets, the prediction bounds around them are so tight that they cannot reach the target points for significantly many of the $50$ points sampled. However, this is not an unavoidable problem unlike the prediction interval issues with sliding windowed parametric models on the data from discontinuous functions or forgetting-factor based parametric models in general. A possible solution to this problem arbitrarily extend the reach of the prediction bounds. This will surely result in an increase in \texttt{SAIW} but since \texttt{SAIW} values are relatively low as seen in \ref{fig:gen_alg_comparison} ($0.33$ for both learners), this interval extension method can be applied to some extent without trouble. From a theoretical point of view, extending the interval between the prediction bounds by multiplying the length between the bounds and the point prediction by some constant is equivalent of asking the prediction bound estimation method to return the bounds with a higher confidence parameter than the \textit{usual} $95\%$.

Following the the idea proposed above, $2$ \texttt{KernelRegression} learners were again tested on $576$ stream simulations after modifying the prediction bound estimation mechanism so that it returns the confidence intervals\footnote{Note that confidence intervals instead of prediction intervals are used as the prediction bounds for \texttt{KernelRegression} learners.} with $99.9\%$ confidence instead of $95\%$. The modified \texttt{KernelRegression} learners are referred to as \texttt{KernelRegression\_HighConf}.

\begin{figure}[htbp]
  \centering
    \includegraphics[width=\linewidth]{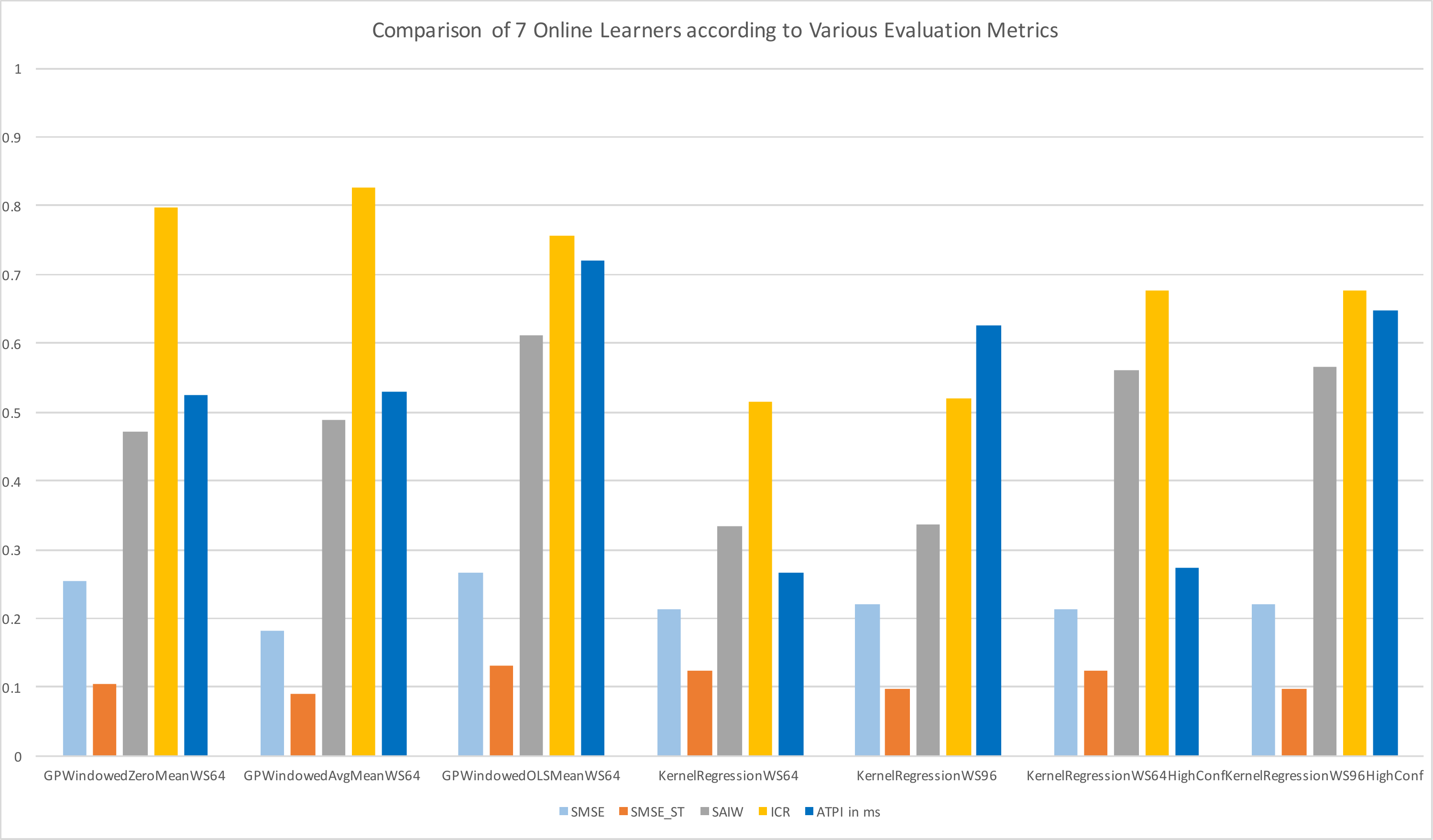}
  \caption{Comparison of \texttt{SMSE}, \texttt{SMSE\_ST}, \texttt{ICR} and \texttt{ATPI} statistics for 7 different learners whose full codenames are shown on the x- axis tested on 576 streams}
  \label{fig:final_7_comparison}
\end{figure}

Figure \ref{fig:final_7_comparison}, shows various statistics of the both original and modified \texttt{KernelRegression} learners along with the \texttt{GPRegression} learners for comparison. In the graph, it appears that all $7$ learners were pretty accurate keeping \texttt{SMSE} and \texttt{SMSE\_ST} scores below $0.25$ and $0.15$ respectively. \texttt{GPRegressionAvgMean} seems to be a bit more accurate than the rest. It is followed by the \texttt{KernelRegressionWS96} (both versions) and \texttt{GPRegressionZeroMean}. \texttt{GPRegressionOLSMean} trails behind the other $6$ learners in terms of accuracy. With the prediction interval modification on the \texttt{KernelRegression} algorithms, target variable coverage scores of \texttt{KernelRegression} learners are brought to the same levels of that of \texttt{GPRegression} learners. This unsurprisingly resulted in increasing \texttt{SAIW} scores although they did not pass $0.6$ meaning the average gap between prediction bounds were still not too high. In terms of the prediction time per item, none of the $7$ learners spend more than $0.75$ ms per item on average. \texttt{KernelRegressionWS64}, processing a stream item $0.27$ ms on average, appears to be the quickest learner while \texttt{GPRegressionOLSMean} is the slowest one with \texttt{APTI} of $0.72$ ms.

Modifying the \texttt{KernelRegression} learners from the list \ref{final_11} as discussed above and eliminating the unsuitable ones, $5$ learners out of the initial $52$ are left. These are listed as follows:

\begin{itemize}
\label{final_5}
\item \texttt{GPRegressionZeroMean\_WS64}
\item \texttt{GPRegressionAvgMean\_WS64}
\item \texttt{GPRegressionOLSMean\_WS64}
\item \texttt{KernelRegression\_HighConf\_WS64}
\item \texttt{KernelRegression\_HighConf\_WS96}
\end{itemize}

Next, using the advantage of having synthetic test data, how the accuracy, prediction bounds quality and time-efficiency change with the changing noise levels, input dimensionality\footnote{Note that analyzing the experiment results by the input dimensionality would be possible with the real measurement data as well.} and discontinuity of the function used for mocking the streaming data.

\begin{figure}[htbp]
  \centering
    \includegraphics[width=\linewidth]{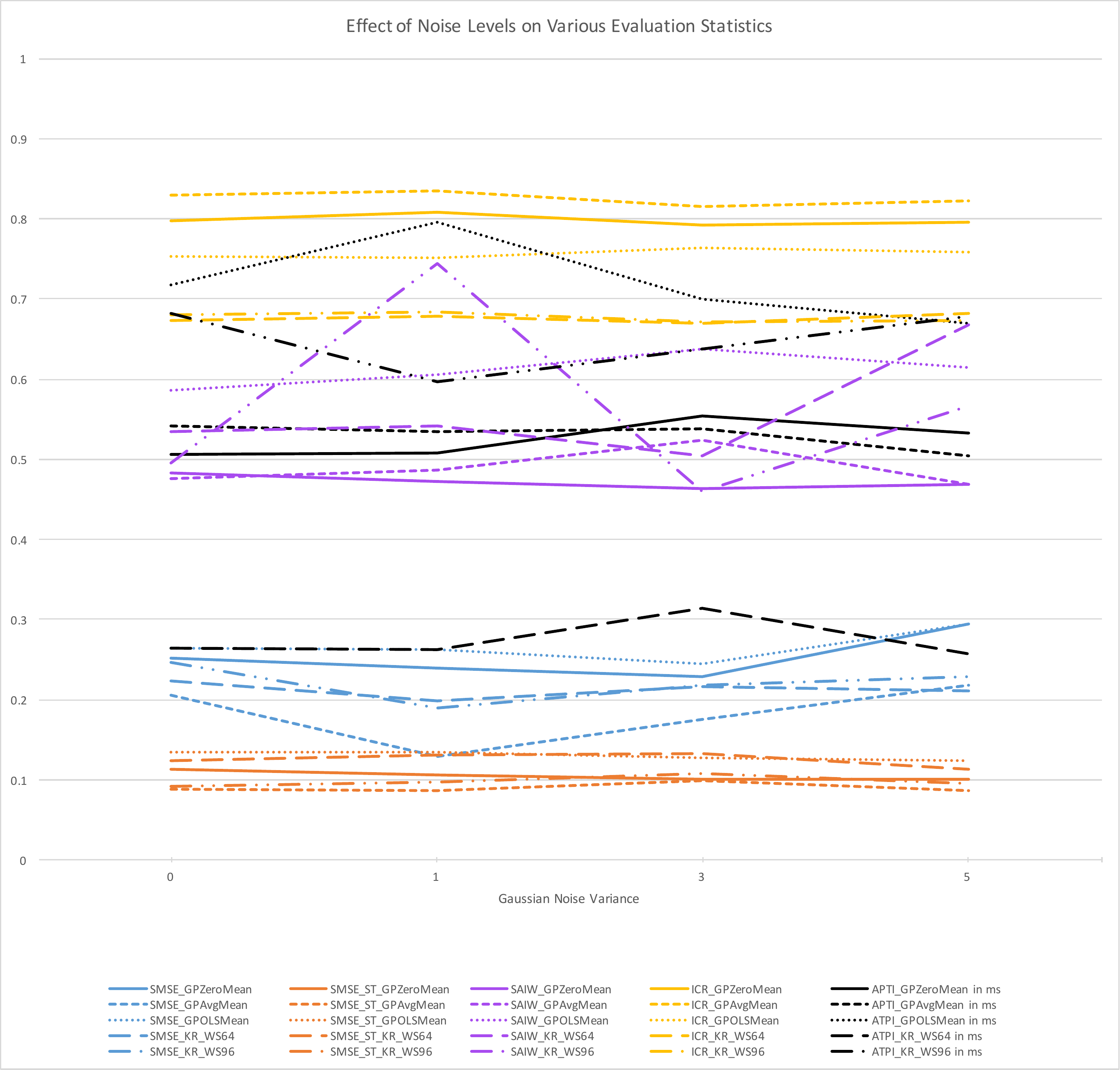}
  \caption{Visualization of how \texttt{SMSE}, \texttt{SMSE\_ST}, \texttt{ICR} and \texttt{ATPI} scores change as the measurement noise increases for 5 different learners listed in \ref{final_5}. The results are aggregated over 144 stream simulations for each noise level}
  \label{fig:noise_effect_on_final_5}
\end{figure}

Figure \ref{fig:noise_effect_on_final_5} shows that the average stable and general accuracy of the learners \textit{remarkably} did not increase with the increasing noise levels. This allows for a quick yet an important conclusion about the online learning algorithms employed: More measurement noise does not mean more error. In other words, the online learners employed are robust to noise.

Average prediction bounds coverage also remained nearly the same when the noise variance is increased from $0.0$ (no noise) to $5.0$. Moreover, in terms of the average width of the intervals set by the prediction bounds, \texttt{GPRegression} learners were almost insensitive to noise exhibiting small fluctuations with the increasing noise. As for \texttt{KernelRegression}, the \texttt{SAIW} statistic of the learner with the sliding-window size of $96$ is funky. It jumps to high levels when switching from no-noise to the minimal noise as indicated by the double dotted dashed line. Moreover, It decreases when the noise is set to a moderate level and then increases again when the noise is high. This jumpy pattern does not allow to draw a confident conclusion about the prediction bounds behavior with the changing noise variance for the \texttt{KernelRegression\_HighConfidence\_WS96}. The other \texttt{KernelRegression} learner from the list \ref{final_5} seems to represent the same pattern but with the smaller amounts of \textit{jumps} from one noise level to another. 

As indicated by the black lines in the Figure \ref{fig:noise_effect_on_final_5}, average time spent for one stream item also fluctuates with the increasing noise. Similar to the prediction bounds behavior of \texttt{KernelRegression}, black lines for the all $5$ learners do not represent a steady trend making it hard to infer a relation between the average processing time per item and the noise level. However, it is not surprising as the underlying mechanism that produced this experiment data, being the execution flow of the learning algorithms that is partially controlled by the drift detection mechanism that modifies the internal state of the learners causing update and tuning operations that increases the item processing time to be fired or inhibited, is complex. 

\begin{figure}[htbp]
  \centering
    \includegraphics[width=\linewidth]{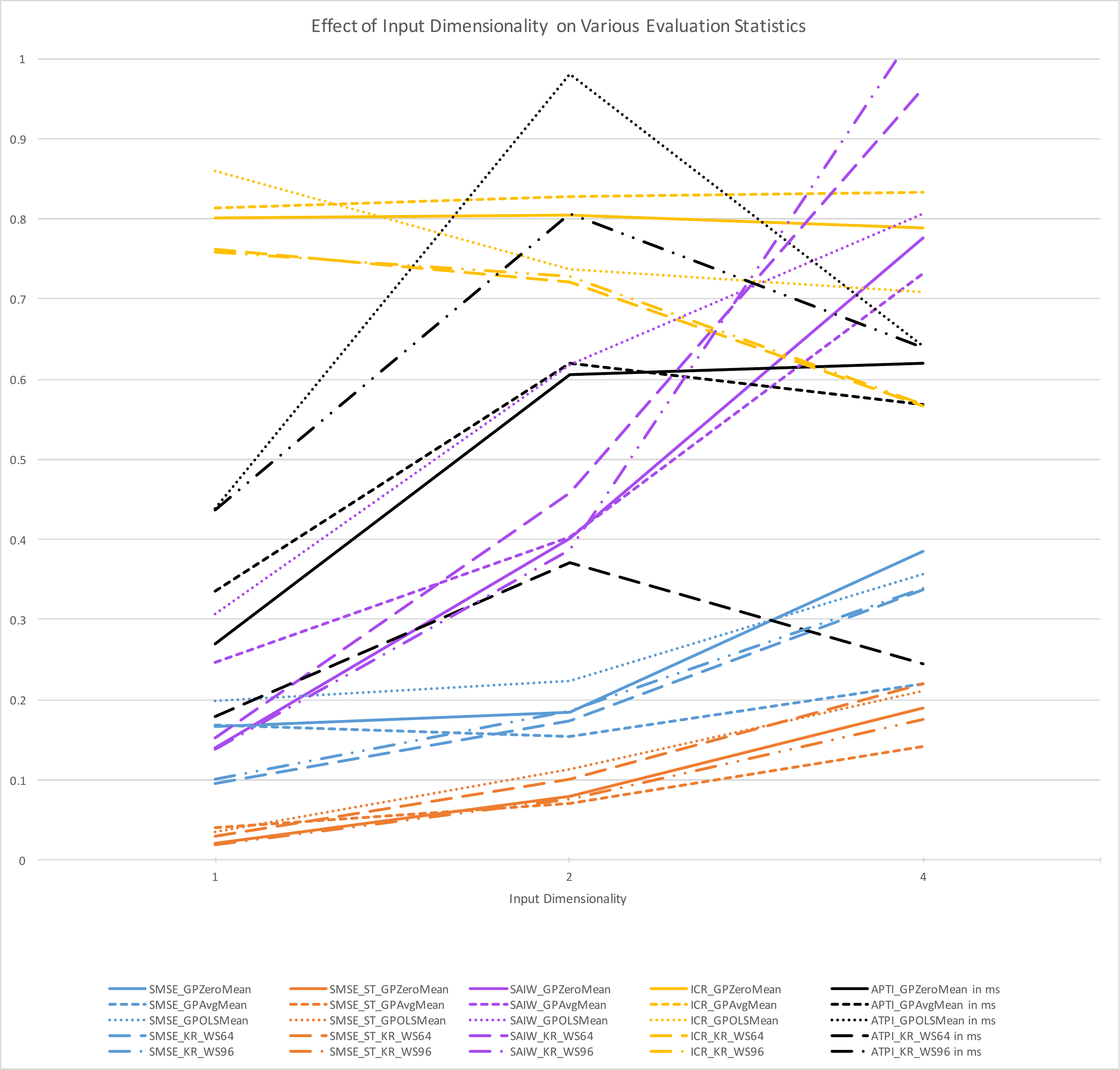}
  \caption{Visualization of how \texttt{SMSE}, \texttt{SMSE\_ST}, \texttt{ICR} and \texttt{ATPI} scores change as the input dimensionality increases for 5 different learners listed in \ref{final_5}. The results are aggregated over 144, 216 and 216 stream simulations for input dimensionalities of 1,2 and 4 respectively}
  \label{fig:inpdim_effect_on_final_5}
\end{figure}

In Figure \ref{fig:inpdim_effect_on_final_5}, as observed in the previous more general experiment results presented in \ref{fig:curse_of_dim}, average general and stable accuracy seem to increase with the growing number of predictor variables.  As for the averaged \texttt{ICR} values, learners except for the \texttt{KernelRegression\_HighConf\_WS64} and \texttt{KernelRegression\_HighConf\_WS96} that appeared to be input dimensionality-oblivious, all exhibited a drop when the input dimensionality increased. This means that the average coverage of the prediction bounds decreases when the number of predictors in the input increases. This drop is more more remarkable in the case of \texttt{KernelRegression} learners and especially when the number of predictors are bumped up to $4$ from $2$. As for, \texttt{GPRegressionOLSMean\_WS64}, the prediction bound coverage drop did not make the learner unemployable but it definitely shows that \texttt{GPRegressionOLSMean\_WS64} is not robust to growing input-dimensionality. This is expected as unlike the other \texttt{GPRegression} learners, it partially uses Ordinary Least Squares algorithm in addition to Gaussian Processes.

A more striking observation than the \texttt{ICR} drop with the growing number of predictors is the dramatic increase in \texttt{SAIW} under the same conditions. Without an exception, the average gap between the upper and the lower prediction bounds that online learners estimate significantly increases as the number of dimensions in the input is raised. This is even more exaggerated in the case of \texttt{GPRegressionZeroMean\_WS64} and \texttt{GPRegressionAvgMean\_WS96}. These results are not totally surprising when the mechanism used to estimate prediction bounds are understood. These are covered in Chapter \ref{Chapter4}.

When the change in the average time spent for each data stream item as the number of predictors increase are examined, a pattern which all 5 learners exhibit is observed. The \texttt{ATPI} first significantly increases when the input dimensionality is raised from $1$ to $2$, then, it either remained almost same or decreased when it the number of predictors are doubled to $4$ from $2$. The first bump is rather understandable as the decrease in the accuracy might have caused the error-triggered update and tuning mechanism to fire more often resulting in increased processing time per stream item. The explanation for the mentioned \textit{drop} in the processing time is somewhat involved and it lies in the implementation details of the error-triggered update and tuning mechanism that is covered in \ref{section:online_learner_semantics}. In short, the magnitude of the increase in error terms was not high enough for error-triggered update and tuning mechanism to kick in in the case of four dimensional input space as frequently as it does with the streams with data points having two predictors.

\begin{figure}[htbp]
  \centering
    \includegraphics[width=\linewidth]{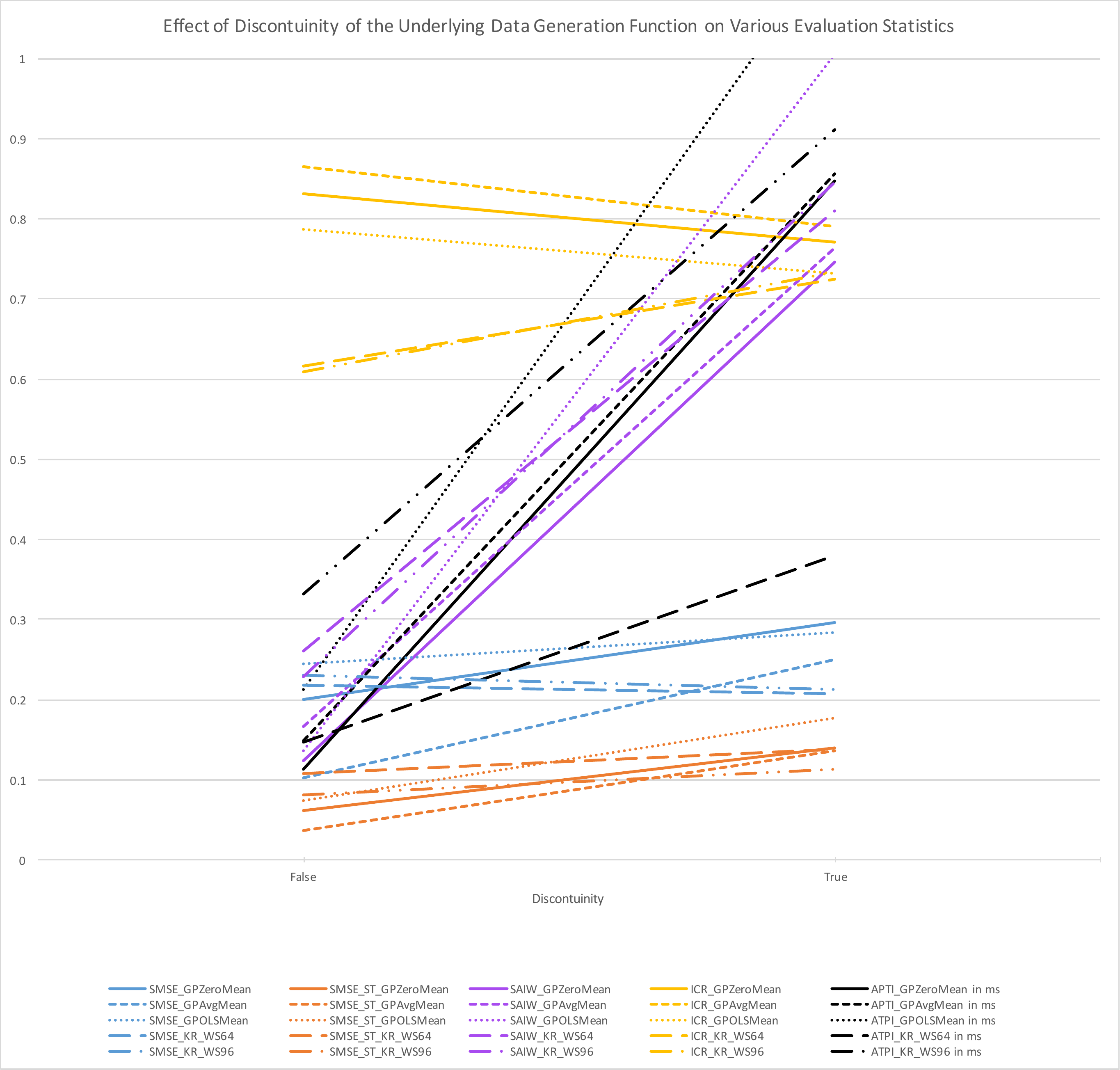}
  \caption{Visualization of how discontinuity of the function used to generate data sets which the streams are simulated from affects \texttt{SMSE}, \texttt{SMSE\_ST}, \texttt{ICR} and \texttt{ATPI} scores for 5 different learners listed in \ref{final_5}. The results are aggregated over 276 discontinuous and 276 continuous cases.}
  \label{fig:disc_effect_on_final_5}
\end{figure}

The Graph \ref{fig:disc_effect_on_final_5} shows that all the statistics except for \texttt{ICR} are higher in the discontinuous case than in the continuous case\footnote{continuous and discontinuous case refer to the the function that generated the data set which the test streams are simulated from having discontinuity or not.}. Increasing \texttt{SAIW} and \texttt{APTI} can be explained by the descent increase in the \texttt{SMSE} and \texttt{SMSE\_ST} as lower accuracy is the result of higher errors that also cause the gaps between the prediction bounds to grow larger to avoid declining \texttt{ICR} scores and activating the error-triggered update-tuning mechanism more often. As for \texttt{ICR} scores, they did not change as dramatically as the other statistics. \texttt{GPRegression} learners had lowered \texttt{ICR} when switched from continuous case to discontinuous case while \texttt{KernelRegression} learners improved their \texttt{ICR} scores. This obviously suggests that the different behaviour of the prediction bounds estimated by the different learners using different learning algorithms is linked to their different prediction bound estimation mechanism. However, why exactly the discontinuity of the data generation function has boosted the prediction bounds coverage of the learners of one learning algorithm family as well as the reason why it dropped that of the learners of the other learning algorithm family remain an open question.

\subsection{Visualization of the cost function approximations by the shortlisted learners on synthetic data sets}

In this subsection, predictions of the learners listed in \ref{final_5} on $2$  simulated streams and their visual comparison to the layout of the data points in the stream together with their corresponding target points are presented. The chosen datasets used for the stream simulations are all generated by discontinuous functions. Moreover, the streams on which the learners are tested feature concept drifts. As a result, when the data points in the test streams are visualized, four partial functions constituting one discontinuous function per stream appear. The motivation for using concept-drifting data streams simulated from datasets generated by discontinuous functions is that the stream learning scenario that Ocelot's runtime operator estimator module will confront are mostly expected to be similar. 

\begin{figure}[htbp]
  \centering
    \includegraphics[width=\linewidth]{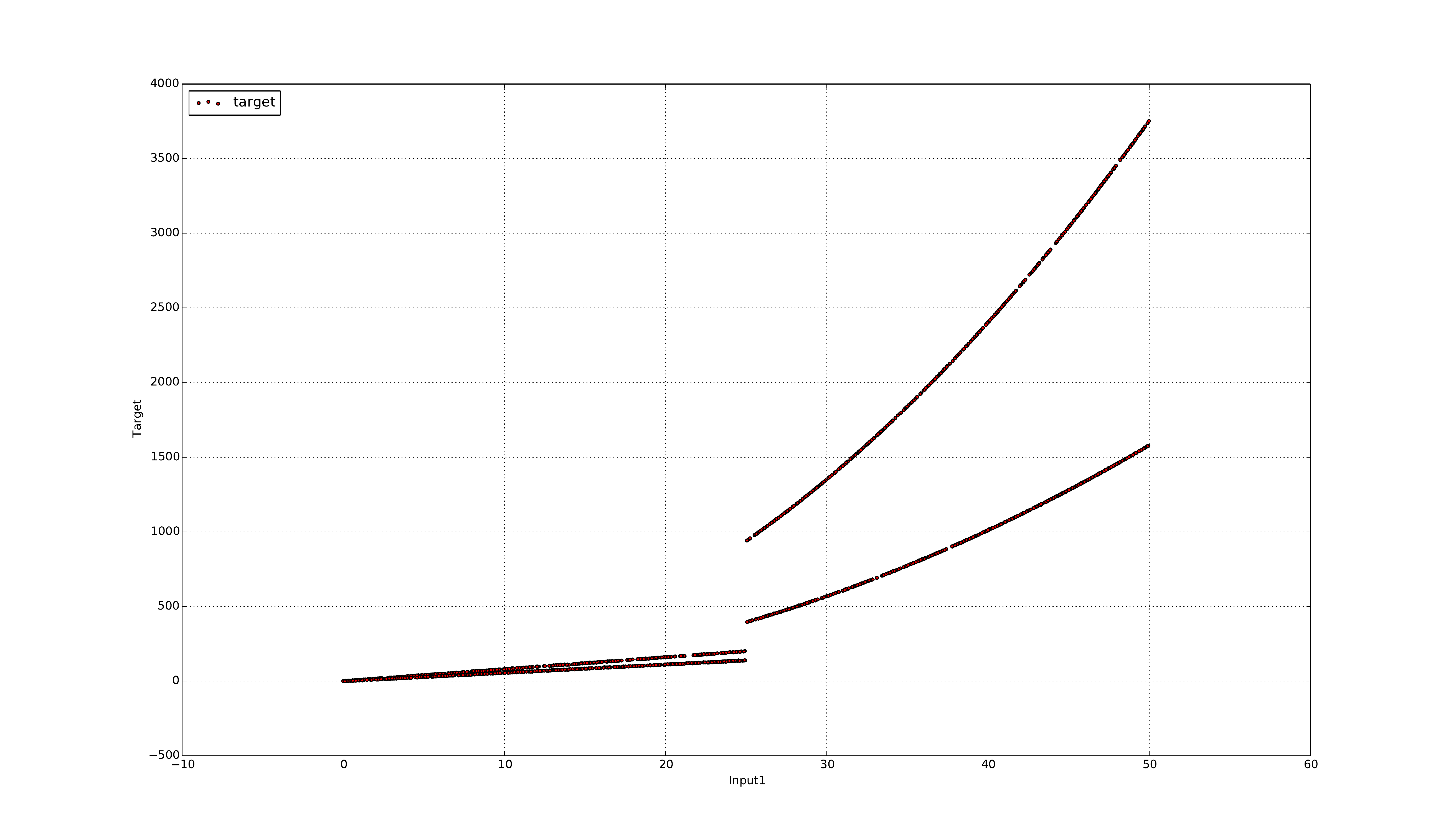}
  \caption{Visualization of the stream simulated from dataset} \texttt{SYNTH\_D\_CD\_2000\_1\_50\_1\_13}
  \label{fig:ref_func_SYNTH_D_CD_2000_1_50_1_13}
\end{figure}

\begin{figure}[htbp]
  \centering
    \includegraphics[width=\linewidth]{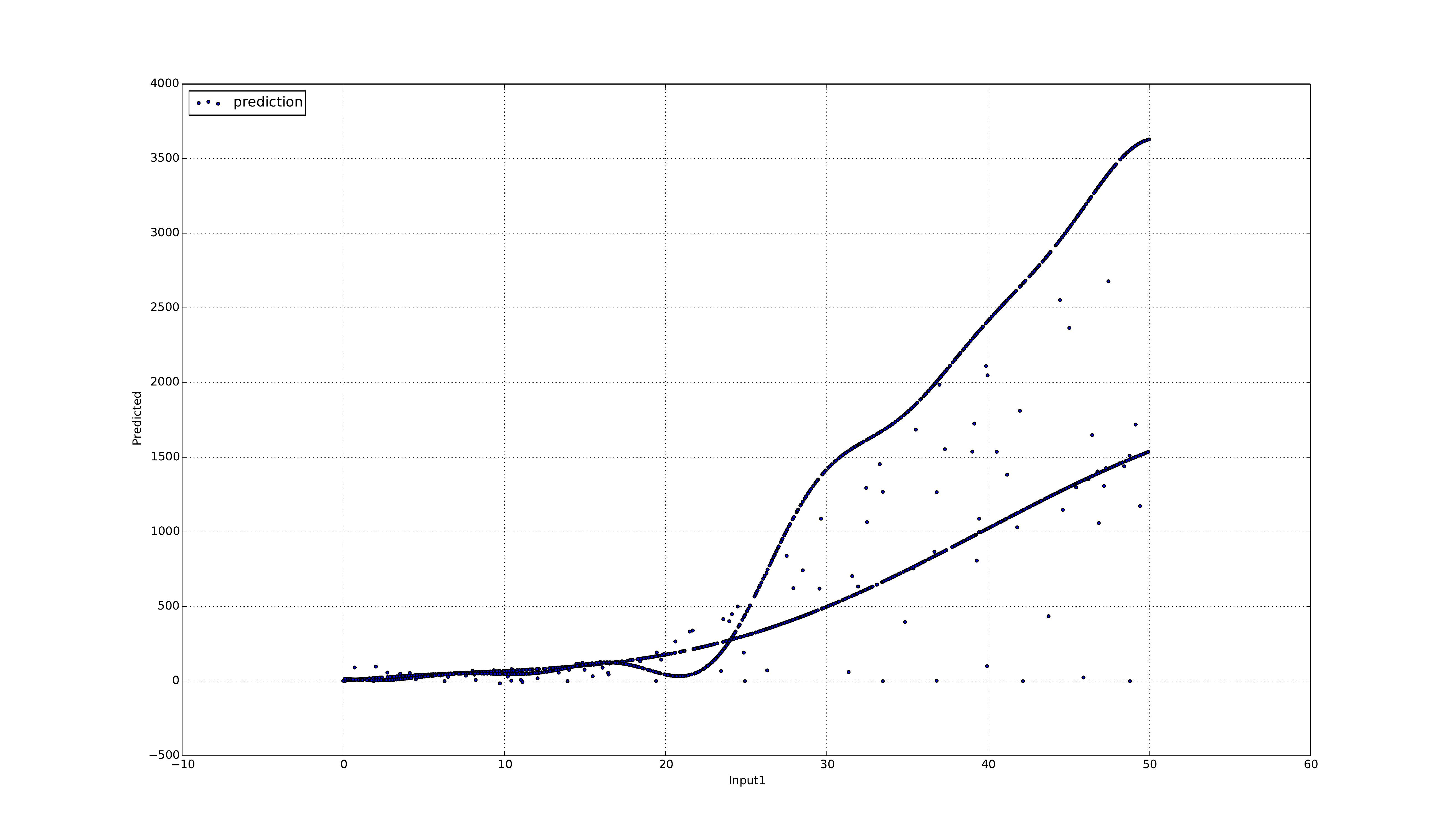}
  \caption{Predictions of \texttt{GPRegressionZeroMean\_WS64} on the stream simulated from the dataset \texttt{SYNTH\_D\_CD\_2000\_1\_50\_1\_13}}
  \label{fig:gpreg_zeromean_ws64_approximated_func_SYNTH_D_CD_2000_1_50_1_13}
\end{figure}

\begin{figure}[htbp]
  \centering
    \includegraphics[width=\linewidth]{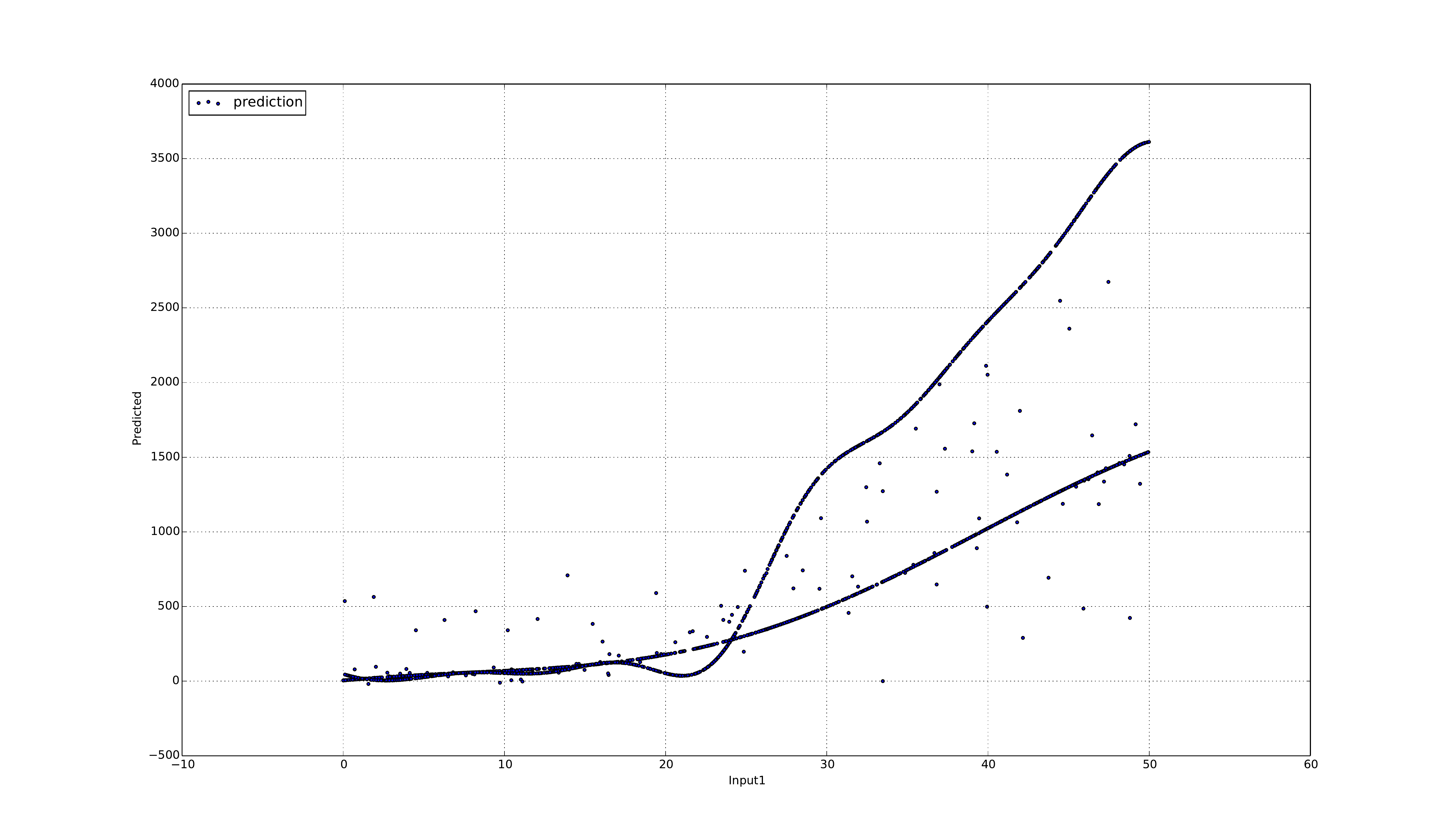}
  \caption{Predictions of \texttt{GPRegressionAvgMean\_WS64} on the stream simulated from the dataset \texttt{SYNTH\_D\_CD\_2000\_1\_50\_1\_13}}
  \label{fig:gpreg_avgmean_ws64_approximated_func_SYNTH_D_CD_2000_1_50_1_13}
\end{figure}

\begin{figure}[htbp]
  \centering
    \includegraphics[width=\linewidth]{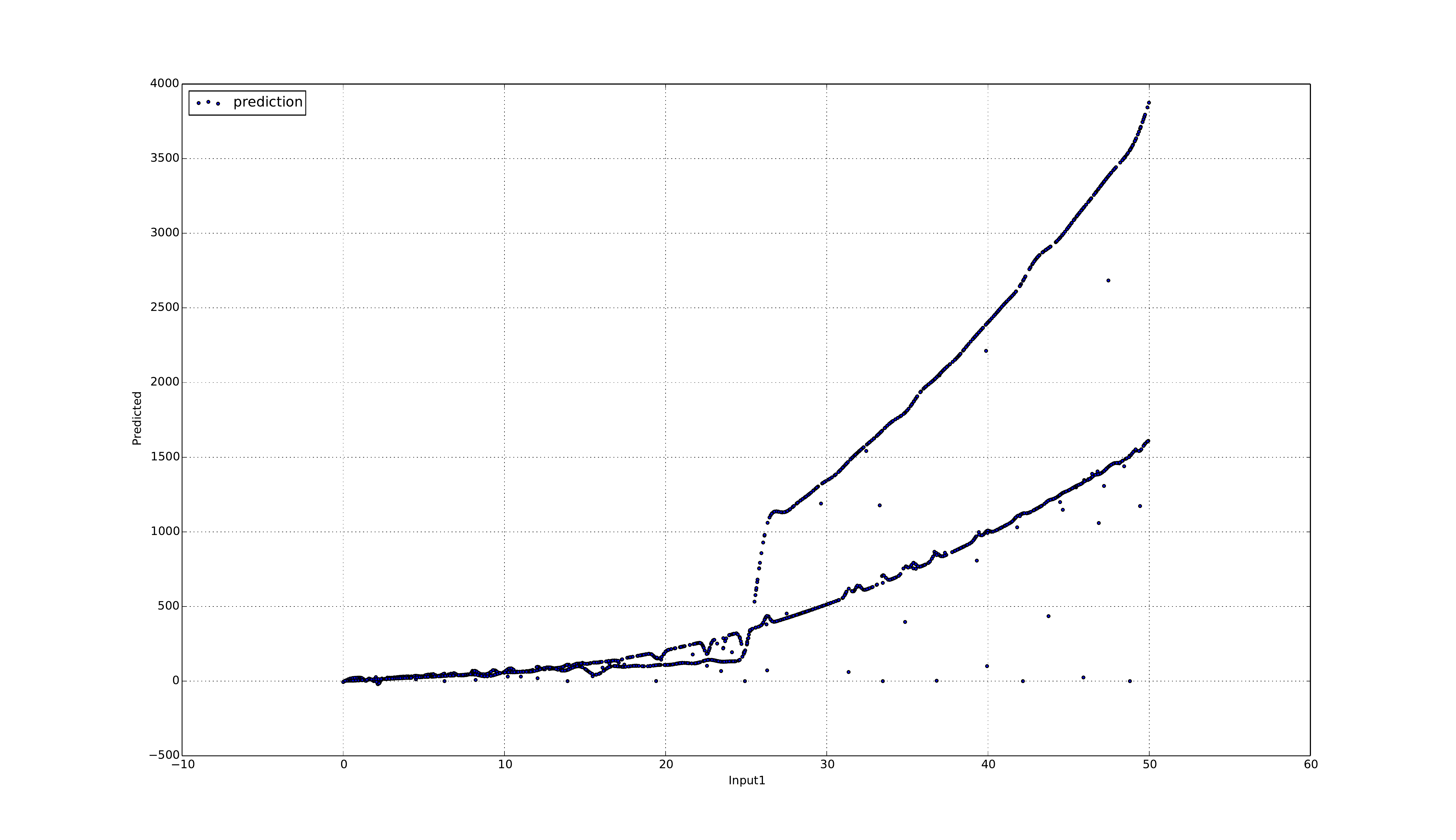}
  \caption{Predictions of \texttt{GPRegressionOLSMean\_WS64} on the stream simulated from the dataset \texttt{SYNTH\_D\_CD\_2000\_1\_50\_1\_13}}
  \label{fig:gpreg_olsmean_ws64_approximated_func_SYNTH_D_CD_2000_1_50_1_13}
\end{figure}

\begin{figure}[htbp]
  \centering
    \includegraphics[width=\linewidth]{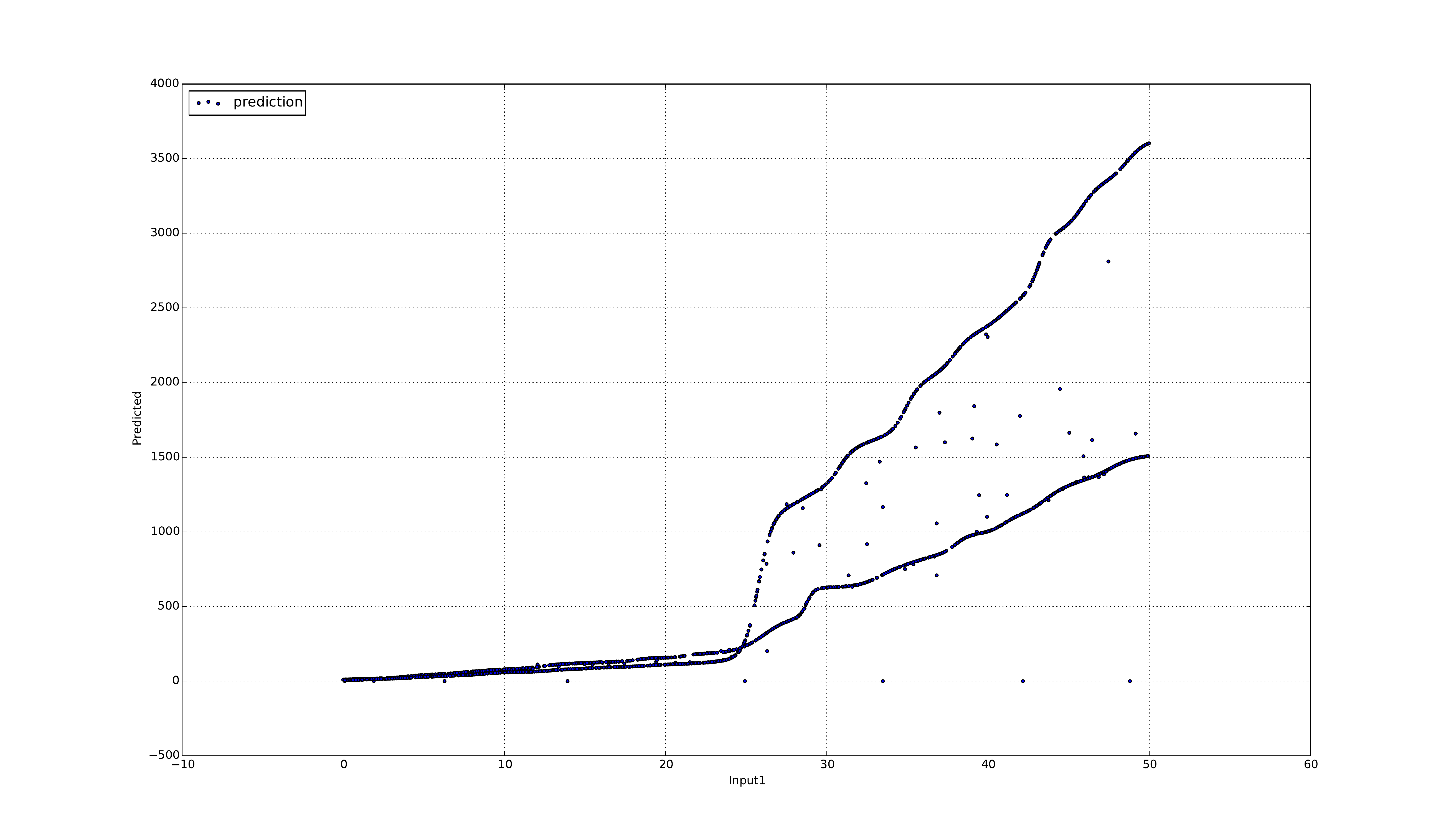}
  \caption{Predictions of \texttt{KernelRegression\_HighConf\_WS64} on the stream simulated from the dataset \texttt{SYNTH\_D\_CD\_2000\_1\_50\_1\_13}}
  \label{fig:kreg_ws64_approximated_func_SYNTH_D_CD_2000_1_50_1_13}
\end{figure}

\begin{figure}[htbp]
  \centering
    \includegraphics[width=\linewidth]{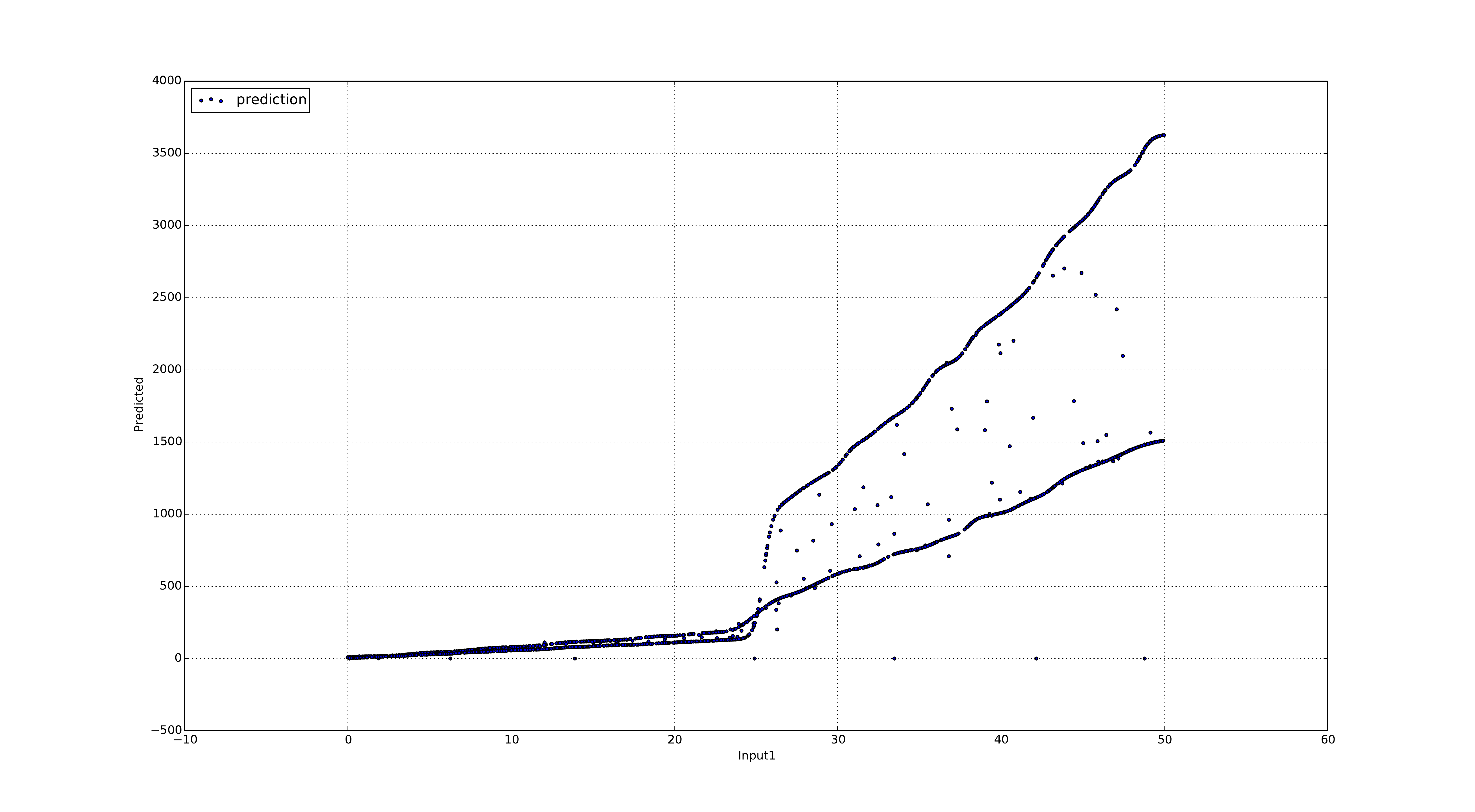}
  \caption{Predictions of \texttt{KernelRegression\_HighConf\_WS96} on the stream simulated from the dataset \texttt{SYNTH\_D\_CD\_2000\_1\_50\_1\_13}}
  \label{fig:kreg_ws96_approximated_func_SYNTH_D_CD_2000_1_50_1_13}
\end{figure}

\begin{figure}[htbp]
  \centering
    \includegraphics[width=\linewidth]{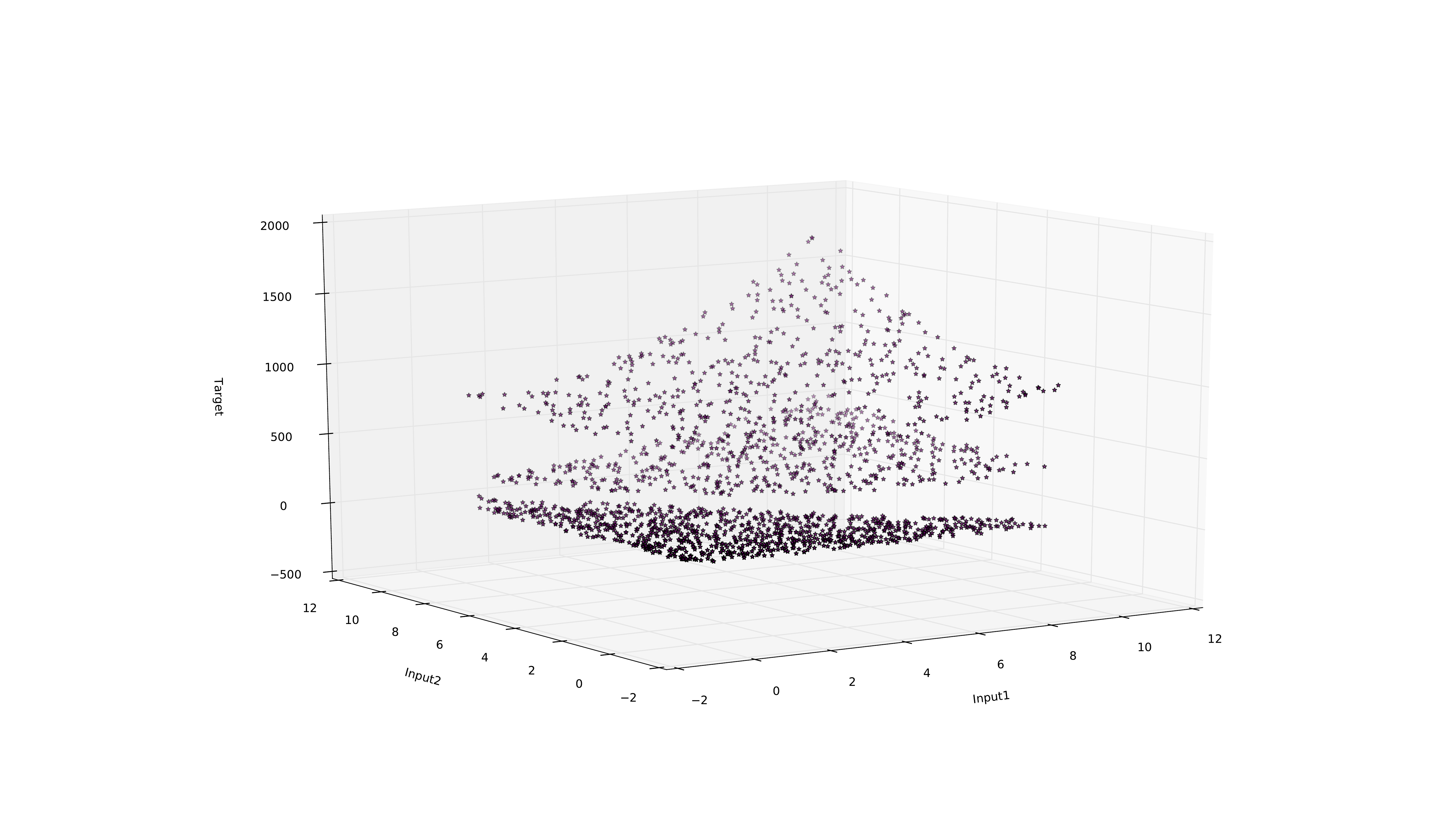}
  \caption{Visualization of the stream simulated from dataset} \texttt{SYNTH\_D\_CD\_2000\_2\_10\_1\_13}
  \label{fig:ref_func_SYNTH_D_CD_2000_2_10_1_13}
\end{figure}

\begin{figure}[htbp]
  \centering
    \includegraphics[width=\linewidth]{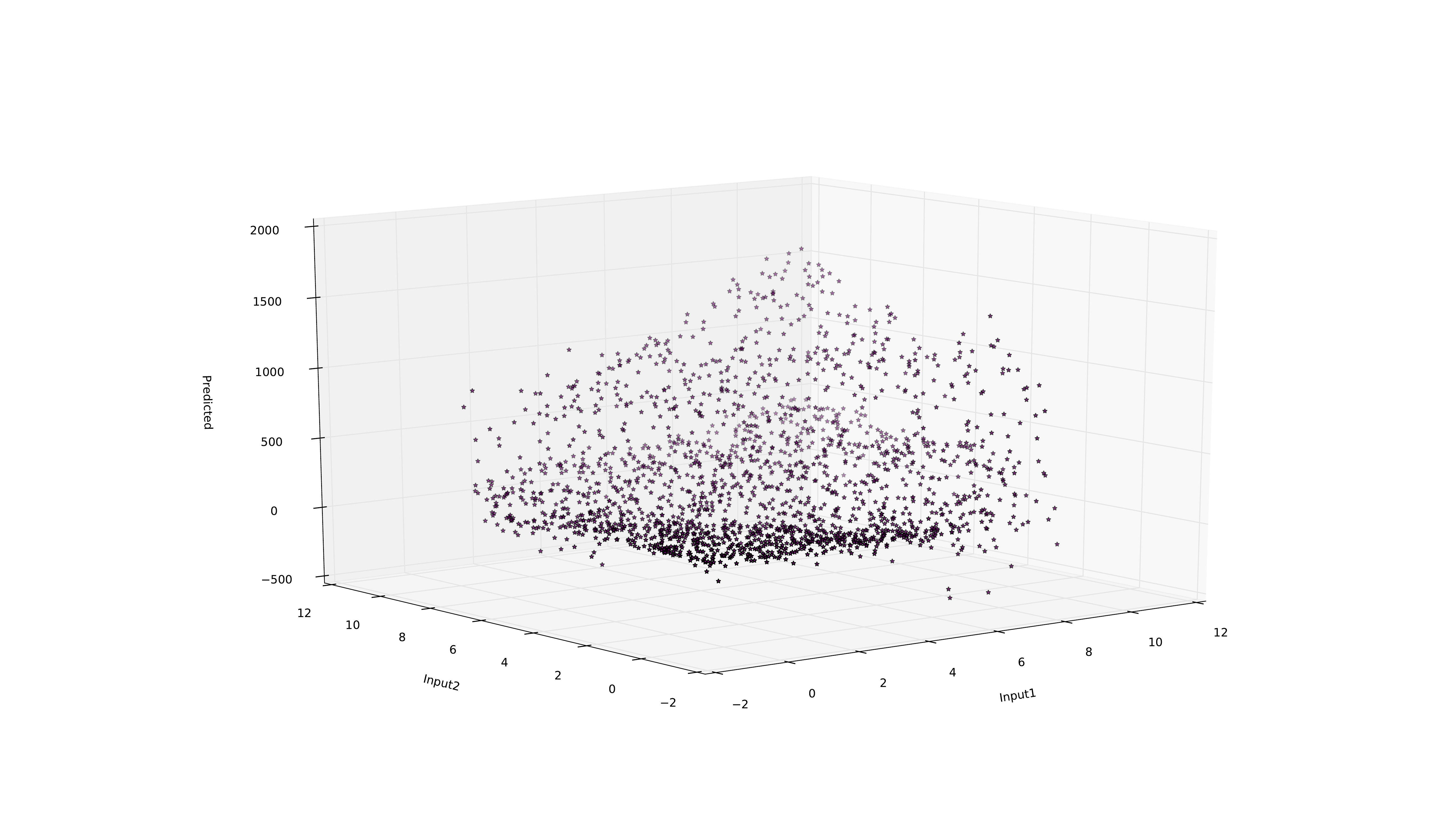}
  \caption{Predictions of \texttt{GPRegressionZeroMean\_WS64} on the stream simulated from the dataset \texttt{SYNTH\_D\_CD\_2000\_2\_10\_1\_13}}
  \label{fig:gpreg_zeromean_ws64_approximated_func_SYNTH_D_CD_2000_2_10_1_13}
\end{figure}

\begin{figure}[htbp]
  \centering
    \includegraphics[width=\linewidth]{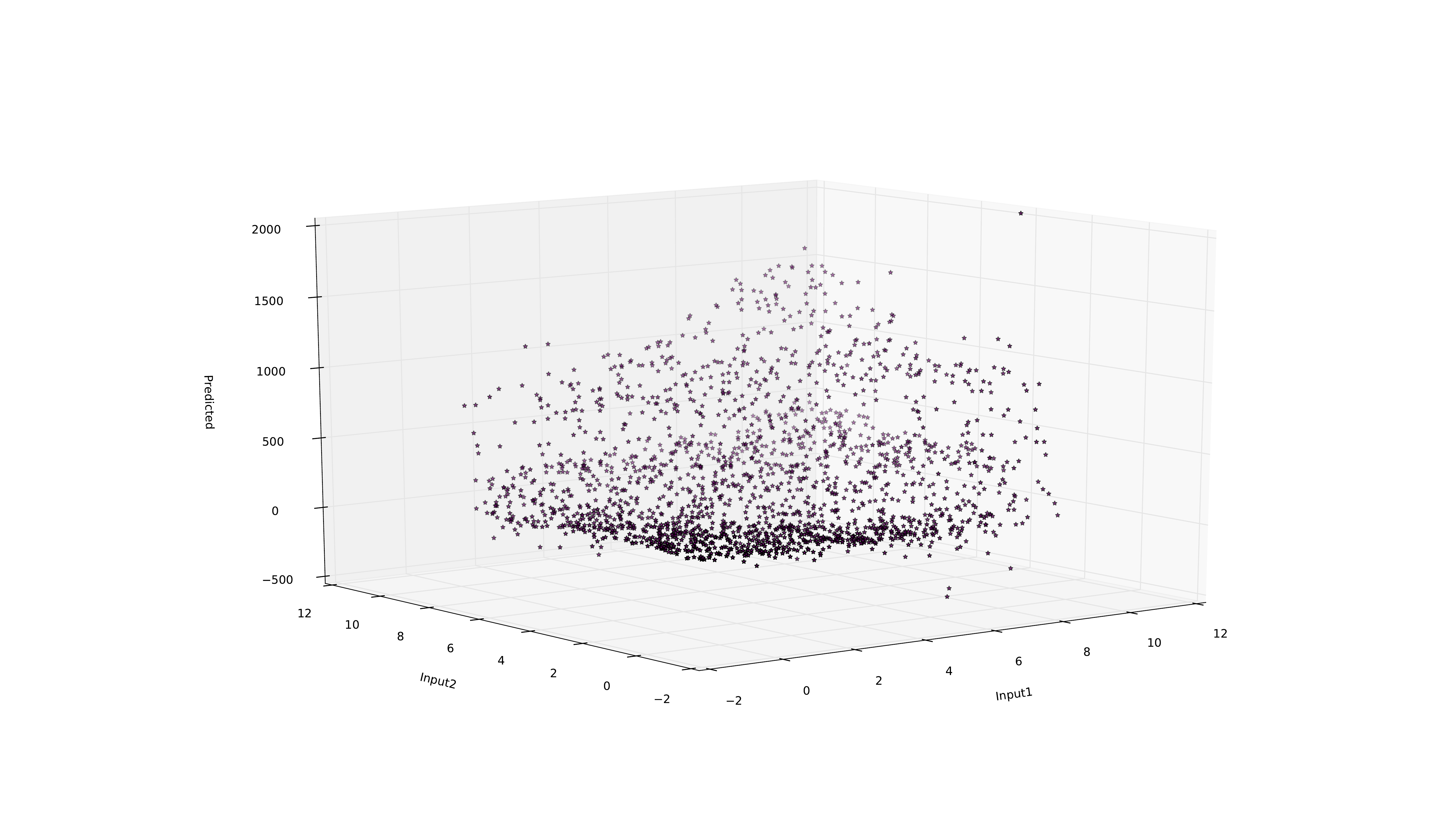}
  \caption{Predictions of \texttt{GPRegressionAvgMean\_WS64} on the stream simulated from the dataset \texttt{SYNTH\_D\_CD\_2000\_2\_10\_1\_13}}
  \label{fig:gpreg_avgmean_ws64_approximated_func_SYNTH_D_CD_2000_2_10_1_13}
\end{figure}

\begin{figure}[htbp]
  \centering
    \includegraphics[width=\linewidth]{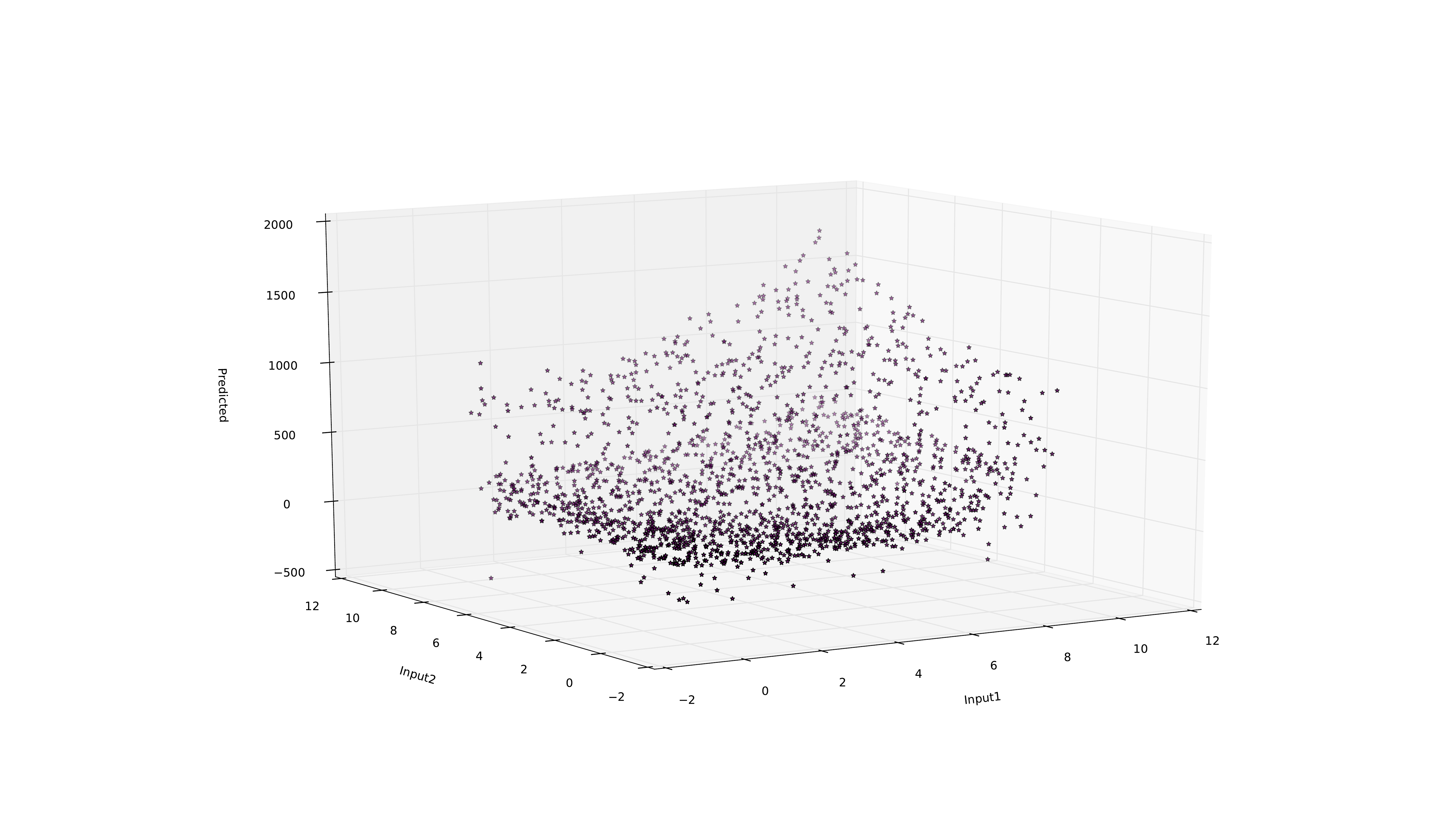}
  \caption{Predictions of \texttt{GPRegressionOLSMean\_WS64} on the stream simulated from the dataset \texttt{SYNTH\_D\_CD\_2000\_2\_10\_1\_13}}
  \label{fig:gpreg_olsmean_ws64_approximated_func_SYNTH_D_CD_2000_2_10_1_13}
\end{figure}

\begin{figure}[htbp]
  \centering
    \includegraphics[width=\linewidth]{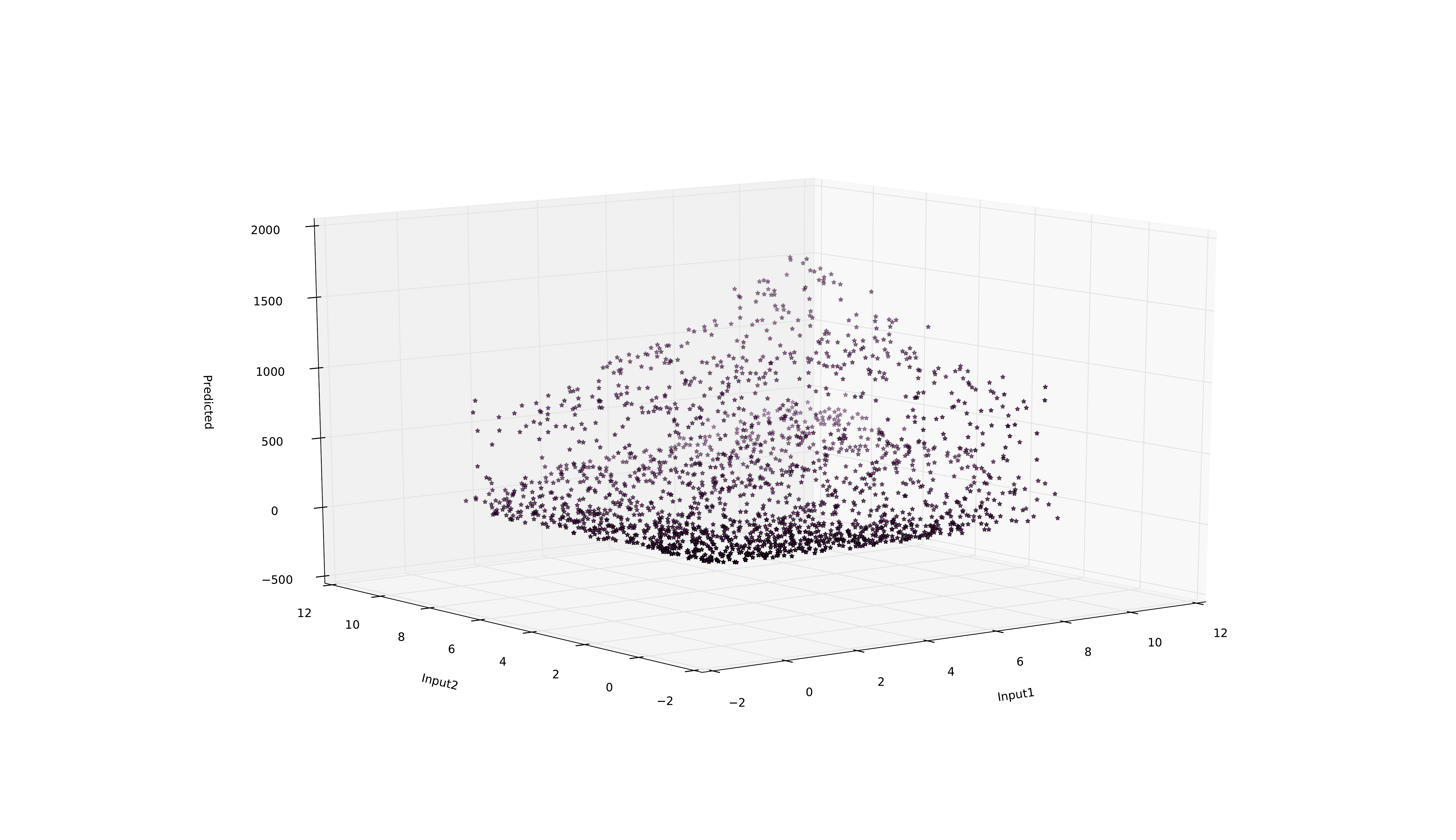}
  \caption{Predictions of \texttt{KernelRegression\_HighConf\_WS64} on the stream simulated from the dataset \texttt{SYNTH\_D\_CD\_2000\_2\_10\_1\_13}}
  \label{fig:kreg_ws64_approximated_func_SYNTH_D_CD_2000_2_10_1_13}
\end{figure}

\begin{figure}[htbp]
  \centering
    \includegraphics[width=\linewidth]{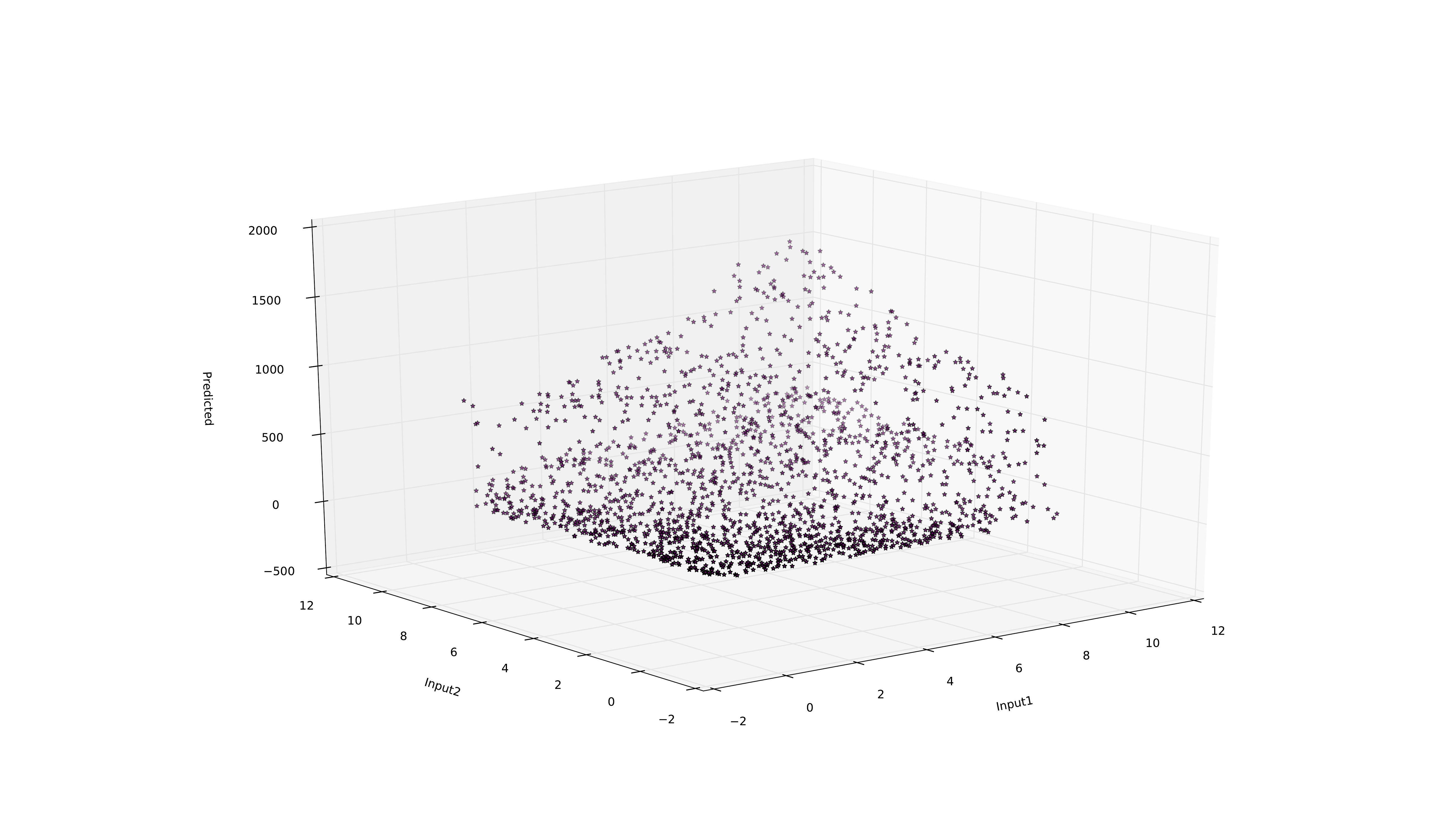}
  \caption{Predictions of \texttt{KernelRegression\_HighConf\_WS96} on the stream simulated from the dataset \texttt{SYNTH\_D\_CD\_2000\_2\_10\_1\_13}}
  \label{fig:kreg_ws96_approximated_func_SYNTH_D_CD_2000_2_10_1_13}
\end{figure}

\clearpage

\subsection{Visualization of the cost function approximations by the shortlisted learners on the measurement data}

In this subsection, predictions of the shortlisted learners listed in \ref{final_5} on the streams simulated from various data sets containing Ocelot runtime measurements.

\begin{figure}[htbp]
  \centering
    \includegraphics[width=\linewidth]{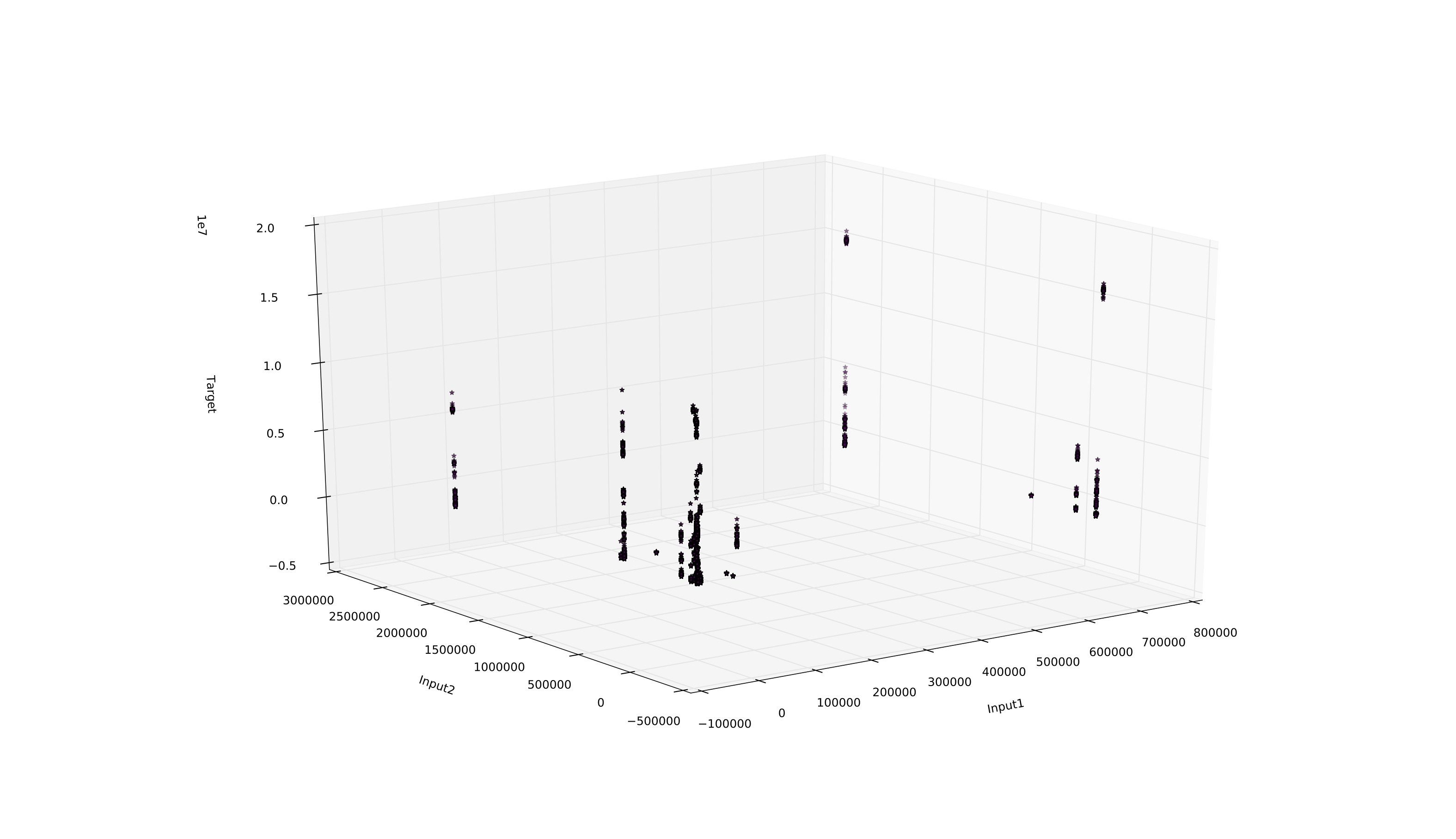}
  \caption{Visualization of the measurement data for the left fetch join operator on GPU. Data set includes 30075 data point-target pairs}
  \label{ref_ocl_leftfetchjoin_on_gpu}
\end{figure}

\begin{figure}[htbp]
  \centering
    \includegraphics[width=\linewidth]{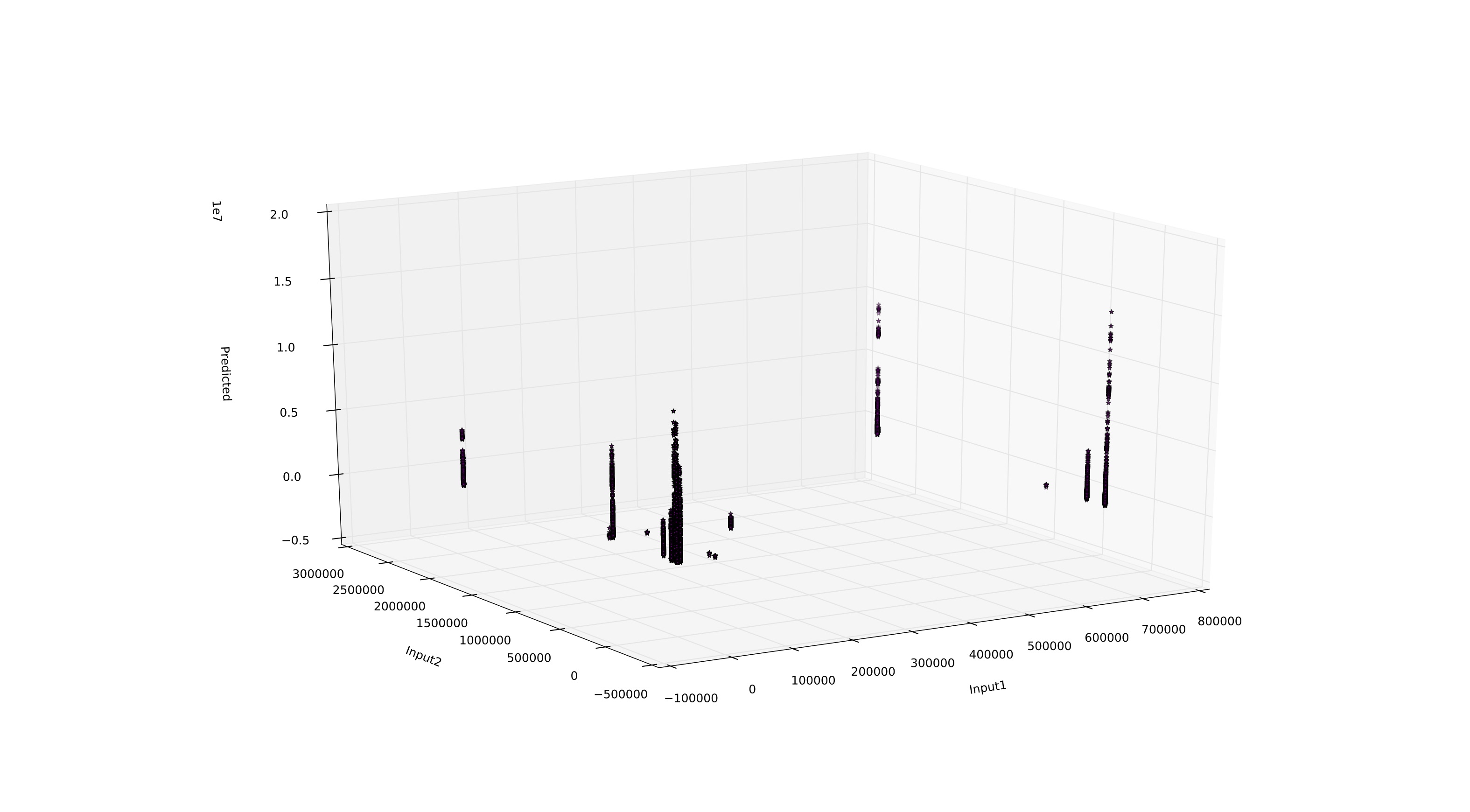}
  \caption{Predictions of \texttt{GPRegressionZeroMean\_WS64} on the runtime measurement data for the left fetch join operator on GPU}
  \label{gpreg_zeromean_ws64_ocl_leftfetchjoin_on_gpu}
\end{figure}

\begin{figure}[htbp]
  \centering
    \includegraphics[width=\linewidth]{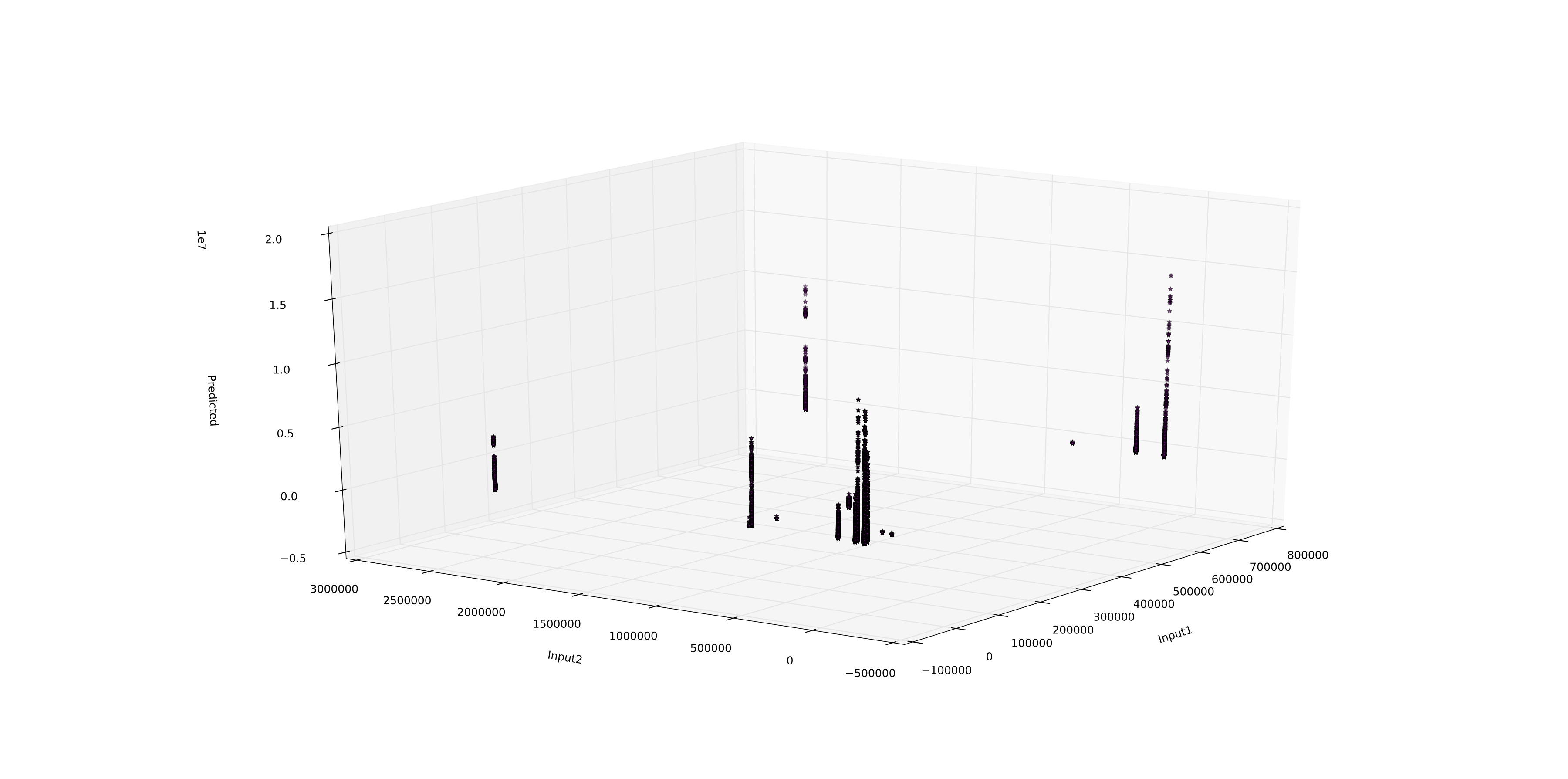}
  \caption{Predictions of \texttt{GPRegressionAvgMean\_WS64} on the runtime measurement data for the left fetch join operator on GPU}
  \label{gpreg_avgmean_ws64_ocl_leftfetchjoin_on_gpu}
\end{figure}

\begin{figure}[htbp]
  \centering
    \includegraphics[width=\linewidth]{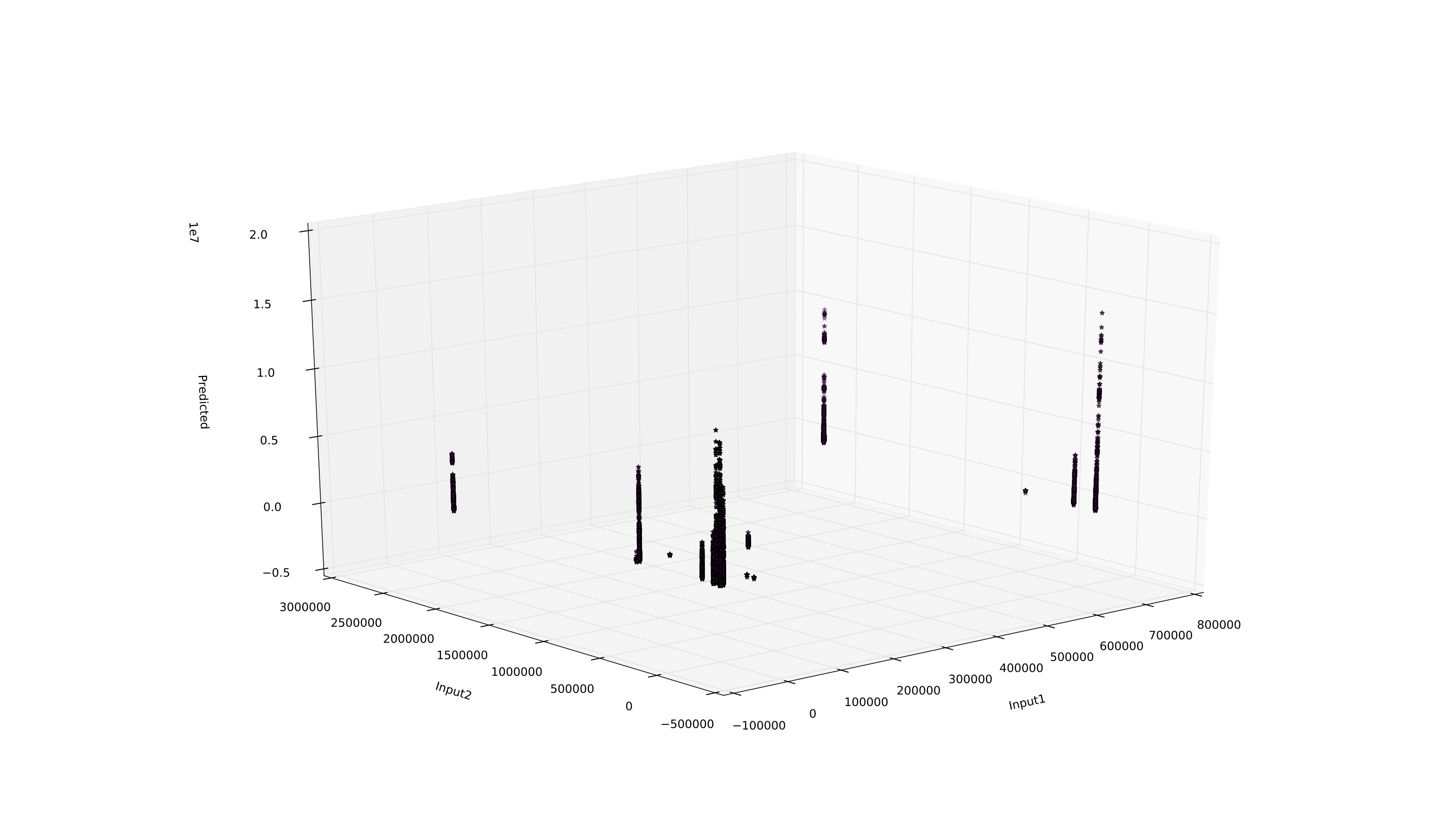}
  \caption{Predictions of \texttt{GPRegressionOLSMean\_WS64} on the runtime measurement data for the left fetch join operator on GPU}
  \label{gpreg_olsmean_ws64_ocl_leftfetchjoin_on_gpu}
\end{figure}

\begin{figure}[htbp]
  \centering
    \includegraphics[width=\linewidth]{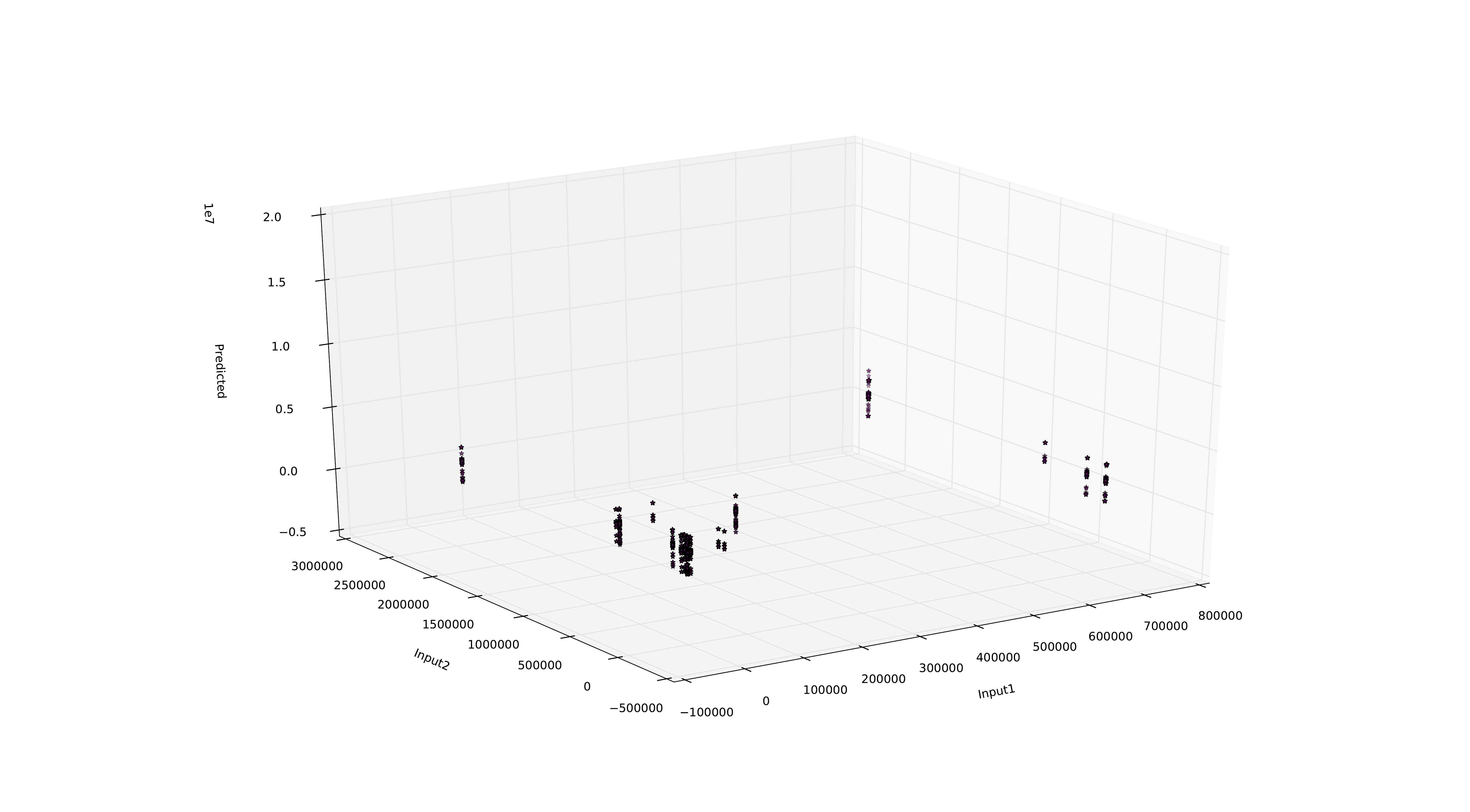}
  \caption{Predictions of \texttt{KernelRegression\_HighConf\_WS64} on the runtime measurement data for the left fetch join operator on GPU}
  \label{kreg_ws64_ocl_leftfetchjoin_on_gpu}
\end{figure}

\begin{figure}[htbp]
  \centering
    \includegraphics[width=\linewidth]{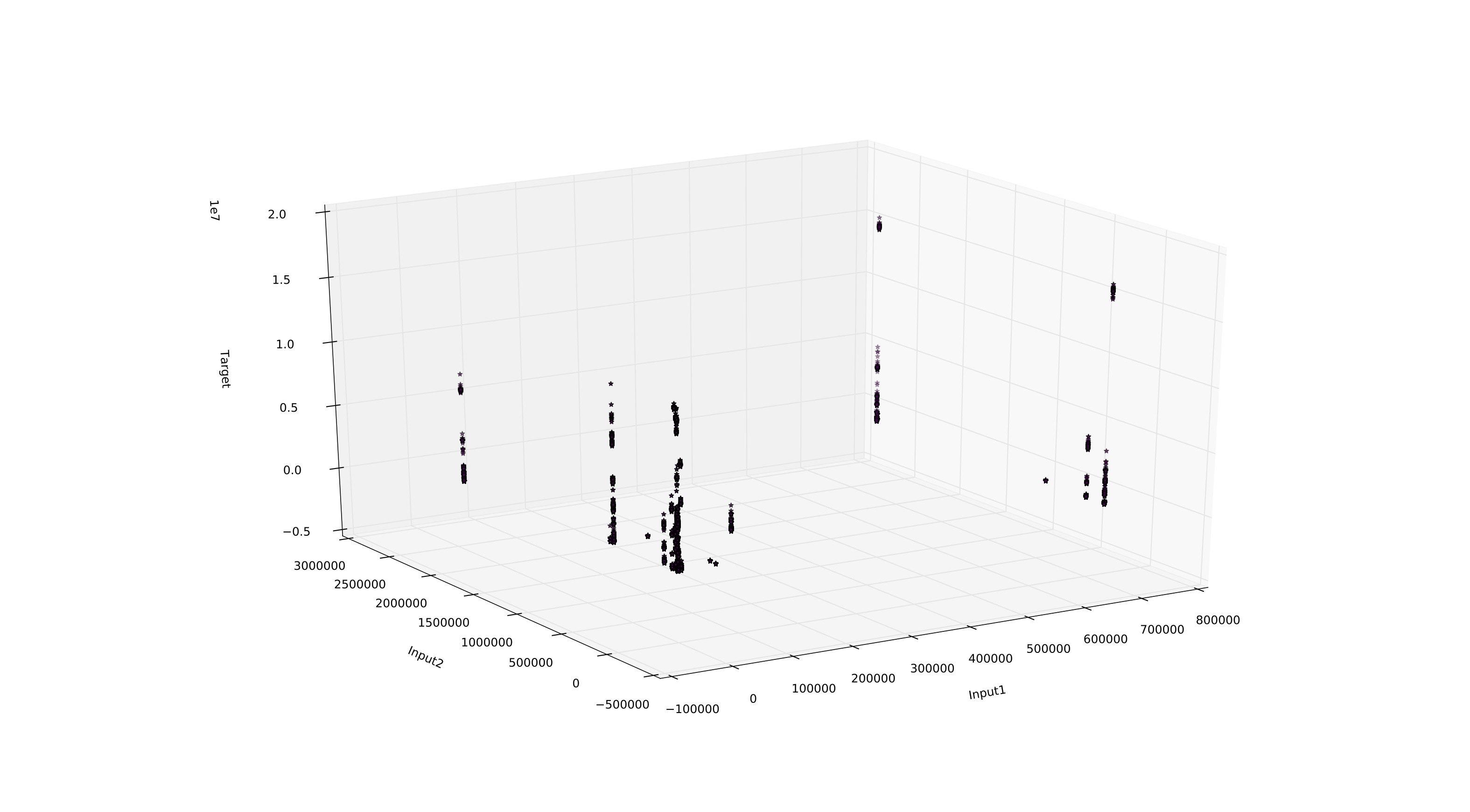}
  \caption{Predictions of \texttt{KernelRegression\_HighConf\_WS96} on the runtime measurement data for the left fetch join operator on GPU}
  \label{kreg_ws96_ocl_leftfetchjoin_on_gpu}
\end{figure}

\begin{figure}[htbp]
  \centering
    \includegraphics[width=\linewidth]{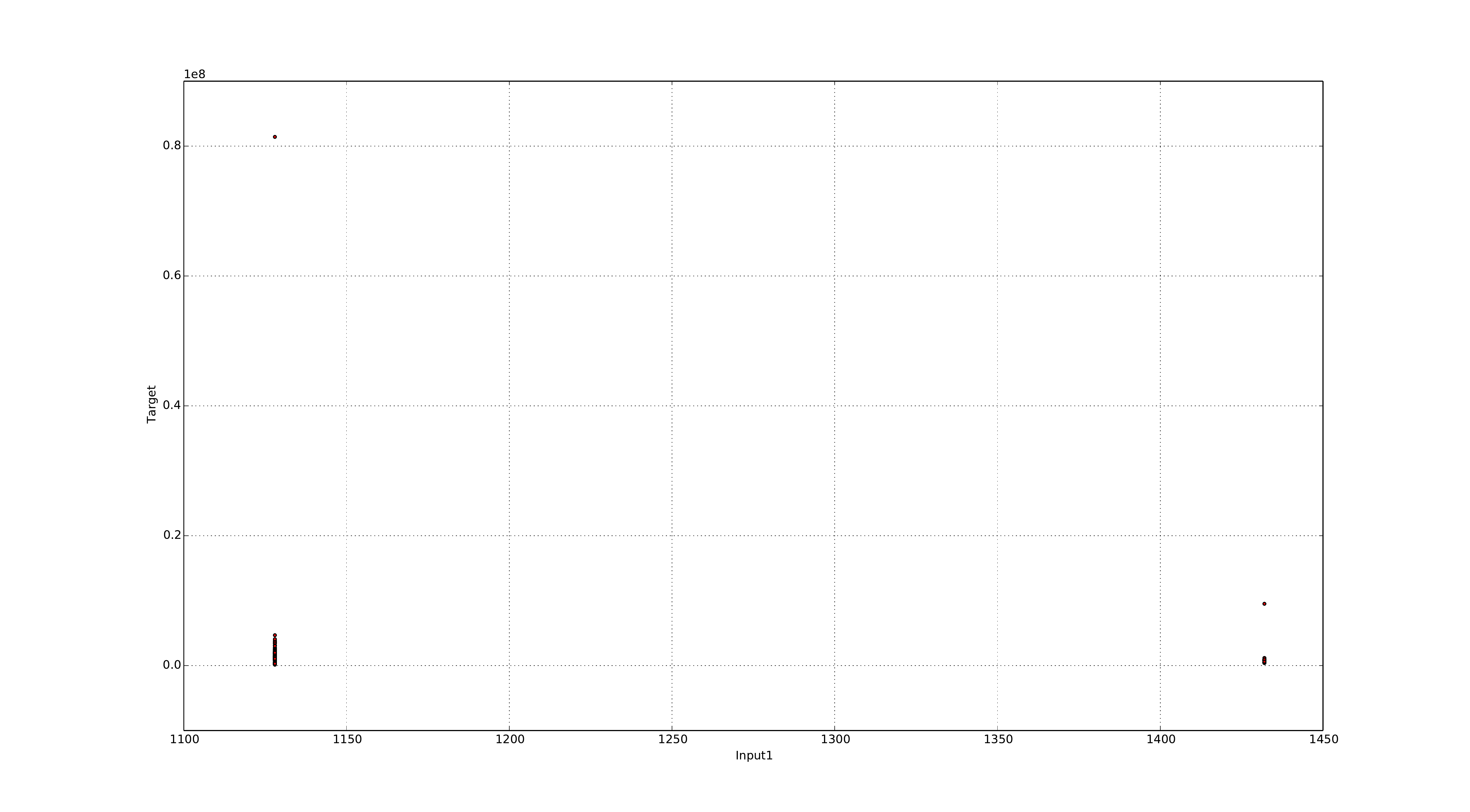}
  \caption{Visualization of the measurement data for the groupby operator on CPU. Data set includes 592 data point-target pairs}
  \label{ref_ocl_groupby_on_device_on_cpu}
\end{figure}

\begin{figure}[htbp]
  \centering
    \includegraphics[width=\linewidth]{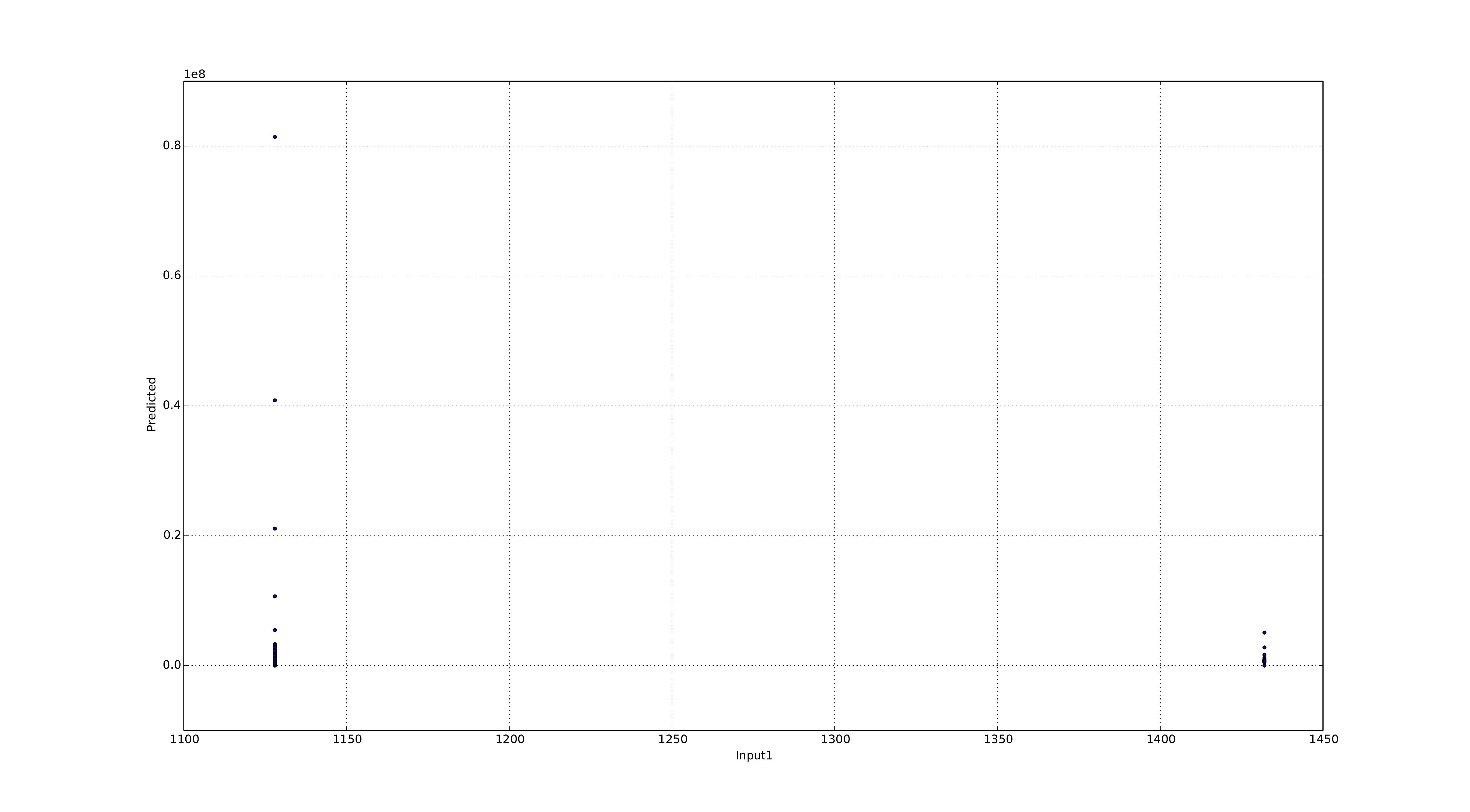}
  \caption{Predictions of \texttt{GPRegressionZeroMean\_WS64} on the runtime measurement data for the groupby operator on CPU}
  \label{gpreg_zeromean_ws64_ocl_groupby_on_device_on_cpu}
\end{figure}

\begin{figure}[htbp]
  \centering
    \includegraphics[width=\linewidth]{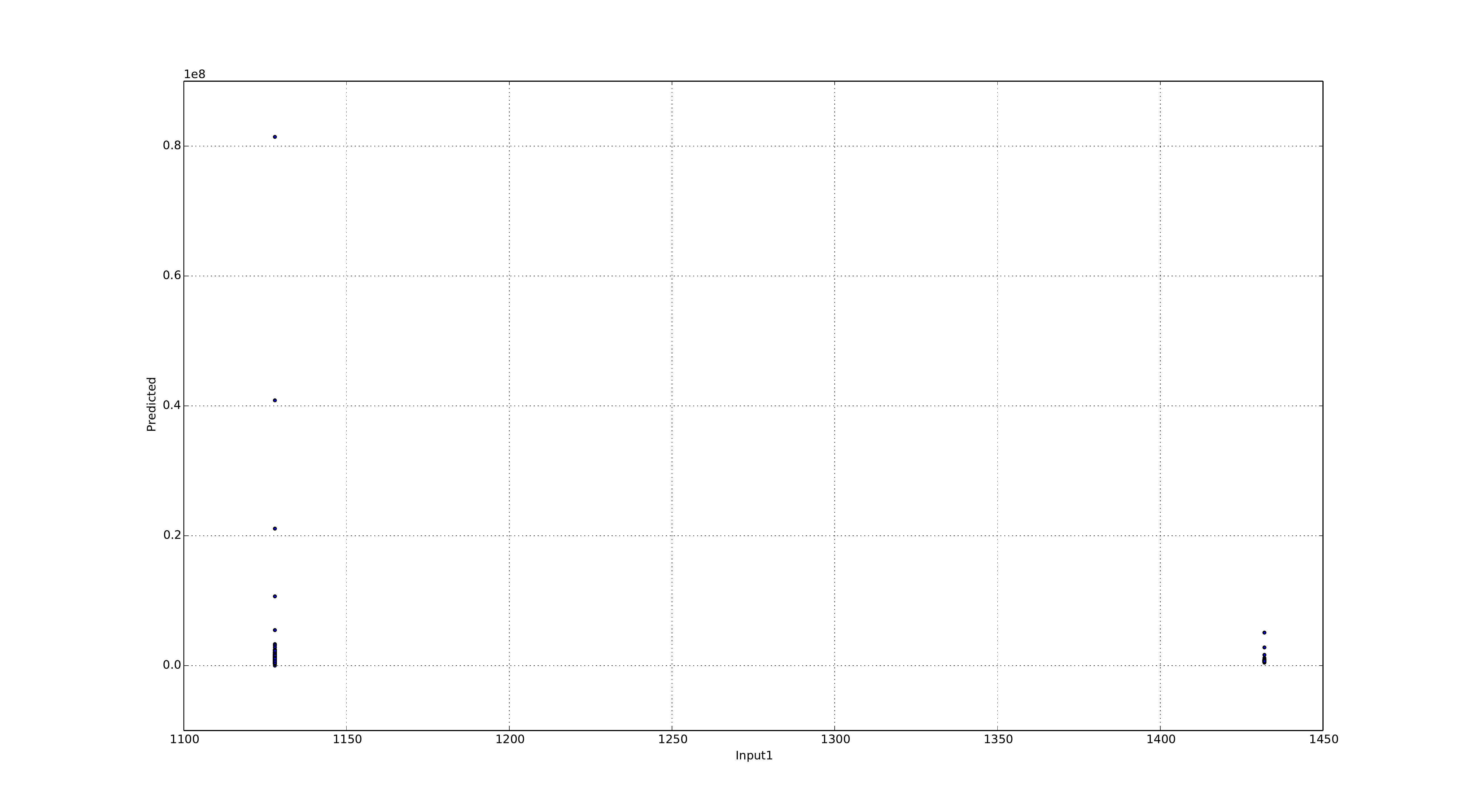}
  \caption{Predictions of \texttt{GPRegressionAvgMean\_WS64} on the runtime measurement data for the groupby operator on CPU}
  \label{gpreg_avgmean_ws64_ocl_groupby_on_device_on_cpu}
\end{figure}

\begin{figure}[htbp]
  \centering
    \includegraphics[width=\linewidth]{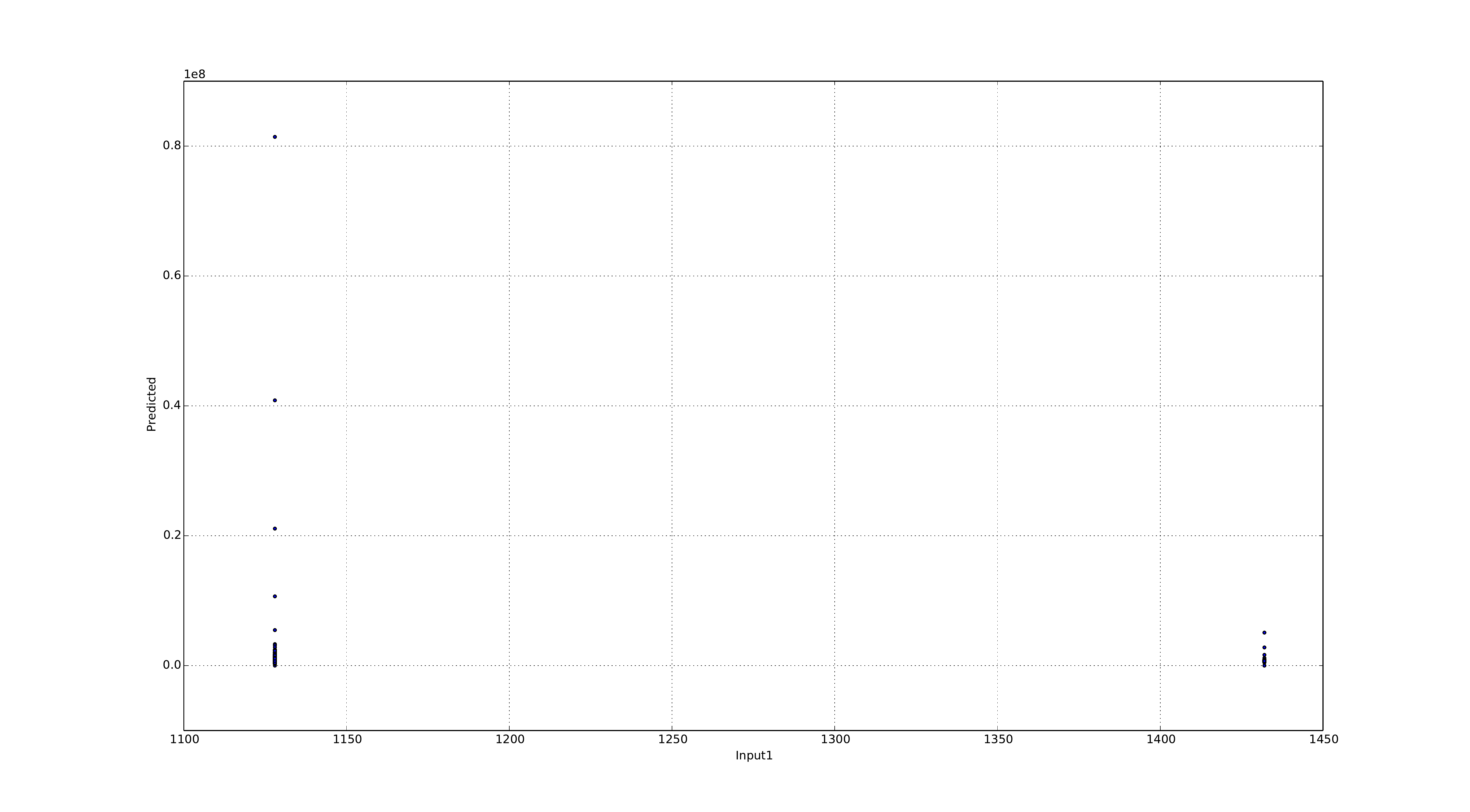}
  \caption{Predictions of \texttt{GPRegressionOLSMean\_WS64} on the runtime measurement data for the groupby operator on CPU}
  \label{gpreg_olsmean_ws64_ocl_groupby_on_device_on_cpu}
\end{figure}

\begin{figure}[htbp]
  \centering
    \includegraphics[width=\linewidth]{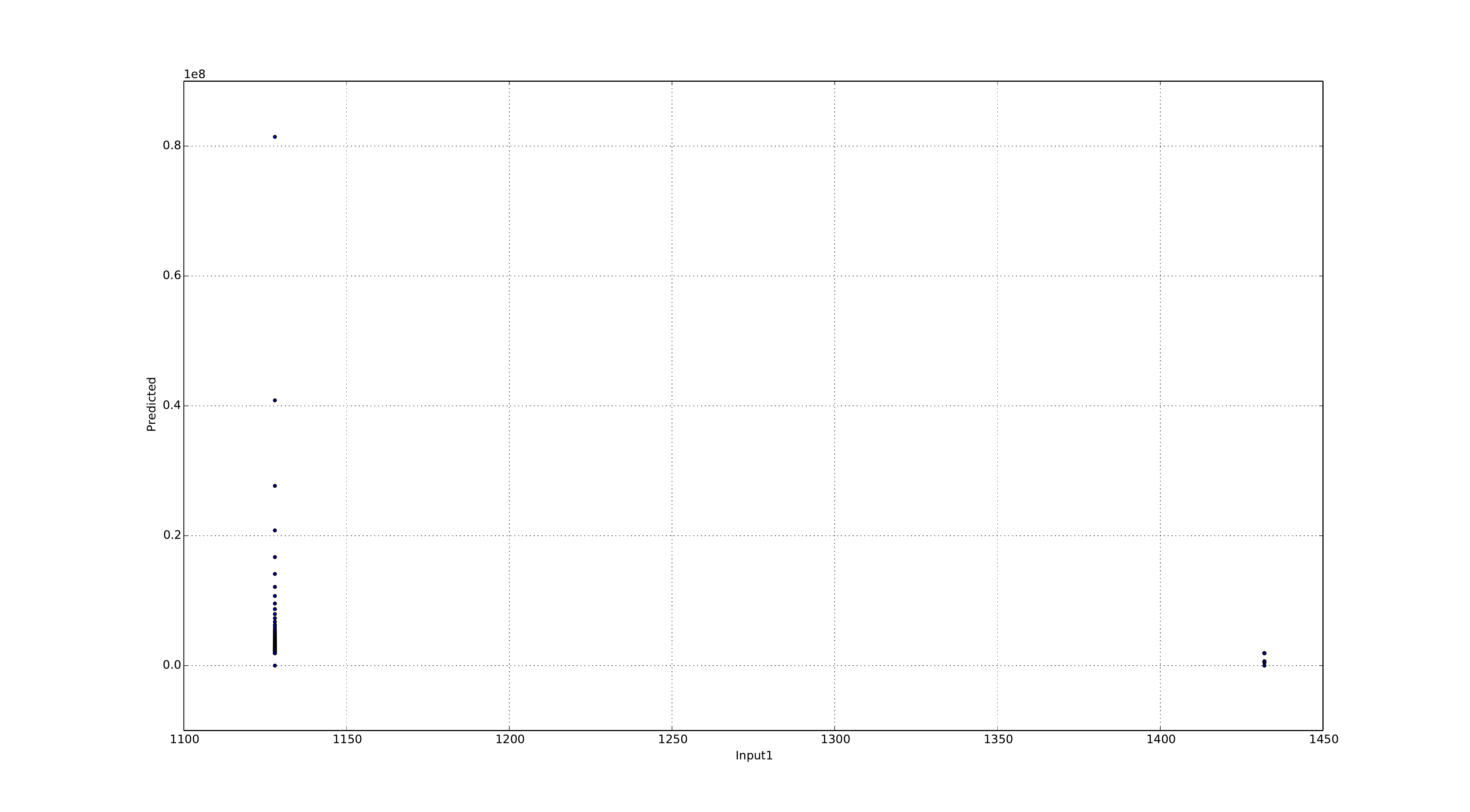}
  \caption{Predictions of \texttt{KernelRegression\_HighConf\_WS64} on the runtime measurement data for the groupby operator on CPU}
  \label{kreg_ws64_ocl_groupby_on_device_on_cpu}
\end{figure}

\begin{figure}[htbp]
  \centering
    \includegraphics[width=\linewidth]{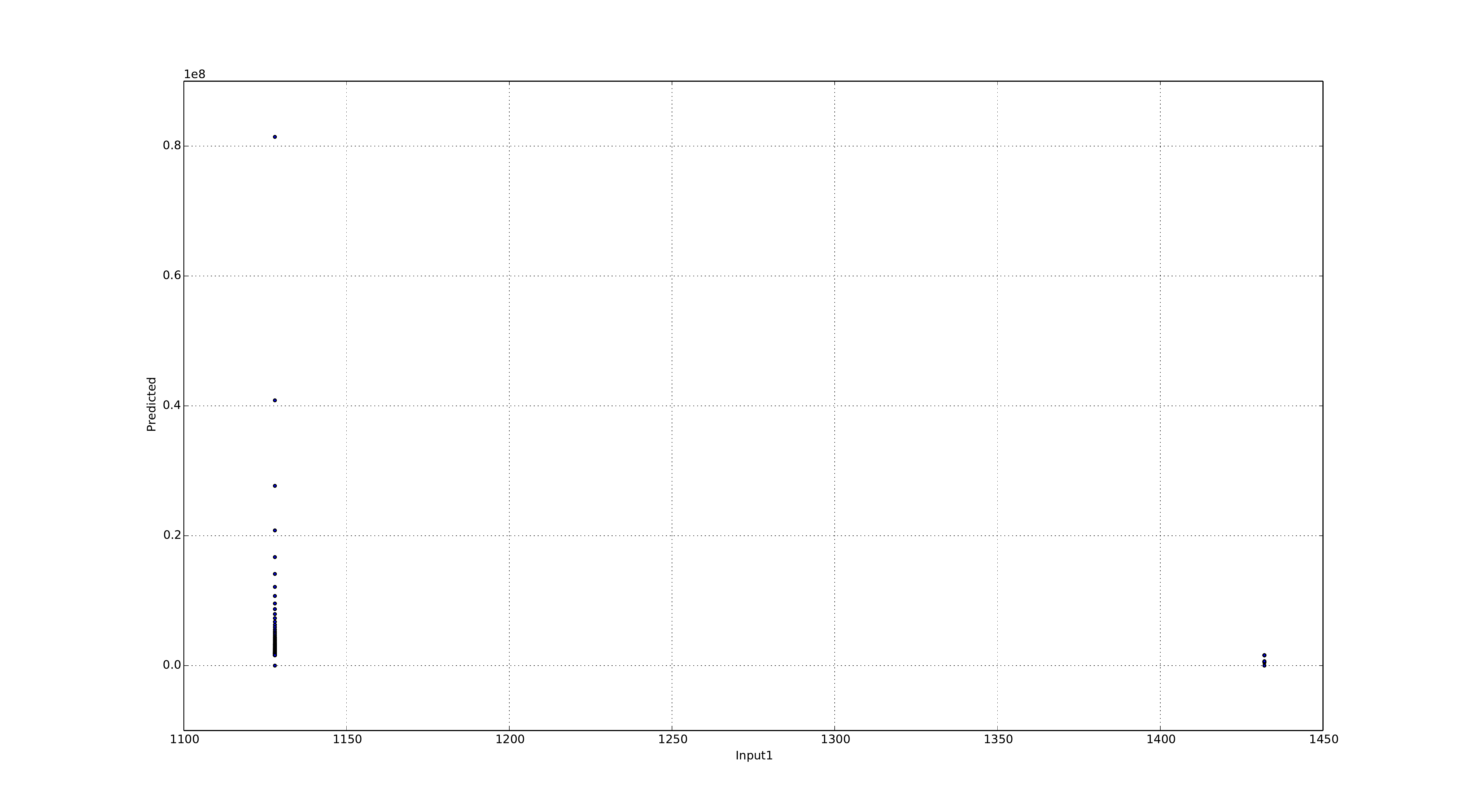}
  \caption{Predictions of \texttt{KernelRegression\_HighConf\_WS96} on the runtime measurement data for the groupby operator on CPU}
  \label{kreg_ws96_ocl_groupby_on_device_on_cpu}
\end{figure}

\chapter{Conclusion} 

\label{Chapter7} 

\lhead{Chapter 7. \emph{Conclusion}} 

\section{Final Remarks on Online Regression Algorithms}

In the light of the experiment results presented and detailed comparison of the implemented online learners using different online regression algorithms in Chapter \ref{Chapter4}, in this section, a high-level summary of the comparison of online regression algorithms explored is presented.

There are $4$ online regression algorithms explored in this thesis namely MLE-method, MAP-method, Gaussian Process Regression and Kernel Regression. Although many different variations of these algorithms are implemented, the performance characteristics of the variations from the same algorithm family appeared to be similar. This makes it possible to make general comments on different algorithms without drilling down to the details of different implementations of them. 

The criteria chosen for evaluating the online learners have three dimensions namely the  accuracy, prediction-bounds quality and time-efficiency. In Chapter \ref{Chapter6}, $40$ online learners implementing $4$ different algorithms are evaluated according to these criteria by the various evaluation various metrics defined. Without referring to any evaluation metrics for the sake of delivering a concise comparison, the table below summarizes the comparison of online regression algorithms with respect to the main dimensions of the evaluation criteria.

\begin{table}[]
\centering
\label{table}
\begin{tabular}{lllll}
                          & \texttt{BayesianMLE} & \texttt{BayesianMAP} & \texttt{GPRegression} & \texttt{KernelRegression} \\
Accuracy                  & +          & +          & +++                         & +++               \\
Prediction Bounds & +          & +          & +++                          & ++               \\
Time Efficiency           & +++        & +++        & +                           & ++               
\end{tabular}
\caption{High-level comparison of online regression algorithms}
\end{table}

\section{Future Work}

Online Machine Learning is a relatively new paradigm. The overwhelming majority of the learning algorithms provided in well-known machine learning libraries implement the traditional batch-learning paradigm. Moreover, in the case of regression algorithms, the regression problem is usually understood as building a predictive model that estimates the responses of an unknown function that is used for generating the data. Therefore, the regression algorithms are generally designed to make point predictions instead of returning prediction bounds along with point predictions. As a result, there is lack of regression algorithms that are incrementally updatable (online-learning ready) and provide prediction bounds at the same time. For example, there are very promising regression algorithms such as Support Vector Regression and Quantile Regression. However, despite the substantial interest in them by the machine learning community, to the best of the author's knowledge, there is no proposed incremental support vector regression algorithm or quantile regression algorithm that provide prediction bounds making them unsuitable to be employed in the Ocelot's runtime prediction module. Another example is Least Squares SVM. It provides prediction intervals that can be used as prediction bounds but it requires retraining every time a new data point is needed to be incorporated. Although recent work on regression algorithms attempt to provide either incremental update mechanism (\cite{poggio_incremental_2001}, \cite{martin_-line_2002}, \cite{ma_accurate_2003}, \cite{moller_time-adaptive_2008}) or prediction bounds estimation mechanism (\cite{chu_bayesian_2001}, \cite{de_brabanter_approximate_2011}, \cite{gao_probabilistic_2002}) for existing regression algorithms, these possible extensions are not usually offered together. In summary, there is significant amount of work yet to be done regarding online regression algorithms featuring prediction uncertainty estimation mechanism.

\addtocontents{toc}{\vspace{2em}} 

\appendix 



\addtocontents{toc}{\vspace{2em}} 

\backmatter


\label{Bibliography}

\lhead{\emph{Bibliography}} 

\bibliographystyle{unsrtnat} 

\bibliography{bibliography} 

\end{document}